\documentclass[english]{article}
\usepackage[T1]{fontenc}
\usepackage[latin9]{inputenc}
\usepackage{geometry}
\usepackage{color}
\usepackage{babel}
\usepackage{float}
\usepackage{mathtools}
\usepackage{bm}
\usepackage{dsfont}
\usepackage{amsmath}
\usepackage{amssymb}
\usepackage{graphicx}
\usepackage[unicode=true,
 bookmarks=true,bookmarksnumbered=false,bookmarksopen=false,
 breaklinks=false,pdfborder={0 0 0},pdfborderstyle={},backref=false,colorlinks=true]
 {hyperref}
\usepackage{subcaption}

\makeatletter

\floatstyle{ruled}
\newfloat{algorithm}{tbp}{loa}
\providecommand{\algorithmname}{Algorithm}
\floatname{algorithm}{\protect\algorithmname}

\usepackage{babel}
\usepackage{algorithmic}
\usepackage{amsthm}

\theoremstyle{plain}

\newtheorem{lemma}{\textbf{Lemma}}
\newtheorem{theorem}{\textbf{Theorem}}

\newtheorem{definition}{\textbf{Definition}}

\theoremstyle{definition}
\newtheorem{remark}{\textbf{Remark}}

\usepackage{color}
\definecolor{cm}{RGB}{0,0,200}
\definecolor{yy}{RGB}{200,0,200}

\usepackage{authblk}

\author{Yuepeng Yang} \author{Cong Ma} \affil{Department of Statistics, University of Chicago} 

\@ifundefined{showcaptionsetup}{}{%
 \PassOptionsToPackage{caption=false}{subfig}}
\makeatother

\begin{document}
\title{Estimating shared subspace with AJIVE: \\ the power and
limitation of multiple data matrices}

\maketitle
\begin{abstract}
Integrative data analysis often requires disentangling joint and individual variations across multiple datasets, a challenge commonly addressed by the Joint and Individual Variation Explained (JIVE) model. While numerous methods have been developed to estimate the shared subspace under JIVE, the theoretical understanding of their performance remains limited, particularly in the context of multiple matrices and varying degrees of subspace misalignment. This paper bridges this gap by providing a systematic analysis of shared subspace estimation in multi-matrix settings.

We focus on the Angle-based Joint and Individual Variation Explained (AJIVE) method, a two-stage spectral approach, and establish new performance guarantees that uncover its strengths and limitations. Specifically, we show that in high signal-to-noise ratio (SNR) regimes, AJIVE's estimation error decreases with the number of matrices, demonstrating the power of multi-matrix integration. Conversely, in low-SNR settings, AJIVE exhibits a non-diminishing error, highlighting fundamental limitations. To complement these results, we derive minimax lower bounds, showing that AJIVE achieves optimal rates in high-SNR regimes. Furthermore, we analyze an oracle-aided spectral estimator to demonstrate that the non-diminishing error in low-SNR scenarios is a fundamental barrier.
Extensive numerical experiments corroborate our theoretical findings, providing insights into the interplay between SNR, the number of matrices, and subspace misalignment. 
\end{abstract}

\section{Introduction}

Modern data analysis is increasingly focused on integrating information from multiple sources. This has sparked a wave of applications using diverse datasets, such as identifying communities within heterogeneous networks~\cite{macdonald2022latent}, analyzing various types of high-dimensional genomic data~\cite{lock2013joint}, and discerning global and local features in federated learning~\cite{shi2024personalized}.
A significant challenge in these analyses lies in separating the joint and individual variations present in such multi-view datasets.

In the seminal paper~\cite{lock2013joint}, Lock et al.~pioneered a matrix decomposition model known as Joint and Individual Variation Explained (JIVE). JIVE models each dataset as a noisy low-rank matrix $\bm{A}_{k} \in \mathbb{R}^{n \times d_k}$, which is further decomposed into 
\begin{align}\label{eq:intro-model}
    \bm{A}_k \approx \bm{U}^{\star}\bm{V}_{k}^{\star\top}+\bm{U}_{k}^{\star}\bm{W}_{k}^{\star\top} \in \mathbb{R}^{n \times d_{k}}, \qquad \text{for }1 \leq k \leq K.
\end{align}
Here, $\bm{U}^{\star}$---an $n \times r$ orthonormal matrix, denotes the shared subspace, 
and $\bm{U}_{k}^{\star} \perp \bm{U}^{\star}$---an $n\times r_{k}$ orthonormal matrix,  denotes the unique subspace. 
In words, JIVE assumes that multiple data matrices share a common subspace $\mathrm{col}(\bm{U}^\star)$, while each matrix also has a unique subspace modeled by $\mathrm{col}(\bm{U}_{k}^\star)$.

Despite its wide adoption and various methodological advancements, the theoretical understanding of shared subspace estimation under the JIVE model remains limited compared to single-matrix subspace estimation~\cite{cai2018rate, Cai2021Unbalanced}.  
Two key challenges hinder the theoretical progress in this area:

\begin{itemize}
    \item {\bf Impact of multiple matrices.}  Theoretical studies on shared subspace estimation often focus on cases with only two matrices (see e.g., \cite{sergazinov2024spectral, ma2024optimal}), leaving the potential advantages and limitations of having multiple matrices unexplored. 
    
    \item {\bf Impact of unique subspaces.} The identifiability of the model requires that the unique subspaces $\{\mathrm{col}(\bm{U}_{k}^\star)\}$ do not have an intersection, i.e., they are misaligned.   
    Conceptually, estimating the shared subspace becomes more challenging when the unique subspaces become more aligned. 
    Existing work~\cite{zheng2022limit, ma2024optimal} typically assumes highly misaligned subspaces, overlooking a broader spectrum of alignment scenarios.
\end{itemize}

\paragraph{Our contributions. }
In this work, we establish fundamental limits for estimating shared subspaces from multiple matrices, highlighting dependencies on the signal-to-noise ratio (SNR), the number of matrices, and the degree of subspace misalignment. Key results include:

\begin{itemize}
    \item Performance guarantees for AJIVE: We derive new statistical guarantees for AJIVE, a two-stage spectral method~\cite{Feng2018}. These results reveal that (1) in high-SNR regimes, estimation error decreases with more matrices, demonstrating the benefits of multi-matrix integration, and (2) in low-SNR regimes, AJIVE's error does not diminish even as the number of matrices grows, illustrating inherent limitations.

    \item Minimax lower bounds: We prove that AJIVE achieves optimal performance in high-SNR settings, confirming its efficiency. 
    More importantly, the optimal rate of convergence is faster when the level of misalignment is lower, confirming our intuition. 

    \item Insights into low-SNR limitations: Numerical experiments corroborate our theoretical findings, showing that AJIVE's non-diminishing error is not merely an artifact of analysis but a fundamental barrier. Furthermore, we provide lower bounds for an oracle-aided spectral estimator, establishing that non-diminishing error persists even under ideal conditions.

\end{itemize}

By addressing these challenges, our work offers a deeper understanding of shared subspace estimation and the role of multiple matrices in integrative data analysis.

\subsection{Related work}
\paragraph{JIVE. } The JIVE model is originally proposed in the paper~\cite{lock2013joint}, in which they also propose a nonconvex least-squares approach to estimate the shared and individual components. 
A similar approach is discussed in the work~\cite{zhou2015group}. 
This optimization-based approach is iterative and computationally intensive. 
As a remedy, Feng et al.~\cite{Feng2018} proposed a two-stage spectral method AJIVE, followed by its robust version RaJIVE~\cite{ponzi2021rajive}. Other spectral approaches include stacked singular value decomposition (SVD)~\cite{ma2024optimal}, and methods based on the product of projection matrices~\cite{sergazinov2024spectral}, though the latter primarily focuses on the case of $K=2$ matrices.

When the matrices $\bm{V}^\star_{k}$ and $\bm{W}^\star_{k}$ are normally distributed, the JIVE model is closely related to the personalized PCA problem~\cite{Shi2023}.  
When there is no personalization, i.e., the unique components are not present, 
this problem reduces to the distributed PCA problem studied in~\cite{Fan2019dist, zheng2022limit}.

The JIVE model has been extended in recent literature to handle more general data structures. Instead of assuming complete noisy observations, recent work~\cite{Shi2023, shi2024triple} has adapted JIVE to accommodate missing data and entrywise outliers in the observed matrices. Moreover, the classical JIVE framework assumes that the shared subspace $\mathrm{col}(\bm{U}^\star)$ is common to all data matrices. However, recent advances have introduced the concept of partially shared subspaces, where different subsets of data matrices share different subspaces. Notable examples include Data Integration via Analysis of Subspaces (DIVAS)~\cite{prothero2024data}, covariate-driven factorization by thresholding for multiblock data~\cite{gao2021covariate}, and Structural Learning and Integrative DEcomposition of Multi-view Data (SLIDE)~\cite{gaynanova2019structural}.

\paragraph{Subspace estimation from a single matrix.}
Our theoretical investigation is closely related to estimating the subspace of a single matrix. 
Wedin's theorem~\cite{wedin1973perturbation}, a classical result in matrix perturbation theory allows one to obtain perturbation bounds for both the left and right singular subspaces. 
However, the bound provided by Wedin is loose when 
we focus exclusively on column subspace estimation and the number of columns is much larger than the number of rows. 
The paper~\cite{cai2018rate} provides rate-optimal guarantees in this scenario. 
Later, Cai et al.~\cite{Cai2021Unbalanced} provides more refined analysis for subspace estimation in the face of missing data. 
We refer interested readers to the recent monograph on this topic~\cite{chen2021spectral}.

\paragraph{Notation.}

For a positive integer $n$, we denote $[n]=\{1,2,\ldots,n\}$. For
any $a,b\in\mathbb{R}$, $a\wedge b$ means the minimum of $a,b$, 
and $a\vee b$ means the maximum of $a,b$. For symmetric matrices
$\bm{A},\bm{B}\in\mathbb{R}^{n\times n}$, $\bm{A}\preceq\bm{B}$
means $\bm{B}-\bm{A}$ is positive semidefinite, i.e., $\bm{v}^{\top}(\bm{B}-\bm{A})\bm{v}\ge0$
for any $\bm{v}\in\mathbb{R}^{n}$. We use $\bm{e}_{i}$ to denote
the standard unit vector with $1$ at $i$-th coordinate and 0 elsewhere.
For any rank-$r$ matrix $\bm{A}$, we use $\sigma_{1}(\bm{A})\ge\sigma_{2}(\bm{A})\ge\ldots\ge\sigma_{r}(\bm{A})>0$
to denote its singular values. For any $n\times d$ matrix $\bm{A}$,
we use $[\bm{A}]_{i,\cdot}$ to denote its $i$-th row in the form
of a $1\times d$ matrix. We use $\mathrm{Trace}(\cdot)$ to denote
the trace of a square matrix. 
For $r \leq d$, we use $\mathcal{O}^{d \times r}$ to denote the set of orthonormal matrices $\bm{U} \in \mathbb{R}^{d \times r}$, i.e., $\bm{U}^\top  \bm{U} = \bm{I}_r$. We use the big-O notation $O(X)$ to indicate any term $Y$ such that $Y\le CX$ for some large enough constant $C$. 

\section{Background\label{sec:Problem_formulation}}
In this section, we introduce the problem of estimating the shared subspace from multiple noisy data matrices, focusing on its identifiability.  
We also provide a review of the AJIVE method, a two-stage spectral method for shared subspace estimation. 

\subsection{Observation models}

Consider $K$ ground truth matrices
$\{\bm{A}_{k}^{\star}\}_{1 \leq k \leq K}$, 
each with a low-rank decomposition
\begin{equation}
\bm{A}_{k}^{\star}=\bm{U}^{\star}\bm{V}_{k}^{\star\top}+\bm{U}_{k}^{\star}\bm{W}_{k}^{\star\top} \in \mathbb{R}^{n \times d_{k}}.\label{eq:def_A*}
\end{equation}
Here, $\bm{U}^{\star} \in \mathcal{O}^{n \times r}$ represents the shared subspace common to all matrices, 
and $\bm{U}_{k}^{\star}\in\mathcal{O}^{n\times r_{k}} \perp \bm{U}^{\star}$ represents the unique subspace specific to the $k$-th matrix. For each $k\in[K]$, $\bm{V}_k^\star \in \mathbb{R}^{d_k\times r}$ and $\bm{W}_k^\star \in \mathbb{R}^{d_k\times r_k}$ are full-rank loading matrices.
Suppose we observe noisy versions of $\{\bm{A}_{k}^{\star}\}_{1 \leq k \leq K}$:
\begin{equation}
\bm{A}_{k}=\bm{A}_{k}^{\star}+\bm{E}_{k},\label{eq:def_A}
\end{equation}
where $\bm{E}_{k}$ denotes additive noise, containing i.i.d.~Gaussian entries with mean 0 and variance $\sigma^{2}$.
The goal is then to estimate the shared subspace $\bm{U}^{\star}$ based on noisy observations $\{\bm{A}_{k}\}_{1 \leq k \leq K}$.

\subsection{Identifiability of the shared subspace}\label{sec:identity}
Before discussing estimation methods, 
we must ensure that the shared subspace $\mathrm{col}(\bm{U}^\star)$ is identifiable from the noiseless matrices $\{ \bm{A}_{k}^\star \}$. 
The identifiability condition requires: 
\begin{align}\label{eq:identifibility}
    \mathrm{col}(\bm{U}^\star) = \cap_{1 \leq k \leq K} \mathrm{col}(\bm{A}_{k}^\star).
\end{align}
Note that the orthogonality constraint $\bm{U}_{k}^{\star}  \perp \bm{U}^{\star}$ alone does not guarantee  identifiability~\eqref{eq:identifibility}. 
Additional assumptions are needed. 

\paragraph{Faithfulness: $\mathrm{col}(\bm{U}^{\star}) \subseteq  \cap_{1 \leq k \leq K} \mathrm{col}(\bm{A}_{k}^{\star})$.} 
Since $\mathrm{col}(\bm{U}^{\star})$ is the shared subspace among $\{\mathrm{col}(\bm{A}_{k}^{\star})\}$, we assume $\mathrm{col}(\bm{U}^{\star}) \subseteq \mathrm{col}(\bm{A}_{k}^{\star})$ for each $k$. 
Combined with the orthogonality constraint $\bm{U}^\star \perp \bm{U}_{k}^{\star}$, this implies $\mathrm{col}(\bm{U}_{k}^{\star}) \subseteq \mathrm{col}(\bm{A}_{k}^{\star})$. 
In fact, these assumptions are equivalent to requiring\footnote{The assumption $\mathrm{rank}(\bm{A}_{k}^\star) = r + r_k$ is adopted in the original JIVE paper~\cite{lock2013joint}, while the assumption~$\mathrm{col}(\bm{U}^{\star}) \subseteq \mathrm{col}(\bm{A}_{k}^{\star})$ is adopted in the later AJIVE paper~\cite{Feng2018}.} $\mathrm{rank}(\bm{A}_{k}^\star) = r + r_k$. 

Throughout the paper, we define $\sigma_{\min}\coloneqq\min_{k}\sigma_{r+r_{k}}(\bm{A}_{k}^{\star})$,
$\sigma_{\max}\coloneqq\max_{k}\sigma_{1}(\bm{A}_{k}^{\star})$
and the condition number to be $\kappa\coloneqq\sigma_{\max}/\sigma_{\min}$.

\paragraph{Exhaustiveness:  $\cap_{1 \leq k \leq K} \mathrm{col}(\bm{A}_{k}^\star) \subseteq \mathrm{col}(\bm{U}^\star).$} While 
faithfulness ensures $\mathrm{col}(\bm{U}^{\star}) \subseteq \cap_{1 \leq k \leq K } \mathrm{col}(\bm{A}_{k}^{\star})$,  
identifiability further requires that  
$\mathrm{col}(\bm{U}^\star)$ captures all shared information in $\cap_{1 \leq k \leq K} \mathrm{col}(\bm{A}_{k}^\star)$. 
A degenerate case arises if $\cap_{1 \leq k \leq K } \mathrm{col}(\bm{U}_{k}^{\star}) \neq \emptyset$,  
meaning unique subspaces share a common direction, making $\mathrm{col}(\bm{U}^{\star})$ a strict subset of $\cap_{1 \leq k \leq K } \mathrm{col}(\bm{A}_{k}^{\star})$, and thus not identifiable. 
To avoid this, one needs to assume that 
$\cap_{1 \leq k \leq K } \mathrm{col}(\bm{U}_{k}^{\star}) = \emptyset$, i.e., 
the unique subspaces are \emph{misaligned}. 

In this paper, we quantify the degree of misalignment among the unique subspaces via the following definition~\cite{shi2024personalized, Shi2023}. 

\begin{definition}{(misalignment)} We say that the collection of subspaces $\{\bm{U}_{k}^{\star}\}$ is $\theta$-misaligned if  
\[
\left\Vert \frac{1}{K}\sum_{k=1}^{K}\bm{U}_{k}^{\star}\bm{U}_{k}^{\star\top}\right\Vert \leq 1-\theta.
\]
\end{definition}

\noindent It is easy to see that $\theta \in [0, 1-1/K]$. We single out several interesting scenarios. 
\begin{itemize}
    \item \emph{Aligned}. One of the extreme cases is when $\theta=0$. 
    This is equivalent to $\cap_{1 \leq k \leq K } \mathrm{col}(\bm{U}_{k}^{\star}) \neq \emptyset$, i.e., the unique subspaces are aligned. In this case, the shared subspace is not identifiable. 
    \item \emph{Misaligned}. The other extreme case is when all the unique subspaces are orthogonal to each other, implying that $\left\Vert \frac{1}{K}\sum_{k=1}^{K}\bm{U}_{k}^{\star}\bm{U}_{k}^{\star\top}\right\Vert = 1/K$, and that $\theta = 1 - 1/K$. 
    This is also the regime considered in the recent work~\cite{ma2024optimal}. 
    \item \emph{Realizability via randomization.} Fix any $\theta \in (0, 1- 1/K]$. One can generate a collection of $\theta$-misaligned subspaces $\{\bm{U}_{k}^{\star}\}$ in a random fashion. To see this, for each $k \in [K]$, generate $\bm{U}_{k}^{\star}$ via  
    \begin{align}
        \bm{U}_{k}^\star = \sqrt{1- \theta} \bm{Z} + \sqrt{\theta} \bm{Z}_{k}
    \end{align}
    for some fixed $\bm{Z} \in \mathcal{O}^{d \times r_k}$, and $\bm{Z_k} \in \mathcal{O}^{d \times r_k}$ drawn uniformly at random from the orthogonal subspace to $\bm{Z}$. 
    Rough calculations show that 
    \begin{align*}
\left\Vert \frac{1}{K}\sum_{k=1}^{K}\bm{U}_{k}^{\star}\bm{U}_{k}^{\star\top}\right\Vert \approx \left\Vert \mathbb{E} \frac{1}{K} \sum_{k=1}^{K}\bm{U}_{k}^{\star}\bm{U}_{k}^{\star\top}\right\Vert =1-\theta.
\end{align*}
\end{itemize}

\subsection{Angle-based joint and individual variation explained (AJIVE)}
Now we are ready to review the AJIVE method put forward in the paper~\cite{Feng2018} 
for extracting the shared and unique subspaces from noisy observations $\{\bm{A}_{k}\}$. 
AJIVE is essentially a two-stage spectral method.  
In the first
stage, for each $k\in[K]$, we estimate the ($r+r_k$)-dimensional column space
of $\bm{A}_{k}^{\star}$ using SVD of the noisy matrix $\bm{A}_{k}$. 
Then in the second stage,
we combine the estimates in the first stage and use SVD again to estimate 
the most prominent (i.e., shared) $r$-dimensional subspace $\mathrm{col}(\bm{U}^{\star})$. 
See Algorithm~\ref{alg:spectral} for a detailed description of AJIVE.

\begin{algorithm}[h]
\caption{Angle-based joint and individual variation explained (AJIVE)\label{alg:spectral}}

Input: $\{\bm{A}_{k}\}_{k=1}^{K},r,\{r_{k}\}_{k=1}^{K}$.
\begin{enumerate}
\item For $k=1,\ldots,K$, Let $\widetilde{\bm{U}}_{k}$ be the top-$(r+r_{k})$
left singular matrix of $\bm{A}_{k}$.
\item Let $\widehat{\bm{U}}$ be the matrix whose columns are the top-$r$
eigenvectors of $\sum_{k=1}^{K}\widetilde{\bm{U}}_{k}\widetilde{\bm{U}}_{k}^{\top}$.
Output $\mathrm{col}(\widehat{\bm{U}})$ as the estimate of $\mathrm{col}(\bm{U}^{\star})$.
\item (Optional) For $k=1,\ldots,K$, let $\widehat{\bm{V}}_{k}=\bm{A}_{k}^{\top}\widehat{\bm{U}}$,
$\widehat{\bm{U}}_{k}$ be the top-$r_{k}$ left singular matrix of
$\bm{I}_{n}-\widehat{\bm{U}}\widehat{\bm{V}}_{k}^{\top}$, and $\widehat{\bm{W}}_{k}=\bm{A}_{k}^{\top}\widehat{\bm{U}}_{k}$.
Output $\widehat{\bm{V}}_{k},\widehat{\bm{U}}_{k},\widehat{\bm{W}}_{k}$
as the estimates of $\bm{V}_{k}^{\star},\bm{U}_{k}^{\star},\bm{W}_{k}^{\star}$,
and
\[
\widehat{\bm{A}}_{k}\coloneqq\widehat{\bm{U}}\widehat{\bm{V}}_{k}^{\top}+\widehat{\bm{U}}_{k}\widehat{\bm{W}}_{k}^{\top}
\]
as the estimate of $\bm{A}_{k}^{\star}$.
\end{enumerate}
\end{algorithm}

AJIVE is one-shot, and hence computationally cheaper than the optimization-based method~\cite{lock2013joint}. In addition, it can be naturally distributed and therefore can be made private. 

\paragraph{A digression: failure of stacked SVD.}
Another method to estimate the shared column
space is to use the top eigenvectors of $\sum_{k=1}^{K}\bm{A}_{k}\bm{A}_{k}^{\top}$.
This is equivalent to taking the left singular vectors of the stacked
matrix $\begin{bmatrix}\bm{A}_{1} \cdots  \bm{A}_{K}\end{bmatrix} \in \mathbb{R}^{n \times \sum_{k=1}^{K}d_k}$.

Stacked SVD is extensively studied in the concurrent work~\cite{ma2024optimal}. 

When there are no unique components, i.e., all the matrices share the same subspace, 
stacked SVD is shown to be an optimal estimator for the shared subspace. 
However, when unique subspaces are present, 
stacked SVD cannot even recover the true subspace in the noiseless case. 
Consider the following simple example with $K=2$, $n=3$, and $r = r_{1}=r_{2}=1$. 
Let $\epsilon>0$ be a scalar. 
Set
\begin{align*}
\bm{A}_{1} & =\begin{bmatrix}1 &
0 &
0
\end{bmatrix}^\top\begin{bmatrix}1 & 0 & 0\end{bmatrix}+\epsilon\begin{bmatrix}0 &
1 &
-1
\end{bmatrix}^\top\begin{bmatrix}1 & 1 & 1\end{bmatrix};\\
\bm{A}_{2} & =\begin{bmatrix}1 &
0 &
0
\end{bmatrix}^\top\begin{bmatrix}1 & 0 & 0\end{bmatrix}+\epsilon\begin{bmatrix}0 &
1 &
1
\end{bmatrix}^\top\begin{bmatrix}1 & 1 & 1\end{bmatrix}.
\end{align*}
It is easy to check that is satisfies the identifiability assumptions
in Section~\ref{sec:identity}. However, the matrix
\[
\frac{1}{2}\left(\bm{A}_{1}\bm{A}_{1}^{\top}+\bm{A}_{2}\bm{A}_{2}^{\top}\right)=\begin{bmatrix}1 & \epsilon & 0\\
\epsilon & 3\epsilon^{2} & 0\\
0 & 0 & 3\epsilon^{2}
\end{bmatrix}
\]
does not have $\begin{bmatrix}1 & 0 & 0\end{bmatrix}^{\top}$
as its eigenvector. 
In the general case, we compute 
\[
\sum_{k=1}^{K}\bm{A}_{k}^\star\bm{A}_{k}^{\star\top}=\sum_{k=1}^{K}\left(\bm{U}^{\star}\bm{V}_{k}^{\star\top}\bm{V}_{k}^{\star}\bm{U}^{\star\top}+\bm{U}^{\star}\bm{V}_{k}^{\star\top}\bm{W}_{k}^{\star}\bm{U}_{k}^{\star\top}+\bm{U}_{k}^{\star}\bm{W}_{k}^{\star\top}\bm{V}_{k}^{\star}\bm{U}^{\star\top}+\bm{U}_{k}^{\star}\bm{W}_{k}^{\star\top}\bm{W}_{k}^{\star}\bm{U}_{k}^{\star\top}\right).
\]
The cross terms $\bm{U}^{\star}\bm{V}_{k}^{\star\top}\bm{W}_{k}^{\star}\bm{U}_{k}^{\star\top}$
and $\bm{U}_{k}^{\star}\bm{W}_{k}^{\star\top}\bm{V}_{k}^{\star}\bm{U}^{\star\top}$
can introduce bias when $\bm{V}_{k}^{\star\top}\bm{W}_{k}^{\star}\neq \bm{0}$, and hence 
stacked SVD estimates the wrong direction. 

\begin{remark}
    Ma et al.~\cite{ma2024optimal} require two extra assumptions to make stacked SVD a correct method in the noiseless case: (1) they require the singular values of $\bm{A}_1^\star$ and $\bm{A}_2^\star$ are distinct, and (2) they need to know the column indices of $\bm{U}^\star$ in the singular vectors of the stacked matrix $[\bm{A}_1^\star, \bm{A}_2^\star]$. In this paper, we do not make such assumptions. 
\end{remark}

\section{Performance guarantees of AJIVE\label{sec:Estimation}}
In this section, we present the performance guarantees of the AJIVE algorithm; See Section~\ref{sec:Proof_main} for the proof of Theorem~\ref{thm:main}. 
From now on, we set $d \coloneqq \max_{k} d_k$, and $N \coloneqq \max\{n, d\}$. 

\begin{theorem}\label{thm:main}Assume $n\ge C_{1}\log N$ for some
sufficiently large constant $C_{1}>0$. 
Further assume the following conditions 
\begin{subequations}\label{eq:signal_size}
    \begin{align}
        \frac{\kappa\sigma\sqrt{n}}{\sigma_{\min}}+\frac{\sigma^{2}\sqrt{nd}}{\sigma_{\min}^{2}}\le c_{1}\sqrt{\theta} \\
        \left(\frac{\sigma\sqrt{n}}{\sigma_{\min}}+\frac{\sigma^{2}\sqrt{nd}}{\sigma_{\min}^{2}}\right)\left(\sqrt{\frac{\log N}{K}}+\frac{\log N}{K}\right) & \le c_{2}\theta
    \end{align}
\end{subequations}
hold for some small enough constants $c_1, c_2 > 0$. 
Then with probability at
least $1-O(KN^{-10})$, the AJIVE estimate $\widehat{\bm{U}}$ output by  Algorithm~\ref{alg:spectral}
satisfies 
\begin{align*}
\left\Vert \widehat{\bm{U}}\widehat{\bm{U}}^{\top}-\bm{U}^{\star}\bm{U}^{\star\top}\right\Vert  & \le C_{2}\log^{5/2}N\left[\frac{\sigma}{\sigma_{\min}}\left(\sqrt{\frac{n}{K}+\frac{r+r_{\mathrm{avg}}}{K\theta}+\frac{r\cdot r_{\mathrm{avg}}}{K\theta}\wedge\frac{r}{K^{2}\theta^{2}}}\right)\right.\\
 & \qquad\qquad\qquad\quad\left.+ \frac{\sigma^{2}}{\sigma_{\min}^{2}}\cdot\frac{\kappa^{2}}{\theta(1\wedge K\theta)}\left(\sqrt{nd}+n\right)\right]
\end{align*}
for some constant $C_{2}>0$. Here  $r_{\mathrm{avg}} \coloneqq \frac{1}{K}\sum_{k=1}^{K}r_{k}$. 
\end{theorem} 

An immediate implication of Theorem~\ref{thm:main} is that AJIVE achieves exact recovery of the shared subspace when there is no observation noise, i.e., when $\sigma=0$. 
As a by-product, this also demonstrates the identifiability of the shared subspace under the assumed conditions in Section~\ref{sec:identity}. 

Now we turn to the performance of AJIVE in the noisy case. 
To simplify the discussion, we focus on the well-conditioned case when $\kappa \asymp 1$, 
$r \asymp r_\mathrm{avg} \asymp 1$, and $d \asymp n$, and also ignore the log factors.  
Under this circumstance, Theorem~\ref{thm:main} asserts that as long as the noise obeys
\begin{align}\label{eq:noise-cond-simple}
    \frac{\sigma\sqrt{n}}{\sigma_{\min}} \ll \min \{ \sqrt{\theta}, \sqrt{K} \theta \}, 
\end{align}
the AJIVE estimate $\hat{\bm{U}}$ satisfies 
\begin{align}\label{eq:upper-bound-simple}
\left\Vert \widehat{\bm{U}}\widehat{\bm{U}}^{\top}-\bm{U}^{\star}\bm{U}^{\star\top}\right\Vert \lesssim
\underbrace{\frac{\sigma}{\sigma_{\min}}\sqrt{\frac{n}{K}+\frac{r}{K\theta}}}_{\eqqcolon \mathcal{E}_1} \quad + \quad \underbrace{\frac{1}{\theta (1\wedge K\theta)}\cdot \frac{\sigma^{2}n}{\sigma_{\min}^{2}}}_{\eqqcolon \mathcal{E}_2}.
\end{align}
Our upper bound consists of two terms---depending on the scaling w.r.t.~the noise $\sigma$: (1) the first-order term $\mathcal{E}_1$, and (2) 
the second-order term $\mathcal{E}_2$. 

\paragraph{First-order optimality when SNR is high.} When the signal-to-noise ratio (SNR) 
$\sigma_{\min} / (\sigma \sqrt{n})$ is high, the first-order term $\mathcal{E}_1$ dominates the upper bound~\eqref{eq:upper-bound-simple}. 
Two terms appear in $\mathcal{E}_1$. 
The first component $(\sigma/\sigma_{\min})\sqrt{n/K}$
is the expected boost of performance by the factor of $1/\sqrt{K}$
over subspace estimation based on a single data matrix. 
The second component $(\sigma/\sigma_{\min})\sqrt{r/K\theta}$
highlights the challenge of distinguishing the shared and unique subspaces. 
The factor $1/\theta$ arises from the eigen-gap
between $\bm{U}^{\star}\bm{U}^{\star\top}$ and $(1/K)\sum_{k=1}^{K}\bm{U}_{k}^{\star}\bm{U}_{k}^{\star\top}$, which scales with $\theta$. Fortunately, this dependency also shrinks at a rate of $1/\sqrt{K}$, and relates to the smaller intrinsic dimension $r$ rather than the ambient dimension $n$. 
Overall, the first-order term demonstrates the power of integrating multiple matrices, as 
the estimation error decays at a rate of $1/\sqrt{K}$. 
Later in Section~\ref{sec:Minimax},
we will show that this rate is minimax-optimal 
when the SNR is high. 

\paragraph{Non-diminishing second-order error.}
When the SNR $\sigma_{\min} / (\sigma \sqrt{n})$ is low, the second-order 
term $\mathcal{E}_2$ dominates. 
Notably, $\mathcal{E}_2$ does
not vanish when the number $K$ of matrices increases, as $\mathcal{E}_2$ will converge to
$\frac{1}{\theta} \frac{\sigma^{2}n}{\sigma_{\min}^{2}}$. 
At first sight, this non-diminishing second-order error is puzzling, as it shows 
the limitation of multiple matrices: 
the benefit of more data matrices in estimating the shared subspace will disappear when 
the number $K$ of matrices goes beyond a certain threshold. 

It turns out that this non-diminishing term is not an analytical artifact about the spectral method. 
In Section~\ref{sec:experiment}, we present numerical examples to demonstrate that the estimation error of AJIVE indeed does not vanish as $K\rightarrow\infty$. 
Analytically, 
this non-diminishing effect is fundamentally tied to the fact that SVD on each matrix $\bm{A}_{k}$ produces a biased estimator of the true singular subspace of $\bm{A}_{k}^{\star}$. 
Consequently, when we average the subspace estimates in the second stage in AJIVE, the bias 
persists, and hence the estimation error does not converge to 0. 

It is natural to wonder if this non-diminishing error is the fundamental limit of this problem, and whether other estimators can improve over the spectral approach AJIVE. 
Although we are unable to deliver an information-theoretic limit against this non-diminishing error, in Section~\ref{sec:oracle}, we provide a performance lower bound of an oracle-aided spectral estimator that leverages extra information about the underlying statistical model. 
It is evident from the algorithm-specific lower bound that even for this oracle estimator, the estimation error does not vanish as $K$ increases in the low SNR regime.

\paragraph*{Tightness of SNR assumption~\eqref{eq:noise-cond-simple} when $K =O(1)$. }
Last but not least, we focus on the case where the number $K$ of matrices is a constant. 
In this case, Theorem~\ref{thm:main} (more specifically Equations~\eqref{eq:noise-cond-simple} and~\eqref{eq:upper-bound-simple}) asserts that the spectral method achieves consistent estimation when 
\begin{align}\label{eq:noise-cond-const-K}
    \frac{\sigma\sqrt{n}}{\sigma_{\min}} \ll  \theta. 
\end{align}
This showcases an interesting interplay between the noise level $\sigma$ and the level $\theta$ of misalignment. 
In fact, such an interplay is tight in the sense that if $\theta \ll \frac{\sigma\sqrt{n}}{\sigma_{\min}}$, no estimator can detect if the shared subspace exists or not. 

To formalize this, let $\bm{u}$, $\bm{w}$, and $\widetilde{\bm{w}}$ be unit vectors in $\mathbb{R}^n$ orthogonal to each other. Consider the following two hypotheses when $K=2$. 
\begin{align*}
    H_{0}:&\begin{cases}
\bm{A}_{1}= \bm{u} \bm{v}_1^{\top}+\bm{E}_{1}\\
\bm{A}_{2}=(\cos\alpha \cdot \bm{u})\cos\alpha\cdot \bm{u}^{\top}+\bm{E}_{2};
\end{cases} \\
H_{1}:&\begin{cases}
\bm{A}_{1}= \bm{u} \bm{v}_1^{\top}+\bm{E}_{1}\\
\bm{A}_{2}=(\cos\alpha\cdot \bm{u}+\sin\alpha\cdot \bm{w})(\cos\alpha\cdot \bm{u}+\sin\alpha\cdot\widetilde{\bm{w}})^{\top}+\bm{E}_{2}, 
\end{cases}
\end{align*}
where $\bm{v}_1\in \mathbb{R}^n$ is an arbitrary unit vector and $\alpha$ is chosen such that $\cos\alpha = 1-2\theta$. 
Under the null hypothesis $H_0$, the two matrices share the same subspace spanned by $\bm{u}$. 
In comparison, under the alternative hypothesis $H_1$, 
the two matrices have two unique components $\bm{u}$, and $\cos\alpha\cdot \bm{u}+\sin\alpha\cdot \bm{w}$. 
In the latter case, the choice of $\alpha$ guarantees that the two unique subspaces are $\theta$-misaligned. 

Suppose that one even knows $\bm{u}$ and $\bm{v}_1$. 
Then the hypothesis testing problem boils down to a simpler one:
\begin{align*}
    H_0: \;&(\bm{I - \bm{u} \bm{u}^\top}) \bm{A}_2 (\bm{I - \bm{u} \bm{u}^\top})  = (\bm{I - \bm{u} \bm{u}^\top}) \bm{E}_2 (\bm{I - \bm{u} \bm{u}^\top}); \\
    H_1: \;&(\bm{I - \bm{u} \bm{u}^\top}) \bm{A}_2 (\bm{I - \bm{u} \bm{u}^\top})  = (\sin\alpha)^2 \bm{w} \widetilde{\bm{w}}^\top + (\bm{I - \bm{u} \bm{u}^\top}) \bm{E}_2 (\bm{I - \bm{u} \bm{u}^\top}).
\end{align*}
This is exactly the detection problem in the high-dimensional spiked
rectangular model~\cite{el2018detection}. 
Leveraging the results therein, we can show that detection is impossible if
\begin{align*}
    \sigma \sqrt{n} > (\sin\alpha)^2.
\end{align*}
This concludes our argument as $\sigma_{\min} = (\cos\alpha)^2 \asymp 1$, and $(\sin\alpha)^2 \asymp \theta$ when $\theta$ is sufficiently small.

\section{Minimax lower bounds\label{sec:Minimax}}

In this section, we develop information-theoretic lower bounds for estimating the shared subspace $\bm{U}^\star$ from noisy matrices $\{\bm{A}_{k}\}$, with the proof deferred to Section~\ref{sec:Proof_minimax}.  

We start with formalizing the parameter space. 
Consider $\bm{U}^{\star}\in\mathbb{R}^{n\times r}$,
$\bm{U}_{k}^{\star}\in\mathbb{R}^{n\times r_{k}}$, $\bm{V}_{k}^{\star}\in\mathbb{R}^{d_{k}\times r}$,
$\bm{W}_{k}^{\star}\in\mathbb{R}^{d_{k}\times r_{k}}$ for $k\in[K]$
and $\bm{A}_{k}^{\star}=\bm{U}^{\star}\bm{V}_{k}^{\star\top}+\bm{U}_{k}^{\star}\bm{W}_{k}^{\star\top}$.
Fix some $\sigma_{\min}>0$ and $\theta\in(0,1)$. We assume the following
conditions for all $k\in[K]$:
\begin{subequations}\label{eq:minimax_req}
\begin{align}
\text{Orthogonality: } & \bm{U}^{\star\top}\bm{U}^{\star}=\bm{I}_{r},\qquad\bm{U}_{k}^{\star\top}\bm{U}_{k}^{\star}=\bm{I}_{r_{k}},\qquad\bm{U}^{\star\top}\bm{U}_{k}^{\star}=\bm{0}_{r\times r_{k}};\label{eq:minimax_ortho}\\
\text{Misalignment: } & \left\Vert \frac{1}{K}\sum_{k=1}^{K}\bm{U}_{k}^{\star}\bm{U}_{k}^{\star\top}\right\Vert \le1-\theta;\label{eq:minimax_misalign}\\
\text{Signal strength: } & \sigma_{r+r_{k}}(\bm{A}_{k}^{\star})\ge\sigma_{\min}.\label{eq:minimax_signal}
\end{align}
\end{subequations}
With these definitions in place, we define the parameter space to be 
\[
\Theta\coloneqq\left\{ \left(\bm{U}^{\star},\{\bm{U}_{k}^{\star}\}_{k=1}^{K},\{\bm{V}_{k}^{\star}\}_{k=1}^{K},\{\bm{W}_{k}^{\star}\}_{k=1}^{K}\right): \text{ conditions }\eqref{eq:minimax_req}\text{ hold}\right\} .
\]
We then have the following minimax lower bound\footnote{In Section~\ref{sec:Proof_minimax} in the appendix, we present a stronger version of this lower bound that is more refined when the number of columns $d$ is much larger than the number $n$ of rows. } when $d_{1}=\cdots=d_{K} =d$ and $r=r_{1}=\cdots=r_{K}$.

\begin{theorem}\label{thm:minimax} Suppose that $\theta\le1/2$, $r\ge8$, and $n \geq 6r$. 
Suppose that $d\ge C_1 \log(K/r) + C_2$ for some large enough constant $C_1, C_2 > 0$. Then we have  
\begin{equation}
\inf_{\widehat{\bm{U}}}\sup_{\{\bm{A}_{k}^{\star}\}\in\Theta}\mathbb{E}\left\Vert \bm{U}^{\star}\bm{U}^{\star\top}-\widehat{\bm{U}}\widehat{\bm{U}}^{\top}\right\Vert \ge \frac{1}{20}\sqrt{\frac{n}{K}+\frac{r}{K\theta}}\cdot\frac{\sigma}{\sigma_{\min}}.\label{eq:minimax_lb}
\end{equation} 
\end{theorem}

\begin{remark}
The assumption $\theta \leq 1/2$ is made without loss of generality. 
Recall that the range of $\theta$ is $(0, 1-1/K]$. Therefore $\theta \leq 1/2$ when $K\geq 2$. 
\end{remark}

If we compare the lower bound~\eqref{eq:minimax_lb} with the simplified upper bound~\eqref{eq:upper-bound-simple} when $\kappa \asymp 1$, $r \asymp r_\mathrm{avg} \asymp 1$, and $n \asymp d$, we see that the lower bound matches the first-order term $\mathcal{E}_1$ in~\eqref{eq:upper-bound-simple}, and hence the lower bound is tight when the SNR $\sigma_{\min} / (\sigma \sqrt{n})$ is high. 
The optimal rate of convergence $\sqrt{\frac{n}{K}+\frac{r}{K\theta}}\cdot\frac{\sigma}{\sigma_{\min}}$ reveals two interesting regimes with different dimensional dependency. 

\begin{itemize}
    \item Large $\theta$: when $\theta$ is large, i.e., when the unique components are not aligned, the estimation error scales with $\frac{\sigma}{\sigma_{\min}} \sqrt{n/K}$. 
    This, in fact, corresponds to the optimal rate when all the matrices share the same column subspace, that is, when the unique components are not present. 
    \item Small $\theta$: when $\theta$ is small, i.e., when the unique components are somewhat aligned, the estimation error scales with $\frac{\sigma}{\sigma_{\min}}\sqrt{\frac{r}{K\theta}}$. In this case, the presence of unique components interferes with the estimation of the shared component, and renders the estimation of the shared one more challenging.   
\end{itemize}

We use Fano's method to establish both terms in the lower bound. 
Admittedly, the first term is relatively easier to establish: 
we simply extend the arguments in~\cite{Cai2013} to accommodate multiple matrices. 
However, establishing the second term with the optimal dependence on $r$ and $\theta$ is considerably more challenging, that relies on an explicit construction of a large packing set of the parameter space $\Theta$. 
Such a construction might be of independent interest when studying learning from multiple matrices. 

\section{Performance lower bound of an oracle-aided spectral estimator\label{sec:oracle}}

The minimax lower bound in Section~\ref{sec:Minimax} matches our error
guarantee in Theorem~\ref{thm:main} in the high SNR regime.
However, it fails to explain the non-diminishing error in the low SNR regime when the number $K$ of matrices approaches infinity. 
In this section, we describe an oracle-aided spectral estimator that leverages extra information about the underlying statistical model, and provide performance lower bound of this estimator. 
It turns out that even for this oracle estimator, its estimation error will not drop when $K$ increases. 
To some extent, this argument provides evidence of the non-diminishing error as the fundamental barrier of this problem. 

\subsection{Oracle spectral estimator}
Suppose that one is given the information about unique components $\bm{U}_{k}^{\star}\bm{W}_{k}^{\star\top}$ for each $k \in [K]$, then the optimal estimator 
is given by the top-$r$ eigenspace of 
\begin{align*}
\frac{1}{K}\sum_{k=1}^{K}\left(\bm{A}_{k}-\bm{U}_{k}^{\star}\bm{W}_{k}^{\star\top}\right)\left(\bm{A}_{k}-\bm{U}_{k}^{\star}\bm{W}_{k}^{\star\top}\right)^{\top}.
\end{align*}
However, perfect knowledge about the unique components is not possible in reality. 
It makes sense to replace $\bm{U}_{k}^{\star}\bm{W}_{k}^{\star\top}$ with a good estimate of it. 
The oracle spectral estimator is precisely doing so: it uses an oracle-aided estimate for $\bm{U}_{k}^{\star}\bm{W}_{k}^{\star\top}$: 
\begin{align*}
\text{top-}r_k\text{ SVD of }\quad \mathcal{P}_{\star}^{\perp}\bm{A}_{k}=\bm{U}_{k}^{\star}\bm{W}_{k}^{\star\top}+\mathcal{P}_{\star}^{\perp}\bm{E}_{k},
\end{align*}
where $\mathcal{P}_{\star}^{\perp}\coloneqq\bm{I}-\bm{U}^{\star}\bm{U}^{\star\top}$. 
See Algorithm~\ref{alg:oracle} for detailed implementations of the oracle spectral estimator. 
In words, the oracle spectral estimator is given a good estimate of the unique component, where the goodness arises from using the oracle knowledge $\bm{U}^{\star}$. 

\begin{algorithm}[h]
\caption{Oracle spectral method for shared singular subspace estimation\label{alg:oracle}}

Input: $\{\bm{A}_{k}\}_{k=1}^{K},r,\{r_{k}\}_{k=1}^{K}$, $\bm{U}^\star$.
\begin{enumerate}
\item Let $\widehat{\bm{U}}_{k}\widehat{\bm{W}}_{k}^{\top}$ be the top-$r_{k}$
SVD of $\mathcal{P}_{\star}^{\perp}\bm{A}_{k}=\bm{U}_{k}^{\star}\bm{W}_{k}^{\star\top}+\mathcal{P}_{\star}^{\perp}\bm{E}_{k}$.
\item Let $\widehat{\bm{U}}$ be the matrix whose columns are the top-$r$
eigenvectors of 
\begin{equation}
\bm{M}\coloneqq\frac{1}{K}\sum_{k=1}^{K}\left(\bm{A}_{k}-\widehat{\bm{U}}_{k}\widehat{\bm{W}}_{k}^{\top}\right)\left(\bm{A}_{k}-\widehat{\bm{U}}_{k}\widehat{\bm{W}}_{k}^{\top}\right)^{\top}.\label{eq:def_M}
\end{equation}
\end{enumerate}
\end{algorithm}

\subsection{Performance lower bounds}
Theorem~\ref{thm:oracle_lb} delivers a performance lower bound of the oracle spectral estimator when $r=r_{1}=\cdots=r_{k}$ and $n = d_{1}=d_{2}=\cdots=d_{k}\eqqcolon d$.  
We defer its proof to Section~\ref{sec:oracle_proof}.

\begin{theorem}\label{thm:oracle_lb} 
Consider $\theta \leq 1/2$. Suppose $n$ is large enough, $r \leq n/3$, and $\sigma \sqrt{n}/\sigma_{\min}\le C_{1}$
for some small enough constant $C_{1}$. 
There exists a configuration of $\bm{U}^{\star}$, $\{\bm{U}_{k}^{\star}\}_{k=1}^{K}$,
$\{\bm{V}_{k}^{\star}\}_{k=1}^{K}$, $\{\bm{W}_{k}^{\star}\}_{k=1}^{K}$
such that with probability at least $1-O(KN^{-10})$, the oracle estimator~$\widehat{\bm{U}}$
output by Algorithm~\ref{alg:oracle} satisfies
\begin{equation}
\left\Vert \widehat{\bm{U}}\widehat{\bm{U}}^{\top}-\bm{U}^{\star}\bm{U}^{\star\top}\right\Vert \ge C_{2}\frac{\sigma^{4}n^2}{\sigma_{\min}^{4}}-C_{3} \frac{\log n}{\sqrt{K}}\cdot\frac{\sigma\sqrt{n}}{\sigma_{\min}}  \label{eq:oracle_lb}
\end{equation}
for some constants $C_{2}, C_3>0$, with the proviso that $K\ge C_{4}\log^{2} n$ for some large constant $C_4 > 0$. 
\end{theorem} 
For a fixed noise level $\sigma$, 
as $K\rightarrow\infty$, this lower bound is dominated by the
first term that is positive and invariant to $K$. It shows that, even with oracle information, the spectral
estimator can yield a non-diminishing error for estimating the shared subspace.

\paragraph{Why oracle estimator fails? }
Here, we present a brief overview of the proof of the performance lower bound, hoping to convey the intuitions regarding its inconsistency when $K \rightarrow\infty$. 

The key in the analysis is the series expansion of SVD, developed in the work~\cite{Xia2021} that allows us to obtain a tight degree-4 polynomial approximation $\bm{Q}$ of the oracle matrix $\bm{M}$. 
To see how this unfolds, 
we first consider a fourth-order approximation of $\bm{A}_{k}-\widehat{\bm{U}}_{k}\widehat{\bm{W}}_{k}^{\top}$ 
\[
\bm{A}_{k}-\widehat{\bm{U}}_{k}\widehat{\bm{W}}_{k}^{\top} \approx \bm{T}_{0,k}+\bm{T}_{1,k}+\bm{T}_{2,k}+\bm{T}_{3,k}+\bm{T}_{4,k}, 
\]
where each $\bm{T}_{i,k}$ is an $i$-th degree polynomial of the noise matrix $\bm{E}_{k}$. 
The approximate error decreases as the noise $\|\bm{E}_k\|$ becomes smaller. 
As a result, we can approximate the oracle spectral estimator as 
\begin{align*}
    \frac{1}{K} \sum_{k=1}^{K} \left(\bm{A}_{k}-\widehat{\bm{U}}_{k}\widehat{\bm{W}}_{k}^{\top}\right)\left(\bm{A}_{k}-\widehat{\bm{U}}_{k}\widehat{\bm{W}}_{k}^{\top}\right)^{\top} \approx \underbrace{\frac{1}{K} \sum_{k=1}^{K} \sum_{i=0}^{4}\sum_{j=0}^{4-i}\bm{T}_{i,k}\bm{T}_{j,k}^{\top}}_{\eqqcolon \bm{Q}} 
    \approx \mathbb{E} [\bm{Q}], 
\end{align*}
where the last approximation arises from concentration of measure, and the approximation is tighter as $K$ increases. 
It now boils down to studying the eigenstructure of $\mathbb{E}[\bm{Q}]$, which takes the form 
\begin{align*}
    \mathbb{E}[\bm{Q}] & =(1+\alpha_{1})\bm{U}^{\star}\bm{U}^{\star\top}+\alpha_{2}\bm{I}_{n}+\alpha_{3}\cdot\frac{1}{K}\sum_{k=1}^{K}\bm{U}_{k}^{\star}\bm{U}_{k}^{\star\top} +\alpha_{4}\cdot\frac{1}{K}\sum_{k=1}^{K}\left(\bm{U}^{\star}\bm{V}_k^{\star\top}\bm{W}_k^\star\bm{U}_{k}^{\star\top}+\bm{U}_{k}^{\star}\bm{W}_k^{\star\top}\bm{V}_k^\star\bm{U}^{\star\top}\right).
\end{align*}
Here, $\alpha_i \asymp (\sigma \sqrt{n} / \sigma_{\min})^{i}$.  
We can choose $\bm{V}_k^{\star}$ and $\bm{W}_k^{\star}$ so that $\bm{V}_k^{\star\top}\bm{W}_k=\bm{I}$ and the cross term $\{\bm{U}^{\star}\bm{U}_{k}^{\star\top}+\bm{U}_{k}^{\star}\bm{U}^{\star\top}\}$ is not negligible. Due to the presence of this cross term, one readily sees that the leading eigenspace of $\mathbb{E}[\bm{Q}]$ is not $\bm{U}^\star \bm{U}^{\star\top}$. 
This together with proper control on the approximation accuracy allows one to conclude the inconsistency of the oracle spectral estimator when $K \rightarrow \infty$.

\subsection{Connections to nonconvex estimator and Neyman and Scott's problem}
While we introduce the oracle estimator from the perspective of the spectral method, 
it in fact bears intimate connections with the nonconvex least-squares approach that solves the following optimization problem 
\begin{align*}
\min_{\bm{U},\bm{U}_{k},\bm{V}_{k},\bm{W}_{k}} & \qquad\sum_{k=1}^{K}\|\bm{U}\bm{V}_{k}^{\top}+\bm{U}_{k}\bm{W}_{k}^{\top}-\bm{A}_{k}\|_{\mathrm{F}}^{2}\\
\text{subject to} & \qquad\bm{U}^{\top}\bm{U}=\bm{I}_{r},\\
 & \qquad\bm{U}_{k}^{\top}\bm{U}_{k}=\bm{I}_{r_{k}},\quad\bm{U}^{\top}\bm{U}_{k}=\bm{0}_{r\times r_{k}}, \qquad \text{for each } k \in [K]. 
\end{align*}
A natural way to solve the nonconvex program is alternating minimization where one alternates between 
\begin{itemize}
    \item Fixing the shared subspace $\bm{U}$, find the unique components $\bm{U}_{k} \bm{W}_{k}^\top$;
    \item Fixing the unique components $\{\bm{U}_{k} \bm{W}_{k}^\top\}$, find the shared component $\bm{U}$. 
\end{itemize}
With some calculations, it is straightforward to see that the oracle spectral estimator we put forward early corresponds exactly to one step of alternating minimization starting from the ground truth $\bm{U}^\star$. 

Since the nonconvex estimator can be viewed as the MLE under Gaussian noise, the inconsistency of MLE is intimately related to the Neyman and Scott's problem~\cite{neyman1948consistent}, also known as the incidental parameter problem~\cite{lancaster2000incidental}.
In the incidental parameter problem, the law of a sequence of independent observations is governed by two sets of parameters: the first set---called structural parameters, appears in the law of every observation, while the second---called incidental parameters, only appears in each individual observation. Neyman and Scott showed that MLE can be inconsistent for estimating the structural parameter in this case. 
In the language of Neyman and Scott, the shared subspace $\bm{U}^\star$ is the structural parameter, while the unique subspaces $\{\bm{U}^\star_{k}\}$ are the incidental parameters. What we prove for the oracle estimator (and hence the nonconvex MLE estimator) can be seen as another realization of the Neyman and Scott problem.

\section{Numerical experiments\label{sec:experiment}}
In this section, we provide numerical support for Theorems~\ref{thm:main} and~\ref{thm:oracle_lb}.

\paragraph{Setup.} Recall the JIVE model 
\[
\bm{A}_{k}=\bm{U}^{\star}\bm{V}_{k}^{\star\top}+\bm{U}_{k}^{\star}\bm{W}_{k}^{\star\top}+\bm{E}_{k},
\]
where $\bm{E}_{k}$ has i.i.d~zero-mean Gaussian entries with variance $\sigma^{2}$. 
Throughout the experiment, we set  $r=r_{1}=\cdots=r_{k}$
and $d=d_{1}=\cdots=d_{k}$.

For each instance in our simulation, we generate the shared component $\bm{U}^{\star}$ as a random $n\times r$ orthogonal matrix. 
Then we generate the unique subspaces $\{\bm{U}_{k}^{\star}\}$ that are $\theta$-misaligned. 
We follow the randomized method proposed in Section~\ref{sec:identity}.
Specifically, we generate $\bm{Z}$
and $\bm{Z}_{k}$ such that 
\begin{itemize}
\item $\bm{Z}$ is a random $d\times r$ orthonormal matrix such that
$\bm{Z}^{\top}\bm{U}=\bm{0}$. 
\item For each $k\in[K]$, $\bm{Z}_{k}$ is a random $d\times r$ orthonormal
matrix such that $\bm{Z}_{k}^{\top}\bm{U}=\bm{0}$ and $\bm{Z}_{k}^{\top}\bm{Z}=\bm{0}$. 
\end{itemize}
Then for each $k \in [K]$, we construct $\bm{U}_{k}^{\star}$ as 
\[
\bm{U}_{k}^{\star}=\sqrt{1-{\theta}}\cdot\bm{Z} +\sqrt{{\theta}}\cdot\bm{Z}_{k}.
\]
This
ensures that $\bm{U}_{k}^{\star}$ is orthonormal and $\bm{U}_{k}^{\star\top}\bm{U}^\star=\bm{0}$.
As we have explained in Section~\ref{sec:identity}, this construction also fulfills the $\theta$-misalignment requirement. 
Last but not least, we introduce two schemes for generating the loading matrices $\bm{V}_{k}^{\star}$ and $\bm{W}_{k}^{\star}$.
\begin{itemize}
\item Random loading: for each $k$, we let $\bm{V}_{k}^{\star}$ and $(1/\gamma)\bm{W}_{k}^{\star}$
be random orthonormal matrices, where $\gamma$ is a parameter controlling the signal strength of the unique components relative to
the shared component. 
\item Shared loading: we let $\bm{V}^{\star}$ and $(1/\gamma)\bm{W}^{\star}$
be random orthonormal matrices. For each $k\in[K]$, we let $\bm{V}_{k}^{\star}=\bm{V}^{\star}$
and $\bm{W}_{k}^{\star}=\bm{W}^{\star}$. 
This selection arises from the hard instance for the oracle-aided spectral estimator. 
\end{itemize}

\begin{figure}[t]
\begin{centering}
 \begin{minipage}[t]{0.45\textwidth}
        \centering
    \includegraphics[scale=0.45]{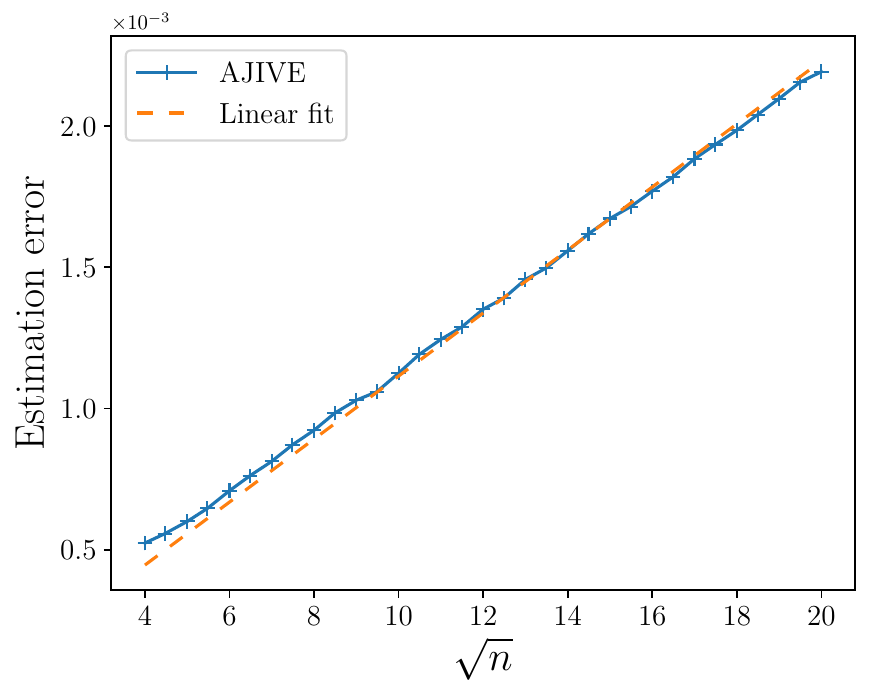}
        \subcaption{\label{fig:large_theta_n}Estimation error $\|\widehat{\bm{U}}\widehat{\bm{U}}^{\top}-\bm{U}^{\star}\bm{U}^{\star\top}\|$
vs $\sqrt{n}$. The parameter is chosen to be $K=100$ and $n$ ranges
from 16 to 400.}
\end{minipage}
\qquad{}
 \begin{minipage}[t]{0.45\textwidth}
        \centering
\includegraphics[scale=0.45]{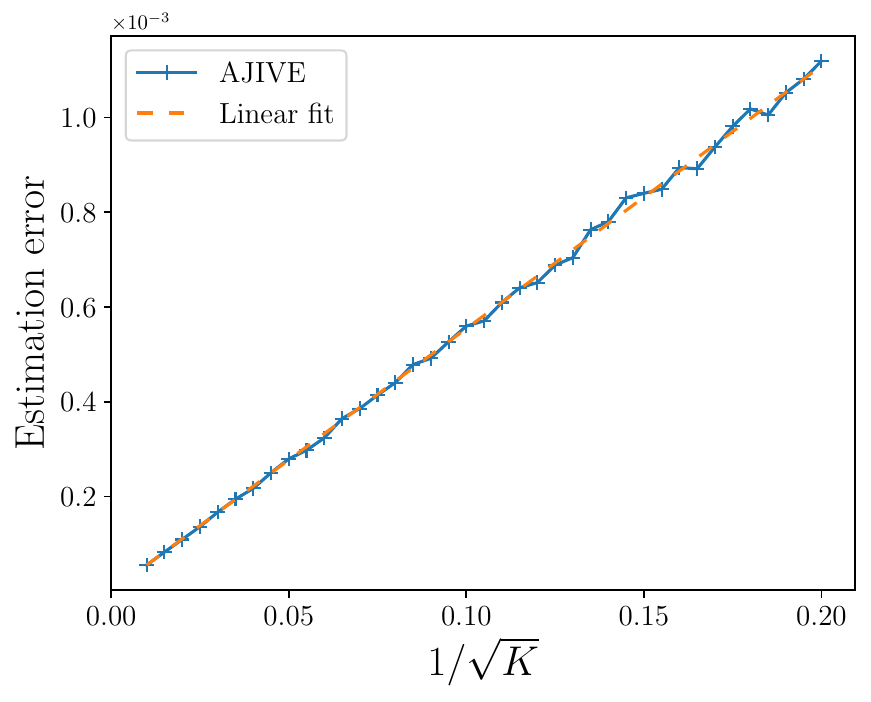}
        \subcaption{\label{fig:large_theta_K}Estimation error $\|\widehat{\bm{U}}\widehat{\bm{U}}^{\top}-\bm{U}^{\star}\bm{U}^{\star\top}\|$
vs $1/\sqrt{K}$. The parameter is chosen to be $n=20$ and $K$ ranges
from 25 to 10000.}
\end{minipage}
\caption{\label{fig:large_theta}Estimation error when ${\theta}=1/2$.
The parameters are chosen to be $d=20$, $r=2$, $\sigma=0.001$,
and $\gamma=0.5$. The loading matrices use the random generation
scheme. Each point is an average of 100 trials.}
\end{centering}

\end{figure}

\begin{figure}
    \centering
    \begin{minipage}[t]{0.45\textwidth}
        \centering
        \includegraphics[scale=0.45]{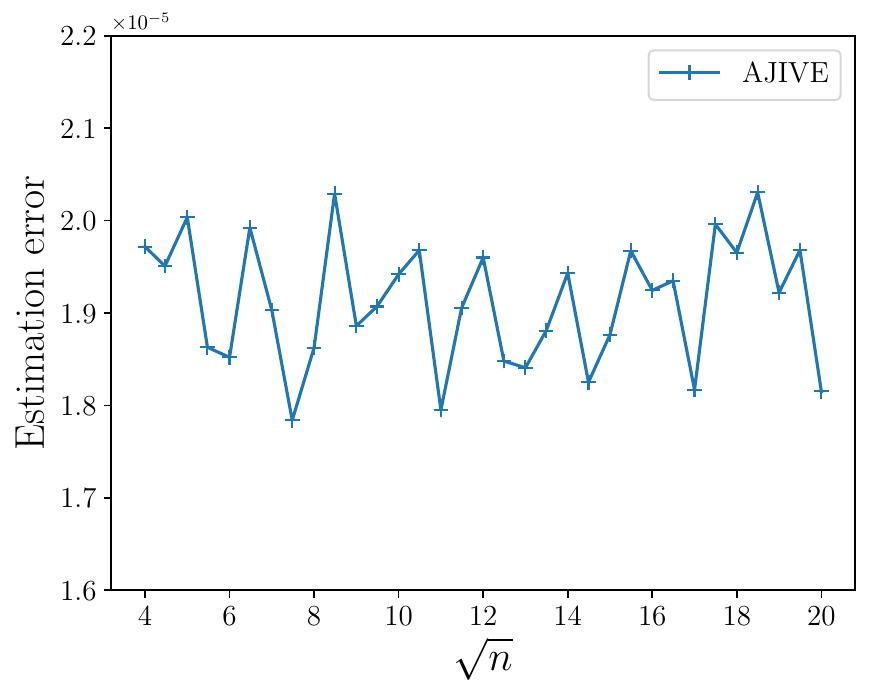} 
        \subcaption{\label{fig:small_theta_n}Estimation error $\|\widehat{\bm{U}}\widehat{\bm{U}}^{\top}-\bm{U}^{\star}\bm{U}^{\star\top}\|$
vs $\sqrt{n}$. }
    \end{minipage}
    \qquad 
    \begin{minipage}[t]{0.45\textwidth}
        \centering
        \includegraphics[scale=0.45]{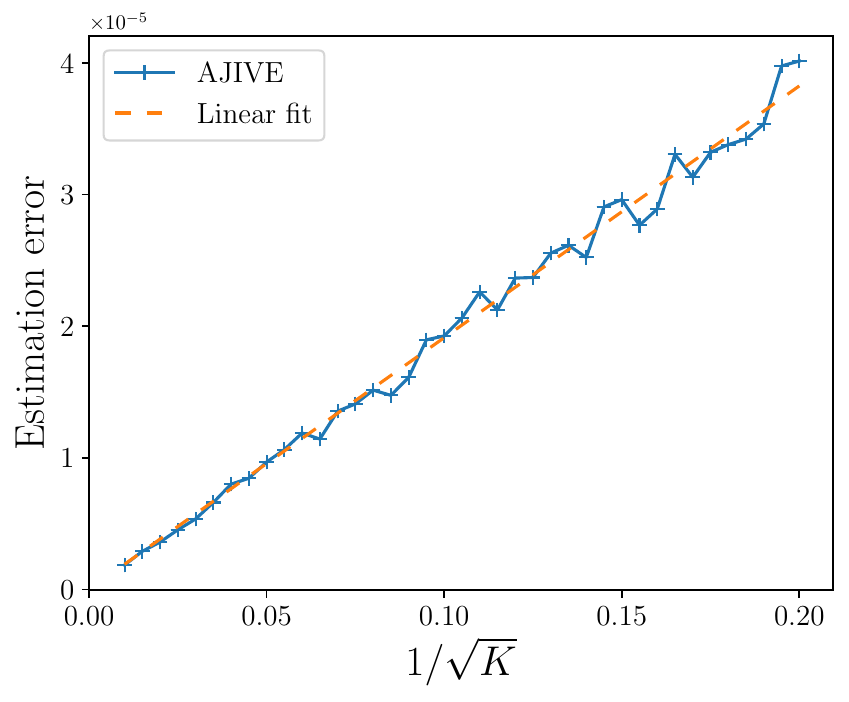} 
        \subcaption{\label{fig:small_theta_K}Estimation error $\|\widehat{\bm{U}}\widehat{\bm{U}}^{\top}-\bm{U}^{\star}\bm{U}^{\star\top}\|$
vs $1/\sqrt{K}$. }
    \end{minipage}

    \vspace{0.5cm}

    \begin{minipage}[t]{0.45\textwidth}
        \centering
        \includegraphics[scale=0.45]{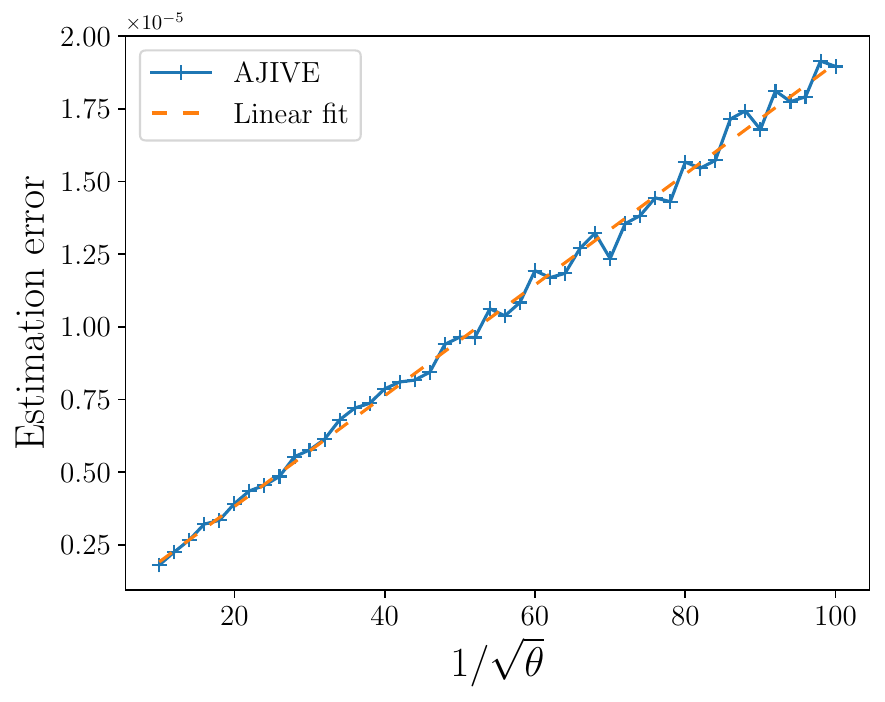} 
        \subcaption{\label{fig:small_theta_theta}Estimation error $\|\widehat{\bm{U}}\widehat{\bm{U}}^{\top}-\bm{U}^{\star}\bm{U}^{\star\top}\|$
vs $1/\sqrt{\theta}$. }
    \end{minipage}
    \qquad 
    \begin{minipage}[t]{0.45\textwidth}
        \vspace{-15em} 
        \caption{\label{fig:small_theta}Estimation error when $\theta$ is small. The loading matrices use the random generation scheme. For all three subfigures, the parameters are $d=20$, $r=2$, $\sigma=10^{-6}$, and $\gamma=0.5$. 
        In (a), $K=100$, $\theta=0.0001$
and $n$ ranges from 16 to 400.
In (b), $n=20$, $\theta=0.0001$,
and $K$ ranges from 25 to 10000.
In (c), $n=20$, $K=100$,
and $\theta$ ranges from $0.01$ to $0.0001$. 
Each point is an average of 100 trials. 
}

    \end{minipage}

\end{figure}

\begin{figure}
\begin{centering}
 \begin{minipage}[t]{0.45\textwidth}
        \centering
        \includegraphics[scale=0.45]{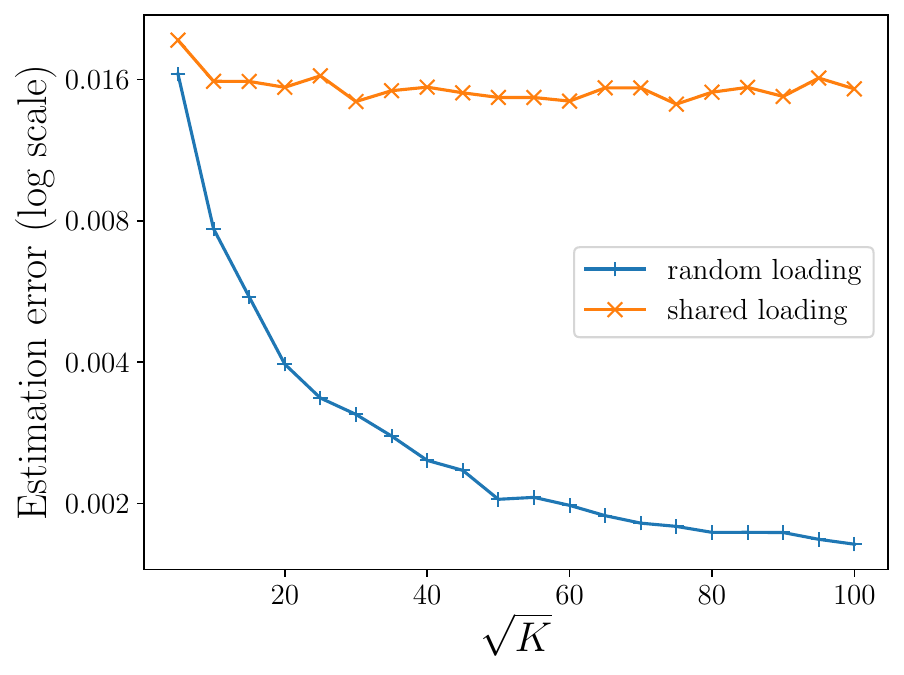}
        \subcaption{\label{fig:random_vs_shared}Estimation error with Algorithm~\ref{alg:spectral}. The noise level is set to be $\sigma=0.01$.}
\end{minipage}
\qquad{}
 \begin{minipage}[t]{0.45\textwidth}
        \centering
    \includegraphics[scale=0.45]{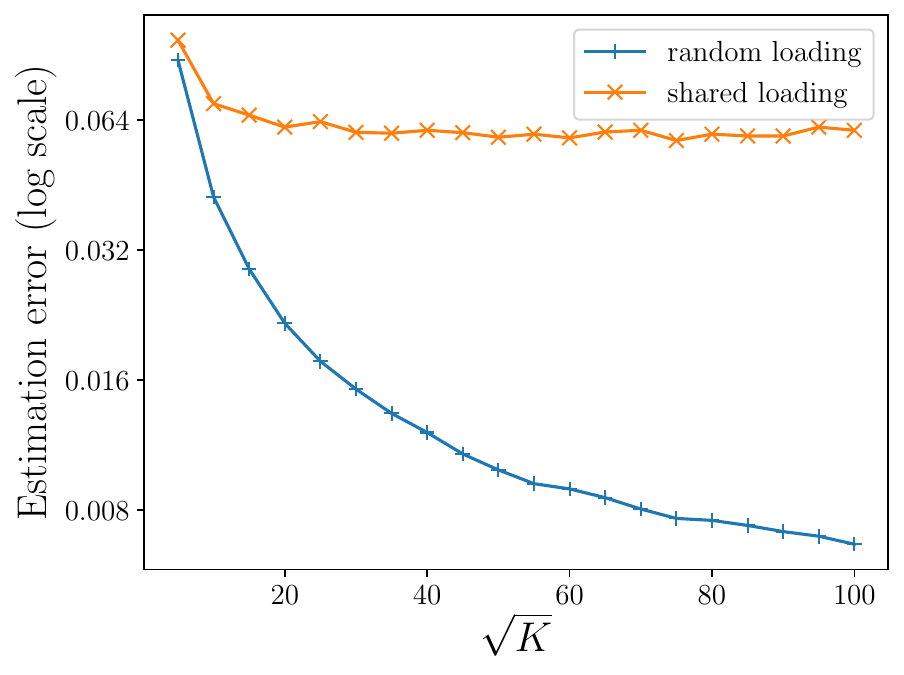}
        \subcaption{\label{fig:oracle} Estimation error with Algorithm~\ref{alg:oracle}. The noise level is set to be $\sigma=0.1$.}
\end{minipage}
\caption{Estimation error $\|\widehat{\bm{U}}\widehat{\bm{U}}^{\top}-\bm{U}^{\star}\bm{U}^{\star\top}\|$
vs $\sqrt{K}$ using random vs shared loading generation scheme. The parameters are chosen to be $n=d=20$, $r=2$, $\gamma=0.5$,
and $K$ ranges from 25 to 10000. Each point is an average of 100
trials. The $y$-axis is in log scale.}
\end{centering}
\end{figure}

\paragraph{Results.}

Theorem~\ref{thm:main} shows that when $r\asymp r_{\mathrm{avg}}\asymp1$,
AJIVE achieves estimation error 
\begin{equation}
\left\Vert \widehat{\bm{U}}\widehat{\bm{U}}^{\top}-\bm{U}^{\star}\bm{U}^{\star\top}\right\Vert \lesssim\frac{\sigma}{\sigma_{\min}}\sqrt{\frac{n}{K}+\frac{r}{K\theta}}.\label{eq:first_order}
\end{equation}
in the high-SNR regime. 
To highlight this error dependence,
we set a small noise level $\sigma$.  
\begin{itemize}
\item Figure~\ref{fig:large_theta} shows the performance of AJIVE when $\theta$ is large; we set $\theta = 1/2$ in the experiment. 
In this case, we see from Figure~\ref{fig:large_theta} that the error scales linearly with $\sqrt{n}$ and $1/\sqrt{K}$. 
\item Figure~\ref{fig:small_theta} shows the performance of AJIVE when $\theta$ is small. 
In particular, Figures~\ref{fig:small_theta_n} and~\ref{fig:small_theta_K} fix a small $\theta= 0.0001$. 
We clearly see that the error does not depend on the ambient dimension $n$, and decays
linearly with $1/\sqrt{K}$. 
Moreover, Figure~\ref{fig:small_theta_theta} fixes $n$ and $K$, and we see a clear linear dependence on $1/\sqrt{\theta}$. 
\end{itemize}
All of these corroborate our theoretical prediction about AJIVE's performance.

Another phenomenon we observe in Theorem~\ref{thm:main} is that
with a high signal-to-noise ratio, AJIVE can be inconsistent, i.e., the estimation error does not go to 0
as $K\rightarrow\infty$. Figure~\ref{fig:random_vs_shared} demonstrates this stagnation under the shared loading scheme. Moreover, Figure~\ref{fig:oracle} shows a similar persisting error with the oracle-aided Algorithm~\ref{alg:oracle}, supporting our oracle lower bound in Theorem~\ref{thm:oracle_lb}. This observation underscores the challenge of estimating shared singular subspaces in this regime.

\section{Discussion}

In this paper, we analyzed the performance of the AJIVE algorithm for estimating the shared subspace from multiple data matrices.  We provided new statistical guarantees on its performance, highlighting the power and potential limitations of multiple matrices in estimating the shared subspace.  We also developed minimax lower bounds, demonstrating the optimality of AJIVE in the high-SNR regime.  
Our analysis revealed that AJIVE achieves exact recovery in the noiseless case and exhibits first-order optimality in the high-SNR regime.  Additionally, we confirmed the benefit of using multiple data matrices for estimation in this regime.  However, we also observed that AJIVE suffers from a non-diminishing error in the low-SNR regime as the number of matrices increases.  
To further investigate this non-diminishing error, we provided a performance lower bound for an oracle-aided spectral estimator.  This analysis suggested that the non-diminishing error might be a fundamental limit of the problem, rather than an artifact of the AJIVE algorithm.  

Our work provides a theoretical understanding of shared subspace estimation from multiple matrices. Future research directions include closing the gap between the upper and lower bounds in the low-SNR regime and exploring other estimation methods in this context. 

\subsection*{Acknowledgements}
YY and CM were partially supported by the National Science Foundation via grant DMS-2311127.

\bibliographystyle{alpha}
\bibliography{All-of-Bibs}

\newcommand{\etalchar}[1]{$^{#1}$}
\begin{thebibliography}{FWWZ19}

\bibitem[CCFM21]{chen2021spectral}
Yuxin Chen, Yuejie Chi, Jianqing Fan, and Cong Ma.
\newblock Spectral methods for data science: A statistical perspective.
\newblock {\em Foundations and Trends{\textregistered} in Machine Learning},
  14(5):566--806, 2021.

\bibitem[CLC{\etalchar{+}}21]{Cai2021Unbalanced}
Changxiao Cai, Gen Li, Yuejie Chi, H.~Vincent Poor, and Yuxin Chen.
\newblock Subspace estimation from unbalanced and incomplete data matrices:
  {$\ell_{2,\infty}$} statistical guarantees.
\newblock {\em Ann. Statist.}, 49(2):944--967, 2021.

\bibitem[CMW13]{Cai2013}
T.~Tony Cai, Zongming Ma, and Yihong Wu.
\newblock Sparse {PCA}: optimal rates and adaptive estimation.
\newblock {\em Ann. Statist.}, 41(6):3074--3110, 2013.

\bibitem[CZ18]{cai2018rate}
T.~Tony Cai and Anru Zhang.
\newblock Rate-optimal perturbation bounds for singular subspaces with
  applications to high-dimensional statistics.
\newblock {\em Ann. Statist.}, 46(1):60--89, 2018.

\bibitem[EAJ18]{el2018detection}
Ahmed El~Alaoui and Michael~I Jordan.
\newblock Detection limits in the high-dimensional spiked rectangular model.
\newblock In {\em Conference On Learning Theory}, pages 410--438. PMLR, 2018.

\bibitem[FJHM18]{Feng2018}
Qing Feng, Meilei Jiang, Jan Hannig, and JS~Marron.
\newblock Angle-based joint and individual variation explained.
\newblock {\em Journal of multivariate analysis}, 166:241--265, 2018.

\bibitem[FWWZ19]{Fan2019dist}
Jianqing Fan, Dong Wang, Kaizheng Wang, and Ziwei Zhu.
\newblock Distributed estimation of principal eigenspaces.
\newblock {\em Ann. Statist.}, 47(6):3009--3031, 2019.

\bibitem[GL19]{gaynanova2019structural}
Irina Gaynanova and Gen Li.
\newblock Structural learning and integrative decomposition of multi-view data.
\newblock {\em Biometrics}, 75(4):1121--1132, 2019.

\bibitem[GLLJ21]{gao2021covariate}
Xing Gao, Sungwon Lee, Gen Li, and Sungkyu Jung.
\newblock Covariate-driven factorization by thresholding for multiblock data.
\newblock {\em Biometrics}, 77(3):1011--1023, 2021.

\bibitem[Iss18]{isserlis1918formula}
Leon Isserlis.
\newblock On a formula for the product-moment coefficient of any order of a
  normal frequency distribution in any number of variables.
\newblock {\em Biometrika}, 12(1/2):134--139, 1918.

\bibitem[JJ12]{jones2012information}
Gareth~A Jones and J~Mary Jones.
\newblock {\em Information and coding theory}.
\newblock Springer Science \& Business Media, 2012.

\bibitem[Lan00]{lancaster2000incidental}
Tony Lancaster.
\newblock The incidental parameter problem since 1948.
\newblock {\em Journal of econometrics}, 95(2):391--413, 2000.

\bibitem[LHMN13]{lock2013joint}
Eric~F Lock, Katherine~A Hoadley, James~Stephen Marron, and Andrew~B Nobel.
\newblock Joint and individual variation explained ({JIVE}) for integrated
  analysis of multiple data types.
\newblock {\em The annals of applied statistics}, 7(1):523, 2013.

\bibitem[MLZ22]{macdonald2022latent}
P.~W. MacDonald, E.~Levina, and J.~Zhu.
\newblock Latent space models for multiplex networks with shared structure.
\newblock {\em Biometrika}, 109(3):683--706, 2022.

\bibitem[MM24]{ma2024optimal}
Zhengchi Ma and Rong Ma.
\newblock Optimal estimation of shared singular subspaces across multiple noisy
  matrices.
\newblock {\em arXiv preprint arXiv:2411.17054}, 2024.

\bibitem[NS48]{neyman1948consistent}
Jerzy Neyman and Elizabeth~L Scott.
\newblock Consistent estimates based on partially consistent observations.
\newblock {\em Econometrica: journal of the Econometric Society}, pages 1--32,
  1948.

\bibitem[NW87]{neudecker1987fourth}
Heinz Neudecker and Tom Wansbeek.
\newblock Fourth-order properties of normally distributed random matrices.
\newblock {\em Linear Algebra and its Applications}, 97:13--21, 1987.

\bibitem[PJH{\etalchar{+}}24]{prothero2024data}
Jack Prothero, Meilei Jiang, Jan Hannig, Quoc Tran-Dinh, Andrew Ackerman, and
  JS~Marron.
\newblock Data integration via analysis of subspaces ({DIVAS}).
\newblock {\em TEST}, pages 1--42, 2024.

\bibitem[PTG21]{ponzi2021rajive}
Erica Ponzi, Magne Thoresen, and Abhik Ghosh.
\newblock {RaJIVE}: Robust angle based {JIVE} for integrating noisy
  multi-source data.
\newblock {\em arXiv preprint arXiv:2101.09110}, 2021.

\bibitem[SFAK24]{shi2024triple}
Naichen Shi, Salar Fattahi, and Raed Al~Kontar.
\newblock Triple component matrix factorization: Untangling global, local, and
  noisy components.
\newblock {\em Journal of Machine Learning Research}, 25(332):1--76, 2024.

\bibitem[SK24]{shi2024personalized}
Naichen Shi and Raed~Al Kontar.
\newblock Personalized {PCA}: Decoupling shared and unique features.
\newblock {\em Journal of machine learning research}, 25:1--82, 2024.

\bibitem[SKF23]{Shi2023}
Naichen Shi, Raed~Al Kontar, and Salar Fattahi.
\newblock Heterogeneous matrix factorization: When features differ by datasets.
\newblock {\em arXiv preprint arXiv:2305.17744}, 2023.

\bibitem[STG24]{sergazinov2024spectral}
Renat Sergazinov, Armeen Taeb, and Irina Gaynanova.
\newblock A spectral method for multi-view subspace learning using the product
  of projections.
\newblock {\em arXiv preprint arXiv:2410.19125}, 2024.

\bibitem[Ver18]{vershynin2018high}
Roman Vershynin.
\newblock {\em High-dimensional probability: An introduction with applications
  in data science}, volume~47.
\newblock Cambridge university press, 2018.

\bibitem[Wed73]{wedin1973perturbation}
Per-{\AA}ke Wedin.
\newblock Perturbation theory for pseudo-inverses.
\newblock {\em BIT Numerical Mathematics}, 13:217--232, 1973.

\bibitem[Xia21]{Xia2021}
Dong Xia.
\newblock Normal approximation and confidence region of singular subspaces.
\newblock {\em Electronic Journal of Statistics}, 15(2):3798--3851, 2021.

\bibitem[Yu97]{Yu1997}
Bin Yu.
\newblock Assouad, {F}ano, and {L}e {C}am.
\newblock In {\em Festschrift for {L}ucien {L}e {C}am}, pages 423--435.
  Springer, New York, 1997.

\bibitem[ZCZM15]{zhou2015group}
Guoxu Zhou, Andrzej Cichocki, Yu~Zhang, and Danilo~P Mandic.
\newblock Group component analysis for multiblock data: Common and individual
  feature extraction.
\newblock {\em IEEE transactions on neural networks and learning systems},
  27(11):2426--2439, 2015.

\bibitem[ZT22]{zheng2022limit}
Runbing Zheng and Minh Tang.
\newblock Limit results for distributed estimation of invariant subspaces in
  multiple networks inference and {PCA}.
\newblock {\em arXiv preprint arXiv:2206.04306}, 2022.

\end{thebibliography}

\appendix

\section{Additional experiments\label{sec:Additional_experiments}}

In this section we show the results of some additional experiment
that is omitted in the main text due to space constraint. We use the
same experiment setting as Section~\ref{sec:experiment}.

\begin{figure}[h]
\begin{centering}
 \begin{minipage}[t]{0.45\textwidth}
        \centering
\includegraphics[scale=0.45]{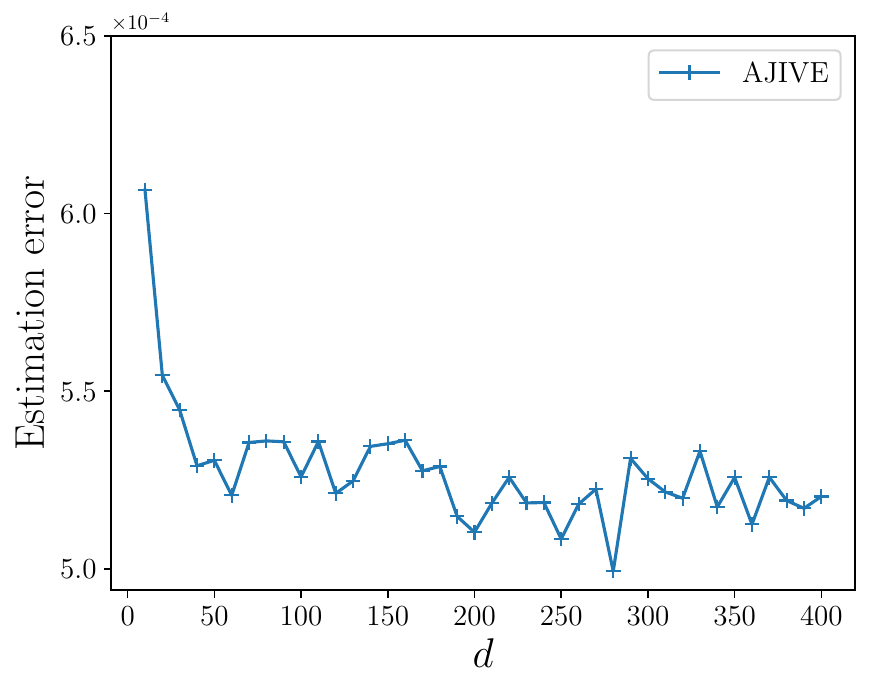}
        \subcaption{\label{fig:large_theta_d}Estimation error $\|\widehat{\bm{U}}\widehat{\bm{U}}^{\top}-\bm{U}^{\star}\bm{U}^{\star\top}\|$
vs $d$. The parameter is chosen to be $\sigma=10^{-3}$ and $d$
ranges from 10 to 400.}
\end{minipage}
\qquad{}
 \begin{minipage}[t]{0.45\textwidth}
        \centering
\includegraphics[scale=0.45]{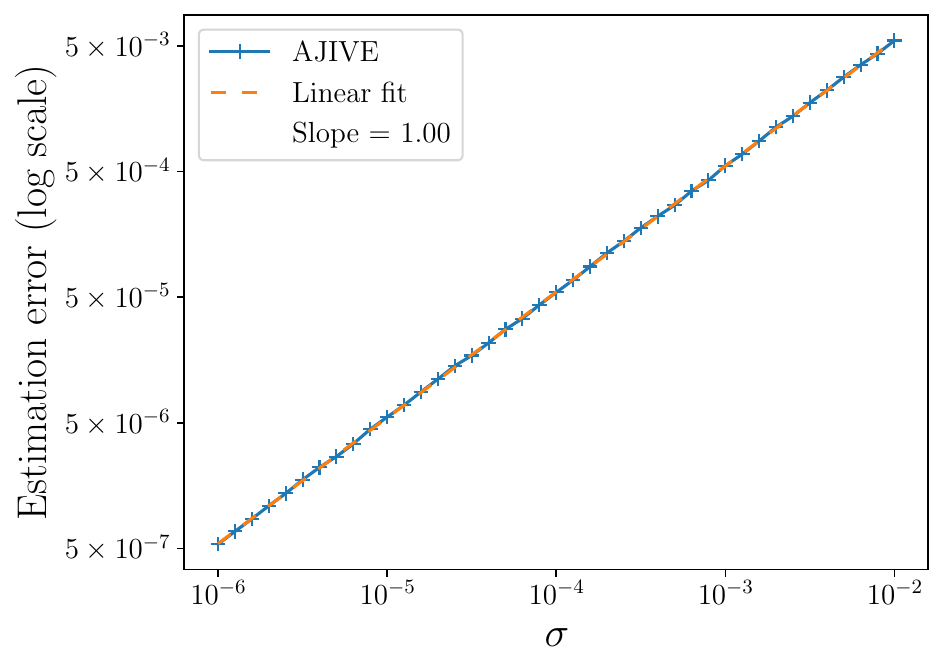}
        \subcaption{\label{fig:large_theta_sigma}Estimation error $\|\widehat{\bm{U}}\widehat{\bm{U}}^{\top}-\bm{U}^{\star}\bm{U}^{\star\top}\|$
vs $\sigma$. The parameter is chosen to be $d=20$ and $\sigma$
ranges from $10^{-6}$ to $10^{-3}$. Both $x$-axis and $y$-axis
are in log scale.}
\end{minipage}
\caption{\label{fig:large_theta_extra}Estimation error when $\widetilde{\theta}=1/2$.
The parameters are chosen to be $n=20$, $r=2$, $K=100$, and $\gamma=0.5$.
The loading matrices use the random generation scheme. Each point
is an average of 100 trials.}
\end{centering}
\end{figure}

We verify the first-order error dependency on $\sigma$ and $d$.
In Figure~\ref{fig:large_theta_d}, we can see that the estimation
errors are about the same for different $d$. In Figure~\ref{fig:large_theta_sigma},
we see that the estimation error scales linearly with the noise level
$\sigma$.

\section{Expectation of monomials of the noise \label{subsec:Supporting_expectation}}

In this section we collect some generic results about the expectation
of monomials of $\bm{E}\in\mathbb{R}^{n_{1}\times n_{2}}$, which
is a random matrix whose entries are i.i.d. Gaussian random variables
with variance $\sigma^{2}$. These results will be helpful in proving
Lemma~\ref{lem:EQ}. As any odd degree monomial of $\bm{E}$ is zero-mean,
we focus on degree 2 and 4 monomials here. By linearity of expectation,
we only need to compute those monomials in the form of $\bm{E}_{1}\bm{A}\bm{E}_{2}$
and $\bm{E}_{1}\bm{A}\bm{E}_{2}\bm{B}\bm{E}_{3}\bm{C}\bm{E}_{4}$,
where $\bm{A},\bm{B},\bm{C}$ are generic matrices and $\bm{E}_{i}$
is either $\bm{E}$ or $\bm{E}^{\top}$. Note that the flip all the
transpose of $\bm{E}$ is equivalent to computing the expectation
as a monomial of $\bm{E}^{\top}$, which is itself a random matrix
with i.i.d. Gaussian noise. Hence we omit those monomials that are
inherently repetitive.

We start with degree-2 monomials $\bm{E}\bm{A}\bm{E}$ and $\bm{E}\bm{A}\bm{E}^{\top}$.
In addition we compute $\mathbb{E}[\mathrm{Trace}(\bm{E}\bm{A})\bm{E}]$
which is useful later in the analysis for the degree-4 monomials.
The proof is deferred to Section~\ref{subsec:Proof_expectation_2}.

\begin{lemma}\label{lem:expectation_2} Let $\bm{E}$ be a $n_{1}\times n_{2}$
matrix where each entry $E_{ij}$ is a zero-mean Gaussian random variable
with variance $\sigma^{2}$. Let $\bm{A}$ be a matrix of appropriate
dimension. Then 
\begin{align*}
\mathbb{E}[\bm{E}\bm{A}\bm{E}] & =\sigma^{2}\bm{A}^{\top};\\
\mathbb{E}[\bm{E}\bm{A}\bm{E}^{\top}] & =\sigma^{2}\mathrm{Trace}(\bm{A})\bm{I};\\
\mathbb{E}[\mathrm{Trace}(\bm{E}\bm{A})\bm{E}] & =\sigma^{2}\bm{A}^{\top}.
\end{align*}

\end{lemma}Similarly we have the following lemma for degree-4 monomials.
The proof is deferred to Section~\ref{subsec:Proof_expectation_4}.

\begin{lemma}\label{lem:expectation_4} Let $\bm{E}$ be a $n_{1}\times n_{2}$
matrix where each entry $E_{ij}$ is a zero-mean Gaussian random variable
with variance $\sigma^{2}$. Let $\bm{A},\bm{B},\bm{C}$ be matrices
of appropriate dimension. Then 

\begin{subequations}\label{eq:expectation_E_fourth}
\begin{align}
\mathbb{E}\left[\bm{E}\bm{A}\bm{E}^{\top}\bm{B}\bm{E}\bm{C}\bm{E}^{\top}\right] & =\sigma^{4}\mathrm{Trace}(\bm{C})\mathrm{Trace}(\bm{A})\bm{B}+\sigma^{4}\mathrm{Trace}(\bm{A}\bm{C}^{\top})\bm{B}^{\top}+\sigma^{4}\mathrm{Trace}(\bm{B})\mathrm{Trace}(\bm{AC})\bm{I}\label{eq:expectation_E_fourth_1}\\
\mathbb{E}\left[\bm{E}\bm{A}\bm{E}\bm{B}\bm{E}^{\top}\bm{C}\bm{E}^{\top}\right] & =\sigma^{4}\bm{A}^{\top}\bm{B}\bm{C}^{\top}+\sigma^{4}\bm{C}\bm{B}\bm{A}+\sigma^{4}\mathrm{Trace}(\bm{B})\mathrm{Trace}(\bm{AC})\bm{I}\label{eq:expectation_E_fourth_2}\\
\mathbb{E}\left[\bm{E}^{\top}\bm{A}\bm{E}\bm{B}\bm{E}\bm{C}\bm{E}^{\top}\right] & =\sigma^{4}\mathrm{Trace}(\bm{C})\mathrm{Trace}(\bm{A})\bm{B}+\sigma^{4}\bm{C}\bm{B}\bm{A}+\sigma^{4}\bm{C}^{\top}\bm{B}\bm{A}^{\top}\label{eq:expectation_E_fourth_3}\\
\mathbb{E}\left[\bm{E}^{\top}\bm{A}\bm{E}\bm{B}\bm{E}\bm{C}\bm{E}\right] & =\sigma^{4}\mathrm{Trace}(\bm{A})\bm{B}\bm{C}^{\top}+\sigma^{4}\bm{C}\bm{A}^{\top}\bm{B}^{\top}+\sigma^{4}\mathrm{Trace}(\bm{A}\bm{B}^{\top}\bm{C})\bm{I}\label{eq:expectation_E_fourth_4}\\
\mathbb{E}\left[\bm{E}\bm{A}\bm{E}^{\top}\bm{B}\bm{E}\bm{C}\bm{E}\right] & =\sigma^{4}\mathrm{Trace}(\bm{A})\bm{B}\bm{C}^{\top}+\sigma^{4}\bm{B}^{\top}\bm{C}^{\top}\bm{A}+\sigma^{4}\mathrm{Trace}(\bm{B})\bm{C}^{\top}\bm{A}^{\top}\label{eq:expectation_E_fourth_5}\\
\mathbb{E}\left[\bm{E}\bm{A}\bm{E}\bm{B}\bm{E}^{\top}\bm{C}\bm{E}\right] & =\sigma^{4}\mathrm{Trace}(\bm{C})\bm{A}^{\top}\bm{B}+\sigma^{4}\bm{C}\bm{A}^{\top}\bm{B}^{\top}+\sigma^{4}\mathrm{Trace}(\bm{B})\bm{C}^{\top}\bm{A}^{\top}\label{eq:expectation_E_fourth_6}\\
\mathbb{E}\left[\bm{E}\bm{A}\bm{E}\bm{B}\bm{E}\bm{C}\bm{E}^{\top}\right] & =\sigma^{4}\mathrm{Trace}(\bm{C})\bm{A}^{\top}\bm{B}+\sigma^{4}\bm{B}^{\top}\bm{C}^{\top}\bm{A}+\sigma^{4}\mathrm{Trace}(\bm{A}\bm{B}^{\top}\bm{C})\bm{I}\label{eq:expectation_E_fourth_7}\\
\mathbb{E}\left[\bm{E}\bm{A}\bm{E}\bm{B}\bm{E}\bm{C}\bm{E}\right] & =\sigma^{4}\bm{A}^{\top}\bm{B}\bm{C}^{\top}+\sigma^{4}\mathrm{Trace}(\bm{A}\bm{C}^{\top})\bm{B}^{\top}+\sigma^{4}\bm{C}^{\top}\bm{B}\bm{A}^{\top}.\label{eq:expectation_E_fourth_8}
\end{align}

\end{subequations}

\end{lemma}

\subsection{Proof of Lemma~\ref{lem:expectation_2} \label{subsec:Proof_expectation_2}}

We prove the equations in Lemma~\ref{lem:expectation_2} in order.

\paragraph{$\mathbb{E}[\bm{E}\bm{A}\bm{E}]$. }

For each $(i,j)\in[n_{1}]\times[n_{2}]$,

\[
\mathbb{E}[\bm{E}\bm{A}\bm{E}]_{ij}=\sum_{k,l}\mathbb{E}\bm{E}_{ik}A_{kl}\bm{E}_{lj}=\mathbb{E}\bm{E}_{ij}A_{ji}\bm{E}_{ij}=\sigma^{2}A_{ji},
\]
where the second equality holds since $\mathbb{E}\bm{E}_{ik}A_{kl}\bm{E}_{lj}=0$
if $i\neq l$ or $k\neq j$. Thus 
\[
\mathbb{E}[\bm{E}\bm{A}\bm{E}]=\sigma^{2}\bm{A}^{\top}.
\]

\paragraph{$\mathbb{E}[\bm{E}\bm{A}\bm{E}^{\top}]$. }

For each $(i,j)\in[n_{1}]\times[n_{2}]$, if $i\neq j$, 
\[
\mathbb{E}[\bm{E}\bm{A}\bm{E}^{\top}]_{ij}=\sum_{k,l}\mathbb{E}\bm{E}_{ik}A_{kl}\bm{E}_{jl}=0
\]
If $i=j$, 

\begin{align*}
\mathbb{E}[\bm{E}\bm{A}\bm{E}^{\top}]_{ii} & =\sum_{k,l}\mathbb{E}\bm{E}_{ik}A_{kl}\bm{E}_{il}=\sum_{k}\mathbb{E}\bm{E}_{ik}A_{kk}\bm{E}_{ik}=\sigma^{2}\mathrm{Trace}(\bm{A}).
\end{align*}
Then $\mathbb{E}[\bm{E}\bm{A}\bm{E}^{\top}]=\sigma^{2}\mathrm{Trace}(\bm{A})\bm{I}$.

\paragraph{$\mathbb{E}[\mathrm{Trace}(\bm{E}\bm{A})\bm{E}]$.}

For each $(i,j)$,

\[
\mathbb{E}[\mathrm{Trace}(\bm{E}\bm{A})\bm{E}]_{ij}=\mathbb{E}[\mathrm{Trace}(\bm{E}\bm{A})E_{ij}]=\mathbb{E}\left(\sum_{k}\sum_{l}E_{kl}A_{lk}\right)E_{ij}=\mathbb{E}E_{ij}A_{ji}E_{ij}=\sigma^{2}A_{ji}.
\]
Then
\[
\mathbb{E}[\mathrm{Trace}(\bm{E}\bm{A})\bm{E}]=\sigma^{2}\bm{A}^{\top}.
\]

\subsection{Proof of Lemma~\ref{lem:expectation_4} \label{subsec:Proof_expectation_4}}

This proof is inspired by results in \cite{neudecker1987fourth},
which gives a formula for $\mathbb{E}\left[\bm{E}\bm{A}\bm{E}^{\top}\bm{B}\bm{E}\bm{C}\bm{E}^{\top}\right]$.
It is proved using the following argument:

Expand $\mathbb{E}\left[\bm{E}\bm{A}\bm{E}^{\top}\bm{B}\bm{E}\bm{C}\bm{E}^{\top}\right]$
as a polynomial of real valued random variables $E_{ij}$. Consider
a degree-four monomial $abcE_{i_{1}j_{1}}E_{i_{2}j_{2}}E_{i_{3}j_{3}}E_{i_{4}j_{4}}$,
where $a,b,c$ are some scalars. by Isserlis's theorem (\cite{isserlis1918formula}),
we have that 
\begin{align*}
abc\mathbb{E}E_{i_{1}j_{1}}E_{i_{2}j_{2}}E_{i_{3}j_{3}}E_{i_{4}j_{4}} & =abc\mathbb{E}\left[E_{i_{1}j_{1}}E_{i_{2}j_{2}}\right]\mathbb{E}\left[E_{i_{3}j_{3}}E_{i_{4}j_{4}}\right]\\
 & \quad+abc\mathbb{E}\left[E_{i_{1}j_{1}}E_{i_{3}j_{3}}\right]\mathbb{E}\left[E_{i_{2}j_{2}}E_{i_{4}j_{4}}\right]\\
 & \quad+abc\mathbb{E}\left[E_{i_{1}j_{1}}E_{i_{4}j_{4}}\right]\mathbb{E}\left[E_{i_{2}j_{2}}E_{i_{3}j_{3}}\right].
\end{align*}
Now we putting this back to the polynomial of $\mathbb{E}\left[\bm{E}\bm{A}\bm{E}^{\top}\bm{B}\bm{E}\bm{C}\bm{E}^{\top}\right]$.
To help with the notation, we label the four occurrences of $\bm{E}$
as $\bm{E}_{1},\bm{E}_{2},\bm{E}_{3},\bm{E}_{4}$ without altering
their meaning. We also label the expectation sign $\mathbb{E}_{ij}$
to indicate that the expectation is taken for $\bm{E}_{i}$ and $\bm{E}_{j}$,
treating all other matrices as constants. Combining all terms, we
have that 
\begin{align*}
\mathbb{E}\left[\bm{E}_{1}\bm{A}\bm{E}_{2}^{\top}\bm{B}\bm{E}_{3}\bm{C}\bm{E}_{4}^{\top}\right] & =\mathbb{E}_{12}\left[\bm{E}_{1}\bm{A}\bm{E}_{2}^{\top}\bm{B}\mathbb{E}_{34}\left[\bm{E}_{3}\bm{C}\bm{E}_{4}^{\top}\right]\right]\\
 & \quad+\mathbb{E}_{13}\left[\bm{E}_{1}\bm{A}\mathbb{E}_{24}\left[\bm{E}_{2}^{\top}\bm{B}\bm{E}_{3}\bm{C}\bm{E}_{4}^{\top}\right]\right]\\
 & \quad+\mathbb{E}_{14}\left[\bm{E}_{1}\bm{A}\mathbb{E}_{23}\left[\bm{E}_{2}^{\top}\bm{B}\bm{E}_{3}\right]\bm{C}\bm{E}_{4}^{\top}\right].
\end{align*}
Using the same idea, we can get \eqref{eq:expectation_E_fourth_1}
to \eqref{eq:expectation_E_fourth_8} by substituting some of the
$\bm{E}_{i}$'s with their transpose and simplifying the equation
with Lemma~\ref{lem:expectation_2}. We include all the detailed
steps in the rest of this section.

\paragraph{Proof of \eqref{eq:expectation_E_fourth_1}.}

\begin{align*}
\mathbb{E}\left[\bm{E}\bm{A}\bm{E}^{\top}\bm{B}\bm{E}\bm{C}\bm{E}^{\top}\right] & =\mathbb{E}\left[\bm{E}_{1}\bm{A}\bm{E}_{2}^{\top}\bm{B}\mathbb{E}_{34}[\bm{E}_{3}\bm{C}\bm{E}_{4}^{\top}]\right]+\mathbb{E}\left[\bm{E}\bm{A}\mathbb{E}_{24}[\bm{E}_{2}^{\top}\bm{B}\bm{E}_{3}\bm{C}\bm{E}_{4}^{\top}]\right]\\
 & \quad+\mathbb{E}\left[\bm{E}_{1}\bm{A}\mathbb{E}_{23}[\bm{E}_{2}^{\top}\bm{B}\bm{E}_{3}]\bm{C}\bm{E}_{4}^{\top}\right]\\
 & =\sigma^{2}\mathrm{Trace}(\bm{C})\mathbb{E}\left[\bm{E}_{1}\bm{A}\bm{E}_{2}^{\top}\bm{B}\right]+\sigma^{2}\mathbb{E}\left[\bm{E}\bm{A}\bm{C}^{\top}\bm{E}_{3}^{\top}\bm{B}^{\top}\right]\\
 & \quad+\sigma^{2}\mathrm{Trace}(\bm{B})\mathbb{E}\left[\bm{E}_{1}\bm{A}\bm{C}\bm{E}_{4}^{\top}\right]\\
 & =\sigma^{4}\mathrm{Trace}(\bm{C})\mathrm{Trace}(\bm{A})\bm{B}+\sigma^{4}\mathrm{Trace}(\bm{A}\bm{C}^{\top})\bm{B}^{\top}+\sigma^{4}\mathrm{Trace}(\bm{B})\mathrm{Trace}(\bm{AC})\bm{I}.
\end{align*}

\paragraph{Proof of \eqref{eq:expectation_E_fourth_2}.}

\begin{align*}
\mathbb{E}\left[\bm{E}\bm{A}\bm{E}\bm{B}\bm{E}^{\top}\bm{C}\bm{E}^{\top}\right] & =\mathbb{E}\left[\bm{E}_{1}\bm{A}\bm{E}_{2}\bm{B}\mathbb{E}_{34}[\bm{E}_{3}^{\top}\bm{C}\bm{E}_{4}^{\top}]\right]+\mathbb{E}\left[\bm{E}\bm{A}\mathbb{E}_{24}[\bm{E}_{2}\bm{B}\bm{E}_{3}^{\top}\bm{C}\bm{E}_{4}^{\top}]\right]\\
 & \quad+\mathbb{E}\left[\bm{E}_{1}\bm{A}\mathbb{E}_{23}[\bm{E}_{2}\bm{B}\bm{E}_{3}^{\top}]\bm{C}\bm{E}_{4}^{\top}\right]\\
 & =\sigma^{2}\mathbb{E}\left[\bm{E}_{1}\bm{A}\bm{E}_{2}\bm{B}\bm{C}^{\top}\right]+\sigma^{2}\mathbb{E}\left[\bm{E}\bm{A}\mathrm{Trace}(\bm{B}\bm{E}_{3}^{\top}\bm{C})\right]\\
 & \quad+\sigma^{2}\mathrm{Trace}(\bm{B})\mathbb{E}\left[\bm{E}_{1}\bm{A}\bm{C}\bm{E}_{4}^{\top}\right]\\
 & =\sigma^{4}\bm{A}^{\top}\bm{B}\bm{C}^{\top}+\sigma^{4}\bm{C}\bm{B}\bm{A}+\sigma^{4}\mathrm{Trace}(\bm{B})\mathrm{Trace}(\bm{AC})\bm{I}.
\end{align*}

\paragraph{Proof of \eqref{eq:expectation_E_fourth_3}.}

\begin{align*}
\mathbb{E}\left[\bm{E}^{\top}\bm{A}\bm{E}\bm{B}\bm{E}\bm{C}\bm{E}^{\top}\right] & =\mathbb{E}\left[\bm{E}_{1}^{\top}\bm{A}\bm{E}_{2}\bm{B}\mathbb{E}_{34}[\bm{E}\bm{C}\bm{E}^{\top}]\right]+\mathbb{E}\left[\bm{E}_{1}^{\top}\bm{A}\mathbb{E}_{24}[\bm{E}\bm{B}\bm{E}\bm{C}\bm{E}^{\top}]\right]\\
 & \quad+\mathbb{E}\left[\bm{E}_{1}^{\top}\bm{A}\mathbb{E}_{23}[\bm{E}\bm{B}\bm{E}]\bm{C}\bm{E}_{4}^{\top}\right]\\
 & =\sigma^{2}\mathrm{Trace}(\bm{C})\mathbb{E}\left[\bm{E}_{1}^{\top}\bm{A}\bm{E}_{2}\bm{B}\right]+\sigma^{2}\mathbb{E}\left[\bm{E}_{1}^{\top}\bm{A}\mathrm{Trace}(\bm{B}\bm{E}_{3}\bm{C})\right]\\
 & \quad+\sigma^{2}\mathbb{E}\left[\bm{E}_{1}^{\top}\bm{A}\bm{B}^{\top}\bm{C}\bm{E}_{4}^{\top}\right]\\
 & =\sigma^{4}\mathrm{Trace}(\bm{C})\mathrm{Trace}(\bm{A})\bm{B}+\sigma^{4}\bm{C}\bm{B}\bm{A}+\sigma^{4}\bm{C}^{\top}\bm{B}\bm{A}^{\top}.
\end{align*}

\paragraph{Proof of \eqref{eq:expectation_E_fourth_4}.}

\begin{align*}
\mathbb{E}\left[\bm{E}^{\top}\bm{A}\bm{E}\bm{B}\bm{E}\bm{C}\bm{E}\right] & =\mathbb{E}\left[\bm{E}_{1}^{\top}\bm{A}\bm{E}_{2}\bm{B}\mathbb{E}_{34}[\bm{E}_{3}\bm{C}\bm{E}_{4}]\right]+\mathbb{E}\left[\bm{E}_{1}^{\top}\bm{A}\mathbb{E}_{24}[\bm{E}_{2}\bm{B}\bm{E}_{3}\bm{C}\bm{E}_{4}]\right]\\
 & \quad+\mathbb{E}\left[\bm{E}_{1}^{\top}\bm{A}\mathbb{E}_{23}[\bm{E}_{2}\bm{B}\bm{E}_{3}]\bm{C}\bm{E}_{4}\right]\\
 & =\sigma^{2}\mathbb{E}\left[\bm{E}_{1}^{\top}\bm{A}\bm{E}_{2}\bm{B}\bm{C}^{\top}\right]+\sigma^{2}\mathbb{E}\left[\bm{E}_{1}^{\top}\bm{A}\bm{C}^{\top}\bm{E}_{3}^{\top}\bm{B}^{\top}\right]\\
 & \quad+\sigma^{2}\mathbb{E}\left[\bm{E}_{1}^{\top}\bm{A}\bm{B}^{\top}\bm{C}\bm{E}_{4}\right]\\
 & =\sigma^{4}\mathrm{Trace}(\bm{A})\bm{B}\bm{C}^{\top}+\sigma^{4}\bm{C}\bm{A}^{\top}\bm{B}^{\top}+\sigma^{4}\mathrm{Trace}(\bm{A}\bm{B}^{\top}\bm{C})\bm{I}.
\end{align*}

\paragraph{Proof of \eqref{eq:expectation_E_fourth_5}.}

\begin{align*}
\mathbb{E}\left[\bm{E}\bm{A}\bm{E}^{\top}\bm{B}\bm{E}\bm{C}\bm{E}\right] & =\mathbb{E}\left[\bm{E}\bm{A}\bm{E}^{\top}\bm{B}\mathbb{E}_{34}[\bm{E}\bm{C}\bm{E}]\right]+\mathbb{E}\left[\bm{E}_{1}^{\top}\bm{A}\mathbb{E}_{24}[\bm{E}_{2}\bm{B}\bm{E}_{3}^{\top}\bm{C}\bm{E}_{4}^{\top}]\right]\\
 & \quad+\mathbb{E}\left[\bm{E}_{1}\bm{A}\mathbb{E}_{23}[\bm{E}_{2}^{\top}\bm{B}\bm{E}_{3}]\bm{C}\bm{E}_{4}\right]\\
 & =\sigma^{2}\mathbb{E}\left[\bm{E}_{1}\bm{A}\bm{E}_{2}^{\top}\bm{B}\bm{C}^{\top}\right]+\sigma^{2}\mathbb{E}\left[\bm{E}_{1}\bm{A}\mathrm{Trace}(\bm{B}\bm{E}_{3}^{\top}\bm{C})\right]\\
 & \quad+\sigma^{2}\mathrm{Trace}(\bm{B})\mathbb{E}\left[\bm{E}_{1}\bm{A}\bm{C}\bm{E}_{4}\right]\\
 & =\sigma^{4}\mathrm{Trace}(\bm{A})\bm{B}\bm{C}^{\top}+\sigma^{4}\bm{B}^{\top}\bm{C}^{\top}\bm{A}+\sigma^{4}\mathrm{Trace}(\bm{B})\bm{C}^{\top}\bm{A}^{\top}.
\end{align*}

\paragraph{Proof of \eqref{eq:expectation_E_fourth_6}.}

\begin{align*}
\mathbb{E}\left[\bm{E}\bm{A}\bm{E}\bm{B}\bm{E}^{\top}\bm{C}\bm{E}\right] & =\mathbb{E}\left[\bm{E}_{1}\bm{A}\bm{E}_{2}\bm{B}\mathbb{E}_{34}[\bm{E}_{3}^{\top}\bm{C}\bm{E}_{4}]\right]+\mathbb{E}\left[\bm{E}_{1}\bm{A}\mathbb{E}_{24}[\bm{E}_{2}\bm{B}\bm{E}_{3}^{\top}\bm{C}\bm{E}_{4}]\right]\\
 & \quad+\mathbb{E}\left[\bm{E}_{1}\bm{A}\mathbb{E}_{23}[\bm{E}_{2}\bm{B}\bm{E}_{3}^{\top}]\bm{C}\bm{E}_{4}\right]\\
 & =\sigma^{2}\mathrm{Trace}(\bm{C})\mathbb{E}\left[\bm{E}_{1}\bm{A}\bm{E}_{2}\bm{B}\right]+\sigma^{2}\mathbb{E}\left[\bm{E}_{1}\bm{A}\bm{C}^{\top}\bm{E}_{3}\bm{B}^{\top}\right]\\
 & \quad+\sigma^{2}\mathrm{Trace}(\bm{B})\mathbb{E}\left[\bm{E}_{1}\bm{A}\bm{C}\bm{E}_{4}\right]\\
 & =\sigma^{4}\mathrm{Trace}(\bm{C})\bm{A}^{\top}\bm{B}+\sigma^{4}\bm{C}\bm{A}^{\top}\bm{B}^{\top}+\sigma^{4}\mathrm{Trace}(\bm{B})\bm{C}^{\top}\bm{A}^{\top}.
\end{align*}

\paragraph{Proof of \eqref{eq:expectation_E_fourth_7}.}

\begin{align*}
\mathbb{E}\left[\bm{E}\bm{A}\bm{E}\bm{B}\bm{E}\bm{C}\bm{E}^{\top}\right] & =\mathbb{E}\left[\bm{E}_{1}\bm{A}\bm{E}_{2}\bm{B}\mathbb{E}_{34}[\bm{E}_{3}\bm{C}\bm{E}_{4}^{\top}]\right]+\mathbb{E}\left[\bm{E}_{1}\bm{A}\mathbb{E}_{24}[\bm{E}_{2}\bm{B}\bm{E}_{3}\bm{C}\bm{E}_{4}^{\top}]\right]\\
 & \quad+\mathbb{E}\left[\bm{E}_{1}\bm{A}\mathbb{E}_{23}[\bm{E}_{2}\bm{B}\bm{E}_{3}]\bm{C}\bm{E}_{4}^{\top}\right]\\
 & =\sigma^{2}\mathrm{Trace}(\bm{C})\mathbb{E}\left[\bm{E}_{1}\bm{A}\bm{E}_{2}\bm{B}\right]+\sigma^{2}\mathbb{E}\left[\bm{E}_{1}\bm{A}\mathrm{Trace}(\bm{B}\bm{E}_{3}\bm{C})\right]\\
 & \quad+\sigma^{2}\mathbb{E}\left[\bm{E}_{1}\bm{A}\bm{B}^{\top}\bm{C}\bm{E}_{4}^{\top}\right]\\
 & =\sigma^{4}\mathrm{Trace}(\bm{C})\bm{A}^{\top}\bm{B}+\sigma^{4}\bm{B}^{\top}\bm{C}^{\top}\bm{A}+\sigma^{4}\mathrm{Trace}(\bm{A}\bm{B}^{\top}\bm{C})\bm{I}.
\end{align*}

\paragraph{Proof of \eqref{eq:expectation_E_fourth_8}.}

\begin{align*}
\mathbb{E}\left[\bm{E}\bm{A}\bm{E}\bm{B}\bm{E}\bm{C}\bm{E}\right] & =\mathbb{E}\left[\bm{E}_{1}\bm{A}\bm{E}_{2}\bm{B}\mathbb{E}_{34}[\bm{E}_{3}\bm{C}\bm{E}_{4}]\right]+\mathbb{E}\left[\bm{E}\bm{A}\mathbb{E}_{24}[\bm{E}_{2}\bm{B}\bm{E}_{3}\bm{C}\bm{E}_{4}]\right]\\
 & \quad+\mathbb{E}\left[\bm{E}_{1}\bm{A}\mathbb{E}_{23}[\bm{E}_{2}\bm{B}\bm{E}_{3}]\bm{C}\bm{E}_{4}\right]\\
 & =\sigma^{2}\mathbb{E}\left[\bm{E}_{1}\bm{A}\bm{E}_{2}\bm{B}\bm{C}^{\top}\right]+\sigma^{2}\mathbb{E}\left[\bm{E}\bm{A}\bm{C}^{\top}\bm{E}_{3}^{\top}\bm{B}^{\top}\right]\\
 & \quad+\sigma^{2}\mathbb{E}\left[\bm{E}_{1}\bm{A}\bm{B}^{\top}\bm{C}\bm{E}_{4}\right]\\
 & =\sigma^{4}\bm{A}^{\top}\bm{B}\bm{C}^{\top}+\sigma^{4}\mathrm{Trace}(\bm{A}\bm{C}^{\top})\bm{B}^{\top}+\sigma^{4}\bm{C}^{\top}\bm{B}\bm{A}^{\top}.
\end{align*}

\section{Proof of Theorem~\ref{thm:main}\label{sec:Proof_main}}

We prove a more detailed theorem here. 

\begin{theorem}\label{thm:full} Instate the assumptions of Theorem~\ref{thm:main},
we have that with probability at least $1-O(KN^{-10})$,
\begin{align*}
\left\Vert \widehat{\bm{U}}\widehat{\bm{U}}^{\top}-\bm{U}^{\star}\bm{U}^{\star\top}\right\Vert \le & C_{1}\left[\frac{\sigma}{\sigma_{\min}}\left(\sqrt{\frac{n}{K}+\frac{r+r_{\mathrm{avg}}}{K\theta}+\frac{r\cdot r_{\mathrm{avg}}}{K\theta}\wedge\frac{r}{K^{2}\theta^{2}}}\right)\log^{5/2}N\right.\\
 & \qquad+\frac{\sigma^{2}}{\sigma_{\min}^{2}}\sqrt{\frac{nd}{K}+\frac{d(r+r_{\mathrm{avg}})}{K\theta}+\left(nd+n^{2}\right)\left(\frac{3r_{\mathrm{avg}}}{K\theta}\wedge\frac{1}{K^{2}\theta^{2}}\right)}\log N.\\
 & \qquad+\frac{1}{\theta}\left(\frac{\kappa^{2}\sigma^{2}n}{\sigma_{\min}^{2}}+\frac{\sigma^{4}(nd+n^{2})}{\sigma_{\min}^{4}}\right)\\
 & \qquad+\frac{\sigma^{2}n}{\sigma_{\min}^{2}}\left(\frac{\log N}{K\theta^{2}}+\frac{\log^{2}N}{K^{2}\theta^{2}}\right)\\
 & \qquad\left.+\frac{\sigma^{4}}{\sigma_{\min}^{4}}\left(\frac{nd\log N}{K\theta^{2}}+\frac{(nd+n^{2})\log^{2}N}{K^{2}\theta^{2}}\right)\right]
\end{align*}
for some constant $C_{1}>0$.

\end{theorem}It is easy to see that the following theorem can be
simplified to Theorem~\ref{thm:main} by using Assumption~\eqref{eq:signal_size}
and taking the worst log dependency. It is worth noting that when
$d\gg n$, this complete theorem offers a better dependency on the
dimension for second-order term that does not scale with $K$. Ignoring
the log factors, the dependency of the upper bound of on $\sigma^{2}/\sigma_{\min}^{2}$
is 
\[
\frac{1}{\theta(K\theta\wedge1)}\cdot\frac{\kappa^{2}\sigma^{2}n}{\sigma_{\min}^{2}}
\]
in Theorem~\ref{thm:full} and 
\[
\frac{1}{\theta(K\theta\wedge1)}\cdot\frac{\kappa^{2}\sigma^{2}(\sqrt{nd}+n)}{\sigma_{\min}^{2}}
\]
in the simplified Theorem~\ref{thm:main}. 

We now begin the proof for Theorem~\ref{thm:full}. The key of the
proof of this theorem is the use of a first-order approximation result
in \cite{Xia2021} on SVD. We first establish a first-order approximation
of $\widetilde{\bm{U}}_{k}\widetilde{\bm{U}}_{k}^{\top}$, and then
a first-order approximation of $\widehat{\bm{U}}\widehat{\bm{U}}^{\top}$.
This then allow us to control the estimation error $\|\bm{U}^{\star}\bm{U}^{\star\top}-\widehat{\bm{U}}\widehat{\bm{U}}^{\top}\|$
by controlling the first-order terms, resulting in a first-order optimal
error bound.

We start with $\widetilde{\bm{U}}_{k}\widetilde{\bm{U}}_{k}^{\top}$.
For any $k\in[K]$, let the top-$(r+r_{k})$ SVD of $\bm{A}_{k}^{\star}$
be $\widetilde{\bm{U}}_{k}^{\star}\widetilde{\bm{\Sigma}}_{k}^{\star}\widetilde{\bm{V}}_{k}^{\star\top}$.
Since $\mathrm{rank}(\bm{A}_{k}^{\star})=r+r_{k}$, $\mathrm{col}(\bm{A}_{k}^{\star})=\mathrm{col}(\widetilde{\bm{U}}_{k}^{\star})=\mathrm{col}(\bm{U}^{\star})+\mathrm{col}(\bm{U}_{k}^{\star})$
and we can write it as 
\[
\widetilde{\bm{U}}_{k}^{\star}\widetilde{\bm{U}}_{k}^{\star\top}=\bm{U}^{\star}\bm{U}^{\star\top}+\bm{U}_{k}^{\star}\bm{U}_{k}^{\star\top}.
\]
We also define the matrices $\bm{P}_{k}^{\perp}$ and $\bm{P}_{k}^{-1}$
as $\bm{P}_{k}^{\perp}\coloneqq\bm{I}_{n}-\widetilde{\bm{U}}_{k}^{\star}\widetilde{\bm{U}}_{k}^{\star\top}$
and $\bm{P}_{k}^{-1}\coloneqq\widetilde{\bm{U}}_{k}^{\star}\widetilde{\bm{\Sigma}}_{k}^{\star-2}\widetilde{\bm{U}}_{k}^{\star\top}$.
We have the following first order approximation of $\widetilde{\bm{U}}_{k}\widetilde{\bm{U}}_{k}^{\top}$. 

\begin{lemma}\label{lem:expansion_U}

Instate the assumptions of Theorem~\ref{thm:main}. Then with probability
at least $1-O(KN^{-11})$, for every $k\in[K]$,

\[
\widetilde{\bm{U}}_{k}\widetilde{\bm{U}}_{k}^{\top}=\widetilde{\bm{U}}_{k}^{\star}\widetilde{\bm{U}}_{k}^{\star\top}+\bm{B}_{k}+\bm{B}_{k}^{\top}+\bm{R}_{k},
\]
where 
\[
\bm{B}_{k}\coloneqq\bm{P}_{k}^{\perp}\bm{E}_{k}\widetilde{\bm{V}}_{k}^{\star}\widetilde{\bm{\Sigma}}_{k}^{\star-1}\widetilde{\bm{U}}_{k}^{\star\top}+\bm{P}_{k}^{\perp}\left(\bm{E}_{k}\bm{E}_{k}^{\top}-d_{k}\sigma^{2}\bm{I}_{n}\right)\bm{P}_{k}^{-1}
\]
and $\bm{R}_{k}$ is some $n\times n$ matrix such that 
\[
\|\bm{R}_{k}\|\le C_{1}\left(\frac{\kappa^{2}\sigma^{2}n}{\sigma_{\min}^{2}}+\frac{\sigma^{4}(nd_{k}+n^{2})}{\sigma_{\min}^{4}}\right)
\]
 for some constant $C_{1}>0$.

\end{lemma}Moreover, the following lemma controls the overall size
of the first-order perturbation terms.

\begin{lemma}\label{lem:B_size} Instate the assumptions of Theorem~\ref{thm:main}.
We have that with probability $1-O(KN^{-10})$,
\begin{align*}
\left\Vert \frac{1}{K}\sum_{k=1}^{K}\bm{B}_{k}\right\Vert  & \le C_{2}\cdot\frac{\sigma\sqrt{n}}{\sigma_{\min}}\left(\sqrt{\frac{\log N}{K}}+\frac{\log N}{K}\right)\\
 & \qquad+C_{3}\cdot\frac{\sigma^{2}}{\sigma_{\min}^{2}}\left(\sqrt{\frac{nd\log N}{K}}+\frac{\left(\sqrt{nd}+n\right)\log N}{K}\right)
\end{align*}
for some constants $C_{2},C_{3}>0$.

\end{lemma}Lemma~\ref{lem:expansion_U} and \ref{lem:B_size} show
that 
\begin{align}
\|\bm{\Delta}\| & \le C_{2}\cdot\frac{\sigma\sqrt{n}}{\sigma_{\min}}\left(\sqrt{\frac{\log N}{K}}+\frac{\log N}{K}\right)\label{eq:Delta_bound}\\
 & \qquad+C_{3}\cdot\frac{\sigma^{2}}{\sigma_{\min}^{2}}\left(\sqrt{\frac{nd\log N}{K}}+\frac{\left(\sqrt{nd}+n\right)\log N}{K}\right)\nonumber \\
 & \qquad+C_{1}\left(\frac{\kappa^{2}\sigma^{2}n}{\sigma_{\min}^{2}}+\frac{\sigma^{4}(nd+n^{2})}{\sigma_{\min}^{4}}\right)\le\theta/8.\nonumber 
\end{align}
Here the last inequality comes from Assumption~\eqref{eq:signal_size}.

We now introduce the notations necessary for the first-order approximation
of $\widehat{\bm{U}}\widehat{\bm{U}}^{\top}$. Let 

\begin{subequations}
\begin{align}
\bm{\Delta} & \coloneqq\frac{1}{K}\sum_{k=1}^{K}\left(\bm{B}_{k}+\bm{B}_{k}^{\top}+\bm{R}_{k}\right);\label{eq:delta_def}\\
\bm{P} & \coloneqq\bm{I}_{n}-\frac{1}{K}\sum_{k=1}^{K}\widetilde{\bm{U}}_{k}^{\star}\widetilde{\bm{U}}_{k}^{\star\top}=\bm{I}_{n}-\bm{U}^{\star}\bm{U}^{\star\top}-\frac{1}{K}\sum_{k=1}^{K}\bm{U}_{k}^{\star}\bm{U}_{k}^{\star\top}.\label{eq:P_def}
\end{align}
\end{subequations}As $\|K^{-1}\sum_{k=1}^{K}\bm{U}_{k}^{\star}\bm{U}_{k}^{\star\top}\|<1$,
$\mathrm{rank}(\bm{P})=n-r$. Let its eigen-decomposition be
\begin{equation}
\bm{P}=\sum_{i=1}^{n-r}\mu_{i}\bm{x}_{i}\bm{x}_{i}^{\top},\label{eq:mu_def}
\end{equation}
where $\mu_{i}\in\mathbb{R}$ and $\{\bm{x}_{i}\}_{i\in[n-r]}$ are
mutually orthogonal unit vectors. We can then define 
\begin{align*}
\bm{P}^{-1} & \coloneqq\sum_{i=1}^{n-r}\mu_{i}^{-1}\bm{x}_{i}\bm{x}_{i}^{\top};\\
\bm{P}^{\perp} & \coloneqq\bm{U}^{\star}\bm{U}^{\star\top}.
\end{align*}
We are now ready to state the following lemma, which gives a first-order
approximation of $\widehat{\bm{U}}\widehat{\bm{U}}^{\top}$.

\begin{lemma}\label{lem:approx_UUT}

Instate the assumptions of Theorem~\ref{thm:main}. Then with probability
at least $1-O(KN^{-10})$,
\begin{equation}
\widehat{\bm{U}}\widehat{\bm{U}}^{\top}=\bm{U}^{\star}\bm{U}^{\star\top}+\bm{P}^{-1}\bm{\Delta}\bm{P}^{\perp}+\bm{P}^{\perp}\bm{\Delta}\bm{P}^{-1}+\bm{R}\label{eq:UU^T_decomp}
\end{equation}
where $\bm{R}$ is some $n\times n$ matrix such that 
\begin{equation}
\|\bm{R}\|\le\frac{32\|\bm{\Delta}\|^{2}}{\theta{}^{2}}.\label{eq:R_bound_Delta}
\end{equation}

\end{lemma}Furthermore, the following lemma controls the size of
first-order perturbation. 

\begin{lemma}\label{lem:size_PDeltaP} Instate the assumptions of
Theorem~\ref{thm:main}. We have that with probability at least $1-O(KN^{-10})$,
\begin{align}
\left\Vert \bm{P}^{-1}\bm{\Delta}\bm{P}^{\perp}\right\Vert  & \le C_{4}\frac{\sigma}{\sigma_{\min}}\left(\sqrt{\frac{n}{K}+\frac{r+r_{\mathrm{avg}}}{K\theta}+\frac{r\cdot r_{\mathrm{avg}}}{K\theta}\wedge\frac{r}{K^{2}\theta^{2}}}\right)\log^{5/2}N\label{eq:PDeltaPperp_bound}\\
 & \qquad+C_{5}\frac{\sigma^{2}}{\sigma_{\min}^{2}}\sqrt{\frac{nd}{K}+\frac{d(r+r_{\mathrm{avg}})}{K\theta}+\left(nd+n^{2}\right)\left(\frac{3r_{\mathrm{avg}}}{K\theta}\wedge\frac{1}{K^{2}\theta^{2}}\right)}\log N.\nonumber \\
 & \qquad+\frac{C_{6}}{\theta}\left(\frac{\kappa^{2}\sigma^{2}n}{\sigma_{\min}^{2}}+\frac{\sigma^{4}(nd+n^{2})}{\sigma_{\min}^{4}}\right).\nonumber 
\end{align}
for some constants $C_{4},C_{5},C_{6}>0$.

\end{lemma}

We are now ready to bound the estimation error $\|\widehat{\bm{U}}\widehat{\bm{U}}^{\top}-\bm{U}^{\star}\bm{U}^{\star\top}\|$. Combining \eqref{eq:Delta_bound}, \eqref{eq:UU^T_decomp}, \eqref{eq:R_bound_Delta}, 
and \eqref{eq:PDeltaPperp_bound}, we reach the conclusion that 
\begin{align*}
\left\Vert \widehat{\bm{U}}\widehat{\bm{U}}^{\top}-\bm{U}^{\star}\bm{U}^{\star\top}\right\Vert  & \le2\left\Vert \bm{P}^{-1}\bm{\Delta}\bm{P}^{\perp}\right\Vert +\|\bm{R}\|\\
 & \le2C_{7}\left[\frac{\sigma}{\sigma_{\min}}\left(\sqrt{\frac{n}{K}+\frac{r+r_{\mathrm{avg}}}{K\theta}+\frac{r\cdot r_{\mathrm{avg}}}{K\theta}\wedge\frac{r}{K^{2}\theta^{2}}}\right)\log^{5/2}N\right.\\
 & \qquad+\frac{\sigma^{2}}{\sigma_{\min}^{2}}\sqrt{\frac{nd}{K}+\frac{d(r+r_{\mathrm{avg}})}{K\theta}+\left(nd+n^{2}\right)\left(\frac{3r_{\mathrm{avg}}}{K\theta}\wedge\frac{1}{K^{2}\theta^{2}}\right)}\log N.\\
 & \qquad+\frac{1}{\theta}\left(\frac{\kappa^{2}\sigma^{2}n}{\sigma_{\min}^{2}}+\frac{\sigma^{4}(nd+n^{2})}{\sigma_{\min}^{4}}\right)\\
 & \qquad+\frac{\sigma^{2}n}{\sigma_{\min}^{2}}\left(\frac{\log N}{K\theta^{2}}+\frac{\log^{2}N}{K^{2}\theta^{2}}\right)\\
 & \qquad\left.+\frac{\sigma^{4}}{\sigma_{\min}^{4}}\left(\frac{nd\log N}{K\theta^{2}}+\frac{\left(\sqrt{nd}+n\right)^{2}\log^{2}N}{K^{2}\theta^{2}}\right)\right].
\end{align*}

\subsection{Proof of Lemma~\ref{lem:expansion_U}}

Consider the Gram matrix $\bm{A}_{k}\bm{A}_{k}^{\top}$. Using the
definition of SVD of $\bm{A}^{\star}$, we have that
\begin{align*}
\bm{A}_{k}\bm{A}_{k}^{\top} & =\left(\widetilde{\bm{U}}_{k}^{\star}\widetilde{\bm{\Sigma}}_{k}^{\star}\widetilde{\bm{V}}_{k}^{\star\top}+\bm{E}_{k}\right)\left(\widetilde{\bm{U}}_{k}^{\star}\widetilde{\bm{\Sigma}}_{k}^{\star}\widetilde{\bm{V}}_{k}^{\star\top}+\bm{E}_{k}\right)^{\top}\\
 & =\widetilde{\bm{U}}_{k}^{\star}\widetilde{\bm{\Sigma}}_{k}^{\star2}\widetilde{\bm{U}}_{k}^{\star\top}+\bm{E}_{k}\widetilde{\bm{V}}_{k}^{\star}\widetilde{\bm{\Sigma}}_{k}^{\star}\widetilde{\bm{U}}_{k}^{\star\top}+\widetilde{\bm{U}}_{k}^{\star}\widetilde{\bm{\Sigma}}_{k}^{\star}\widetilde{\bm{V}}_{k}^{\star\top}\bm{E}_{k}^{\top}+\bm{E}_{k}\bm{E}_{k}^{\top}.
\end{align*}
Recall that the columns of $\widetilde{\bm{U}}_{k}$ are the top-$(r+r_{k})$
left singular vectors of $\bm{A}_{k}$. Then equivalently, these columns
are also the top-$(r+r_{k})$ eigenvectors of $\bm{A}_{k}\bm{A}_{k}^{\top}-d_{k}\sigma^{2}\bm{I}_{n}$. 

We will now first present an first-order approximation $\widetilde{\bm{U}}_{k}\widetilde{\bm{U}}_{k}^{\top}$
based on the eigen-decomposition. Treat $\widetilde{\bm{U}}_{k}^{\star}\widetilde{\bm{\Sigma}}_{k}^{\star2}\widetilde{\bm{U}}_{k}^{\star\top}$
as the ground truth and 
\[
\bm{\Delta}_{k}\coloneqq\bm{E}_{k}\widetilde{\bm{V}}_{k}^{\star}\widetilde{\bm{\Sigma}}_{k}^{\star}\widetilde{\bm{U}}_{k}^{\star\top}+\widetilde{\bm{U}}_{k}^{\star}\widetilde{\bm{\Sigma}}_{k}^{\star}\widetilde{\bm{V}}_{k}^{\star\top}\bm{E}_{k}^{\top}+\bm{E}_{k}\bm{E}_{k}^{\top}-d_{k}\sigma^{2}\bm{I}_{n}
\]
as the perturbation. 

We first control $\|\bm{\Delta}_k\|$. We claim that as long as $n\ge C_{1}\log N$
for some large enough constant $C_{1}>0$, with probability at least
$1-O(N^{-11})$,
\begin{subequations}\label{eq:AAT_perturbation_size}
\begin{align}
\|\bm{E}_{k}\bm{E}_{k}^{\top}-d_{k}\sigma^{2}\bm{I}_{n}\| & \le C_{2}\sigma^{2}\left(\sqrt{nd_{k}}+n\right)\label{eq:EET-dI}\\
\|\bm{E}_{k}\widetilde{\bm{V}}_{k}^{\star}\| & \le C_{3}\sigma\sqrt{n}.\label{eq:EV}
\end{align}
\end{subequations} Moreover, this truncation is tight in the sense
and the expectation of the truncated part is small. We have that with
probability at least $1-O(N^{-11})$,
\begin{subequations}\label{eq:AAT_trunc_size}
\begin{align}
\mathbb{E}\left[\|\bm{E}_{k}\bm{E}_{k}^{\top}-d_{k}\sigma^{2}\bm{I}_{n}\|;\|\bm{E}_{k}\bm{E}_{k}^{\top}-d_{k}\sigma^{2}\bm{I}_{n}\|>C_{2}\sigma^{2}\left(\sqrt{nd_{k}}+n\right)\right] & \le C_{4}\sigma^{2}N^{-11}\label{eq:EET-dI_trunc}\\
\left\Vert \mathbb{E}\left[\bm{E}_{k}\widetilde{\bm{V}}_{k}^{\star};\|\bm{E}_{k}\widetilde{\bm{V}}_{k}^{\star}\|>C_{3}\sigma\sqrt{n}\right]\right\Vert  & =0\label{eq:EV_trunc}
\end{align}
\end{subequations}for some large enough constants $C_{4},C_{5}$.
Here $\mathbb{E}[A;B]$ denotes the conditional expectation of $A$ given $B$. Both of \eqref{eq:AAT_perturbation_size} and \eqref{eq:AAT_trunc_size} come from standard matrix concentration inequalities and
we prove them in Section~\ref{subsec:Proof_truncation}. Moreover,
Using \eqref{eq:AAT_perturbation_size} and the triangle inequality,
we have that
\begin{align}
\|\bm{\Delta}_{k}\| & \le2\left\Vert \bm{E}_{k}\widetilde{\bm{V}}_{k}^{\star}\widetilde{\bm{\Sigma}}_{k}^{\star}\widetilde{\bm{U}}_{k}^{\star\top}\right\Vert +\left\Vert \bm{E}_{k}\bm{E}_{k}^{\top}-d_{k}\sigma^{2}\bm{I}_{n}\right\Vert \label{eq:Deltak_bound}\\
 & \le2\left\Vert \bm{E}_{k}\widetilde{\bm{V}}_{k}^{\star}\right\Vert \left\Vert \widetilde{\bm{\Sigma}}_{k}^{\star}\right\Vert \left\Vert \widetilde{\bm{U}}_{k}^{\star\top}\right\Vert +\|\bm{E}_{k}\bm{E}_{k}^{\top}-d_{k}\sigma^{2}\bm{I}_{n}\|\nonumber \\
 & \le2C_{3}\sigma\cdot\sigma_{\max}\sqrt{n}+C_{2}\sigma^{2}\left(\sqrt{nd_{k}}+n\right)\nonumber 
\end{align}
Then Assumption~\eqref{eq:signal_size} implies that 
\begin{equation}
\|\bm{\Delta}_{k}\|\le\sigma_{\min}^{2}/8\le\sigma_{r+r_{k}}(\widetilde{\bm{U}}_{k}^{\star}\widetilde{\bm{\Sigma}}_{k}^{\star2}\widetilde{\bm{U}}_{k}^{\star\top})/8.\label{eq:Delta_sigma_bound}
\end{equation}

Recall that $\bm{P}_{k}^{-1}=\widetilde{\bm{U}}_{k}^{\star}\widetilde{\bm{\Sigma}}_{k}^{\star-2}\widetilde{\bm{U}}_{k}^{\star\top}$
and $\bm{P}_{k}^{\perp}=\bm{I}_{n}-\widetilde{\bm{U}}_{k}^{\star}\widetilde{\bm{U}}_{k}^{\star\top}$.
We can then invoke Theorem~1 of \cite{Xia2021} to see that
\begin{equation}
\widetilde{\bm{U}}_{k}\widetilde{\bm{U}}_{k}^{\top}=\widetilde{\bm{U}}_{k}^{\star}\widetilde{\bm{U}}_{k}^{\star\top}+\bm{P}_{k}^{\perp}\bm{\Delta}_{k}\bm{P}_{k}^{-1}+\bm{P}_{k}^{-1}\bm{\Delta}_{k}\bm{P}_{k}^{\perp}+\bm{R}_{k},\label{eq:Uk_tilde_expansion}
\end{equation}
where $\bm{R}_{k}$ is some $n\times n$ matrix such that 
\begin{align*}
\|\bm{R}_{k}\| & \le\sum_{i=2}^{\infty}\left(\frac{4\|\bm{\Delta}_{k}\|}{\sigma_{r+r_{k}}(\widetilde{\bm{U}}_{k}^{\star}\widetilde{\bm{\Sigma}}_{k}^{\star2}\widetilde{\bm{U}}_{k}^{\star\top})}\right)^{i}\le\frac{32\|\bm{\Delta}_{k}\|^{2}}{\sigma_{\min}^{4}}.
\end{align*}
Here the second inequality holds because of \eqref{eq:Delta_sigma_bound}.
Substituting \eqref{eq:Deltak_bound} in the above inequality, we
reach that 
\begin{align*}
\|\bm{R}_{k}\| & \le\frac{256C_{3}^{2}n\sigma^{2}\cdot\sigma_{\max}^{2}+4C_{2}^{2}\sigma^{4}(nd_{k}+n^{2})}{\sigma_{\min}^{4}}\\
 & \le C_{4}\left(\frac{\kappa^{2}\sigma^{2}n}{\sigma_{\min}^{2}}+\frac{\sigma^{4}(nd_{k}+n^{2})}{\sigma_{\min}^{4}}\right).
\end{align*}
for some large enough constant $C_{4}>0$. Here we use the inequality
$(a+b)^{2}\le2a^{2}+2b^{2}$.

Expand $\bm{P}_{k}^{\perp}\bm{\Delta}_{k}\bm{P}_{k}^{-1}$, we have
that 
\begin{align*}
\bm{P}_{k}^{\perp}\bm{\Delta}_{k}\bm{P}_{k}^{-1} & =\bm{P}_{k}^{\perp}\left(\bm{E}_{k}\widetilde{\bm{V}}_{k}^{\star}\widetilde{\bm{\Sigma}}_{k}^{\star}\widetilde{\bm{U}}_{k}^{\star\top}+\widetilde{\bm{U}}_{k}^{\star}\widetilde{\bm{\Sigma}}_{k}^{\star}\widetilde{\bm{V}}_{k}^{\star\top}\bm{E}_{k}^{\top}+\bm{E}_{k}\bm{E}_{k}^{\top}-d_{k}\sigma^{2}\bm{I}_{n}\right)\bm{P}_{k}^{-1}\\
 & =\bm{P}_{k}^{\perp}\bm{E}_{k}\widetilde{\bm{V}}_{k}^{\star}\widetilde{\bm{\Sigma}}_{k}^{\star-1}\widetilde{\bm{U}}_{k}^{\star\top}+\bm{P}_{k}^{\perp}\left(\bm{E}_{k}\bm{E}_{k}^{\top}-d_{k}\sigma^{2}\bm{I}_{n}\right)\bm{P}_{k}^{-1}.
\end{align*}
This equals $\bm{B}_{k}$ by definition. Plugging it back to \eqref{eq:Uk_tilde_expansion}
finishes the proof.

\subsubsection{Proof of \eqref{eq:AAT_perturbation_size} and \eqref{eq:AAT_trunc_size}\label{subsec:Proof_truncation}}

By (4.22) in \cite{vershynin2018high}, for any $t>0$, 
\begin{equation}
\mathbb{P}\left[\|\bm{E}_{k}\bm{E}_{k}^{\top}-d_{k}\sigma^{2}\bm{I}_{n}\|\ge C_{1}\sigma^{2}\left(\sqrt{nd_{k}}+t\sqrt{d_{k}}+n+t^{2}\right)\right]\le2\exp(-2t^{2}),\label{eq:EE-dI_t_bound}
\end{equation}
where $C_{1}>0$ is some large enough constant. 

Set $t=\sqrt{6\log N}$, as long as $n\ge C_{2}\log N$ for some large
enough constant $C_{2}>0$, we have that with probability at least
$1-2N^{-12}$, 
\begin{align*}
\|\bm{E}_{k}\bm{E}_{k}^{\top}-d_{k}\sigma^{2}\bm{I}_{n}\| & \le C_{1}\sigma^{2}\left(\sqrt{nd_{k}}+\sqrt{d_{k}\log N}+n+\log N\right)\\
 & \le2C_{1}\sigma^{2}\left(\sqrt{nd_{k}}+n\right).
\end{align*}
Within this proof, let $\alpha\coloneqq2C_{1}\sigma^{2}(\sqrt{nd_{k}}+n)$.
Using integration by parts, we have that 
\begin{align*}
 & \mathbb{E}\left[\|\bm{E}_{k}\bm{E}_{k}^{\top}-d_{k}\sigma^{2}\bm{I}_{n}\|;\|\bm{E}_{k}\bm{E}_{k}^{\top}-d_{k}\sigma^{2}\bm{I}_{n}\|>\alpha\right]\\
 & \quad=\alpha\mathbb{P}\left[\|\bm{E}_{k}\bm{E}_{k}^{\top}-d_{k}\sigma^{2}\bm{I}_{n}\|\ge\alpha\right]+\int_{\alpha}^{\infty}\mathbb{P}\left[\|\bm{E}_{k}\bm{E}_{k}^{\top}-d_{k}\sigma^{2}\bm{I}_{n}\|\ge x\right]\mathrm{d}x.
\end{align*}
For the first term, 
\begin{align*}
\alpha\mathbb{P}\left[\|\bm{E}_{k}\bm{E}_{k}^{\top}-d_{k}\sigma^{2}\bm{I}_{n}\|\ge\alpha\right] & \le2C_{1}\sigma^{2}(\sqrt{nd_{k}}+n)\cdot2N^{-12}\\
 & \le4C_{1}\sigma^{2}N^{-11}.
\end{align*}
For the second term, using the change of variable $x=C_{1}\sigma^{2}(\sqrt{nd_{k}}+t\sqrt{d_{k}}+n+t^{2})$
and \eqref{eq:EE-dI_t_bound}, we have that 
\begin{align*}
\int_{\alpha}^{\infty}\mathbb{P}\left[\|\bm{E}_{k}\bm{E}_{k}^{\top}-d_{k}\sigma^{2}\bm{I}_{n}\|\ge x\right]\mathrm{d}x & \le C_{1}\sigma^{2}\int_{\sqrt{6\log N}}^{\infty}\left(2t+\sqrt{d_{k}}\right)\cdot2\exp(-2t^{2})\mathrm{d}t\\
 & \le C_{3}\sigma^{2}\sqrt{d_{k}}\exp(-12\log N)=C_{2}\sigma^{2}N^{-11}
\end{align*}
for some large enough constant $C_{3}>0$. The last inequality here
follows from standard Gaussian tail bound. Combining the bounds for
these two terms completes the proof for \eqref{eq:EET-dI_trunc}.

Now consider $\bm{E}_{k}\widetilde{\bm{V}}_{k}^{\star}$ which is
a $n\times(r+r_{k})$ matrix, since $\widetilde{\bm{V}}_{k}^{\star\top}\widetilde{\bm{V}}_{k}^{\star}=\bm{I}_{r+r_{k}}$,
the rows $[\bm{E}_{k}\widetilde{\bm{V}}_{k}^{\star}]_{i,\cdot}=[\bm{E}_{k}]_{i,\cdot}\widetilde{\bm{V}}_{k}^{\star}$
are independent isotropic Gaussian vectors with covariance $\sigma^{2}\bm{I}_{n}$.
Then by Theorem~4.6.1 in \cite{vershynin2018high}, for any $t>0$,
\[
\mathbb{P}\left[\|\bm{E}_{k}\widetilde{\bm{V}}_{k}^{\star}\|>C_{4}\sigma\left(\sqrt{n}+\sqrt{r+r_{k}}+t\right)\right]\le2\exp(-t^{2})
\]
for some large enough constant $C_{4}>0$. Now let $t=\sqrt{6\log(n)}$,
we have that as long as $n\ge C_{2}\log N$ for some large enough
constant $C_{2}$, with probability at least $1-2n^{-12}$,
\[
\|\bm{E}_{k}\widetilde{\bm{V}}_{k}^{\star}\|\le4C_{4}\sigma\sqrt{n}.
\]
Finally, \eqref{eq:EV_trunc} is obtained by observing the symmetry
of Gaussian random variable.

\subsection{Proof of Lemma~\ref{lem:B_size}}

Recall that 
\begin{equation}
\bm{B}_{k}\coloneqq\underbrace{\bm{P}_{k}^{\perp}\bm{E}_{k}\widetilde{\bm{V}}_{k}^{\star}\widetilde{\bm{\Sigma}}_{k}^{\star-1}\widetilde{\bm{U}}_{k}^{\star\top}}_{\eqqcolon\bm{B}_{k,1}}+\underbrace{\bm{P}_{k}^{\perp}\left(\bm{E}_{k}\bm{E}_{k}^{\top}-d_{k}\sigma^{2}\bm{I}_{n}\right)\bm{P}_{k}^{-1}}_{\eqqcolon\bm{B}_{k,2}}.\label{eq:Bk_12_def}
\end{equation}
We define $\bm{B}_{k,1}$ and $\bm{B}_{k,2}$ as in \eqref{eq:Bk_12_def}.
In Section~\ref{subsec:proof_Bk1} and \ref{subsec:proof_Bk2}, We will show that with probability
at least $1-O(KN^{-10})$, 

\begin{subequations}
\begin{align}
\left\Vert \frac{1}{K}\sum_{k=1}^{K}\bm{B}_{k,1}\right\Vert  & \le C_{1}\cdot\frac{\sigma\sqrt{n}}{\sigma_{\min}}\left(\sqrt{\frac{\log N}{K}}+\frac{\log N}{K}\right)\label{eq:avg_Bk1_bound}\\
\left\Vert \frac{1}{K}\sum_{k=1}^{K}\bm{B}_{k,2}\right\Vert  & \le C_{2}\cdot\frac{\sigma^{2}}{\sigma_{\min}^{2}}\left(\sqrt{\frac{nd_{k}\log N}{K}}+\frac{\left(\sqrt{nd_{k}}+n\right)\log N}{K}\right)\label{eq:avg_Bk2_bound}
\end{align}
\end{subequations}Then the proof is completed by triangular inequality. 

\subsubsection{Proof of \eqref{eq:avg_Bk1_bound}\label{subsec:proof_Bk1}}

We will use the truncated matrix Bernstein inequality to control $\|K^{-1}\sum_{k=1}^{K}\bm{B}_{k,1}\|$.
We first show that the following inequalities. Recall $r_{\mathrm{avg}}=K^{-1}\sum_{k=1}^{K}r_{k}$.
For all $k\in[K]$, with probability at least $1-O(N^{-11})$, as
long as $n\ge C_{1}\log N$ for some constant $C_{1}>0$, 

\begin{subequations}
\begin{equation}
\|\bm{B}_{k,1}\|\le C_{2}\cdot\frac{\sigma\sqrt{n}}{\sigma_{\min}}\label{eq:ub_Bk1}
\end{equation}
and 
\begin{equation}
\left\Vert \mathbb{E}\left[\bm{B}_{k,1};\|\bm{B}_{k,1}\|>C_{2}\cdot\frac{\sigma\sqrt{n}}{\sigma_{\min}}\right]\right\Vert =0.\label{eq:trunc_Bk1}
\end{equation}
for some constant $C_{2}>0.$Moreover,
\begin{align}
\left\Vert \sum_{k=1}^{K}\mathbb{E}\bm{B}_{k,1}\bm{B}_{k,1}^{\top}\right\Vert  & \le\frac{K(r+r_{\mathrm{avg}})\sigma^{2}}{\sigma_{\min}^{2}}\label{eq:ub_Bk1Bk1T}\\
\left\Vert \sum_{k=1}^{K}\mathbb{E}\bm{B}_{k,1}^{\top}\bm{B}_{k,1}\right\Vert  & \le\frac{Kn\sigma^{2}}{\sigma_{\min}^{2}}.\label{eq:ub_Bk1TBk1}
\end{align}
 \end{subequations}The proof of these bounds is deferred to the end
of this section. Now invoking the truncated matrix Bernstein inequality
(see \cite{chen2021spectral}, Corollary~3.2), we have that with
probability at least $1-O(KN^{-10})$, 
\[
\left\Vert \frac{1}{K}\sum_{k=1}^{K}\bm{B}_{k}\right\Vert \le C_{3}\frac{\sigma\sqrt{n}}{\sigma_{\min}}\left(\sqrt{\frac{\log N}{K}}+\frac{\log N}{K}\right)
\]
for some large enough constant $C_{3}>0$.

\paragraph{Proof of \eqref{eq:ub_Bk1} and \eqref{eq:trunc_Bk1}.}

For each $k\in[K]$,
\begin{align*}
\|\bm{B}_{k,1}\| & =\left\Vert \bm{P}_{k}^{\perp}\bm{E}_{k}\widetilde{\bm{V}}_{k}^{\star}\widetilde{\bm{\Sigma}}_{k}^{\star-1}\widetilde{\bm{U}}_{k}^{\star\top}\right\Vert \le\left\Vert \bm{P}_{k}^{\perp}\right\Vert \left\Vert \bm{E}_{k}\widetilde{\bm{V}}_{k}^{\star}\right\Vert \left\Vert \widetilde{\bm{\Sigma}}_{k}^{\star-1}\right\Vert \left\Vert \widetilde{\bm{U}}_{k}^{\star}\right\Vert .
\end{align*}
Recall that $\bm{P}_{k}^{-1}=\widetilde{\bm{U}}_{k}^{\star}\widetilde{\bm{\Sigma}}_{k}^{\star-2}\widetilde{\bm{U}}_{k}^{\star\top}$
so and $\|\bm{P}_{k}^{\perp}\|\le1$. Combining this with \eqref{eq:AAT_perturbation_size},
we have that with probability at least $1-O(N^{-11})$, 
\begin{align*}
\|\bm{B}_{k,1}\| & \le C_{1}\cdot\frac{\sigma\sqrt{n}}{\sigma_{\min}},
\end{align*}
for some large enough constant $C_{1}>0$. For the truncated expectation,
similar to \eqref{eq:EV_trunc}, the symmetry of Gaussian random variable
implies that the expectation is $\bm{0}$.

\paragraph{Proof of \eqref{eq:ub_Bk1Bk1T}.}

For each $k\in[K]$, 
\begin{align*}
\mathbb{E}\left[\bm{B}_{k,1}\bm{B}_{k,1}^{\top}\right] & =\bm{P}_{k}^{\perp}\mathbb{E}\left[\bm{E}_{k}\widetilde{\bm{V}}_{k}^{\star}\widetilde{\bm{\Sigma}}_{k}^{\star-2}\widetilde{\bm{V}}_{k}^{\star\top}\bm{E}_{k}^{\top}\right]\bm{P}_{k}^{\perp}.\\
 & \overset{\text{(i)}}{=}\sigma^{2}\mathrm{Trace}\left(\widetilde{\bm{V}}_{k}^{\star}\widetilde{\bm{\Sigma}}_{k}^{\star-2}\widetilde{\bm{V}}_{k}^{\star\top}\right)\bm{P}_{k}^{\perp} \overset{\text{(ii)}}{=}\frac{(r+r_{k})\sigma^{2}}{\sigma_{\min}^{2}}\bm{P}_{k}^{\perp}.
\end{align*}
Here (i) follows from Lemma~\ref{lem:expectation_2} and (ii) follows
from the fact
\[
\mathrm{Trace}\left(\widetilde{\bm{V}}_{k}^{\star}\widetilde{\bm{\Sigma}}_{k}^{\star-2}\widetilde{\bm{V}}_{k}^{\star\top}\right)\le\mathrm{rank}\left(\widetilde{\bm{V}}_{k}^{\star}\widetilde{\bm{\Sigma}}_{k}^{\star-2}\widetilde{\bm{V}}_{k}^{\star\top}\right)\cdot\left\Vert \widetilde{\bm{V}}_{k}^{\star}\widetilde{\bm{\Sigma}}_{k}^{\star-2}\widetilde{\bm{V}}_{k}^{\star\top}\right\Vert \le\frac{(r+r_{k})}{\sigma_{\min}^{2}}.
\]
Then combining triangular inequality with the fact that $\|\bm{P}_{k}^{\perp}\|\le1$
finishes the proof.

\paragraph{Proof of \eqref{eq:ub_Bk1TBk1}. }

Expanding $\bm{B}_{k}^{\top}\bm{B}_{k}$, we have that 
\begin{align*}
\mathbb{E}\bm{B}_{k,1}^{\top}\bm{B}_{k,1} & =\widetilde{\bm{U}}_{k}^{\star}\widetilde{\bm{\Sigma}}_{k}^{\star-1}\widetilde{\bm{V}}_{k}^{\star\top}\mathbb{E}\left[\bm{E}_{k}^{\top}\bm{P}_{k}^{\perp}\bm{E}_{k}\right]\widetilde{\bm{V}}_{k}^{\star}\widetilde{\bm{\Sigma}}_{k}^{\star-1}\widetilde{\bm{U}}_{k}^{\star\top}\\
 & \overset{\text{(i)}}{=}\sigma^{2}\mathrm{Trace}\left(\bm{P}_{k}^{\perp}\right)\widetilde{\bm{U}}_{k}^{\star}\widetilde{\bm{\Sigma}}_{k}^{\star-1}\widetilde{\bm{V}}_{k}^{\star\top}\widetilde{\bm{V}}_{k}^{\star}\widetilde{\bm{\Sigma}}_{k}^{\star-1}\widetilde{\bm{U}}_{k}^{\star\top}\\
 & =\sigma^{2}(n-r-r_{k})\widetilde{\bm{U}}_{k}^{\star}\widetilde{\bm{\Sigma}}_{k}^{\star-2}\widetilde{\bm{U}}_{k}^{\star\top}.
\end{align*}
Here (i) follows from Lemma~\ref{lem:expectation_2}. Then by triangular
inequality

\begin{align*}
\left\Vert \sum_{k=1}^{K}\mathbb{E}\bm{B}_{k,1}^{\top}\bm{B}_{k,1}\right\Vert  & \le\sum_{k=1}^{K}\left\Vert \mathbb{E}\bm{B}_{k,1}^{\top}\bm{B}_{k,1}\right\Vert \le K\cdot n\sigma^{2}\cdot\|\widetilde{\bm{\Sigma}}_{k}^{\star-2}\|=\frac{Kn\sigma^{2}}{\sigma_{\min}^{2}}.
\end{align*}

\subsubsection{Proof of \eqref{eq:avg_Bk2_bound}\label{subsec:proof_Bk2}}

We will use the truncated matrix Bernstein inequality to control $\|K^{-1}\sum_{k=1}^{K}\bm{B}_{k,2}\|$.
We first show the following inequalities. For each $k\in[K]$,
with probability at least $1-O(N^{-11})$, as long as $n\ge C_{1}\log N$
for some constant $C_{12}>0$, 

\begin{subequations}
\begin{equation}
\|\bm{B}_{k,2}\|\le\frac{C_{2}\sigma^{2}\left(\sqrt{nd}+n\right)}{\sigma_{\min}^{2}}\label{eq:ub_Bk2}
\end{equation}
and 
\begin{equation}
\left\Vert \mathbb{E}\left[\bm{B}_{k,2};\|\bm{B}_{k,2}\|>\frac{C_{2}\sigma^{2}\left(\sqrt{nd}+n\right)}{\sigma_{\min}^{2}}\right]\right\Vert \le\frac{C_{2}\sigma^{2}}{\sigma_{\min}^{2}}N^{-11}\label{eq:trunc_Bk2}
\end{equation}
for some constant $C_{2}>0$. Moreover,
\begin{align}
\left\Vert \sum_{k=1}^{K}\mathbb{E}\bm{B}_{k,2}\bm{B}_{k,2}^{\top}\right\Vert  & \le\frac{K\sigma^{4}d (r+r_{\mathrm{avg}})}{\sigma_{\min}^{4}}\label{eq:ub_Bk2Bk2T}\\
\left\Vert \sum_{k=1}^{K}\mathbb{E}\bm{B}_{k,2}^{\top}\bm{B}_{k,2}\right\Vert  & \le\frac{K\sigma^{4}d n}{\sigma_{\min}^{4}}.\label{eq:ub_Bk2TBk2}
\end{align}
\end{subequations}The proof of these bounds is deferred to the end
of this section. Now invoking the truncated matrix Bernstein inequality
(see \cite{chen2021spectral}, Corollary~3.2), with probability at
least $1-O(KN^{-10})$, 
\begin{align*}
\left\Vert \frac{1}{K}\sum_{k=1}^{K}\bm{B}_{k}\right\Vert  & \le C_{3}\frac{\sigma^{2}}{\sigma_{\min}^{2}}\left(\sqrt{\frac{nd\log N}{K}}+\frac{\left(\sqrt{nd}+n\right)\log N}{K}\right)
\end{align*}
for some large enough constant $C_{3}>0$.

\paragraph{Proof of \eqref{eq:ub_Bk2} and \eqref{eq:trunc_Bk2}.}

For each $k\in[K]$,
\begin{align*}
\|\bm{B}_{k,2}\| & \le\left\Vert \bm{P}_{k}^{\perp}\left(\bm{E}_{k}\bm{E}_{k}^{\top}-d_{k}\sigma^{2}\bm{I}_{n}\right)\bm{P}_{k}^{-1}\right\Vert \le\left\Vert \bm{P}_{k}^{\perp}\right\Vert \left\Vert \bm{E}_{k}\bm{E}_{k}^{\top}-d_{k}\sigma^{2}\bm{I}_{n}\right\Vert \left\Vert \bm{P}_{k}^{-1}\right\Vert .
\end{align*}
Recall that $\bm{P}_{k}^{-1}=\widetilde{\bm{U}}_{k}^{\star}\widetilde{\bm{\Sigma}}_{k}^{\star-2}\widetilde{\bm{U}}_{k}^{\star\top}$
and $\bm{P}_{k}^{\perp}=\bm{I}_{n}-\widetilde{\bm{U}}_{k}^{\star}\widetilde{\bm{U}}_{k}^{\star\top}$.
Then $\|\bm{P}_{k}^{-1}\|\le1/\sigma_{\min}^{2}$ and $\|\bm{P}_{k}^{\perp}\|\le1$.
Combining this with \eqref{eq:AAT_perturbation_size}, we have that
with probability at least $1-O(N^{-11})$, 
\begin{align*}
\|\bm{B}_{k,2}\| & \le C_{1}\frac{\sigma^{2}\left(\sqrt{nd_{k}}+n\right)}{\sigma_{\min}^{2}},
\end{align*}
for some large enough constant $C_{1}>0$. For the truncated expectation,
by \eqref{eq:EV_trunc}, for some large enough constant $C_{2}>0$,
\begin{align*}
&\mathbb{E}\left[\|\bm{B}_{k,2}\|;\|\bm{B}_{k,2}\|>\frac{C_{1}\sigma^{2}\left(\sqrt{nd_{k}}+n\right)}{\sigma_{\min}^{2}} \right] \\
&\quad \le\left\Vert \bm{P}_{k}^{-1}\right\Vert \cdot\mathbb{E}\left[\|\bm{E}_{k}\bm{E}_{k}^{\top}-d_{k}\sigma^{2}\bm{I}_{n}\|;\|\bm{E}_{k}\bm{E}_{k}^{\top}-d_{k}\sigma^{2}\bm{I}_{n}\|>C_{1}\cdot\sigma\sqrt{n} \right]
 \le\frac{C_{2}\sigma^{2}}{\sigma_{\min}^{2}}N^{-11}.
\end{align*}
Union bound on $k\in[K]$ completes the proof.

\paragraph{Proof of \eqref{eq:ub_Bk2Bk2T}.}

For each $k\in[K]$, 
\[
\mathbb{E}\left[\bm{B}_{k,2}\bm{B}_{k,2}^{\top}\right]=\bm{P}_{k}^{\perp}\mathbb{E}\left[\left(\bm{E}_{k}\bm{E}_{k}^{\top}-d_{k}\sigma^{2}\bm{I}_{n}\right)\bm{P}_{k}^{-1}\bm{P}_{k}^{-1}\left(\bm{E}_{k}\bm{E}_{k}^{\top}-d_{k}\sigma^{2}\bm{I}_{n}\right)\right]\bm{P}_{k}^{\perp}.
\]
By Lemma~\ref{lem:expectation_2}, $\mathbb{E}\bm{E}_{k}\bm{E}_{k}^{\top}=d_{k}\sigma^{2}\bm{I}_{n}$,
therefore we can simplify the equation above to 
\begin{align}
\mathbb{E}\left[\bm{B}_{k,2}\bm{B}_{k,2}^{\top}\right] & =\bm{P}_{k}^{\perp}\mathbb{E}\left[\bm{E}_{k}\bm{E}_{k}^{\top}\bm{P}_{k}^{-1}\bm{P}_{k}^{-1}\bm{E}_{k}\bm{E}_{k}^{\top}\right]\bm{P}_{k}^{\perp}-\sigma^{4}d_{k}^{2}\bm{P}_{k}^{\perp}\bm{P}_{k}^{-1}\bm{P}_{k}^{-1}\bm{P}_{k}^{\perp}\label{eq:EB2B2T}\\
 & =\sigma^{4}\mathrm{Trace}(\bm{I}_{d_{k}})\mathrm{Trace}(\bm{P}_{k}^{-1}\bm{P}_{k}^{-1})\bm{P}_{k}^{\perp}.\nonumber 
\end{align}
Here the second equality follows from Lemma~\ref{lem:expectation_4}
and the fact that $\bm{P}_{k}^{-1}\bm{P}_{k}^{\perp}=\bm{0}$. Then
\begin{align*}
\left\Vert \mathbb{E}\left[\bm{B}_{k,2}\bm{B}_{k,2}^{\top}\right]\right\Vert  & \le\sigma^{4}d_{k}(r+r_{k})\left\Vert \bm{P}_{k}^{-1}\bm{P}_{k}^{-1}\right\Vert =\sigma^{4}d_{k}(r+r_{k})\left\Vert \widetilde{\bm{U}}_{k}^{\star}\widetilde{\bm{\Sigma}}_{k}^{\star-4}\widetilde{\bm{U}}_{k}^{\star\top}\right\Vert 
 \le\frac{\sigma^{4}d_{k}(r+r_{k})}{\sigma_{\min}^{4}}.
\end{align*}
Summing it up over $k\in[K]$ with triangular inequality completes
the proof.

\paragraph{Proof of \eqref{eq:ub_Bk2TBk2}.}
Expanding $\bm{B}_{k}^{\top}\bm{B}_{k}$, we have that 
\begin{align*}
\mathbb{E}\bm{B}_{k,2}^{\top}\bm{B}_{k,2} & =\bm{P}_{k}^{-1}\mathbb{E}\left[\left(\bm{E}_{k}\bm{E}_{k}^{\top}-d_{k}\sigma^{2}\bm{I}_{n}\right)\bm{P}_{k}^{\perp}\left(\bm{E}_{k}\bm{E}_{k}^{\top}-d_{k}\sigma^{2}\bm{I}_{n}\right)\right]\bm{P}_{k}^{-1}\\
 & \overset{\text{(i)}}{=}\bm{P}_{k}^{-1}\mathbb{E}\left[\bm{E}_{k}\bm{E}_{k}^{\top}\bm{P}_{k}^{\perp}\bm{E}_{k}\bm{E}_{k}^{\top}\right]\bm{P}_{k}^{-1}\\
 & \overset{\text{(ii)}}{=}\sigma^{4}\mathrm{Trace}(\bm{I}_{d_{k}})\cdot\mathrm{Trace}(\bm{P}_{k}^{\perp})\bm{P}_{k}^{-1}\bm{P}_{k}^{-1}.
\end{align*}
Here (i) uses the same simplification as \eqref{eq:EB2B2T} and (ii)
follows from Lemma~\ref{lem:expectation_4}. Then 
\[
\left\Vert \mathbb{E}\left[\bm{B}_{k,2}\bm{B}_{k,2}^{\top}\right]\right\Vert \le\frac{\sigma^{4}d_{k}n}{\sigma_{\min}^{4}}.
\]
Summing it up over $k\in[K]$ with triangular inequality completes
the proof.

\subsection{Proof of Lemma~\ref{lem:approx_UUT}}

Suppose Lemma~\ref{lem:expansion_U} and \ref{lem:B_size} hold.
Consider $\bm{I}_{n}-\widehat{\bm{U}}\widehat{\bm{U}}^{\top}$. It
is the top-$(n-r)$ eigenspace of $\bm{I}_{n}-K^{-1}\sum_{k=1}^{K}\widetilde{\bm{U}}_{k}\widetilde{\bm{U}}_{k}^{\top}$.
Using the decomposition in Lemma~\ref{lem:expansion_U}, we have
that 
\begin{align*}
\bm{I}_{n}-\frac{1}{K}\sum_{k=1}^{K}\widetilde{\bm{U}}_{k}\widetilde{\bm{U}}_{k}^{\top} & =\bm{I}_{n}-\frac{1}{K}\sum_{k=1}^{K}\widetilde{\bm{U}}_{k}^{\star}\widetilde{\bm{U}}_{k}^{\star\top}-\frac{1}{K}\sum_{k=1}^{K}\left(\bm{B}_{k}+\bm{B}_{k}^{\top}+\bm{R}_{k}\right)\\
 & =\bm{I}_{n}-\frac{1}{K}\sum_{k=1}^{K}\widetilde{\bm{U}}_{k}^{\star}\widetilde{\bm{U}}_{k}^{\star\top}-\bm{\Delta}.
\end{align*}
Then \eqref{eq:Delta_bound} imples $\|\bm{\Delta}\|\le\theta/8.$
Now consider 
\[
\bm{I}_{n}-\frac{1}{K}\sum_{k=1}^{K}\widetilde{\bm{U}}_{k}^{\star}\widetilde{\bm{U}}_{k}^{\star\top}=\bm{I}_{n}-\bm{U}^{\star}\bm{U}^{\star\top}-\frac{1}{K}\sum_{k=1}^{K}\bm{U}_{k}^{\star}\bm{U}_{k}^{\star\top}
\]
as the ground truth and $\bm{\Delta}$ as the perturbation. Since
\[
\left\Vert \frac{1}{K}\sum_{k=1}^{K}\bm{U}_{k}^{\star}\bm{U}_{k}^{\star\top}\right\Vert =1-\theta,
\]
the rank of $\bm{I}_{n}-\frac{1}{K}\sum_{k=1}^{K}\widetilde{\bm{U}}_{k}^{\star}\widetilde{\bm{U}}_{k}^{\star\top}$
is $n-r$ and its top-$(n-r)$ subspace is $\bm{I}_{n}-\bm{U}^{\star}\bm{U}^{\star\top}$.
This and the fact that $\theta/8$ allow us to invoke Theorem~1 in
\cite{Xia2021}, and arrive at 
\[
\bm{I}_{n}-\widehat{\bm{U}}\widehat{\bm{U}}^{\top}=\bm{I}_{n}-\bm{U}^{\star}\bm{U}^{\star\top}-\bm{P}^{-1}\bm{\Delta}\bm{P}^{\perp}-\bm{P}^{\perp}\bm{\Delta}\bm{P}^{-1}-\bm{R}
\]
for some $\bm{R}\in\mathbb{R}^{n\times n}$ such that $\|\bm{R}\|\le32\|\bm{\Delta}\|^{2}/\theta^{2}.$
Then 
\[
\widehat{\bm{U}}\widehat{\bm{U}}^{\top}=\bm{U}^{\star}\bm{U}^{\star\top}+\bm{P}^{-1}\bm{\Delta}\bm{P}^{\perp}+\bm{P}^{\perp}\bm{\Delta}\bm{P}^{-1}+\bm{R}.
\]

\subsection{Proof of Lemma~\ref{lem:size_PDeltaP}}

Recall that 
\begin{align*}
\bm{\Delta} & =\frac{1}{K}\sum_{k=1}^{K}\left(\bm{B}_{k}+\bm{B}_{k}^{\top}+\bm{R}_{k}\right),
\end{align*}
where 
\begin{align*}
\bm{B}_{k} & =\bm{P}_{k}^{\perp}\bm{E}_{k}\widetilde{\bm{V}}_{k}^{\star}\widetilde{\bm{\Sigma}}_{k}^{\star-1}\widetilde{\bm{U}}_{k}^{\star\top}+\bm{P}_{k}^{\perp}\left(\bm{E}_{k}\bm{E}_{k}^{\top}-d_{k}\sigma^{2}\bm{I}_{n}\right)\bm{P}_{k}^{-1}\\
 & =\bm{B}_{k,1}+\bm{B}_{k,2}
\end{align*}
with $\bm{P}_{k}^{\perp}=\bm{I}_{n}-\widetilde{\bm{U}}_{k}^{\star}\widetilde{\bm{U}}_{k}^{\star\top}$
and $\bm{P}_{k}^{-1}=\widetilde{\bm{U}}_{k}^{\star\top}\widetilde{\bm{\Sigma}}_{k}^{\star-2}\widetilde{\bm{U}}_{k}^{\star\top}$.
Moreover,
\[
\bm{P}=\sum_{i=1}^{n-r}\mu_{i}\bm{x}_{i}\bm{x}_{i}^{\top},\qquad\bm{P}^{-1}=\sum_{i=1}^{n-r}\mu_{i}^{-1}\bm{x}_{i}\bm{x}_{i}^{\top},\qquad\text{and}\qquad\bm{P}^{\perp}=\bm{U}^{\star}\bm{U}^{\star\top}.
\]
Since 
\[
\bm{P}^{\perp}\bm{P}_{k}^{\perp}=\bm{U}^{\star}\bm{U}^{\star\top}(\bm{I}_{n}-\widetilde{\bm{U}}_{k}^{\star}\widetilde{\bm{U}}_{k}^{\star\top})=\bm{U}^{\star}\bm{U}^{\star\top}(\bm{I}_{n}-\bm{U}^{\star}\bm{U}^{\star\top}-\bm{U}_{k}^{\star}\bm{U}_{k}^{\star\top})=\bm{0},
\]
we have that for all $k\in[K]$, $\bm{P}^{-1}\bm{B}_{k}^{\top}\bm{P}^{\perp}=\bm{0}$.
Then it suffices to control $K^{-1}\sum_{k=1}^{K}\bm{P}^{-1}\bm{B}_{k}\bm{P}^{\perp}$
and $K^{-1}\sum_{k=1}^{K}\bm{P}^{-1}\bm{R}_{k}\bm{P}^{\perp}$. In
the rest of this proof, we will show that with probability at least
$1-O(KN^{-10})$, the following inequalities hold:

\begin{subequations}
\begin{align}
\left\Vert \frac{1}{K}\sum_{k=1}^{K}\bm{P}^{-1}\bm{R}_{k}\bm{P}^{\perp}\right\Vert  & \le\frac{C_{1}}{\theta}\left(\frac{\kappa^{2}\sigma^{2}n}{\sigma_{\min}^{2}}+\frac{\sigma^{4}(nd_{k}+n^{2})}{\sigma_{\min}^{4}}\right)\label{eq:PRP_bound}\\
\left\Vert \frac{1}{K}\sum_{k=1}^{K}\bm{P}^{-1}\bm{B}_{k,1}\bm{P}^{\perp}\right\Vert  & \le C_{2}\frac{\sigma}{\sigma_{\min}}\left(\sqrt{\frac{n}{K}+\frac{r+r_{\mathrm{avg}}}{K\theta}+\frac{r\cdot r_{\mathrm{avg}}}{K\theta}\wedge\frac{r}{K^{2}\theta^{2}}}\right)\log^{5/2}N\label{eq:PBP_bound}\\
\left\Vert \frac{1}{K}\sum_{k=1}^{K}\bm{P}^{-1}\bm{B}_{k,2}\bm{P}^{\perp}\right\Vert  & \le C_{3}\frac{\sigma^{2}}{\sigma_{\min}^{2}}\sqrt{\frac{nd}{K}+\frac{d(r+r_{\mathrm{avg}})}{K\theta}+\left(nd+n^{2}\right)\left(\frac{3r_{\mathrm{avg}}}{K\theta}\wedge\frac{1}{K^{2}\theta^{2}}\right)}\log N.\label{eq:PB2P_bound}
\end{align}
\end{subequations}for some constants $C_{1},C_{2},C_{3}>0$. These
combined imply the result of Lemma~\eqref{lem:size_PDeltaP}.

\paragraph{Proof of \eqref{eq:PRP_bound}.}

Recall that Lemma~\ref{lem:expansion_U} says that with probability
at least $1-O(KN^{-10})$, for any $k\in[K]$,
\[
\|\bm{R}_{k}\|\le C_{1}\left(\frac{\kappa^{2}\sigma^{2}n}{\sigma_{\min}^{2}}+\frac{\sigma^{4}(nd_{k}+n^{2})}{\sigma_{\min}^{4}}\right)
\]
for some constant $C_{1}>0$. Then since $\mu_{i}\ge\theta$ and $\bm{P}^{\perp}$
is a projection,
\begin{align*}
\left\Vert \frac{1}{K}\sum_{k=1}^{K}\bm{P}^{-1}\bm{R}_{k}\bm{P}^{\perp}\right\Vert  & \le\frac{1}{K}\sum_{k=1}^{K}\|\bm{P}^{-1}\|\cdot\|\bm{R}_{k}\|\cdot\|\bm{P}^{\perp}\|\\
 & \le\frac{C_{1}}{\theta}\left(\frac{\kappa^{2}\sigma^{2}n}{\sigma_{\min}^{2}}+\frac{\sigma^{4}(nd_{k}+n^{2})}{\sigma_{\min}^{4}}\right).
\end{align*}

\paragraph{Proof of \eqref{eq:PBP_bound}.}

Within the scope of this proof, for any $k\in[K]$, let 
\begin{align*}
\bm{Y}_{k,1} & \coloneqq\bm{P}^{-1}\bm{B}_{k,1}\bm{P}^{\perp}=\bm{P}^{-1}\bm{P}_{k}^{\perp}\bm{E}_{k}\widetilde{\bm{V}}_{k}^{\star}\widetilde{\bm{\Sigma}}^{\star-1}\widetilde{\bm{U}}_{k}^{\star\top}\bm{P}^{\perp}.
\end{align*}
It is clear that $\bm{Y}_{k,1}$ is zero-mean. We will prove \eqref{eq:PBP_bound}
with truncated matrix Bernstein inequality. We first claim the following
inequalities hold. For some constant $C_{1}>0$, with probability
at least for each $k\in[K]$, with probability at least $1-O(N^{-11})$,

\begin{subequations}
\begin{equation}
\|\bm{Y}_{k,1}\|\le C_{1}\frac{\sigma}{\sigma_{\min}}\left(\sqrt{\left(n+\frac{Kr_{\mathrm{avg}}}{\theta}+\frac{Kr\cdot r_{\mathrm{avg}}}{\theta}\wedge\frac{r}{\theta^{2}}\right)\log^{3}N}\right)\label{eq:ub_Y}
\end{equation}
and 
\begin{equation}
\left\Vert \mathbb{E}\left[\bm{Y}_{k,1};\|\bm{Y}_{k,1}\|>C_{1}\frac{\sigma}{\sigma_{\min}}\left(\sqrt{\left(n+\frac{Kr_{\mathrm{avg}}}{\theta}+\frac{Kr\cdot r_{\mathrm{avg}}}{\theta}\wedge\frac{r}{\theta^{2}}\right)\log^{3}N}\right)\right]\right\Vert =0.\label{eq:trunc_Y1}
\end{equation}
Moreover,
\begin{align}
\left\Vert \mathbb{E}\sum_{k=1}^{K}\bm{Y}_{k,1}\bm{Y}_{k,1}^{\top}\right\Vert  & \le\frac{rK\sigma^{2}}{\theta\sigma_{\min}^{2}}\label{eq:ub_YY^T}\\
\left\Vert \mathbb{E}\sum_{k=1}^{K}\bm{Y}_{k,1}^{\top}\bm{Y}_{k,1}\right\Vert  & \le2\left(n+\frac{r_{\mathrm{avg}}}{\theta}\right)\frac{K\sigma^{2}}{\sigma_{\min}^{2}}.\label{eq:ub_Y^TY}
\end{align}

\end{subequations} The identity \eqref{eq:trunc_Y1} holds since
Gaussian random variable is symmetric and the whole expectation is
$\bm{0}$. We defer the proof of \eqref{eq:ub_Y}, \eqref{eq:ub_YY^T}
and \eqref{eq:ub_Y^TY} to Section~\ref{subsec:Proof_ub_Y}, \ref{subsec:Proof_YYT},
and \ref{subsec:Proof_YTY}, respectively. Now invoking the truncated
matrix Bernstein inequality (see \cite{chen2021spectral}, Corollary~3.2),
we have that with probability at least $1-O(KN^{-10})$,
\begin{align*}
\left\Vert \frac{1}{K}\sum_{k=1}^{K}\bm{P}^{-1}\bm{B}_{k,1}\bm{P}^{\perp}\right\Vert =\left\Vert \frac{1}{K}\sum_{k=1}^{K}\bm{Y}_{k,1}\right\Vert  & \le C_{4}\frac{\sigma}{\sigma_{\min}}\left(\sqrt{\frac{n}{K}+\frac{r+r_{\mathrm{avg}}}{K\theta}+\frac{r\cdot r_{\mathrm{avg}}}{K\theta}\wedge\frac{r}{K^{2}\theta^{2}}}\right)\log^{5/2}N
\end{align*}
for some constants $C_{4}>0$.

\paragraph{Proof of \eqref{eq:PB2P_bound}.}

Within the scope of this proof, for any $k\in[K]$, let 
\begin{align*}
\bm{Y}_{k,2} & \coloneqq\bm{P}^{-1}\bm{B}_{k,2}\bm{P}^{\perp}=\bm{P}^{-1}\bm{P}_{k}^{\perp}\left(\bm{E}_{k}\bm{E}_{k}^{\top}-d_{k}\sigma^{2}\bm{I}_{n}\right)\bm{P}_{k}^{-1}\bm{P}^{\perp}.
\end{align*}
By Lemma~\ref{lem:expectation_2}, $\bm{E}_{k}\bm{E}_{k}^{\top}-d_{k}\sigma^{2}\bm{I}_{n}$
is zero-mean. Then $\bm{Y}_{k,2}$ is zero-mean as well. We will prove
\eqref{eq:PBP_bound} with truncated matrix Bernstein inequality.
We first claim the following inequalities hold. For some constant
$C_{1}>0$, with probability at least for each $k\in[K]$, with probability
at least $1-O(N^{-11})$,

\begin{subequations}
\begin{equation}
\|\bm{Y}_{k,2}\|\le\frac{C_{1}\sigma^{2}\left(\sqrt{nd_{k}}+n\right)}{\sigma_{\min}^{2}}\sqrt{\frac{3Kr_{\mathrm{avg}}}{\theta}\wedge\frac{1}{\theta^{2}}}\label{eq:ub_Y-2}
\end{equation}
and 
\begin{equation}
\left\Vert \mathbb{E}\left[\bm{Y}_{k,2};\|\bm{Y}_{k,2}\|>\frac{C_{1}\sigma^{2}\left(\sqrt{nd_{k}}+n\right)}{\sigma_{\min}^{2}}\sqrt{\frac{3Kr_{\mathrm{avg}}}{\theta}\wedge\frac{1}{\theta^{2}}}\right]\right\Vert \le C_{4}\frac{\sigma^{2}}{\sigma_{\min}^{2}}N^{-11}\sqrt{\frac{3Kr_{\mathrm{avg}}}{\theta}\wedge\frac{1}{\theta^{2}}}.\label{eq:trunc_Y2}
\end{equation}
Moreover,
\begin{align}
\left\Vert \mathbb{E}\sum_{k=1}^{K}\bm{Y}_{k,2}\bm{Y}_{k,2}^{\top}\right\Vert  & \le\frac{Krd\sigma^{4}}{\theta\sigma_{\min}^{4}}\label{eq:ub_YY^T-2}\\
\left\Vert \mathbb{E}\sum_{k=1}^{K}\bm{Y}_{k,2}^{\top}\bm{Y}_{k,2}\right\Vert  & \le\frac{2\sigma^{4}}{\sigma_{\min}^{4}}Kd\left(n+\frac{r_{\mathrm{avg}}}{\theta}\right).\label{eq:ub_Y^TY-2}
\end{align}

\end{subequations} The identity \eqref{eq:trunc_Y2} holds since
Gaussian random variable is symmetric and the whole expectation is
$\bm{0}$. We defer the proof of \eqref{eq:ub_Y-2}, \eqref{eq:ub_YY^T-2}
and \eqref{eq:ub_Y^TY-2} to Section~\ref{subsec:Proof_ub_Y}, \ref{subsec:Proof_YYT},
and \ref{subsec:Proof_YTY}, respectively. Now invoking the truncated
matrix Bernstein inequality (see \cite{chen2021spectral}, Corollary~3.2),
we have that with probability at least $1-O(KN^{-10})$,
\begin{align*}
\left\Vert \frac{1}{K}\sum_{k=1}^{K}\bm{P}^{-1}\bm{B}_{k,2}\bm{P}^{\perp}\right\Vert =\left\Vert \frac{1}{K}\sum_{k=1}^{K}\bm{Y}_{k,2}\right\Vert  & \le C_{4}\frac{\sigma^{2}}{\sigma_{\min}^{2}}\sqrt{\frac{nd}{K}+\frac{d(r+r_{\mathrm{avg}})}{K\theta}+\left(nd+n^{2}\right)\left(\frac{3r_{\mathrm{avg}}}{K\theta}\wedge\frac{1}{K^{2}\theta^{2}}\right)}\log N.
\end{align*}
for some constants $C_{4}>0$. 

Before we proceed with the proof of the these conditions, we present
several useful lemmas. The first one controls the number of $\mu_{i}$
that is small and gives a upper bound on $\sum_{i=1}^{n-r}\mu_{i}^{-1}$.
The proof is deferred to Section~\ref{subsec:Proo_reciprocal}.

\begin{lemma}\label{lem:sum_mu_reciprocal} Let $\{\mu_{i}\}_{i\in[n-r]}$
be defined as in \eqref{eq:mu_def}. Then \begin{subequations}
\begin{equation}
\left|\{i:\mu_{i}\le1/2\}\right|\le2r_{\mathrm{avg}}\label{eq:mu_less_2}
\end{equation}
 and
\begin{align}
\sum_{i=1}^{n-r}\mu_{i} & =n-r-r_{\mathrm{avg}}\label{eq:mu_sum}\\
\sum_{i=1}^{n-r}\mu_{i}^{-1} & \le2n+\frac{2r_{\mathrm{avg}}}{\theta}.\label{eq:mu_reciprocal_sum}
\end{align}
\end{subequations}

\end{lemma}

The second lemma concerns with the result on alignment of eigenvectors
of $\bm{P}$ and $\bm{P}_{k}^{\perp}$. We defer the proof to Section~\ref{subsec:Proof_P_proj}.

\begin{lemma}\label{lem:P_proj_norm}Recall $\bm{x}_{i}$ is defined
by the eigen-decomposition $\bm{P}=\sum_{i=1}^{n-r}\mu_{i}\bm{x}_{i}\bm{x}_{i}^{\top}.$
We claim that for every $k\in[K]$ and $i\in[n-r]$,

\begin{subequations}
\begin{equation}
\bm{x}_{i}^{\top}\bm{P}_{k}^{\perp}\bm{x}_{i}\le(K\mu_{i})\wedge1,\label{eq:x^T_Pk_x}
\end{equation}
and 
\begin{equation}
\left\Vert \bm{P}^{-1}\bm{P}_{k}^{\perp}\right\Vert ^{2}\le\frac{3Kr_{\mathrm{avg}}}{\theta}\wedge\frac{1}{\theta^{2}}.\label{eq:P-1_proj_norm}
\end{equation}

\end{subequations}\end{lemma}

\subsubsection{Proof of \eqref{eq:ub_Y}\label{subsec:Proof_ub_Y}}

Fix $k\in[K]$. Recall that 
\begin{align*}
\bm{Y}_{k,1} & =\bm{P}^{-1}\bm{B}_{k,1}\bm{P}^{\perp}=\bm{P}^{-1}\bm{P}_{k}^{\perp}\bm{E}_{k}\widetilde{\bm{V}}_{k}^{\star}\widetilde{\bm{\Sigma}}_{k}^{\star-1}\widetilde{\bm{U}}_{k}^{\star\top}\bm{U}^{\star}\bm{U}^{\star\top}.\\
 & =\sum_{ij}\underbrace{[\bm{E}_{k}]_{ij}\bm{P}^{-1}\bm{P}_{k}^{\perp}\bm{e}_{i}\bm{e}_{j}^{\top}\widetilde{\bm{V}}_{k}^{\star}\widetilde{\bm{\Sigma}}_{k}^{\star-1}\widetilde{\bm{U}}_{k}^{\star\top}\bm{U}^{\star}\bm{U}^{\star\top}}_{\eqqcolon\bm{S}_{ij}}.
\end{align*}

We will bound $\|\bm{Y}_{k}\|$ with truncated matrix Bernstein inequality.
We first claim that for each $(i,j)\in[n]\times[d_{k}]$, with probability
at least $1-O(N^{-14})$, \begin{subequations}\label{eq:S_bernstein_cond}
\begin{equation}
\|\bm{S}_{ij}\|\le\frac{C_{1}\sigma\sqrt{\log N}}{\sigma_{\min}}\sqrt{\frac{3Kr_{\mathrm{avg}}}{\theta}\wedge\frac{1}{\theta^{2}}}\label{eq:S_norm}
\end{equation}
and
\begin{equation}
\left\Vert \mathbb{E}\left[\bm{S}_{ij};\|\bm{S}_{ij}\|>\frac{C_{1}\sigma\sqrt{\log N}}{\sigma_{\min}}\sqrt{\frac{3Kr_{\mathrm{avg}}}{\theta}\wedge\frac{1}{\theta^{2}}}\right]\right\Vert =0.\label{eq:S_trunc}
\end{equation}
Moreover,
\begin{align}
\left\Vert \mathbb{E}\sum_{ij}\bm{S}_{ij}\bm{S}_{ij}^{\top}\right\Vert  & \le\frac{r\sigma^{2}}{\sigma_{\min}^{2}}\cdot\left(\frac{3Kr_{\mathrm{avg}}}{\theta}\wedge\frac{1}{\theta^{2}}\right)\label{eq:SS^T}\\
\left\Vert \mathbb{E}\sum_{ij}\bm{S}_{ij}^{\top}\bm{S}_{ij}\right\Vert  & \le\frac{\sigma^{2}}{\sigma_{\min}^{2}}\left(4n+\frac{2Kr_{\mathrm{avg}}}{\theta}\right).\label{eq:S^TS}
\end{align}

\end{subequations}The equality \eqref{eq:S_trunc} follows from the
symmetry of Gaussian random variable. We defer the proof of the other
inequalities to the end of this section. Now invoking the truncated
matrix Bernstein inequality (see \cite{chen2021spectral}, Corollary~3.2),
we have that with probability at least $1-O(N^{-11})$,
\begin{align*}
\|\bm{Y}_{k}\| & \le C_{2}\frac{\sigma}{\sigma_{\min}}\left(\sqrt{\left(n+\frac{Kr_{\mathrm{avg}}}{\theta}+\frac{Kr\cdot r_{\mathrm{avg}}}{\theta}\wedge\frac{r}{\theta^{2}}\right)\log^{3}N}\right)
\end{align*}
for some constant $C_{2}>0$. 

\paragraph{Proof of \eqref{eq:S_norm}. }

Using Gaussian tail bound, we have that for each $(i,j)\in[n]\times[d_{k}]$,
with probability at least $1-N^{-14}$, 
\[
|[\bm{E}_{k}]_{ij}|\le C_{1}\sigma\sqrt{\log N}
\]
 for some constant $C_{1}>0$. Moreover, 

\begin{align*}
\|\bm{S}_{ij}\| & \le\left|[\bm{E}_{k}]_{ij}\right|\cdot\|\widetilde{\bm{\Sigma}}_{k}^{\star-1}\|\cdot\left\Vert \bm{P}^{-1}\bm{P}_{k}^{\perp}\right\Vert \\
 & \le\frac{C_{1}\sigma\sqrt{\log N}}{\sigma_{\min}}\sqrt{\frac{3Kr_{\mathrm{avg}}}{\theta}\wedge\frac{1}{\theta^{2}}}
\end{align*}
for some constant $C_{1}>0$. The last inequality uses \eqref{eq:P-1_proj_norm}
in Lemma~\ref{lem:P_proj_norm}.

\paragraph{Proof of \eqref{eq:SS^T}.}

Consider $\mathbb{E}\sum_{ij}\bm{S}_{ij}\bm{S}_{ij}^{\top}$. We organize
it to be
\begin{align*}
\mathbb{E}\sum_{ij}\bm{S}_{ij}\bm{S}_{ij}^{\top} & =\mathbb{E}[\bm{E}_{k}]_{ij}^{2}\bm{P}^{-1}\bm{P}_{k}^{\perp}\left(\sum_{ij}\alpha_{j}\bm{e}_{i}\bm{e}_{i}^{\top}\right)\bm{P}_{k}^{\perp}\bm{P}^{-1}\\
 & =\sigma^{2}\bm{P}^{-1}\bm{P}_{k}^{\perp}\left(\sum_{j=1}^{d_{k}}\alpha_{j}\bm{I}_{n}\right)\bm{P}_{k}^{\perp}\bm{P}^{-1}\\
 & =\left(\sigma^{2}\sum_{j=1}^{d_{k}}\alpha_{j}\right)\bm{P}^{-1}\bm{P}_{k}^{\perp}\bm{P}^{-1},
\end{align*}
where 
\[
\alpha_{j}\coloneqq\bm{e}_{j}^{\top}\widetilde{\bm{V}}_{k}^{\star}\widetilde{\bm{\Sigma}}_{k}^{\star-1}\widetilde{\bm{U}}_{k}^{\star\top}\bm{U}^{\star}\bm{U}^{\star\top}\widetilde{\bm{U}}_{k}^{\star}\widetilde{\bm{\Sigma}}_{k}^{\star-1}\widetilde{\bm{V}}_{k}^{\star\top}\bm{e}_{j}.
\]
Since $\bm{U}^{\star}\bm{U}^{\star\top}$ is rank-$r$,
\begin{align*}
\sum_{j=1}^{d_{k}}\alpha_{j} & =\mathrm{Trace}\left(\widetilde{\bm{V}}_{k}^{\star}\widetilde{\bm{\Sigma}}_{k}^{\star-1}\widetilde{\bm{U}}_{k}^{\star\top}\bm{U}^{\star}\bm{U}^{\star\top}\widetilde{\bm{U}}_{k}^{\star}\widetilde{\bm{\Sigma}}_{k}^{\star-1}\widetilde{\bm{V}}_{k}^{\star\top}\right)\\
 & \le r\left\Vert \widetilde{\bm{V}}_{k}^{\star}\widetilde{\bm{\Sigma}}_{k}^{\star-1}\widetilde{\bm{U}}_{k}^{\star\top}\bm{U}^{\star}\bm{U}^{\star\top}\widetilde{\bm{U}}_{k}^{\star}\widetilde{\bm{\Sigma}}_{k}^{\star-1}\widetilde{\bm{V}}_{k}^{\star\top}\right\Vert \\
 & \le r\|\widetilde{\bm{\Sigma}}_{k}^{\star-2}\|\le\frac{r}{\sigma_{\min}^{2}}.
\end{align*}
Then with Lemma~\eqref{lem:P_proj_norm}, we have that
\begin{align*}
\left\Vert \mathbb{E}\sum_{ij}\bm{S}_{ij}\bm{S}_{ij}^{\top}\right\Vert  & \le\frac{r\sigma^{2}}{\sigma_{\min}^{2}}\left\Vert \bm{P}^{-1}\bm{P}_{k}^{\perp}\bm{P}^{-1}\right\Vert \\
 & =\frac{r\sigma^{2}}{\sigma_{\min}^{2}}\left\Vert \bm{P}^{-1}\bm{P}_{k}^{\perp}\right\Vert ^{2} 
 \le\frac{r\sigma^{2}}{\sigma_{\min}^{2}}\cdot\left(\frac{3Kr_{\mathrm{avg}}}{\theta}\wedge\frac{1}{\theta^{2}}\right).
\end{align*}

\paragraph{Proof of for \eqref{eq:S^TS}. }

Within the scope of this proof we define $\bm{M}$ with 
\begin{align*}
\bm{M}\coloneqq & \sum_{ij}\bm{e}_{j}\bm{e}_{i}^{\top}\bm{P}_{k}^{\perp}\bm{P}^{-2}\bm{P}_{k}^{\perp}\bm{e}_{i}\bm{e}_{j}^{\top}\\
= & \sum_{j=1}^{d_{k}}\bm{e}_{j}\left(\sum_{i=1}^{n}\bm{e}_{i}^{\top}\bm{P}_{k}^{\perp}\bm{P}^{-2}\bm{P}_{k}^{\perp}\bm{e}_{i}\right)\bm{e}_{j}^{\top}\\
= & \mathrm{Trace}\left(\bm{P}_{k}^{\perp}\bm{P}^{-2}\bm{P}_{k}^{\perp}\right)\bm{I}_{d_{k}}
\end{align*}
To control this, we have 
\begin{align*}
\mathrm{Trace}\left(\bm{P}_{k}^{\perp}\bm{P}^{-2}\bm{P}_{k}^{\perp}\right) & =\mathrm{Trace}\left(\bm{P}^{-2}\bm{P}_{k}^{\perp}\right)\\
 & =\mathrm{Trace}\left(\sum_{i}^{n-r}\mu_{i}^{-2}\bm{x}_{i}\bm{x}_{i}^{\top}\bm{P}_{k}^{\perp}\right)\\
 & =\sum_{i=1}^{n-r}\mu_{i}^{-2}\bm{x}_{i}^{\top}\bm{P}_{k}^{\perp}\bm{x}_{i}
 \le\sum_{i=1}^{n-r}\left(K\mu_{i}^{-1}\wedge\mu_{i}^{-2}\right).
\end{align*}
The last inequality uses \eqref{eq:x^T_Pk_x}. 

By Lemma~\ref{lem:sum_mu_reciprocal}, $\left|\{i:\mu_{i}\le1/2\}\right|\le2r_{\mathrm{avg}}$.
For any $i$ such that $\mu_{i}\le1/2$, $K\mu_{i}^{-1}\wedge\mu_{i}^{-2}\le K/\theta$,
otherwise $K\mu_{i}^{-1}\wedge\mu_{i}^{-2}\le4$. Therefore,
\begin{equation}
\mathrm{Trace}\left(\bm{P}_{k}^{\perp}\bm{P}^{-2}\bm{P}_{k}^{\perp}\right)\le\sum_{i=1}^{n-r}\left(K\mu_{i}^{-1}\wedge\mu_{i}^{-2}\right)\le4n+\frac{2Kr_{\mathrm{avg}}}{\theta}.\label{eq:Tr_PkP}
\end{equation}
Combining these results, we have 

\begin{align*}
\mathbb{E}\sum_{ij}\bm{S}_{ij}^{\top}\bm{S}_{ij} & =\sigma^{2}\bm{U}^{\star}\bm{U}^{\star\top}\widetilde{\bm{U}}_{k}^{\star}\widetilde{\bm{\Sigma}}^{\star-1}\widetilde{\bm{V}}_{k}^{\star\top}\bm{M}\widetilde{\bm{V}}_{k}^{\star}\widetilde{\bm{\Sigma}}^{\star-1}\widetilde{\bm{U}}_{k}^{\star\top}\bm{U}^{\star}\bm{U}^{\star\top}\\
 & \preceq\sigma^{2}\left(4n+\frac{2Kr_{\mathrm{avg}}}{\theta}\right)\bm{U}^{\star}\bm{U}^{\star\top}\widetilde{\bm{U}}_{k}^{\star}\widetilde{\bm{\Sigma}}^{\star-1}\widetilde{\bm{V}}_{k}^{\star\top}\widetilde{\bm{V}}_{k}^{\star}\widetilde{\bm{\Sigma}}^{\star-1}\widetilde{\bm{U}}_{k}^{\star\top}\bm{U}^{\star}\bm{U}^{\star\top}
\end{align*}
Then 
\begin{align*}
\left\Vert \mathbb{E}\sum_{ij}\bm{S}_{ij}^{\top}\bm{S}_{ij}\right\Vert  & \le\frac{\sigma^{2}}{\sigma_{\min}^{2}}\left(4n+\frac{2Kr_{\mathrm{avg}}}{\theta}\right).
\end{align*}

\subsubsection{Proof of \eqref{eq:ub_YY^T}\label{subsec:Proof_YYT}}

Recall
\[
\bm{Y}_{k,1}\bm{Y}_{k,1}^{\top}=\bm{P}^{-1}\bm{B}_{k,1}\bm{P}^{\perp}\bm{B}_{k,1}^{\top}\bm{P}^{-1},
\]
where $\bm{B}_{k,1}=\bm{P}_{k}^{\perp}\bm{E}_{k}\widetilde{\bm{V}}_{k}^{\star}\widetilde{\bm{\Sigma}}_{k}^{\star-1}\widetilde{\bm{U}}_{k}^{\star\top}$.
Invoke Lemma~\ref{lem:expectation_2}, we have that 
\begin{align*}
\mathbb{E}\bm{Y}_{k,1}\bm{Y}_{k,1}^{\top} & =\bm{P}^{-1}\bm{P}_{k}^{\perp}\mathbb{E}\left[\bm{E}_{k}\widetilde{\bm{V}}_{k}^{\star}\widetilde{\bm{\Sigma}}_{k}^{\star-1}\widetilde{\bm{U}}_{k}^{\star\top}\bm{P}^{\perp}\widetilde{\bm{U}}_{k}^{\star}\widetilde{\bm{\Sigma}}_{k}^{\star-1}\widetilde{\bm{V}}_{k}^{\star\top}\bm{E}_{k}^{\top}\right]\bm{P}_{k}^{\perp}\bm{P}^{-1}\\
 & =\mathrm{Trace}(\widetilde{\bm{V}}_{k}^{\star}\widetilde{\bm{\Sigma}}_{k}^{\star-1}\widetilde{\bm{U}}_{k}^{\star\top}\bm{P}^{\perp}\widetilde{\bm{U}}_{k}^{\star}\widetilde{\bm{\Sigma}}_{k}^{\star-1}\widetilde{\bm{V}}_{k}^{\star\top})\bm{P}^{-1}\bm{P}_{k}^{\perp}\bm{P}^{-1}.
\end{align*}
As $\bm{P}^{\perp}$ is rank-$r$ and $\|\widetilde{\bm{\Sigma}}_{k}^{\star-1}\|\le\sigma_{\min}^{-1}$,
\begin{equation}
\mathbb{E}\bm{Y}_{k,1}\bm{Y}_{k,1}^{\top}\preceq\frac{r\sigma^{2}}{\sigma_{\min}^{2}}\bm{P}^{-1}\bm{P}_{k}^{\perp}\bm{P}^{-1}.\label{eq:YYT}
\end{equation}
Summing up over $k$ and substituting $\bm{P}_{k}^{\perp}=\bm{I}_{n}-\widetilde{\bm{U}}_{k}^{\star}\widetilde{\bm{U}}_{k}^{\star\top}$,
we have 
\begin{align}
\sum_{k=1}^{K}\mathbb{E}\bm{Y}_{k,1}\bm{Y}_{k,1}^{\top} & \preceq\frac{r\sigma^{2}}{\sigma_{\min}^{2}}\bm{P}^{-1}\sum_{k=1}^{K}\left(\bm{I}_{n}-\widetilde{\bm{U}}_{k}^{\star}\widetilde{\bm{U}}_{k}^{\star\top}\right)\bm{P}^{-1}\nonumber \\
 & =\frac{r\sigma^{2}}{\sigma_{\min}^{2}}\left(\sum_{i=1}^{n-r}\mu_{i}^{-1}\bm{x}_{i}\bm{x}_{i}^{\top}\right)\left(K\sum_{i=1}^{n-r}\mu_{i}\bm{x}_{i}\bm{x}_{i}^{\top}\right)\left(\sum_{i=1}^{n-r}\mu_{i}^{-1}\bm{x}_{i}\bm{x}_{i}^{\top}\right)\nonumber \\
 & =\frac{Kr\sigma^{2}}{\sigma_{\min}^{2}}\left(\sum_{i=1}^{n-r}\mu_{i}^{-1}\bm{x}_{i}\bm{x}_{i}^{\top}\right).\label{eq:sum_YYT}
\end{align}
The first equality follows from the definition of $\bm{P}^{-1}$,
$\{\mu_{i}\}$ and $\{\bm{x}_{i}\}$. Then as $\min_{i}\mu_{i}\ge\theta$,
\[
\left\Vert \sum_{k=1}^{K}\mathbb{E}[\bm{Y}_{k}\bm{Y}_{k}^{\top}]\right\Vert \le\frac{Kr\sigma^{2}}{\theta\sigma_{\min}^{2}}.
\]

\subsubsection{Proof of \eqref{eq:ub_Y^TY}\label{subsec:Proof_YTY}}

Let $\bm{v}\in\mathbb{R}^{n}$ be an arbitrary vector such that $\|\bm{v}\|\le1$.
By linearity of expectation,
\begin{align*}
\bm{v}^{\top}\left(\mathbb{E}\sum_{k=1}^{K}\bm{Y}_{k}^{\top}\bm{Y}_{k}\right)\bm{v} & =\mathrm{Trace}\left(\sum_{k=1}^{K}\mathbb{E}\left[\bm{Y}_{k}\bm{v}\bm{v}^{\top}\bm{Y}_{k}^{\top}\right]\right).
\end{align*}
Similar to \eqref{eq:sum_YYT}, since $\bm{v}\bm{v}^{\top}$is rank
1 and $\|\bm{v}\|\le1$, we have that 
\[
\sum_{k=1}^{K}\mathbb{E}[\bm{Y}_{k}\bm{v}\bm{v}^{\top}\bm{Y}_{k}^{\top}]\preceq\frac{K\sigma^{2}}{\sigma_{\min}^{2}}\left(\sum_{i=1}^{n-r}\mu_{i}^{-1}\bm{x}_{i}\bm{x}_{i}^{\top}\right).
\]
Therefore
\begin{align*}
\bm{v}^{\top}\left(\mathbb{E}\sum_{k=1}^{m}\bm{Y}_{k}^{\top}\bm{Y}_{k}\right)\bm{v} & =\mathrm{Trace}\left(\sum_{k=1}^{K}\mathbb{E}\left[\bm{Y}_{k}\bm{v}\bm{v}^{\top}\bm{Y}_{k}^{\top}\right]\right)\\
 & \le\mathrm{Trace}\left(\sum_{i=1}^{n-r}\mu_{i}^{-1}\bm{x}_{i}\bm{x}_{i}^{\top}\right)\cdot\frac{K\sigma^{2}}{\sigma_{\min}^{2}}
 =\sum_{i=1}^{n-r}\mu_{i}^{-1}\cdot\frac{K\sigma^{2}}{\sigma_{\min}^{2}}.
\end{align*}

To this end we take supremum over $\bm{v}$ such that $\|\bm{v}\|\le1$
and invoke the Lemma~\ref{lem:sum_mu_reciprocal} to reach \eqref{eq:ub_Y^TY}. 

\subsubsection{Proof of \eqref{eq:ub_Y-2} and \eqref{eq:trunc_Y2}}

Fix $k\in[K]$. By \eqref{eq:EET-dI} and \eqref{eq:P-1_proj_norm},
with probability at least $1-O(N^{-11})$, for some large enough constant
$C_{1}$.
\begin{align*}
\left\Vert \bm{Y}_{k,2}\right\Vert  & \le\left\Vert \bm{P}^{-1}\bm{P}_{k}^{\perp}\right\Vert \left\Vert \bm{E}_{k}\bm{E}_{k}^{\top}-d_{k}\sigma^{2}\bm{I}_{n}\right\Vert \left\Vert \bm{P}_{k}^{-1}\bm{P}^{\perp}\right\Vert \\
 & \le\sqrt{\frac{3Kr_{\mathrm{avg}}}{\theta}\wedge\frac{1}{\theta^{2}}}\cdot C_{1}\sigma^{2}\left(\sqrt{nd_{k}}+n\right)\cdot\sigma_{\min}^{-2}\\
 & =\frac{C_{1}\sigma^{2}\left(\sqrt{nd_{k}}+n\right)}{\sigma_{\min}^{2}}\sqrt{\frac{3Kr_{\mathrm{avg}}}{\theta}\wedge\frac{1}{\theta^{2}}}\cdot
\end{align*}
Moreover, 
\begin{align*}
 & \left\Vert \mathbb{E}\left[\bm{Y}_{k,2};\|\bm{Y}_{k,2}\|>\frac{C_{1}\sigma^{2}\left(\sqrt{nd_{k}}+n\right)}{\sigma_{\min}^{2}}\sqrt{\frac{3Kr_{\mathrm{avg}}}{\theta}\wedge\frac{1}{\theta^{2}}}\right]\right\Vert \\
 & \quad\le\left\Vert \bm{P}^{-1}\bm{P}_{k}^{\perp}\right\Vert \left\Vert \bm{P}_{k}^{-1}\bm{P}^{\perp}\right\Vert \left\Vert \mathbb{E}\left[\|\bm{E}_{k}\bm{E}_{k}^{\top}-d_{k}\sigma^{2}\bm{I}_{n}\|;\|\bm{E}_{k}\bm{E}_{k}^{\top}-d_{k}\sigma^{2}\bm{I}_{n}\|>C_{2}\sigma^{2}\left(\sqrt{nd_{k}}+n\right)\right]\right\Vert \\
 & \quad\le\frac{C_{2}\sigma^{2}N^{-10}}{\sigma_{\min}^{2}}\sqrt{\frac{3Kr_{\mathrm{avg}}}{\theta}\wedge\frac{1}{\theta^{2}}}
\end{align*}
for some large enough constant $C_{2}$.

\subsubsection{Proof of \eqref{eq:ub_YY^T-2}}

By Lemma~\ref{lem:expectation_2}, $\mathbb{E}[\bm{E}_{k}\bm{E}_{k}^{\top}]=d_{k}\sigma^{2}\bm{I}_{n}$.
Then we can simplify $\mathbb{E}\bm{Y}_{k,2}\bm{Y}_{k,2}^{\top}$
as 
\begin{align*}
\mathbb{E}\bm{Y}_{k,2}\bm{Y}_{k,2}^{\top} & =\bm{P}^{-1}\bm{P}_{k}^{\perp}\mathbb{E}\left[\bm{E}_{k}\bm{E}_{k}^{\top}\bm{P}_{k}^{-1}\bm{P}^{\perp}\bm{P}_{k}^{-1}\bm{E}_{k}\bm{E}_{k}^{\top}\right]\bm{P}_{k}^{\perp}\bm{P}^{-1}-d_{k}^{2}\sigma^{4}\bm{P}^{-1}\bm{P}_{k}^{\perp}\bm{P}_{k}^{-1}\bm{P}^{\perp}\bm{P}_{k}^{-1}\bm{P}_{k}^{\perp}\bm{P}^{-1}\\
 & =\bm{P}^{-1}\bm{P}_{k}^{\perp}\mathbb{E}\left[\bm{E}_{k}\bm{E}_{k}^{\top}\bm{P}_{k}^{-1}\bm{P}^{\perp}\bm{P}_{k}^{-1}\bm{E}_{k}\bm{E}_{k}^{\top}\right]\bm{P}_{k}^{\perp}\bm{P}^{-1}.
\end{align*}
Now using Lemma~\ref{lem:expectation_4}, we can simplify this to
be 
\begin{align*}
\mathbb{E}\bm{Y}_{k,2}\bm{Y}_{k,2}^{\top} & =\sigma^{4}d_{k}\mathrm{Trace}(\bm{P}_{k}^{-1}\bm{P}^{\perp}\bm{P}_{k}^{-1})\bm{P}^{-1}\bm{P}_{k}^{\perp}\bm{P}^{-1} \preceq\frac{rd_{k}\sigma^{4}}{\sigma_{\min}^{2}}\bm{P}^{-1}\bm{P}_{k}^{\perp}\bm{P}^{-1}.
\end{align*}
The last line uses the fact that $\mathrm{rank}(\bm{P}_{k}^{-1}\bm{P}^{\perp}\bm{P}_{k}^{-1})\le r$
and $\|\bm{P}_{k}^{-1}\bm{P}^{\perp}\bm{P}_{k}^{-1}\|\le\sigma_{\min}^{4}$.
Summing up over $k\in[K]$, by the definition of $\bm{P}$ in \eqref{eq:P_def}
and $\bm{P}_{k}^{\perp}\coloneqq\bm{I}_{n}-\widetilde{\bm{U}}_{k}^{\star}\widetilde{\bm{U}}_{k}^{\star\top}$,
we have that
\begin{align*}
\sum_{k=1}^{K}\mathbb{E}\bm{Y}_{k,2}\bm{Y}_{k,2}^{\top} & \preceq\frac{rd\sigma^{4}}{\sigma_{\min}^{4}}\bm{P}^{-1}\left(\sum_{k=1}^{K}\bm{P}_{k}^{\perp}\right)\bm{P}^{-1} =
\frac{Krd\sigma^{4}}{\sigma_{\min}^{4}}\bm{P}^{-1}\left(\bm{I}_{n}-\frac{1}{K}\sum_{k=1}^{K}\widetilde{\bm{U}}^{\star}\widetilde{\bm{U}}^{\star\top}\right)\bm{P}^{-1} =
\frac{Krd\sigma^{4}}{\sigma_{\min}^{4}}\bm{P}^{-1}.
\end{align*}
Then 
\[
\left\Vert \sum_{k=1}^{K}\mathbb{E}\bm{Y}_{k,2}\bm{Y}_{k,2}^{\top}\right\Vert \le\frac{Krd\sigma^{4}}{\sigma_{\min}^{4}}\|\bm{P}^{-1}\|\le\frac{Krd\sigma^{4}}{\theta\sigma_{\min}^{4}}.
\]

\subsubsection{Proof of \eqref{eq:ub_Y^TY-2}}

By Lemma~\ref{lem:expectation_2}, $\mathbb{E}[\bm{E}_{k}\bm{E}_{k}^{\top}]=d_{k}\sigma^{2}\bm{I}_{n}$.
Then we can simplify $\mathbb{E}\bm{Y}_{k,2}^{\top}\bm{Y}_{k,2}$
as 
\begin{align*}
\mathbb{E}\bm{Y}_{k,2}^{\top}\bm{Y}_{k,2} & =\bm{P}^{\perp}\bm{P}_{k}^{-1}\mathbb{E}\left[\bm{E}_{k}\bm{E}_{k}^{\top}\bm{P}_{k}^{\perp}\bm{P}^{-2}\bm{P}_{k}^{\perp}\bm{E}_{k}\bm{E}_{k}^{\top}\right]\bm{P}_{k}^{-1}\bm{P}^{\perp}-d_{k}^{2}\sigma^{4}\bm{P}^{\perp}\bm{P}_{k}^{-1}\bm{P}_{k}^{\perp}\bm{P}^{-1}\bm{P}_{k}^{\perp}\bm{P}_{k}^{-1}\bm{P}^{\perp}\\
 & =\bm{P}^{\perp}\bm{P}_{k}^{-1}\mathbb{E}\left[\bm{E}_{k}\bm{E}_{k}^{\top}\bm{P}_{k}^{\perp}\bm{P}^{-2}\bm{P}_{k}^{\perp}\bm{E}_{k}\bm{E}_{k}^{\top}\right]\bm{P}_{k}^{-1}\bm{P}^{\perp}.
\end{align*}
Now using Lemma~\ref{lem:expectation_4} and \eqref{eq:Tr_PkP},
we can simplify the summation over $k$ be 
\begin{align*}
\sum_{k=1}^{K}\mathbb{E}\bm{Y}_{k,2}^{\top}\bm{Y}_{k,2} & =\sum_{k=1}^{K}\sigma^{4}d_{k}\mathrm{Trace}(\bm{P}_{k}^{\perp}\bm{P}^{-2}\bm{P}_{k}^{\perp})\bm{P}^{\perp}\bm{P}_{k}^{-1}\bm{P}_{k}^{-1}\bm{P}^{\perp}\\
 & \preceq\sigma^{4}d\mathrm{Trace}\left(\bm{P}^{-2}\sum_{k=1}^{K}\bm{P}_{k}^{\perp}\right)\bm{P}^{\perp}\bm{P}_{k}^{-1}\bm{P}_{k}^{-1}\bm{P}^{\perp}\\
 & =\sigma^{4}d\mathrm{Trace}\left(K\bm{P}^{-1}\right)\bm{P}^{\perp}\bm{P}_{k}^{-1}\bm{P}_{k}^{-1}\bm{P}^{\perp}
\end{align*}
The last line uses the definition of $\bm{P}$ in \eqref{eq:P_def}
and $\bm{P}_{k}^{\perp}\coloneqq\bm{I}_{n}-\widetilde{\bm{U}}_{k}^{\star}\widetilde{\bm{U}}_{k}^{\star\top}$.
Now invoke Lemma~\ref{lem:sum_mu_reciprocal}, we have that 
\[
\mathrm{Trace}\left(\bm{P}^{-1}\right)=\sum_{i=1}^{n-r}\mu_{i}^{-1}\le2n+\frac{2r_{\mathrm{avg}}}{\theta}.
\]
Then we conclude that
\begin{align*}
\left\Vert \sum_{k=1}^{K}\mathbb{E}\bm{Y}_{k,2}^{\top}\bm{Y}_{k,2}\right\Vert  & \le\sigma^{4}Kd\left(2n+\frac{2r_{\mathrm{avg}}}{\theta}\right)\left\Vert \bm{P}^{\perp}\bm{P}_{k}^{-1}\bm{P}_{k}^{-1}\bm{P}^{\perp}\right\Vert \\
 & \le\frac{\sigma^{4}}{\sigma_{\min}^{4}}Kd\left(2n+\frac{2r_{\mathrm{avg}}}{\theta}\right).
\end{align*}

\subsubsection{Proof of Lemma~\ref{lem:sum_mu_reciprocal}\label{subsec:Proo_reciprocal}}

We prove \eqref{eq:mu_sum}, \eqref{eq:mu_less_2}, and \eqref{eq:mu_reciprocal_sum}
in order.

For \eqref{eq:mu_sum}, recall that 
\[
\bm{I}_{n}-\bm{U}^{\star}\bm{U}^{\star\top}-\frac{1}{K}\sum_{k=1}^{K}\bm{U}_{k}^{\star}\bm{U}_{k}^{\star\top}=\sum_{i=1}^{n-r}\mu_{i}\bm{x}_{i}\bm{x}_{i}^{\top}.
\]
We take the trace of both sides. For the left hand side, 
\begin{align*}
\mathrm{Trace}\left(\bm{I}_{n}-\bm{U}^{\star}\bm{U}^{\star\top}-\frac{1}{K}\sum_{k=1}^{K}\bm{U}_{k}^{\star}\bm{U}_{k}^{\star\top}\right) & =\frac{1}{K}\sum_{k=1}^{K}\mathrm{Trace}\left(\bm{I}_{n_{1}}-\bm{U}^{\star}\bm{U}^{\star\top}-\bm{U}_{k}^{\star}\bm{U}_{k}^{\star\top}\right)\\
 & =\frac{1}{K}\sum_{k=1}^{K}\left(n-r-r_{k}\right)=n-r-r_{\mathrm{avg}}.
\end{align*}
The last line follows from the fact that $\bm{I}_{n_{1}}-\bm{U}^{\star}\bm{U}^{\star\top}-\bm{U}_{k}^{\star}\bm{U}_{k}^{\star\top}$
has eigenvalue 1 with $n-r-r_{k}$ algorithmic multiplicity. For the
right hand side, since the trace equals the sum of eigenvalues, 
\[
\mathrm{Trace}\left(\sum_{i=1}^{n-r}\mu_{i}\bm{x}_{i}\bm{x}_{i}^{\top}\right)=\sum_{i=1}^{n-r}\mu_{i}.
\]
Combining the two identities yields \eqref{eq:mu_sum}. 

For \eqref{eq:mu_less_2}, let $\alpha\coloneqq\left|\{i:\mu_{i}\le1/2\}\right|$.
Since $\mu_{i}\le1$, then 
\[
n-r-r_{\mathrm{avg}}=\sum_{i=1}^{n-r}\mu_{i}\ge n-r-\left|\{i:\mu_{i}\le1/2\}\right|+\frac{1}{2}\left|\{i:\mu_{i}\le1/2\}\right|.
\]
Therefore $\alpha\le2r_{\mathrm{avg}}$. 

Finally we prove \eqref{eq:mu_reciprocal_sum}. For all $i$ such
that $\mu_{i}\le1/2$, we use the upper bound $\mu_{i}^{-1}\le1/\theta$,
otherwise we use $\mu_{i}^{-1}\le2$, then 
\[
\sum_{i=1}^{n-r}\mu_{i}^{-1}\le\frac{2r_{\mathrm{avg}}}{\theta}+2n.
\]

\subsubsection{Proof of Lemma~\ref{lem:P_proj_norm}\label{subsec:Proof_P_proj}}

We start with the proof of \eqref{eq:x^T_Pk_x}. As $\bm{x}_{i}$
is a unit vector and $\bm{I}_{n}-\bm{U}^{\star}\bm{U}^{\star\top}-\bm{U}_{k}^{\star}\bm{U}_{k}^{\star\top}$
is a projection, 
\[
\bm{x}_{i}^{\top}\left(\bm{I}_{n}-\widetilde{\bm{U}}_{k}^{\star}\widetilde{\bm{U}}_{k}^{\star\top}\right)\bm{x}_{i}\le1.
\]
By the definition of $\bm{x}_{i}$, 
\[
\bm{x}_{i}^{\top}\left(\bm{I}_{n}-\bm{U}^{\star}\bm{U}^{\star\top}-\frac{1}{K}\sum_{j=1}^{K}\bm{U}_{j}^{\star}\bm{U}_{j}^{\star\top}\right)\bm{x}_{i}=\mu_{i}.
\]
Then since $\bm{x}_{i}^{\top}(\bm{I}_{n}-\widetilde{\bm{U}}_{k}^{\star}\widetilde{\bm{U}}_{k}^{\star\top})\bm{x}_{k}\ge0$
for all $k\in[K]$, 
\begin{align*}
\frac{1}{K}\bm{x}_{i}^{\top}\left(\bm{I}_{n}-\widetilde{\bm{U}}_{k}^{\star}\widetilde{\bm{U}}_{k}^{\star\top}\right)\bm{x}_{i}\le\bm{x}_{i}^{\top}\left(\bm{I}_{n}-\bm{U}^{\star}\bm{U}^{\star\top}-\frac{1}{K}\sum_{j=1}^{K}\bm{U}_{j}^{\star}\bm{U}_{j}^{\star\top}\right)\bm{x}_{i} & =\mu_{i}.
\end{align*}
Combining the two inequalities we have \eqref{eq:x^T_Pk_x}. 

Now we consider \eqref{eq:P-1_proj_norm}. Let $\bm{v}\in\mathbb{R}^{n}$
be a unit vector. Let $\alpha_{i}\coloneqq\bm{x}_{i}^{\top}(\bm{I}_{n}-\widetilde{\bm{U}}_{k}^{\star}\widetilde{\bm{U}}_{k}^{\star\top})\bm{v}$.
Since $\{\bm{x}_{i}\}$ is a basis of $\mathrm{col}(\bm{I}_{n}-\bm{U}^{\star}\bm{U}^{\star\top})$,
$\sum_{i}\alpha_{i}^{2}\le1$. Moreover, 
\begin{align}
\alpha_{i} & \le\|\bm{x}_{i}^{\top}(\bm{I}_{n}-\widetilde{\bm{U}}_{k}^{\star}\widetilde{\bm{U}}_{k}^{\star\top})\| \le\sqrt{\bm{x}_{i}^{\top}\left(\bm{I}_{n}-\widetilde{\bm{U}}_{k}^{\star}\widetilde{\bm{U}}_{k}^{\star\top}\right)\bm{x}_{i}} \le\sqrt{K\mu_{i}} .
\label{eq:alpha_mu}
\end{align}
We also have that 
\begin{align*}
\bm{P}^{-1}(\bm{I}_{n}-\widetilde{\bm{U}}_{k}^{\star}\widetilde{\bm{U}}_{k}^{\star\top})\bm{v} & =\sum_{i=1}^{n-r}\mu_{i}^{-1}\bm{x}_{i}\bm{x}_{i}^{\top}(\bm{I}_{n}-\widetilde{\bm{U}}_{k}^{\star}\widetilde{\bm{U}}_{k}^{\star\top})\bm{v}=\sum_{i=1}^{n-r}\frac{\alpha_{i}}{\mu_{i}}\bm{x}_{i}.
\end{align*}
Therefore 
\[
\left\Vert \bm{P}^{-1}(\bm{I}_{n}-\widetilde{\bm{U}}_{k}^{\star}\widetilde{\bm{U}}_{k}^{\star\top})\bm{v}\right\Vert ^{2}=\sum_{i=1}^{n-r}\frac{\alpha_{i}^{2}}{\mu_{i}^{2}}.
\]
As $\mu_{i}\ge\theta$, we immediately has 
\[
\left\Vert \bm{P}^{-1}(\bm{I}_{n}-\widetilde{\bm{U}}_{k}^{\star}\widetilde{\bm{U}}_{k}^{\star\top})\bm{v}\right\Vert ^{2}\le\frac{1}{\theta^{2}}\sum_{i=1}^{n-r}\alpha_{i}^{2}\le\frac{1}{\theta^{2}}.
\]
By Lemma~\ref{lem:sum_mu_reciprocal}, we have that $|\{i:\mu_{i}\le1/2\}|\le2r_{\mathrm{avg}}$.
Then for all $i$ such that $\mu_{i}\le1/2$,
we have the upper bound $\alpha_{i}^{2}/\mu_{i}^{2}\le K/\mu_{i}\le K/\theta$  from \eqref{eq:alpha_mu}. Otherwise we use $\mu_{1}>1/2$ to get $\alpha_{i}^{2}/\mu_{i}^{2}\le4\alpha_{i}^{2}$.
Then 
\[
\sum_{i=1}^{n-r}\frac{\alpha_{i}^{2}}{\mu_{i}^{2}}\le\frac{2Kr_{\mathrm{avg}}}{\theta}+1\le\frac{3Kr_{\mathrm{avg}}}{\theta}.
\]
Combining the two bounds and taking supremum over $\bm{v}$, we have
that
\begin{align*}
\left\Vert \bm{P}^{-1}\bm{P}_{k}^{\perp}\right\Vert ^{2} & =\sup_{\bm{v}:\|\bm{v}\|\le1}\left\Vert \bm{P}^{-1}(\bm{I}_{n}-\widetilde{\bm{U}}_{k}^{\star}\widetilde{\bm{U}}_{k}^{\star\top})\bm{v}\right\Vert ^{2}\le\frac{3Kr_{\mathrm{avg}}}{\theta}\wedge\frac{1}{\theta^{2}}.
\end{align*}

\section{Proof of Theorem~\ref{thm:minimax}\label{sec:Proof_minimax}}
In this section, we prove a stronger version of Theorem~\ref{thm:minimax} which allows for unbalanced matrices, i.e., the number of columns can be much larger than the number $n$ of rows.

\begin{theorem}\label{thm:minimax_full}
    Suppose that $\theta\le1/2$, $r\ge8$, and $n \geq 6r$. Moreover, suppose $d\ge C_1 \log(K/r) + C_2$ for some large enough constant $C_1, C_2 > 0$. Then we have  
\begin{equation}
\inf_{\widehat{\bm{U}}}\sup_{\{\bm{A}_{k}^{\star}\}\in\Theta}\mathbb{E}\left\Vert \bm{U}^{\star}\bm{U}^{\star\top}-\widehat{\bm{U}}\widehat{\bm{U}}^{\top}\right\Vert \ge  \frac{\sigma}{20\sigma_{\min}} \sqrt{\frac{n}{K}+\frac{r}{K\theta}} + \frac{\sigma^2}{50\sigma_{\min}^2} \sqrt{\frac{nd}{K}+\frac{rd}{K\theta}}.
\end{equation}
The infimum on $\widehat{\bm{U}}$ is taken over all measurable functions of the data.
\end{theorem}

\noindent Compared to the vanilla lower bound in Theorem~\ref{thm:minimax}, Theorem~\ref{thm:minimax_full} has an additional term that captures the second-order dependency on the noise level $\sigma$. More importantly, it is identical to two terms in the upper bound in Theorem~\ref{thm:full}.

\subsection{Proof of Theorem~\ref{thm:minimax_full}}\label{subsec:proof_minimax_full}
Assume without loss of generality that 
$\sigma_{\min} = 1$. Consider a random realization of $\bm{A}_k^\star$ given fixed $\bm{U}^\star,\{\bm{U}_k^\star\}$ using the following procedure:
\begin{enumerate}
    \item Draw $\bm{V}^\star_k$ and $\bm{W}^\star_k$ to be independent random matrices with i.i.d.~Gaussian entries of variance $1/d$. Let $\widetilde{\bm{A}}_k^\star = \bm{U}^\star \bm{V}_k^\star + \bm{U}_k^\star \bm{W}_k^\star$ and $\widetilde{\bm{A}}_k = \widetilde{\bm{A}}_k^\star + \bm{E}_k$.
    \item Take $\bm{A}_k^\star = \widetilde{\bm{A}}_k^\star$ and $\bm{A}_k = \widetilde{\bm{A}}_k^\star + \bm{E}_k$ if $\sigma_{2r}(\widetilde{\bm{A}}_k^\star)\ge 0.9$, and re-draw Step~1 otherwise.
\end{enumerate}
We will show in  Section~\ref{subsec:KL_proximity} that the KL-divergence between two truncated instances is close to the KL-divergence between their non-truncated counterpart.

By construction this randomly generated $\{\bm{A}_k^\star\}$ lies in $\Theta$ as long as $\{\bm{U},\bm{U}_{k}^{\star}\}$ satisfies \eqref{eq:minimax_ortho} and \eqref{eq:minimax_misalign}. Therefore
\[\inf_{\widehat{\bm{U}}}\sup_{\{\bm{A}_{k}^{\star}\}\in\Theta}\mathbb{E}\left\Vert \bm{U}^{\star}\bm{U}^{\star\top}-\widehat{\bm{U}}\widehat{\bm{U}}^{\top}\right\Vert  \ge \inf_{\widehat{\bm{U}}}\sup_{\{\bm{U},\bm{U}_{k}^{\star}\}\in\Theta'
}\mathbb{E}\left\Vert \bm{U}^{\star}\bm{U}^{\star\top}-\widehat{\bm{U}}\widehat{\bm{U}}^{\top}\right\Vert, \]
where $\Theta'$ is all $\{\bm{U},\bm{U}_{k}^{\star}\}$ that satisfies \eqref{eq:minimax_ortho} and \eqref{eq:minimax_misalign}. Here the first expectation is taken w.r.t.~$\{\bm{E}_k\}$ and the second expectation is taken w.r.t.~${\{\bm{E}_k, \bm{V}_k^\star, \bm{W}_k^\star\}}$.

It now suffices to prove the following bounds:
\begin{subequations}\label{eq:minimax_all}
\begin{align}
\inf_{\widehat{\bm{U}}}\sup_{\{\bm{U},\bm{U}_{k}^{\star}\}\in\Theta'}\mathbb{E}\left\Vert \bm{U}^{\star}\bm{U}^{\star\top}-\widehat{\bm{U}}\widehat{\bm{U}}^{\top}\right\Vert  & 
\ge  \frac{\sigma}{20}\sqrt{\frac{n}{K}}\label{eq:minimax_first}\\
\inf_{\widehat{\bm{U}}}\sup_{\{\bm{U},\bm{U}_{k}^{\star}\}\in\Theta'}
\mathbb{E}\left\Vert \bm{U}^{\star}\bm{U}^{\star\top}-\widehat{\bm{U}}\widehat{\bm{U}}^{\top}\right\Vert  & 
\ge  \frac{\sigma^2}{20} \sqrt{\frac{nd}{K}}\label{eq:minimax_first_sigma2}\\
\inf_{\widehat{\bm{U}}}\sup_{\{\bm{U},\bm{U}_{k}^{\star}\}\in\Theta'}
\mathbb{E}\left\Vert \bm{U}^{\star}\bm{U}^{\star\top}-\widehat{\bm{U}}\widehat{\bm{U}}^{\top}\right\Vert  & 
\ge \frac{\sigma}{50}\sqrt{\frac{r}{K\theta}}\label{eq:minimax_second}\\
\inf_{\widehat{\bm{U}}}\sup_{\{\bm{U},\bm{U}_{k}^{\star}\}\in\Theta'}
\mathbb{E}\left\Vert \bm{U}^{\star}\bm{U}^{\star\top}-\widehat{\bm{U}}\widehat{\bm{U}}^{\top}\right\Vert  & 
\ge \frac{\sigma^2}{50}\sqrt{\frac{rd}{K\theta}}\label{eq:minimax_second_sigma2}.
\end{align}
\end{subequations}
We prove the first two bounds in Section~\ref{subsec:minimax_first} and the last two in Section~\ref{subsec:minimax_second}.

\subsection{KL-divergence between truncated instances}\label{subsec:KL_proximity}
Consider two sets of subspace parameters $(\bm{U}^\star,\{\bm{U}_k^\star\})$ and $({\bm{U}}^{\star'},\{{\bm{U}}_k^{\star'}\})$. 
We use $[\cdot]_{\bm{U}^\star}$ or $[\cdot]_{{\bm{U}}^{\star'}}$ to distinguish the matrices associated with different parameters. 
Fix $k\in [K]$ and consider $\bm{A}_k^\star$. We denote the law of $[\bm{A}_k]_{\bm{U}^\star}$ as $\mathbb{P}_{k}$ and its probability density to be $P_k(\cdot)$. 
Similarly we denote the law of $[\bm{A}_k]_{{\bm{U}}^{\star'}}$ as ${\mathbb{P}}^{'}_{k}$ and its probability density to be ${P}^{'}_k(\cdot)$. 
On the other hand, we denote the law of $[\widetilde{\bm{A}}_k]_{\bm{U}^\star}$ as $\widetilde{\mathbb{P}}_{k}$ and its probability density to be $\widetilde{{P}}_{k}$. Similarly we denote the law of $[\widetilde{\bm{A}}_k]_{\bm{U}^{\star'}}$ as $\widetilde{\mathbb{P}}_{k}'$ and its probability density to be $\widetilde{{P}}_{k}'$. We then have the following lemma.
\begin{lemma}\label{lem:KL_trunc} Let $k\in[K]$, suppose $(\bm{U}^\star,\{\bm{U}_k^\star\})$ and $({\bm{U}}^{\star'},\{{\bm{U}}_k^{\star'}\})$ satisfy \eqref{eq:minimax_ortho} and \eqref{eq:minimax_misalign}. Then
    \[\mathrm{KL}(\mathbb{P}_{k} \parallel {\mathbb{P}}_{k}') \le \mathrm{KL}(\widetilde{\mathbb{P}}_{k} \parallel \widetilde{\mathbb{P}}_{k}') + 4e^{-C_1d}, \]
    where $C_1>0$ is a universal constant. 
\end{lemma}

\subsubsection{Proof of Lemma~\ref{lem:KL_trunc}}
We first present a lemma that shows the probability that $\widetilde{\bm{A}}_k^\star$ is rejected is small. The proof is deferred to Section~\ref{subsec:proof_rej_rate}.
\begin{lemma}\label{lem:rejection_rate}
    Let $k\in[K]$, suppose $(\bm{U}^\star,\{\bm{U}_k^\star\})$ and $({\bm{U}}^{\star'},\{{\bm{U}}_k^{\star'}\})$ satisfy \eqref{eq:minimax_ortho} and \eqref{eq:minimax_misalign}. Then
    \[\mathbb{P}\left[\sigma_{2r}(\widetilde{\bm{A}}_k^\star) < 0.9\right] \le 2\exp(-C_1 d)\]for some universal constant $C_1$.
\end{lemma}

Given the generation process, it is straightforward to see that for
any matrix of the form $\bm{B}=\bm{U}\bm{V}_{k}^{\top}+\bm{U}_{k}\bm{W}^{\top}$,
we have
\[
\tilde{P}_{k}(\bm{B})=\frac{P_{k}(\bm{B})1\{\sigma_{2r}(\bm{B})\geq0.9\}}{\int P_{k}(\bm{B})1\{\sigma_{2r}(\bm{B})\geq0.9\}\mathrm{d}\bm{B}}.
\]
As a result, we obtain the inequalities 
\[
\tilde{P}_{k}(\bm{B})\leq\frac{P_{k}(\bm{B})}{\mathbb{P}_{k}(\sigma_{2r}(\bm{B})\geq0.9)}\leq\frac{P_{k}(\bm{B})}{1-2e^{-C_{1}d}}.
\]
Moreover, if $\sigma_{2r}(\bm{B})\geq0.9$, we have $P_{k}(\bm{B})\leq\tilde{P}_{k}(\bm{B})$. 

Now consider the KL-divergence $\mathrm{KL}(\mathbb{P}_{k} \parallel {\mathbb{P}}_{k}')$, using the inequalities above, we have
\begin{align*}
    \mathrm{KL}(\mathbb{P}_{k} \parallel {\mathbb{P}}_{k}') &= \int_{\sigma_{2r}(\bm{B})\ge 0.9} P_k(\bm{B}) \log \frac{P_k(\bm{B})}{P_k'(\bm{B})} \\
    &\le \int_{\sigma_{2r}(\bm{B})\ge 0.9}  \widetilde{P}_k(\bm{B})  \log \frac{ \widetilde{P}_k(\bm{B}) }{ (1-2e^{-C_1 d})\widetilde{P}_k'(\bm{B}) }\\
    &\le \int \widetilde{P}_k(\bm{B})  \log \frac{ \widetilde{P}_k(\bm{B}) }{ \widetilde{P}_k'(\bm{B}) } + \int_{\sigma_{2r}(\bm{B})\ge 0.9}  \widetilde{P}_k(\bm{B})  \log \frac{1}{ 1-2e^{-C_1 d} }\\
    &\le  \mathrm{KL}(\widetilde{\mathbb{P}}_{k} \parallel \widetilde{\mathbb{P}}_{k}') -\log (1-2e^{-C_1d}). 
\end{align*}
It remains to control the auxiliary term. As long as $d\ge 2/C_1$, 
\[-\log (1-2e^{-C_1d}) \le 4e^{-C_1d}.\]
The proof is now completed.

\subsubsection{Proof of Lemma~\ref{lem:rejection_rate}}\label{subsec:proof_rej_rate}

Each column of $\widetilde{\bm{A}}_k^\star$  can be written as $\bm{U}^\star \bm{v}+\bm{U}^\star_k \bm{w}$ where $\bm{v},\bm{w}\sim\mathcal{N}(\bm{0},\bm{I}_r/d)$. Therefore these columns are i.i.d. Gaussian vectors with mean 0 and covariance $(\bm{U}^\star\bm{U}^{\star\top}+\bm{U}_k^\star\bm{U}_k^{\star\top})/d$. It is then equivalent to write $\widetilde{\bm{A}}_k^\star$ as $\widetilde{\bm{U}}_k \bm{Z}^\top$, where $\widetilde{\bm{U}}_k$ is a $n\times 2r$ matrix such that $\widetilde{\bm{U}}_k\widetilde{\bm{U}}_k^\top = \bm{U}^\star\bm{U}^{\star\top}+\bm{U}_k^\star\bm{U}_k^{\star\top}$, and $\bm{Z}$ is a $d\times 2r$ random matrix whose entries are i.i.d. Gaussian with mean 0 and variance $1/d$.

It now suffices to control the minimum singular value of $\widetilde{\bm{A}}_k^\star\widetilde{\bm{A}}_k^{\star\top} = \widetilde{\bm{U}}_k\bm{Z}^\top\bm{Z}\widetilde{\bm{U}}_k^\top$, which has expectation $\bm{I}_{2r}$.
By (4.22) in \cite{vershynin2018high}, for any $t>0$, 
\begin{equation}
\mathbb{P}\left[\left\|\bm{Z}^\top\bm{Z} - \bm{I}_{2r}\right\|\ge C_{1}\cdot\frac{1}{d}\left(\sqrt{rd}+t\sqrt{d}+r+t^{2}\right)\right]\le 2\exp(-2t^{2}),
\end{equation}
where $C_{1}>0$ is some large enough constant. Set $t=C_2\sqrt{d}$ for some small enough constant $C_2>0$,  we have that with probability at least
$1-2e^{-2C_2^2 d}$, 
\begin{align*}
\left\|\widetilde{\bm{U}}_k\bm{Z}^\top\bm{Z}\widetilde{\bm{U}}_k^\top - \widetilde{\bm{U}}_k\widetilde{\bm{U}}_k^\top\right\| = \left\|\widetilde{\bm{U}}_k\left(\bm{Z}^\top\bm{Z} - \bm{I}_{2r}\right)\widetilde{\bm{U}}_k^\top \right\| \le \left\|\bm{Z}^\top\bm{Z} - \bm{I}_{2r}\right\| \le 0.19.
\end{align*}
By Weryl's inequality, 
\[\sigma_{2r}(\widetilde{\bm{A}}_k^\star\widetilde{\bm{A}}_k^{\star\top} )\ge 1 - \left\|\widetilde{\bm{U}}_k\bm{Z}^\top\bm{Z}\widetilde{\bm{U}}_k^\top - \widetilde{\bm{U}}_k\widetilde{\bm{U}}_k^\top\right\| \ge 0.81 \]
and 
$$\sigma_{2r}(\widetilde{\bm{A}}_k^\star) =\sqrt{\sigma_{2r}(\widetilde{\bm{A}}_k^\star\widetilde{\bm{A}}_k^{\star\top} )}\ge 0.9.$$

\subsection{Proof of \eqref{eq:minimax_first} and \eqref{eq:minimax_first_sigma2}\label{subsec:minimax_first}}

We employ the generalized Fano's method. We construct a hypotheses class $\Gamma\subset\Theta$,
where each $\gamma\in\Gamma$ represents a hypothesis consisting
of $\bm{U}^{\star}, \bm{U}_{k}^{\star}\in\mathbb{R}^{n\times r}$
for $k\in[K]$. We also assume $\bm{V}_k^\star$ and $\bm{W}_k^\star$ to be sampled with the procedure described in Section~\ref{subsec:proof_minimax_full}. We use subscript $\gamma$ (e.g. $[\bm{U}^{\star}\bm{U}^{\star\top}]_{\gamma}$)
to denote the elements associated with the hypothesis
$\gamma$. We also use $\mathbb{P}^{(\gamma)}$ to denote the probability
distribution of $\{\bm{A}_{k}\}_{k=1}^{K}$ induced by \eqref{eq:def_A}.

Let $\epsilon>0$ be a scalar to be specified later. 
We will construct a hypothesis class $\Gamma$ of cardinality $|\Gamma|\ge e^{nr/6}$
such that for all $\gamma_{1},\gamma_{2}\in\Gamma$, if $\gamma_{1}\neq\gamma_{2}$,
\begin{equation}
\left\Vert \left[\bm{U}^{\star}\bm{U}^{\star\top}\right]_{\gamma_{1}}-\left[\bm{U}^{\star}\bm{U}^{\star\top}\right]_{\gamma_{2}}\right\Vert \ge\frac{C_{1}\epsilon}{\sqrt{2r}}\label{eq:gamma_sep_1}
\end{equation}
and
\begin{equation}
\mathrm{KL}\left(\mathbb{P}^{(\gamma_{1})}\parallel\mathbb{P}^{(\gamma_{2})}\right)\le
\frac{C_1^2 Kd\epsilon^2}{2\sigma^2 d(\sigma^2 d+1)} + 4Ke^{-C_2 d},\label{eq:gamma_KL_1}
\end{equation}
where $C_1$ and $C_2$ are some universal constants. We will construct $\Gamma$
in Section~\ref{subsec:Instance_construction_sqrtK} and prove \eqref{eq:gamma_sep_1} and \eqref{eq:gamma_KL_1} in Section~\ref{subsec:proof_sep_1}.

We now select two different $\epsilon$ to prove \eqref{eq:minimax_first} and \eqref{eq:minimax_first_sigma2}.  

\paragraph{Proof of \eqref{eq:minimax_first}.}
Let  $\epsilon=\sqrt{rn\sigma^{2}/(24C_1^2 K)}$. For any $\gamma_{1},\gamma_{2}\in\Gamma$ such that $\gamma_1\neq\gamma_2$, as long as $d\ge C_2^{-1}\log (192K/nr)$,
\[
\mathrm{KL}\left(\mathbb{P}^{(\gamma_{1})}\parallel\mathbb{P}^{(\gamma_{2})}\right)
\le\frac{C_1^2 Kd\epsilon^2}{2\sigma^2 d} + 4Ke^{-C_2 d}\le\frac{nr}{48} + 4Ke^{-C_2 d}\le\frac{\log|\Gamma|}{4}.
\]
Meanwhile,
\[
\left\Vert \left[\bm{U}^{\star}\bm{U}^{\star\top}\right]_{\gamma_{1}}-\left[\bm{U}^{\star}\bm{U}^{\star\top}\right]_{\gamma_{2}}\right\Vert 
\ge \frac{C_{1}\epsilon}{\sqrt{2r}}\ge \sigma\sqrt{\frac{n}{24K}}.
\]
Then by generalized Fano method (Lemma~3 in \cite{Yu1997}), as long
as $|\Gamma|\ge16$,
\begin{align*}
\inf_{\widehat{\bm{U}}}\sup_{\gamma\in\Theta'}\mathbb{E}\|\bm{U}^{\star}\bm{U}^{\star\top}-\widehat{\bm{U}}\widehat{\bm{U}}^{\top}\| 
& \ge\frac{\sigma}{2}\sqrt{\frac{n}{24K}}\left(1-\frac{\log|\Gamma|/4+\log2}{\log|\Gamma|}\right)\\
 & \ge \frac{\sigma}{20}\sqrt{\frac{n}{K}}.
\end{align*}

\paragraph{Proof of \eqref{eq:minimax_first_sigma2}.}
Let  $\epsilon= \sigma^{2}\sqrt{rnd/(24C_1^2 K)}$. For any $\gamma_{1},\gamma_{2}\in\Gamma$ such that $\gamma_1\neq\gamma_2$, as long as $d\ge C_2^{-1}\log (192K/nr)$,
\[
\mathrm{KL}\left(\mathbb{P}^{(\gamma_{1})}\parallel\mathbb{P}^{(\gamma_{2})}\right)
\le\frac{C_1^2 Kd\epsilon^2}{2\sigma^2 d \cdot \sigma^2 d} + 4Ke^{-C_2 d}\le\frac{nr}{48} + 4Ke^{-C_2 d}\le\frac{\log|\Gamma|}{4}.
\]
Meanwhile,
\[
\left\Vert \left[\bm{U}^{\star}\bm{U}^{\star\top}\right]_{\gamma_{1}}-\left[\bm{U}^{\star}\bm{U}^{\star\top}\right]_{\gamma_{2}}\right\Vert 
\ge \frac{C_{1}\epsilon}{\sqrt{2r}}\ge \sigma^2\sqrt{\frac{nd}{24K}}.
\]
Then by generalized Fano method (Lemma~3 in \cite{Yu1997}), as long
as $|\Gamma|\ge16$,
\begin{align*}
\inf_{\widehat{\bm{U}}}\sup_{\gamma\in\Theta'}\mathbb{E}\|\bm{U}^{\star}\bm{U}^{\star\top}-\widehat{\bm{U}}\widehat{\bm{U}}^{\top}\| 
& \ge\frac{\sigma^2}{2}\sqrt{\frac{nd}{24K}}\left(1-\frac{\log|\Gamma|/4+\log2}{\log|\Gamma|}\right)\\
 & \ge \frac{\sigma^2}{20} \sqrt{\frac{nd}{K}} .
\end{align*}

\subsubsection{Instance construction\label{subsec:Instance_construction_sqrtK}}

Without loss of generality, assume $n$ is divisible by 3 and $K$
is even. For each $\gamma\in\Gamma$, we will associate it with an
orthogonal matrix $\bm{X}\in\mathbb{R}^{n/3\times r}$. Let $\bm{X}_{0}\in\mathbb{R}^{n/3\times r}$
be an arbitrary orthogonal matrix, i.e., $\bm{X}_{0}^{\top}\bm{X}_{0}=\bm{I}_{r}$.
Consider the Frobenius norm on the projection matrix, i.e., for any
orthogonal matrices $\bm{X}_{1}$ and $\bm{X}_{2}$, $d(\bm{X}_{1},\bm{X}_{2})=\|\bm{X}_{1}\bm{X}_{1}^{\top}-\bm{X}_{2}\bm{X}_{2}^{\top}\|_\mathrm{F}$.
Let 
\[
B_{d}(\bm{X}_{0},\epsilon)\coloneqq\left\{ \bm{X}\in\mathbb{R}^{n/3\times r}:\bm{X}^{\top}\bm{X}=\bm{I}_{r},d(\bm{X},\bm{X}_{0})\le\epsilon\right\} .
\]
Then Lemma~1 in \cite{Cai2013} implies that for some small enough
constant $C_{1}>0$, the packing number 
satisfies 
\[
\mathcal{M}(B_{d}(\bm{X}_{0},\epsilon),C_{1}\epsilon,d)\ge e^{r(n/3-r)}\ge e^{rn/6}.
\]
The last inequality comes from the assumption that $6r\le n$. By the definition of a packing number, there exists a set $A_{\epsilon}\subset B_{d}(\bm{X}_{0},\epsilon)$
with $|A_{\epsilon}|\ge e^{nr/6}$ such that for all $\bm{X}_{1},\bm{X}_{2}\in A_{\epsilon}$,
if $\bm{X}_{1}\neq\bm{X}_{2}$, 
\begin{equation}
\|\bm{X}_{1}\bm{X}_{1}^{\top}-\bm{X}_{2}\bm{X}_{2}^{\top}\|\ge\frac{1}{\sqrt{2r}}\|\bm{X}_{1}\bm{X}_{1}^{\top}-\bm{X}_{2}\bm{X}_{2}^{\top}\|_{\mathrm{F}}\ge\frac{C_{1}\epsilon}{\sqrt{2r}}.\label{eq:X_sep}
\end{equation}
For any $\bm{X}$, we design a $\gamma$ uniquely associated with
it. We construct the hypothesis set $\Gamma$ by doing this for all
$\bm{X}\in A_{\epsilon}$. 

\paragraph{Construction of $\gamma$.}

Fix $\bm{X}\in A_{\epsilon}$. For the shared component, let $\bm{U}^{\star}$, $\bm{U}_{+}^{\star}$ and $\bm{U}_{-}^{\star}$ be 
\[
    \bm{U}^{\star}=\begin{bmatrix}\bm{X}\vspace{5pt}\\
        \bm{0}_{n/3\times r}\vspace{5pt}\\
         \bm{0}_{n/3\times r}
        \end{bmatrix}
    \text{,}\qquad
\bm{U}_{+}^{\star}=\begin{bmatrix}\bm{0}_{n/3\times r}\vspace{5pt}\\
\sqrt{1-\theta}\bm{X}\vspace{5pt}\\
\sqrt{\theta}\bm{X}
\end{bmatrix}\qquad\text{and}\qquad\bm{U}_{-}^{\star}=\begin{bmatrix}\bm{0}_{n/3\times r}\vspace{5pt}\\
\sqrt{1-\theta}\bm{X}\vspace{5pt}\\
-\sqrt{\theta}\bm{X}
\end{bmatrix}.
\]
We then let $\bm{U}_{k}^{\star}=\bm{U}_{+}^{\star}$ for all odd $k$
and $\bm{U}_{k}^{\star}=\bm{U}_{-}^{\star}$ for all even $k$.

We now verify the first two conditions in \eqref{eq:minimax_req}:
\begin{enumerate}
    \item Orthogonality \eqref{eq:minimax_ortho}: Since $\bm{U}^{\star}\in B_{d}(\bm{X}_{0},\epsilon)$, we have 
$\bm{U}^{\star\top}\bm{U}^{\star}=\bm{X}^{\top}\bm{X}=\bm{I}_{r}$.
For any $k\in[K]$, 
\[
\bm{U}_{k}^{\star\top}\bm{U}_{k}^{\star}=\bm{0}_{n/3\times r}^{\top}\bm{0}_{n/3\times r}+(1-\theta)\bm{X}^{\top}\bm{X}+\theta\bm{X}^{\top}\bm{X}=\bm{X}^{\top}\bm{X}=\bm{I}_{r},
\]
and $\bm{U}^{\star\top}\bm{U}_{k}^{\star}=\bm{0}_{r\times r}$.
\item Misalignment \eqref{eq:minimax_misalign}: Compute the sum of outer products:\begin{align*}
\bm{U}_{+}^{\star}\bm{U}_{+}^{\star\top}+\bm{U}_{-}^{\star}\bm{U}_{-}^{\star\top} & =2\begin{bmatrix}\bm{0}_{n/3\times n/3} & \bm{0}_{n/3\times n/3} & \bm{0}_{n/3\times n/3}\vspace{5pt}\\
\bm{0}_{n/3\times n/3} & (1-\theta)\bm{X}\bm{X}^{\top} & \bm{0}_{n/3\times n/3} \vspace{5pt}\\
\bm{0}_{n/3\times n/3} & \bm{0}_{n/3\times n/3} & \theta\bm{X}\bm{X}^{\top}
\end{bmatrix}.
\end{align*}
Then
\[
\left\Vert \frac{1}{K}\sum_{k=1}^{K}\bm{U}_{k}^{\star}\bm{U}_{k}^{\star\top}\right\Vert =\left\Vert \frac{1}{2}\left(\bm{U}_{+}^{\star}\bm{U}_{+}^{\star\top}+\bm{U}_{-}^{\star}\bm{U}_{-}^{\star\top}\right)\right\Vert =\theta\vee(1-\theta)=1-\theta.
\]

\end{enumerate}

\subsubsection{Proof of \eqref{eq:gamma_sep_1} and \eqref{eq:gamma_KL_1} \label{subsec:proof_sep_1}}

Let $\gamma_{1},\gamma_{2}$ be two hypotheses and let $\bm{X}_{1}$,
$\bm{X}_{2}$ be the corresponding matrces in $A_{\epsilon}$. For
the separation gap \eqref{eq:gamma_sep_1}, it follows directly from \eqref{eq:X_sep} that
\[
\left\Vert \left[\bm{U}^{\star}\bm{U}^{\star\top}\right]_{\gamma_{1}}-\left[\bm{U}^{\star}\bm{U}^{\star\top}\right]_{\gamma_{2}}\right\Vert =\left\Vert \bm{X}_{1}\bm{X}_{1}^{\top}-\bm{X}_{2}\bm{X}_{2}^{\top}\right\Vert \ge\frac{C_{1}\epsilon}{\sqrt{2r}}.
\]
We now turn to the KL-divergence bound \eqref{eq:gamma_KL_1}. For $i=1,2$, let $\mathbb{P}_{k}^{(\gamma_{i})}$ be the
law of $\bm{A}_{k}$ under $\gamma_{i}$ and $\mathbb{P}^{(\gamma_{i})}$
be the joint law of $\{\bm{A}_{k}\}_{k=1}^{K}$ under $\gamma_{i}$. Due to independence and symmetry, we have
\begin{equation}
\mathrm{KL}\left(\mathbb{P}^{(\gamma_{1})}\parallel\mathbb{P}^{(\gamma_{2})}\right)=\sum_{k=1}^{K}\mathrm{KL}\left(\mathbb{P}_{k}^{(\gamma_{1})}\parallel\mathbb{P}_{k}^{(\gamma_{2})}\right)=K\cdot\mathrm{KL}\left(\mathbb{P}_{1}^{(\gamma_{1})}\parallel\mathbb{P}_{1}^{(\gamma_{2})}\right).\label{eq:KLall_KL1_1}
\end{equation}
Moreover, for $i=1,2$, let $\widetilde{\mathbb{P}}_{k}^{(\gamma_{i})}$ be the
law of $\widetilde{\bm{A}}_{k}$ under $\gamma_{i}$ as defined by the sampling procedure in Section~\ref{subsec:proof_minimax_full}. Lemma~\ref{lem:KL_trunc} shows that 
\begin{equation}\label{eq:KL1_trunc}
\mathrm{KL}\left(\mathbb{P}_{1}^{(\gamma_{1})}\parallel\mathbb{P}_{1}^{(\gamma_{2})}\right) \le \mathrm{KL}\left(\widetilde{\mathbb{P}}_{1}^{(\gamma_{1})}\parallel\widetilde{\mathbb{P}}_{1}^{(\gamma_{2})}\right) + 4 e^{-C_2 d},
\end{equation}
where $C_2$ is a universal constant. 

It now suffices to compute $\mathrm{KL}(\widetilde{\mathbb{P}}_{1}^{(\gamma_{1})}\parallel \widetilde{\mathbb{P}}_{1}^{(\gamma_{2})})$.
Observe that the $j$-th column of $\widetilde{\bm{A}}_k$ is 
\[[\widetilde{\bm{A}}_k]_{\cdot,j} = \bm{U}^\star ([\bm{V}_k^\star]_{j,\cdot})^{\top} + \bm{U}_k^\star ([\bm{W}_k^\star]_{j,\cdot})^{\top}+ [\bm{E}_k]_{\cdot,j}.\]
Recall that $\bm{V}_k^\star$ and $\bm{W}_k^\star$ are entrywise i.i.d. Gaussian with covariance $1/d$  and  $\bm{E}_k$ are entrywise i.i.d. Gaussian with covariance $1$. Then each column of $\bm{A}_k$ is independent with 
\[[\widetilde{\bm{A}}_k]_{\cdot,j}\sim \mathcal{N}\left(\bm{0}, \bm{U}^\star \bm{U}^{\star\top} /d + \bm{U}_1^\star \bm{U}_1^{\star\top} /d + \sigma^2\bm{I}_{n}\right)\]for all $j\in [d]$.
As the covariance of each column is the same, we may denote it as
\[\bm{\Sigma}_i \coloneqq \sigma^2\left[\bm{I}_{n} + \frac{\bm{U}^\star \bm{U}^{\star\top} + \bm{U}_1^\star \bm{U}_1^{\star\top}}{\sigma^2 d} \right]_{\gamma_i}\]
for $i = 1, 2$. It is also easy to see that 
\[\bm{\Sigma}_i^{-1} = \frac{1}{\sigma^2}\left[\bm{I}_n - \frac{\bm{U}^\star \bm{U}^{\star\top} + \bm{U}_1^\star \bm{U}_1^{\star\top}}{\sigma^2 d + 1} \right]_{\gamma_i}.\]

By independence and the KL-divergence of multivariate Gaussian distributions, 
\[\mathrm{KL}(\mathbb{P}_{1}^{(\gamma_{1})}\parallel\mathbb{P}_{1}^{(\gamma_{2})}) 
= \frac{1}{2}\sum_{j = 1}^d \left(\mathrm{Trace}(\bm{\Sigma}_2^{-1} \bm{\Sigma}_1) - n - \log\frac{\mathrm{det}(\bm{\Sigma}_1)}{\mathrm{det}(\bm{\Sigma}_2)}\right).\]
We claim that 
\begin{equation}
    \mathrm{Trace}(\bm{\Sigma}_2^{-1} \bm{\Sigma}_1) = n + \frac{C_1^2 \epsilon^2}{\sigma^2 d(\sigma^2 d+1)}. \label{eq:Trace_KL}    
\end{equation}
The proof is deferred to the end of this section. It is also clear from the form of $\bm{\Sigma}_1$ and $\bm{\Sigma}_2$ that they have the same eigenvalues and thus the same determinant. Then the KL-divergence can be simplified to
\[\mathrm{KL}(\widetilde{\mathbb{P}}_{1}^{(\gamma_{1})}\parallel \widetilde{\mathbb{P}}_{1}^{(\gamma_{2})}) = \frac{1}{2} d \left(n +\frac{C_1^2 \epsilon^2}{\sigma^2 d(\sigma^2 d+1)} - n\right)
= \frac{C_1^2 d\epsilon^2}{2\sigma^2 d(\sigma^2 d+1)}.\]
This combined with \eqref{eq:KLall_KL1_1} and \eqref{eq:KL1_trunc} finishes the proof.

\paragraph{Proof of \eqref{eq:Trace_KL}.}
For $\gamma_{i}$, $i=1,2$,
\begin{align}
\bm{Y}_i \eqqcolon \bm{U}^{\star}\bm{U}^{\star\top}+\bm{U}_{1}^{\star}\bm{U}_{1}^{\star\top} =\begin{bmatrix}\bm{X}_{i}\bm{X}_{i}^{\top} & \bm{0}_{n/3\times n/3} & \bm{0}_{n/3\times n/3}\vspace{5pt}\\
\bm{0}_{n/3\times n/3} & (1-\theta)\bm{X}_{i}\bm{X}_{i}^{\top} & \sqrt{\theta(1-\theta)}\bm{X}_{i}\bm{X}_{i}^{\top}\vspace{5pt}\\
\bm{0}_{n/3\times n/3} & \sqrt{\theta(1-\theta)}\bm{X}_{i}\bm{X}_{i}^{\top} & \theta\bm{X}_{i}\bm{X}_{i}^{\top}
\end{bmatrix}.\label{eq:minimax_Y_form}
\end{align}
Then we have that
\begin{equation}
    \bm{\Sigma}_2^{-1} \bm{\Sigma}_1 = \bm{I}_n 
- \frac{\bm{Y}_2}{\sigma^2 d +1}
+ \frac{\bm{Y}_1}{\sigma^2 d }
- \frac{\bm{Y}_2\bm{Y}_1}{\sigma^2 d (\sigma^2 d +1) }\label{eq:minimax_KL_Sigma}
\end{equation}
Now we compute the trace of each term separately. It is clear that $\mathrm{Trace}(\bm{I}_n) = n$. 
For $\bm{Y}_i$, $i=1, 2$, we can see from \eqref{eq:minimax_Y_form} that
\begin{equation}
    \mathrm{Trace}(\bm{Y}_i) = 2\cdot \mathrm{Trace}(\bm{X}_i\bm{X}_i^\top) 
    = 2\cdot \mathrm{Trace}(\bm{X}_i^\top \bm{X}_i) = 2\cdot\mathrm{Trace}(\bm{I}_r) = 2r. \label{eq:minimax_trace_Y}
\end{equation}
For the trace of $\bm{Y_2}\bm{Y_1}$, 
\begin{align}
    \mathrm{Trace}(\bm{Y}_2 \bm{Y}_1) &= 2\cdot \mathrm{Trace}(\bm{X}_1\bm{X}_1^\top \bm{X}_2\bm{X}_2^\top) \label{eq:minimax_trace_YY}\\
    &= \mathrm{Trace}(\bm{X}_1\bm{X}_1^\top \bm{X}_1\bm{X}_1^\top) + \mathrm{Trace}(\bm{X}_2\bm{X}_2^\top \bm{X}_2\bm{X}_2^\top) - \mathrm{Trace}\left( (\bm{X}_1\bm{X}_1^\top - \bm{X}_2\bm{X}_2^\top)^2\right) \nonumber\\
    &\ge 2r - 2r \|\bm{X}_1\bm{X}_1^\top - \bm{X}_2\bm{X}_2^\top\|^2 \nonumber\\
    &\ge 2r  - C_1^2 \epsilon^2,\nonumber
\end{align}
where the last line comes from \eqref{eq:X_sep}. 
Combining \eqref{eq:minimax_KL_Sigma} with \eqref{eq:minimax_trace_Y} and \eqref{eq:minimax_trace_YY}, we have 
\begin{align*}
    \mathrm{Trace}(\bm{\Sigma}_2^{-1} \bm{\Sigma}_1) \le n - \frac{2r}{\sigma^2 d +1} + \frac{2r}{\sigma^2 d} - \frac{2r - C_1^2 \epsilon^2}{\sigma^2 d(\sigma^2 d+1)} 
    = n + \frac{C_1^2 \epsilon^2}{\sigma^2 d(\sigma^2 d+1)}.
\end{align*}

\subsection{Proof of \eqref{eq:minimax_second} and \eqref{eq:minimax_second_sigma2}\label{subsec:minimax_second}}

We use generalized Fano's method. We will construct a hypotheses class $\Gamma\subset\Theta$,
where each $\gamma\in\Gamma$ represents a hypothesis consisting
of $\bm{U}^{\star}, \bm{U}_{k}^{\star}, \bm{V}_{k}^{\star}, \bm{W}_{k}^{\star}\in\mathbb{R}^{n\times r}$
for $k\in[K]$. We use the subscript $\gamma$ (e.g.  $[\bm{U}^{\star}\bm{U}^{\star\top}]_{\gamma}$)
to denote the elements associated with the hypothesis
$\gamma$. We also use $\mathbb{P}^{(\gamma)}$ to denote the probability
distribution of $\{\bm{A}_{k}\}_{k=1}^{K}$ induced by \eqref{eq:def_A}.

Let $\epsilon>0$ be a scalar to be specified later. 
We will construct $\Gamma$ with cardinality $|\Gamma|\ge2^{r/8}$
such that for all $\gamma,\gamma'\in\Gamma$, if $\gamma\neq\gamma'$,
\begin{equation}
\left\Vert \left[\bm{U}^{\star}\bm{U}^{\star\top}\right]_{\gamma}-\left[\bm{U}^{\star}\bm{U}^{\star\top}\right]_{\gamma'}\right\Vert ^{2}\ge\frac{\epsilon}{16}\label{eq:gamma_sep_2}
\end{equation}
and
\begin{equation}
    \mathrm{KL}\left(\mathbb{P}^{(\gamma)}\parallel\mathbb{P}^{(\gamma')}\right)\le \frac{4K d\epsilon \theta}{\sigma^2 d(\sigma^2 d+1)} + 4Ke^{-C_1 d},\label{eq:gamma_KL_2}
\end{equation}
where $C_1$ is a universal constant. We will construct $\Gamma$
in Section~\ref{subsec:gamma_construction} and prove \eqref{eq:gamma_sep_2}
and \eqref{eq:gamma_KL_2} in Section~\ref{subsec:gamma_separation_KL}. 

We now select two different $\epsilon$ to prove \eqref{eq:minimax_second} and \eqref{eq:minimax_second_sigma2}.  

\paragraph{Proof of \eqref{eq:minimax_second}.}

Let  $\epsilon=r\sigma^{2}/(512 K\theta)$. For any $\gamma_{1},\gamma_{2}\in\Gamma$ such that $\gamma_1\neq\gamma_2$, as long as $d\ge C_1^{-1}\log(512K/r)$,
\[
\mathrm{KL}\left(\mathbb{P}^{(\gamma_{1})}\parallel\mathbb{P}^{(\gamma_{2})}\right)
\le\frac{4K d\epsilon \theta}{\sigma^2 d} + 4Ke^{-C_1 d}\le\frac{r}{128} + 4Ke^{-C_1 d}\le\frac{\log|\Gamma|}{4}.
\]
Meanwhile,
\[
\left\Vert \left[\bm{U}^{\star}\bm{U}^{\star\top}\right]_{\gamma_{1}}-\left[\bm{U}^{\star}\bm{U}^{\star\top}\right]_{\gamma_{2}}\right\Vert^2 
\ge \frac{\epsilon}{16}\ge \frac{r\sigma^2}{128K\theta}.
\]
Then by generalized Fano method (Lemma~3 in \cite{Yu1997}), as long
as $|\Gamma|\ge16$,
\begin{align*}
\inf_{\widehat{\bm{U}}}\sup_{\gamma\in\Theta'}\mathbb{E}\|\bm{U}^{\star}\bm{U}^{\star\top}-\widehat{\bm{U}}\widehat{\bm{U}}^{\top}\| 
& \ge  \frac{1}{2}\sqrt{\frac{r\sigma^2}{128K\theta}}\left(1-\frac{\log|\Gamma|/4+\log2}{\log|\Gamma|}\right)\\
 & \ge  \frac{\sigma}{50}\sqrt{\frac{r}{K\theta}}.
\end{align*}

\paragraph{Proof of \eqref{eq:minimax_second_sigma2}.}
Let  $\epsilon= rd \sigma^4 /(512 K\theta)$. For any $\gamma_{1},\gamma_{2}\in\Gamma$ such that $\gamma_1\neq\gamma_2$, $d\ge C_1^{-1}\log(512K/r)$,
\[
\mathrm{KL}\left(\mathbb{P}^{(\gamma_{1})}\parallel\mathbb{P}^{(\gamma_{2})}\right)
\le\frac{4K d\epsilon \theta}{ \sigma^2 d \cdot \sigma^2 d} + 4Ke^{-C_1 d}\le\frac{r}{128}+ 4Ke^{-C_1 d}\le\frac{\log|\Gamma|}{4}.
\]
Meanwhile,
\[
\left\Vert \left[\bm{U}^{\star}\bm{U}^{\star\top}\right]_{\gamma_{1}}-\left[\bm{U}^{\star}\bm{U}^{\star\top}\right]_{\gamma_{2}}\right\Vert^2 
\ge \frac{\epsilon}{16}\ge \frac{rd \sigma^4}{128K\theta}.
\]
Then by generalized Fano method (Lemma~3 in \cite{Yu1997}), as long
as $|\Gamma|\ge16$,
\begin{align*}
    \inf_{\widehat{\bm{U}}}\sup_{\gamma\in\Theta'}\mathbb{E}\|\bm{U}^{\star}\bm{U}^{\star\top}-\widehat{\bm{U}}\widehat{\bm{U}}^{\top}\| 
    & \ge  \frac{1}{2}\sqrt{\frac{rd \sigma^4}{128K\theta}}\left(1-\frac{\log|\Gamma|/4+\log2}{\log|\Gamma|}\right)\\
     & \ge  \frac{\sigma^2}{50}\sqrt{\frac{rd}{K\theta}}.
\end{align*}

\subsubsection{Instance construction\label{subsec:gamma_construction}}

Consider the set of binary vectors $\{\bm{h}  \in \{-1,1\}^{r} \}$. 
Let $H$ be a subset of it such that for any $\bm{h} \neq \bm{h}'\in H$, the hamming distance $|\{i:h_{i}\neq h'_{i}\}|\ge r/8$.
By Gilbert-Varshamov bound (see Theorem 6.21, \cite{jones2012information}), $|H|$ can be at least $2^{r/8}$.
In this construction, the parameters in $\Gamma$ are indexed by the vectors $\bm{h}$ in $H$. 
Let $\epsilon\in(0,1/2)$, and denote 
$\delta = \frac{1 - \sqrt{1-\epsilon}}{r}$. 
We specify the parameters indexed by $\bm{h}$ as follows. 

For the shared component, define 
\[
\bm{U}^\star(\bm{h}) = \left[\begin{array}{c}
\bm{I}_{r}-\delta r\bm{e}_{1}\bm{e}_{1}^{\top} \vspace{5pt}\\
\sqrt{\frac{\epsilon}{r}}\bm{h} \bm{e}_1^\top \vspace{5pt}\\
\bm{0}_{r}\\
\bm{0}_{(n-3r)\times r}
\end{array}\right].
\]
For the unique component, let $\bm{U}_{k}^{\star}(\bm{h})=\bm{U}_{+}^{\star}(\bm{h})$ for odd $k$  and $\bm{U}_{k}^{\star}(\bm{h})=\bm{U}_{-}^{\star}(\bm{h})$
for even $k$. 
Moreover, we set 
\[
\bm{U}_{+}^{\star}(\bm{h}) = \left[\begin{array}{c}
-\sqrt{\frac{\epsilon(1-\theta)}{r}}\bm{e}_1\bm{h}^{\top} \vspace{5pt}\\
\sqrt{1-\theta}\left(\bm{I}_{r}-\delta\bm{h}\bm{h}^{\top}\right) \vspace{5pt}\\
\sqrt{\theta}\bm{I}_{r} \\
\bm{0}_{(n-3r)\times r}
\end{array}\right], \qquad \text{and}\qquad \bm{U}_{-}^{\star}(\bm{h}) = \left[\begin{array}{c}
-\sqrt{\frac{\epsilon(1-\theta)}{r}}\bm{e}_1\bm{h}^{\top} \vspace{5pt}\\
\sqrt{1-\theta}\left(\bm{I}_{r}-\delta\bm{h}\bm{h}^{\top}\right) \vspace{5pt}\\
-\sqrt{\theta}\bm{I}_{r}\\
\bm{0}_{(n-3r)\times r}
\end{array}\right].
\]
In essence, the shared and unique components are all 3-block matrices, with each block having size $r \times r$.

We now verify the first two conditions in \eqref{eq:minimax_req}.
It is straightforward with basic algebra to verify the  orthogonality condition~\eqref{eq:minimax_ortho}. 
 For the misalignment condition~\eqref{eq:minimax_misalign}, we have 
 \begin{align*}
\left\Vert \frac{1}{K}\sum_{k=1}^{K}\bm{U}_{k}^{\star}(\bm{h})\bm{U}_{k}^{\star\top}(\bm{h})\right\Vert  & =\left\Vert \frac{1}{2}\left(\bm{U}_{+}^{\star}(\bm{h})\bm{U}_{+}^{\star\top}(\bm{h})+\bm{U}_{-}^{\star}(\bm{h})\bm{U}_{-}^{\star\top}(\bm{h})\right)\right\Vert \\
 & = \frac{1 + \left\|\bm{U}_{+}^{\star\top}(\bm{h}) \bm{U}_{-}^{\star}(\bm{h}) \right\|}{2}.
\end{align*}
By construction, we have $\bm{U}_{+}^{\star\top}(\bm{h}) \bm{U}_{-}^{\star}(\bm{h}) = (1 - 2 \theta ) \bm{I}_r$. Hence the misalignment condition holds. 

\subsubsection{Proof of \eqref{eq:gamma_sep_2} and \eqref{eq:gamma_KL_2}\label{subsec:gamma_separation_KL}}

Denote $\bm{h},\bm{h}'\in H$ as the two distinct vectors associated with two hypotheses $\gamma,\gamma'$.

\paragraph{Proof of \eqref{eq:gamma_sep_2}.}

Observe that $[\bm{U}^{\star}\bm{U}^{\star\top}]_{\gamma}$ and $[\bm{U}^{\star}\bm{U}^{\star\top}]_{\gamma'}$ differ only in the first column and they are both projection matrices.
Thus 
\begin{align*}
\left[\bm{U}^{\star}\bm{U}^{\star\top}\right]_{\gamma}-\left[\bm{U}^{\star}\bm{U}^{\star\top}\right]_{\gamma'} & =\left[\left[\bm{U}^{\star}\right]_{\cdot1}\left[\bm{U}^{\star}\right]_{\cdot1}^{\top}\right]_{\gamma}-\left[\left[\bm{U}^{\star}\right]_{\cdot1}\left[\bm{U}^{\star}\right]_{\cdot1}^{\top}\right]_{\gamma'}. 
\end{align*}
By Lemma~2.6 in \cite{chen2021spectral}, we have 
\begin{align*}
    \left\Vert \left[\left[\bm{U}^{\star}\right]_{\cdot1}\left[\bm{U}^{\star}\right]_{\cdot1}^{\top}\right]_{\gamma}-\left[\left[\bm{U}^{\star}\right]_{\cdot1}\left[\bm{U}^{\star}\right]_{\cdot1}^{\top}\right]_{\gamma'}\right\Vert^2   &\ge \frac{1}{2}\left\Vert \left[\left[\bm{U}^{\star}\right]_{\cdot1}\right]_{\gamma}-\left[\left[\bm{U}^{\star}\right]_{\cdot1}\right]_{\gamma'}\right\Vert^2\wedge
\frac{1}{2}\left\Vert \left[\left[\bm{U}^{\star}\right]_{\cdot1}\right]_{\gamma}+\left[\left[\bm{U}^{\star}\right]_{\cdot1}\right]_{\gamma'}\right\Vert^2\\
&\ge \frac{\epsilon}{2r}\cdot \left|\{i:h_{i}\neq h'_{i}\}\right|\wedge 2(1-\epsilon).
\end{align*}
Using $|\{i:h_{i}\neq h'_{i}\}\ge r/8$ and $\epsilon<1/2$, this can be simplified to   
\[  \left\Vert \left[\left[\bm{U}^{\star}\right]_{\cdot1}\left[\bm{U}^{\star}\right]_{\cdot1}^{\top}\right]_{\gamma}-\left[\left[\bm{U}^{\star}\right]_{\cdot1}\left[\bm{U}^{\star}\right]_{\cdot1}^{\top}\right]_{\gamma'}\right\Vert^2 \ge \frac{\epsilon}{16}.\]

\paragraph{Proof of \eqref{eq:gamma_KL_2}.}

Let $\mathbb{P}_{k}^{(\gamma)}$ be the law of $\bm{A}_{k}$ under hypothesis $\gamma$
and $\mathbb{P}^{(\gamma)}$ be the joint law of $\{\bm{A}_{k}\}_{k=1}^{K}$
under $\gamma$. By independence and symmetry, it follows that for any $\gamma,\gamma'\in\Gamma$
\begin{equation}
\mathrm{KL}(\mathbb{P}^{(\gamma)}\parallel\mathbb{P}^{(\gamma')})=\sum_{k=1}^{K}\mathrm{KL}\left(\mathbb{P}_{k}^{(\gamma)}\parallel\mathbb{P}_{k}^{(\gamma')}\right)=K\cdot\mathrm{KL}\left(\mathbb{P}_{1}^{(\gamma)}\parallel\mathbb{P}_{1}^{(\gamma')}\right).\label{eq:KLall_KL1_2}
\end{equation}
Moreover, let $\widetilde{\mathbb{P}}_{k}^{(\gamma)}$ and $\widetilde{\mathbb{P}}_{k}^{(\gamma')}$ be the
law of $\widetilde{\bm{A}}_{k}$ under $\gamma$ and $\gamma'$ respectively. Lemma~\ref{lem:KL_trunc} shows that 
\begin{equation}\label{eq:KL2_trunc}
\mathrm{KL}\left(\mathbb{P}_{1}^{(\gamma)}\parallel\mathbb{P}_{1}^{(\gamma')}\right) \le \mathrm{KL}\left(\widetilde{\mathbb{P}}_{1}^{(\gamma)}\parallel\widetilde{\mathbb{P}}_{1}^{(\gamma')}\right) + 4 e^{-C_1 d},
\end{equation}
where $C_1$ is a universal constant. 

It now suffices to compute $\mathrm{KL}(\widetilde{\mathbb{P}}_{1}^{(\gamma)}\parallel \widetilde{\mathbb{P}}_{1}^{(\gamma')})$. Observe that the $j$-th column of $\bm{A}_k$ is 
\[[\widetilde{\bm{A}}_{k}]_{\cdot,j} = \bm{U}^\star ([\bm{V}_k^\star]_{j,\cdot})^{\top} + \bm{U}_k^\star ([\bm{W}_k^\star]_{j,\cdot})^{\top}+ [\bm{E}_k]_{\cdot,j}.\]
Recall that $\bm{V}_k^\star$ and $\bm{W}_k^\star$ are entrywise i.i.d. Gaussian with covariance $1/d$  and  $\bm{E}_k$ are entrywise i.i.d. Gaussian with covariance $1$. Then each column of $\widetilde{\bm{A}}_{k}$ is independent with 
\[[\widetilde{\bm{A}}_{k}]_{\cdot,j}\sim \mathcal{N}\left(\bm{0}, \bm{U}^\star \bm{U}^{\star\top} /d + \bm{U}_1^\star \bm{U}_1^{\star\top} /d + \sigma^2\bm{I}_{n}\right).\] 
The covariance of each column is the same, we may denote 
\[\bm{\Sigma}_i \coloneqq \sigma^2\left[\bm{I}_{n} + \frac{\bm{U}^\star \bm{U}^{\star\top} + \bm{U}_1^\star \bm{U}_1^{\star\top}}{\sigma^2 d} \right]_{\gamma_i}\]
for $i = 1, 2$. It is also easy to see that 
\[\bm{\Sigma}_i^{-1} = \frac{1}{\sigma^2}\left[\bm{I}_n - \frac{\bm{U}^\star \bm{U}^{\star\top} + \bm{U}_1^\star \bm{U}_1^{\star\top}}{\sigma^2 d + 1} \right]_{\gamma_i}.\]

By independence and the KL-divergence of multivariate Gaussian distributions, 
\[\mathrm{KL}(\widetilde{\mathbb{P}}_{1}^{(\gamma)}\parallel\widetilde{\mathbb{P}}_{1}^{(\gamma')}) 
= \frac{1}{2}\sum_{j = 1}^d \left(\mathrm{Trace}(\bm{\Sigma}_2^{-1} \bm{\Sigma}_1) - n - \log\frac{\mathrm{det}(\bm{\Sigma}_1)}{\mathrm{det}(\bm{\Sigma}_2)}\right).\]
We claim that 
\begin{equation}
    \mathrm{Trace}(\bm{\Sigma}_2^{-1} \bm{\Sigma}_1) \le n + \frac{8\epsilon \theta}{\sigma^2 d(\sigma^2 d+1)}. \label{eq:Trace_KL_2}    
\end{equation}
The proof is deferred to the end of this section. It is also clear from the form of $\bm{\Sigma}_1$ and $\bm{\Sigma}_2$ that they have the same eigenvalues and thus the same determinant. Then the KL-divergence can be simplified to
\[\mathrm{KL}(\widetilde{\mathbb{P}}_{1}^{(\gamma)}\parallel\widetilde{\mathbb{P}}_{1}^{(\gamma')}) 
 = \frac{1}{2} d \left(n + \frac{\epsilon \theta}{4\sigma^2 d(\sigma^2 d+1)}- n\right)
=   \frac{4 d\epsilon \theta}{\sigma^2 d(\sigma^2 d+1)}. \]
This combined with \eqref{eq:KLall_KL1_2} and \eqref{eq:KL2_trunc} finishes the proof.

\paragraph{Proof of \eqref{eq:Trace_KL_2}.}
For $\gamma_{i}$, $i=1,2$, denote $\bm{Y}_i \eqqcolon \bm{U}^{\star}\bm{U}^{\star\top}+\bm{U}_{1}^{\star}\bm{U}_{1}^{\star\top}$. Then we have that
\begin{equation}
    \bm{\Sigma}_2^{-1} \bm{\Sigma}_1 = \bm{I}_n 
- \frac{\bm{Y}_2}{\sigma^2 d +1}
+ \frac{\bm{Y}_1}{\sigma^2 d }
- \frac{\bm{Y}_2\bm{Y}_1}{\sigma^2 d (\sigma^2 d +1) }\label{eq:minimax_KL_Sigma_2}
\end{equation}
We now compute the trace of each term separately. It is clear that $\mathrm{Trace}(\bm{I}_n) = n$. 

 Without loss of generality consider $\bm{Y}_1$. It takes the form
\[\bm{Y}_1 = \left[\begin{array}{c|c}
\begin{array}{ccc}
    \bm{I}_r - \epsilon\theta \bm{e}_1\bm{e}_1^\top &  \theta\sqrt{\frac{\epsilon(1-\epsilon)}{r}}\bm{e}_1 \bm{h}^\top & -\sqrt{\frac{\epsilon\theta(1-\theta)}{r}}\bm{e}_1 \bm{h}^\top \\
    \theta\sqrt{\frac{\epsilon(1-\epsilon)}{r}} \bm{h} \bm{e}_1^\top & (1-\theta)\bm{I}_r - \frac{\epsilon\theta}{r}\bm{h}\bm{h}^\top & \sqrt{\theta(1-\theta)} (\bm{I}_r - \delta\bm{h}\bm{h}^\top)\\
    -\sqrt{\frac{\epsilon\theta(1-\theta)}{r}} \bm{h} \bm{e}_1^\top & \sqrt{\theta(1-\theta)} (\bm{I}_r - \delta\bm{h}\bm{h}^\top) & \theta\bm{I}_r \\
\end{array}
    & \bm{0}_{3r\times (d-3r)}\\
     \hline
    \bm{0}_{(n-3r)\times3r} & \bm{0}_{(n-3r)\times (d-3r)}\\
    
\end{array}\right].\]
For $\bm{Y}_i$, $i=1, 2$, we can see from the form of $\bm{Y}_i$ that
\begin{equation}
    \mathrm{Trace}(\bm{Y}_i) = (1-\epsilon\theta) + (r-1) + r\cdot (1-\theta+\theta\epsilon/r) + r \theta = 2r. \label{eq:minimax_trace_Y_2}
\end{equation}
For the trace of $\bm{Y_1}\bm{Y_2}$, we have
\begin{align*}
    \mathrm{Trace}(\bm{Y}_2 \bm{Y}_1) &= \left\langle \bm{Y}_1, \bm{Y}_2\right\rangle\\
    &= (1-\epsilon\theta)^2 + (r-1) + \frac{2\theta^2\epsilon (1-\epsilon) + 2\epsilon\theta(1-\theta)}{r}\langle \bm{h}, \bm{h}'\rangle 
+ r \cdot (1-\theta)^2 -2 \epsilon\theta(1-\theta) \\
    &\quad +\left(\frac{\epsilon \theta}{r}\right)^2\langle \bm{h}, \bm{h}' \rangle^2 + 2r  - 4 \delta r+ 2 \theta (1-\theta) \delta^2\langle \bm{h}, \bm{h}' \rangle^2 + r\theta^2.
\end{align*}
Observe that $\mathrm{Trace}(\bm{Y}_1 \bm{Y}_2) = 2r$ if $\bm{h} = \bm{h}'$. Then using the fact that $\langle \bm{h}, \bm{h}'\rangle\ge -r$ and $\langle \bm{h}, \bm{h}'\rangle^2 \ge 0$, we can rewrite the above equation as 
\begin{align*}
    \mathrm{Trace}(\bm{Y}_2 \bm{Y}_1) &\ge 2r -4 r \cdot \frac{\theta^2\epsilon (1-\epsilon)}{r} - 4 r\cdot \frac{\epsilon \theta(1-\theta)}{r} - r^2 \cdot \left(\frac{\epsilon \theta}{r}\right)^2 - 2r^2 \cdot \frac{\epsilon \theta (1-\theta)}{r^2}\ge  2r - 8 \epsilon\theta,
\end{align*}
which uses the fact that $\delta r = 1-\sqrt{1-\epsilon} \le \epsilon$. 
Combining \eqref{eq:minimax_KL_Sigma_2} with \eqref{eq:minimax_trace_Y_2} and the above inequality, we have \eqref{eq:Trace_KL_2}.

\section{Proof of Theorem~\ref{thm:oracle_lb} \label{sec:oracle_proof}}

In this section, we give an analysis that leads to an oracle algorithmic
lower bound. For notational
convenience we use the shorthand  $\rho\coloneqq \sigma_{\min}/(\sigma\sqrt{N})$ and abuse the big-O notation $O(X)$ to indicate any residual term $Y$ such that $\|Y\|\le CX$ for some large enough constant $C$. 
We prove a more general theorem that assumes $d_1 =\ldots = d_k \eqqcolon d$ but $d$ can differ from $n$. Theorem~\ref{thm:oracle_lb} is the special case when $n = d$.
\begin{theorem}\label{thm:oracle_full}
    Consider $\theta \leq 1/2$. Suppose $n$ is large enough, $r \leq n/3$, and $\rho\ge C_{1}$
for some large enough constant $C_{1}$. 
There exists a configuration of $\bm{U}^{\star}$, $\{\bm{U}_{k}^{\star}\}_{k=1}^{K}$,
$\{\bm{V}_{k}^{\star}\}_{k=1}^{K}$, $\{\bm{W}_{k}^{\star}\}_{k=1}^{K}$
such that with probability at least $1-O(KN^{-10})$, the oracle estimator~$\widehat{\bm{U}}$
output by Algorithm~\ref{alg:oracle} satisfies
\begin{equation}
\left\Vert \widehat{\bm{U}}\widehat{\bm{U}}^{\top}-\bm{U}^{\star}\bm{U}^{\star\top}\right\Vert \ge C_{2}\sigma^{4}nd-C_{3}\left[\frac{\log N}{\sqrt{K}}\cdot\rho^{-1}+\rho^{-5}\right] \label{eq:oracle_lb_full}
\end{equation}
for some constants $C_{2}, C_3>0$, with the proviso that $K\ge C_{4}\log^{2} n$ for some large constant $C_4 > 0$. 
\end{theorem}  

The proof can be summarized as five steps:
\begin{enumerate}
    \item We define a configuration of $\bm{U}^{\star}$, $\{\bm{U}_{k}^{\star}\}_{k=1}^{K}$,
$\{\bm{V}_{k}^{\star}\}_{k=1}^{K}$, $\{\bm{W}_{k}^{\star}\}_{k=1}^{K}$.

\item We use the singular subspace expansion in \cite{Xia2021}
to identify a fourth-order approximation $\bm{Q}$ such that $\bm{M}=\bm{Q}+O(\rho^{-5})$.
\item We show that SVD on $\mathbb{E}\bm{Q}$ results in a biased
estimation of $\bm{U}^{\star}$. 
\item We observe $\bm{Q}\rightarrow\mathbb{E}\bm{Q}$ as $K\rightarrow\infty$
and deduce that $\bm{M}=\mathbb{E}\bm{Q}+O(\rho^{-5})+O(K^{-1/2}\rho^{-1})$.
\item We combine Step~3 and 4 to reach the algorithmic
estimation lower bound.
\end{enumerate}

\paragraph{Specifying the configuration.}
Without loss of generality assume $k$ is even. Let $\bm{U}^{\star}$ be an arbitrary orthogonal matrix. For each $k\in[K]$, Define $\bm{U}_k^\star$ to be 
\[\bm{U}_{k}^{\star}=\begin{cases} \sqrt{1-\theta}\bm{Z}_{2}+\sqrt{\theta}\bm{Z}_{3}, &k\text{ is odd}\\
\sqrt{1-\theta}\bm{Z}_{2}-\sqrt{\theta}\bm{Z}_{3}, &k\text{ is even,}\end{cases}\]
where $\bm{Z}_{2},\bm{Z}_{3}$ are some orthogonal
matrices such that $\bm{Z}_{2}^{\top}\bm{Z}_{3}=\bm{U}^{\star\top}\bm{Z}_{2}=\bm{U}^{\star\top}\bm{Z}_{3}=\bm{0}_{r\times r}$.
It is straightforward to verify that $\|K^{-1}\sum_{k=1}^{K}\bm{U}_{k}^{\star}\bm{U}_{k}^{\star\top}\|=1-\theta$.

For the loading matrices, we let $\bm{V}^{\star}$ be an arbitrary orthogonal matrix and $\bm{W}^{\star}$
be $\bm{W}^{\star}=0.6\bm{V}^{\star}+0.8\bm{Z}_{1}$, where $\bm{Z}_{1}$
is some orthogonal matrix such that $\bm{V}^{\star\top}\bm{Z}_1=\bm{0}$.
This construction makes $\bm{V}^{\star\top}\bm{W}^{\star}=0.6\bm{I}_{r}$. For all
$k\in[K]$, we set $\bm{V}_{k}^{\star}=\sigma_{\min}\bm{V}^{\star}/\sqrt{0.4}$
and $\bm{W}_{k}^{\star}=\sigma_{\min}\bm{W}^{\star}/\sqrt{0.4}$.
It can be verified that $\sigma_{2r}(\bm{A}_{k}^{\star})=\sigma_{\min}.$

From this point forward, we assume without loss of generality that $\sigma_{\min}=\sqrt{0.4}$,
so that $\bm{V}_{k}^{\star}=\bm{V}^{\star}$ and
$\bm{W}_{k}^{\star}=\bm{W}^{\star}$ for all $k\in[K]$. 

\paragraph{Identifying $\bm{Q}$.}

We establish a fourth-order approximation for $\widehat{\bm{U}}_{k}\widehat{\bm{W}}_{k}^{\top}$
by utilizing the expansion presented in \cite{Xia2021} (see Section~\ref{subsec:SVD_approx}
for a brief introduction). 

\begin{lemma}\label{lem:order4_approx} Suppose $\rho\le C_{1}$
for some small enough constant $C_{1}$. Then with probability at
least $1-O(KN^{-10})$,
\begin{equation}
\bm{A}_{k}-\widehat{\bm{U}}_{k}\widehat{\bm{W}}_{k}^{\top}=\bm{T}_{0,k}+\bm{T}_{1,k}+\bm{T}_{2,k}+\bm{T}_{3,k}+\bm{T}_{4,k}+O(\rho^{-5}),\label{eq:AK-UW_decomp}
\end{equation}
where $\bm{T}_{i,k}$ is an $i$-th degree polynomial of the noise
matrix $\bm{E}_{k}$ and $\|\bm{T}_{i,k}\|=O(\rho^{-i})$.

\end{lemma} 
The proof of this lemma is deferred to Section~\ref{subsec:poly4_approx}. Using this lemma, we can see that for each $k\in[K]$,
\begin{equation}
\left(\bm{A}_{k}-\widehat{\bm{U}}_{k}\widehat{\bm{W}}_{k}^{\top}\right)\left(\bm{A}_{k}-\widehat{\bm{U}}_{k}\widehat{\bm{W}}_{k}^{\top}\right)^{\top}=\sum_{i=0}^{4}\sum_{j=0}^{4-i}\bm{T}_{i,k}\bm{T}_{j,k}^{\top}+O(\rho^{-5}).\label{eq:A-UW_outer_approx}
\end{equation}
We may now define the fourth-order approximation to be 
\begin{equation}
\bm{Q}\coloneqq\frac{1}{K}\sum_{k=1}^{K}\sum_{i=0}^{4}\sum_{j=0}^{4-i}\bm{T}_{i,k}\bm{T}_{j,k}^{\top}.\label{eq:Q_def}
\end{equation}
From \eqref{eq:A-UW_outer_approx} and \eqref{eq:def_M}, we deduce
that $\bm{M}=\bm{Q}+O(\rho^{-5})$.

\paragraph{Biased estimation of $\mathbb{E}\bm{Q}$.}

We characterize what SVD would achieve on the expectation
of $\bm{Q}$. We start with a precise characterization of $\mathbb{E}\bm{Q}$
in the following lemma. The proof is deferred to Section~\ref{sec:Analysis_Q}.

\begin{lemma}\label{lem:EQ} Instate the assumptions of Theorem~\ref{thm:oracle_full}.
Let $\bm{Q}$ be defined as in \eqref{eq:Q_def}. Then
\begin{align}
\mathbb{E}\bm{Q} & =(1+\alpha_{1})\bm{U}^{\star}\bm{U}^{\star\top}+\alpha_{2}\bm{I}_{n}+\alpha_{3}\cdot\frac{1}{K}\sum_{k=1}^{K}\bm{U}_{k}^{\star}\bm{U}_{k}^{\star\top}\label{eq:EQ}\\
 & \quad+\alpha_{4}\cdot\frac{1}{K}\sum_{k=1}^{K}\left(\bm{U}^{\star}\bm{U}_{k}^{\star\top}+\bm{U}_{k}^{\star}\bm{U}^{\star\top}\right),\nonumber 
\end{align}
where $\{\alpha_{i}\}_{i=1}^{4}$ are real numbers satisfying $|\alpha_{i}|\le C_{1}\rho^{-2}$
for $i=1,2,3$ and $\alpha_{4}\in[C_{2}\sigma^{4}nd,C_{3}\sigma^{4}nd]$
for some constants $C_{1},C_{2},C_{3}>0$ and $C_{2}<C_{3}$.

\end{lemma}

The cross term $\bm{U}^{\star}\bm{U}_{k}^{\star\top}$ 
is simplified from $\bm{U}^{\star}\bm{V}^{\star\top}\bm{W}^{\star}\bm{U}_{k}^{\star\top}$.
By picking $\bm{V}^{\star},\bm{W}^{\star}$ not orthogonal to each other, we introduce a non-trivial
cross term that leads to biased estimation. Let $\widetilde{\bm{U}}\in\mathbb{R}^{n\times r}$
be the matrix whose columns are the top-$r$ eigenvectors of
$\mathbb{E}\bm{Q}$. The following lemma formally
demonstrates this induced bias. 

\begin{lemma}\label{lem:eigen_EQ} Instate the assumptions of Theorem~\ref{thm:oracle_full}.
Let $\widetilde{\bm{U}}\widetilde{\bm{U}}^{\top}$ be the eigen space
of $\mathbb{E}\bm{Q}$ and $\alpha_{1},\alpha_{4}$ be the scalars
appeared in Lemma~\ref{lem:EQ}. Then 
\[
\widetilde{\bm{U}}\widetilde{\bm{U}}^{\top}=\bm{U}^{\star}\bm{U}^{\star\top}+\frac{\alpha_{4}}{1+\alpha_{1}}\cdot\frac{1}{K}\sum_{k=1}^{K}\left(\bm{U}^{\star}\bm{U}_{k}^{\star\top}+\bm{U}_{k}^{\star}\bm{U}^{\star\top}\right)+O(\rho^{-6}).
\]

\end{lemma}

\paragraph{Proximity of $\bm{M}$ and $\mathbb{E}\bm{Q}$.}

The following lemma shows that $\bm{M}$ is close to $\mathbb{E}\bm{Q}$.

\begin{lemma}\label{lem:dist_M_EQ}Instate the assumptions of Theorem~\ref{thm:oracle_full},
with probability at least $1-O(KN^{-10})$,
\[
\left\Vert \bm{M}-\mathbb{E}\bm{Q}\right\Vert \le C_{1}\left[\frac{\log N}{\sqrt{K}}\cdot\rho^{-1}+\rho^{-5}\right]
\]
for some constant $C_{1}>0$.

\end{lemma} 

\paragraph{Proof of the lower bound \eqref{eq:oracle_lb_full}.}

We assume that the event that Lemma~\ref{lem:order4_approx},
\ref{lem:eigen_EQ}, and \ref{lem:dist_M_EQ} hold, which happens
with probability at least $1-O(KN^{-10})$. We first bound $\|\widehat{\bm{U}}\widehat{\bm{U}}^{\top}-\widetilde{\bm{U}}\widetilde{\bm{U}}^{\top}\|$
and then $\|\widetilde{\bm{U}}\widetilde{\bm{U}}^{\top}-\bm{U}^{\star}\bm{U}^{\star\top}\|$.

By Wedin's theorem, 
\begin{equation}
\left\Vert \widehat{\bm{U}}\widehat{\bm{U}}^{\top}-\widetilde{\bm{U}}\widetilde{\bm{U}}^{\top}\right\Vert \le\frac{\sqrt{2}\left\Vert \bm{M}-\mathbb{E}\bm{Q}\right\Vert }{\sigma_{r}(\mathbb{E}\bm{Q})-\sigma_{r+1}(\mathbb{E}\bm{Q})-\left\Vert \bm{M}-\mathbb{E}\bm{Q}\right\Vert }.\label{eq:wedin}
\end{equation}
Take $\bm{U}^{\star}\bm{U}^{\star\top}$
as the ground truth matrix and $\mathbb{E}\bm{Q}-\bm{U}^{\star}\bm{U}^{\star\top}$
as the perturbation. By Weyl's inequality and \eqref{eq:EQ}, 
\begin{align*}
\sigma_{r}(\mathbb{E}\bm{Q})-\sigma_{r+1}(\mathbb{E}\bm{Q}) & \ge1-2\left\Vert \mathbb{E}\bm{Q}-\bm{U}^{\star}\bm{U}^{\star\top}\right\Vert \\
 & \ge1-2\left(\alpha_{1}+\alpha_{2}+\alpha_{3}+2\alpha_{4}\right)\\
 & \ge\frac{3}{4},
\end{align*}
where the last line holds as long as $\rho\ge C_{1}$ for
some large enough constant $C_{1}$. More over, by assumption of Theorem~\ref{thm:oracle_full},
$K\ge C_{1}\log^{2}N$ and $\rho\ge C_{2}$ for some large
enough constant $C_{1},C_{2}$. Then
\[
\|\bm{M}-\mathbb{E}\bm{Q}\|\le C_{1}\left[\frac{\log N}{\sqrt{K}}\cdot\rho^{-1}+\rho^{-5}\right]\le\frac{1}{8}+\frac{1}{8}\le1/4.
\]
Applying these bounds and Lemma~\ref{lem:dist_M_EQ} to \eqref{eq:wedin},
we have
\[
\left\Vert \widehat{\bm{U}}\widehat{\bm{U}}^{\top}-\widetilde{\bm{U}}\widetilde{\bm{U}}^{\top}\right\Vert \le\frac{\sqrt{2}\left\Vert \bm{M}-\mathbb{E}\bm{Q}\right\Vert }{1/4}\le C_{3}\left[\frac{\log N}{\sqrt{K}}\cdot\rho^{-1}+\rho^{-5}\right]
\]
for some constant $C_{3}>0$, 

On the other hand, by Lemma~\ref{lem:eigen_EQ}, 
\begin{align*}
\left\Vert \widetilde{\bm{U}}\widetilde{\bm{U}}^{\top}-\bm{U}^{\star}\bm{U}^{\star\top}\right\Vert  & \ge\left\Vert \frac{\alpha_{4}}{1+\alpha_{1}}\cdot\frac{1}{K}\sum_{k=1}^{K}\left(\bm{U}^{\star}\bm{U}_{k}^{\star\top}+\bm{U}_{k}^{\star}\bm{U}^{\star\top}\right)\right\Vert -O(\rho^{-6}).
\end{align*}
We lower bound the main term:
\begin{align*}
\left\Vert \frac{1}{K}\sum_{k=1}^{K}\left(\bm{U}^{\star}\bm{U}_{k}^{\star\top}+\bm{U}_{k}^{\star}\bm{U}^{\star\top}\right)\right\Vert  & \ge\sup_{\bm{v}\in\mathbb{R}^{r}:\|\bm{v}\|\le1}\left\Vert \frac{1}{K}\sum_{k=1}^{K}\left(\bm{U}^{\star}\bm{U}_{k}^{\star\top}+\bm{U}_{k}^{\star}\bm{U}^{\star\top}\right)\bm{U}^{\star}\bm{v}\right\Vert \\
 & =\sup_{\bm{v}\in\mathbb{R}^{r}:\|\bm{v}\|\le1}\left\Vert \frac{1}{K}\sum_{k=1}^{K}\bm{U}_{k}^{\star}\bm{v}\right\Vert \\
 & \overset{\text{(i)}}{=}\sup_{\bm{v}\in\mathbb{R}^{r}:\|\bm{v}\|\le1}\left\Vert \sqrt{1-\theta}\bm{Z}_{2}\bm{v}\right\Vert \\
 & =\sqrt{1-\theta}.
\end{align*}
Here (i) follows from the definition of $\bm{U}_{k}^{\star}$.
Recall that $|\alpha_{1}|\le C_{4}\rho^{-2}$ and $|\alpha_{4}|\ge C_{5}\sigma^{4}nd$
for some constants $C_{4},C_{5}$. Assuming $\theta\le1/2$ and $\rho\ge C_{2}$
for some large enough constant $C_{2}$, we have
\begin{align*}
\left\Vert \widetilde{\bm{U}}\widetilde{\bm{U}}^{\top}-\bm{U}^{\star}\bm{U}^{\star\top}\right\Vert  & \ge\frac{\alpha_{4}}{1+\alpha_{1}}\sqrt{1-\theta}-O(\rho^{-6})\\
 & \ge C_{6}\sigma^{4}nd-O(\rho^{-6})
\end{align*}
for some constant $C_{6}>0$.

Combining the lower bound of $\|\widehat{\bm{U}}\widehat{\bm{U}}^{\top}-\widetilde{\bm{U}}\widetilde{\bm{U}}^{\top}\|$
and $\|\widetilde{\bm{U}}\widetilde{\bm{U}}^{\top}-\bm{U}^{\star}\bm{U}^{\star\top}\|$,
we conclude that 
\begin{align*}
\left\Vert \widehat{\bm{U}}\widehat{\bm{U}}^{\top}-\bm{U}^{\star}\bm{U}^{\star\top}\right\Vert  & \ge\left\Vert \widetilde{\bm{U}}\widetilde{\bm{U}}^{\top}-\bm{U}^{\star}\bm{U}^{\star\top}\right\Vert -\left\Vert \widehat{\bm{U}}\widehat{\bm{U}}^{\top}-\widetilde{\bm{U}}\widetilde{\bm{U}}^{\top}\right\Vert \\
 & \ge C_{6}\sigma^{4}nd-O(\rho^{-6})-C_{3}\left[\frac{\log N}{\sqrt{K}}\cdot\rho^{-1}+\rho^{-5}\right]\\
 & \ge C_{6}\sigma^{4}nd-C_{3}\left[\frac{\log N}{\sqrt{K}}\cdot\rho^{-1}+\rho^{-5}\right]
\end{align*}
as long as $\rho\ge C_{2}$ for some large enough constant
$C_{2}$.

\subsection{Approximation of SVD\label{subsec:SVD_approx}}

In this section, we give a brief explanation of the
approximation of SVD in \cite{Xia2021}. This is based on Theorem~1
and Section~3 of \cite{Xia2021} Here we introduce it for asymmetric
matrices. The symmetric version is more striaightforward since Theorem~1
in \cite{Xia2021} would be applicable. All the notations in this
section works as a generic result and is not related to our problem
setting. 

Let $\bm{A}=\bm{X}\bm{\Sigma}\bm{Y}^{\top}\in\mathbb{R}^{n\times d}$
be a rank-$r$ matrix and its SVD. Let 
\[
\bm{P}^{-1}=\begin{bmatrix}\bm{0} & \bm{X}\bm{\Sigma}^{-1}\bm{Y}^{\top}\\
\bm{Y}\bm{\Sigma}^{-1}\bm{X}^{\top} & \bm{0}
\end{bmatrix},\qquad\text{and}\qquad\bm{P}^{\perp}=\begin{bmatrix}\bm{I}-\bm{X}\bm{X}^{\top} & \bm{0}\\
\bm{0} & \bm{I}-\bm{Y}\bm{Y}^{\top}
\end{bmatrix}.
\]
Overloading the notation we also use $\bm{P}^{0}$ to denote $\bm{P}^{\perp}$.
Furthermore 
\[
\bm{P}^{-2k}=\begin{bmatrix}\bm{X}\bm{\Sigma}^{-2k}\bm{X}^{\top} & \bm{0}\\
\bm{0} & \bm{Y}\bm{\Sigma}^{-2k}\bm{Y}^{\top}
\end{bmatrix},\qquad\text{and}\qquad\bm{P}^{-2k-1}=\begin{bmatrix}\bm{0} & \bm{X}\bm{\Sigma}^{-2k-1}\bm{Y}^{\top}\\
\bm{Y}\bm{\Sigma}^{-2k-1}\bm{X}^{\top} & \bm{0}
\end{bmatrix}
\]
for any integer $k$. Let $\bm{E}$ be error matrix and 
\[
\widetilde{\bm{E}}=\begin{bmatrix}\bm{0} & \bm{E}\\
\bm{E}^{\top} & \bm{0}
\end{bmatrix}
\]
be its dilation. Then let $\widehat{\bm{X}}\widehat{\bm{\Sigma}}\widehat{\bm{Y}}^{\top}$
be the SVD of $\bm{M}+\bm{E}$, we have that $\mathcal{S}_{i}$
\begin{align*}
\begin{bmatrix}\widehat{\bm{X}}\widehat{\bm{X}}^{\top} & \bm{0}\\
\bm{0} & \widehat{\bm{Y}}\widehat{\bm{Y}}^{\top}
\end{bmatrix} & =\begin{bmatrix}\bm{X}\bm{X}^{\top} & \bm{0}\\
\bm{0} & \bm{Y}\bm{Y}^{\top}
\end{bmatrix}+\sum_{i=1}^{\infty}\mathcal{S}_{i}
\end{align*}
where 
\begin{align*}
\mathcal{S}_{i} & =\sum_{\bm{\alpha}\in\mathbb{N}^{i+1},\sum_{j=1}^{i+1}\alpha_{j}=i}(-1)^{|\{j:\alpha_{j}\neq0\}|+1}\bm{P}^{-\alpha_{1}}\widetilde{\bm{E}}\bm{P}^{-\alpha_{2}}\ldots\bm{P}^{-\alpha_{i}}\widetilde{\bm{E}}\bm{P}^{-\alpha_{i+1}}.
\end{align*}
In addition, it can be computed that 
\[
\|\mathcal{S}_{i}\|\le\left(\frac{4\|\bm{E}\|}{\lambda_{r}(\bm{A})}\right)^{i}.
\]

\subsection{Proof of Lemma~\ref{lem:order4_approx} \label{subsec:poly4_approx}}

In this section we give an overview of the fourth order expansion
of $\bm{A}_{k}-\widehat{\bm{U}}_{k}\widehat{\bm{W}}_{k}^{\top}$.
To apply the analysis in Section~\ref{subsec:SVD_approx}, we define
the following matrices. Let 
\[
\widetilde{\bm{E}}_{k}\coloneqq\begin{bmatrix}\bm{0} & \bm{E}_{k}\\
\bm{E}_{k}^{\top} & \bm{0}
\end{bmatrix}\qquad\text{and}\qquad\bm{P}_{k}^{0}\coloneqq\bm{P}_{k}^{\perp}=\begin{bmatrix}\bm{I}-\bm{U}_{k}^{\star}\bm{U}_{k}^{\star\top} & \bm{0}\\
\bm{0} & \bm{I}-\bm{W}^{\star}\bm{W}^{\star\top}
\end{bmatrix}.
\]
By Theorem 4.4.5 in \cite{vershynin2018high}, as long as $\log N\le C_{1}N$
for some constant $C_{1}$, with probability at least $1-N^{-10}$,
\[
\|\widetilde{\bm{E}}_{k}\|=\|\bm{E}_{k}\|\le C_{2}\sigma\sqrt{n\vee d_{k}}\le C_{2}\sigma\sqrt{N}=O(\rho^{-1})
\]
for some large enough constant $C_{2}>0$. For any positive integer
$s$, let 
\[
\bm{P}_{k}^{-2s}=\begin{bmatrix}\bm{U}_{k}^{\star}\bm{U}_{k}^{\star\top} & \bm{0}\\
\bm{0} & \bm{W}^{\star}\bm{W}^{\star\top}
\end{bmatrix}\qquad\text{and}\qquad\bm{P}_{k}^{-2s-1}=\begin{bmatrix}\bm{0} & \bm{U}_{k}^{\star}\bm{W}^{\star\top}\\
\bm{W}^{\star}\bm{U}_{k}^{\star\top} & \bm{0}
\end{bmatrix}.
\]
Note that $\|\bm{P}_{k}^{-2s}\|=\|\bm{P}_{k}^{-2s-1}\|=1$. Using
the approximation of SVD in \cite{Xia2021} we described in Section~\ref{subsec:SVD_approx}
on the ground truth matrix $\bm{U}_{k}^{\star}\bm{W}^{\star\top}$
and its perturbation $\mathcal{P}_{U}^{\perp}\bm{A}_{k}=\bm{U}_{k}^{\star}\bm{W}^{\star\top}+\mathcal{P}_{U}^{\perp}\bm{E}_{k}$,
we have that $\widehat{\bm{U}}_{k}\widehat{\bm{U}}_{k}^{\top}=\widetilde{\bm{U}}_{k}\widetilde{\bm{U}}_{k}^{\top}+O(\rho^{-5})$
for 
\[
\widehat{\bm{U}}_{k}\widehat{\bm{U}}_{k}^{\top}=\bm{U}_{k}^{\star}\bm{U}_{k}^{\star\top}+\sum_{i=1}^{4}\bm{N}_{i,k}+O(\rho^{-5})
\]
where 
\begin{equation}
\bm{N}_{i,k}\coloneqq\left[\sum_{\bm{\alpha}\in\mathbb{N}^{i+1},\sum_{j=1}^{i+1}\alpha_{j}=i}(-1)^{|\{j:\alpha_{j}\neq0\}|+1}\bm{P}_{k}^{-\alpha_{1}}\widetilde{\bm{E}}_{k}\bm{P}_{k}^{-\alpha_{2}}\ldots\bm{P}_{k}^{-\alpha_{i}}\widetilde{\bm{E}}_{k}\bm{P}_{k}^{-\alpha_{i+1}}\right]_{1:n,1:n}.\label{eq:def_N}
\end{equation}
Since $\|\bm{P}_{k}^{-2s}\|=\|\bm{P}_{k}^{-2s-1}\|=1$, $\|\bm{N}_{i,k}\|\le\|\widetilde{\bm{E}}_{k}\|=O(\rho^{-i})$.
Now consider $\bm{A}_{k}-\widehat{\bm{U}}_{k}\widehat{\bm{W}}_{k}^{\top}$
. We use the approximation to reach that
\begin{align*}
\bm{A}_{k}-\widehat{\bm{U}}_{k}\widehat{\bm{W}}_{k}^{\top} & =\bm{A}_{k}-\widehat{\bm{U}}_{k}\widehat{\bm{U}}_{k}^{\top}\mathcal{P}_{U}^{\perp}\bm{A}_{k}\\
 & =\bm{U}^{\star}\bm{V}^{\star\top}+\bm{U}_{k}^{\star}\bm{W}^{\star\top}+\bm{E}_{k}-\left(\bm{U}_{k}^{\star}\bm{U}_{k}^{\star\top}+\sum_{i=1}^{4}\bm{N}_{i,k}\right)\left(\bm{U}_{k}^{\star}\bm{W}^{\star\top}+\mathcal{P}_{U}^{\perp}\bm{E}_{k}\right)+O(\rho^{-5}).
\end{align*}
Renaming the terms with their respective degrees (number of $\bm{E}_{k}$),
we can write this approximation as 
\[
\bm{A}_{k}-\widehat{\bm{U}}_{k}\widehat{\bm{W}}_{k}^{\top}=\bm{T}_{0,k}+\bm{T}_{1,k}+\bm{T}_{2,k}+\bm{T}_{3,k}+\bm{T}_{4,k}+O(\rho^{-5}),
\]
where 
\begin{align}
\bm{T}_{0,k} & \coloneqq\bm{U}^{\star}\bm{V}^{\star\top}\nonumber \\
\bm{T}_{1,k} & \coloneqq\mathcal{P}_{U_{k}}^{\perp}\bm{E}_{k}-\bm{N}_{1,k}\bm{U}_{k}^{\star}\bm{W}^{\star\top}\nonumber \\
\bm{T}_{2,k} & \coloneqq-\bm{N}_{1,k}\mathcal{P}_{U}^{\perp}\bm{E}_{k}-\bm{N}_{2,k}\bm{U}_{k}^{\star}\bm{W}^{\star\top}\label{eq:def_T}\\
\bm{T}_{3,k} & \coloneqq-\bm{N}_{2,k}\mathcal{P}_{U}^{\perp}\bm{E}_{k}-\bm{N}_{3,k}\bm{U}_{k}^{\star}\bm{W}^{\star\top}\nonumber \\
\bm{T}_{4,k} & \coloneqq-\bm{N}_{3,k}\mathcal{P}_{U}^{\perp}\bm{E}_{k}-\bm{N}_{4,k}\bm{U}_{k}^{\star}\bm{W}^{\star\top}.\nonumber 
\end{align}
Note that the term $-\bm{N}_{4,k}\mathcal{P}_{U}^{\perp}\bm{E}_{k}$
is hidden in the residual term $O(\rho^{-5})$. Moreover we
have that $\|\bm{T}_{i,k}\|=O(\rho^{-i})$.

\subsection{Proof of Lemma~\ref{lem:eigen_EQ} \label{subsec:proof_eigenspace}}

Lemma~\ref{lem:EQ} tells us that 
\begin{align*}
\mathbb{E}\bm{Q} & =(1+\alpha_{1})\bm{U}^{\star}\bm{U}^{\star\top}+\alpha_{2}\bm{I}_{n}+\alpha_{3}\cdot\frac{1}{K}\sum_{k=1}^{K}\bm{U}_{k}^{\star}\bm{U}_{k}^{\star\top}\\
 & \quad+\alpha_{4}\cdot\frac{1}{K}\sum_{k=1}^{K}\left(\bm{U}^{\star}\bm{U}_{k}^{\star\top}+\bm{U}_{k}^{\star}\bm{U}^{\star\top}\right),
\end{align*}
where $\{\alpha_{i}\}_{i=1}^{4}$ are real numbers such that $|\alpha_{i}|\le C_{1}\sigma^{2}(\sqrt{nd}+n)$
for $i=1,2,3$ and $\alpha_{4}\in[C_{2}\sigma^{4}nd,C_{3}\sigma^{4}nd]$
for some constants $C_{1},C_{2},C_{3}>0$ and $C_{2}<C_{3}$. Treat
$(1+\alpha_{1})\bm{U}^{\star}\bm{U}^{\star\top}$ as the ground truth
and 
\begin{align*}
\bm{E}_{U} & \coloneqq\alpha_{2}\bm{I}_{n}+\alpha_{3}\cdot\frac{1}{K}\sum_{k=1}^{K}\bm{U}_{k}^{\star}\bm{U}_{k}^{\star\top}+\alpha_{4}\cdot\frac{1}{K}\sum_{k=1}^{K}\left(\bm{U}^{\star}\bm{U}_{k}^{\star\top}+\bm{U}_{k}^{\star}\bm{U}^{\star\top}\right)
\end{align*}
as the perturbation. It is easy to see that $\|\bm{E}_{U}\|\le C_{3}\rho^{-2}\le1/8$
as long as $\rho^{-1}\le C_{4}$ for some large enough constant
$C_{3},C_{4}>0$. Moreover, for $j=1,2$, let $\bm{P}^{-j}\coloneqq(1+\alpha_{1})^{-j}\bm{U}^{\star}\bm{U}^{\star\top}$
and $\bm{P}^{\perp}\coloneqq\bm{I}-\bm{U}^{\star}\bm{U}^{\star\top}$.
Now we may apply Theorem~1 in \cite{Xia2021} to get that 
\begin{align}
\widetilde{\bm{U}}\widetilde{\bm{U}}^{\top} & =\bm{U}^{\star}\bm{U}^{\star\top}+\bm{P}^{-1}\bm{E}_{U}\bm{P}^{\perp}+\bm{P}^{\perp}\bm{E}_{U}\bm{P}^{-1}\label{eq:EQ_SVD_expansion}\\
 & \quad+\bm{P}^{-2}\bm{E}_{U}\bm{P}^{\perp}\bm{E}_{U}\bm{P}^{\perp}+\bm{P}^{\perp}\bm{E}_{U}\bm{P}^{-2}\bm{E}_{U}\bm{P}^{\perp}+\bm{P}^{\perp}\bm{E}_{U}\bm{P}^{\perp}\bm{E}_{U}\bm{P}^{-2}\nonumber \\
 & \quad-\bm{P}^{-1}\bm{E}_{U}\bm{P}^{-1}\bm{E}_{U}\bm{P}^{\perp}-\bm{P}^{\perp}\bm{E}_{U}\bm{P}^{-1}\bm{E}_{U}\bm{P}^{-1}-\bm{P}^{-1}\bm{E}_{U}\bm{P}^{\perp}\bm{E}_{U}\bm{P}^{-1}+O(\|\bm{E}_{U}\|^{3}).\nonumber 
\end{align}
Observe that for $j=1,2$,
\[
\left\Vert \bm{P}^{-j}\bm{E}_{U}\bm{P}^{\perp}\right\Vert =\left\Vert \frac{\alpha_{4}}{(1+\alpha_{1})^{j}}\cdot\frac{1}{K}\sum_{k=1}^{K}\bm{U}^{\star}\bm{U}_{k}^{\star\top}\right\Vert \le\alpha_{4}.
\]
Then we can simplify the expansion \eqref{eq:EQ_SVD_expansion} as
\begin{align*}
\widetilde{\bm{U}}\widetilde{\bm{U}}^{\top} & =\bm{U}^{\star}\bm{U}^{\star\top}+\frac{\alpha_{4}}{1+\alpha_{1}}\cdot\frac{1}{K}\sum_{k=1}^{K}\left(\bm{U}^{\star}\bm{U}_{k}^{\star\top}+\bm{U}_{k}^{\star}\bm{U}^{\star\top}\right)+O(\alpha_{4}\|\bm{E}_{U}\|)+O(\|\bm{E}_{U}\|^{3})\\
 & =\bm{U}^{\star}\bm{U}^{\star\top}+\frac{\alpha_{4}}{1+\alpha_{1}}\cdot\frac{1}{K}\sum_{k=1}^{K}\left(\bm{U}^{\star}\bm{U}_{k}^{\star\top}+\bm{U}_{k}^{\star}\bm{U}^{\star\top}\right)+O\left(\rho^{-6}\right).
\end{align*}
The proof is now completed.

\subsection{Proof of Lemma~\ref{lem:dist_M_EQ} \label{subsec:proof_concentration}}

Recall that 
\begin{align*}
\bm{M} & =\frac{1}{K}\sum_{k=1}^{K}\left(\bm{A}_{k}-\widehat{\bm{U}}_{k}\widehat{\bm{W}}_{k}^{\top}\right)\left(\bm{A}_{k}-\widehat{\bm{U}}_{k}\widehat{\bm{W}}_{k}^{\top}\right)^{\top}.\\
 & =\bm{Q}+O(\rho^{-5}),
\end{align*}
where 
\[
\bm{Q}=\frac{1}{K}\sum_{k=1}^{K}\sum_{i=0}^{4}\sum_{j=0}^{4-i}\bm{T}_{i,k}\bm{T}_{j,k}^{\top}.
\]
It suffices to show $\|\bm{Q}-\mathbb{E}\bm{Q}\|=O(1/\sqrt{K})$.
The matrix $\bm{Q}$ is a sum of independent random matrices $\sum_{i=0}^{4}\sum_{j=0}^{4-i}\bm{T}_{i,k}\bm{T}_{j,k}^{\top}$,
so we may use the matrix Bernstein inequality. We claim that for each
$k\in[K]$, as long as $N\ge C_{1}\log N$ for some large enough constant
$C_{1}>0$, with probability at least $1-O(N^{-11})$,\begin{subequations}

\begin{equation}
\left\Vert \sum_{i=0}^{4}\sum_{j=0}^{4-i}\left(\bm{T}_{i,k}\bm{T}_{j,k}^{\top}-\mathbb{E}\bm{T}_{i,k}\bm{T}_{j,k}^{\top}\right)\right\Vert \le C_{2}\rho^{-1};\label{eq:norm_Q_Bern}
\end{equation}
and 
\begin{equation}
\left\Vert \sum_{k=1}^{K}\sum_{i=0}^{4}\sum_{j=0}^{4-i}\left(\bm{T}_{i,k}\bm{T}_{j,k}^{\top}-\mathbb{E}\bm{T}_{i,k}\bm{T}_{j,k}^{\top}\right);\left\Vert \sum_{i=0}^{4}\sum_{j=0}^{4-i}\left(\bm{T}_{i,k}\bm{T}_{j,k}^{\top}-\mathbb{E}\bm{T}_{i,k}\bm{T}_{j,k}^{\top}\right)\right\Vert >C_{2}\rho^{-1}\right\Vert \le C_{2}\rho^{-1}N^{-11}.\label{eq:Q_Bern_trunc}
\end{equation}
for some constant $C_{2}>0$. Moreover,
\begin{equation}
\left\Vert \mathbb{E}\sum_{k=1}^{K}\left[\sum_{i=0}^{4}\sum_{j=0}^{4-i}\left(\bm{T}_{i,k}\bm{T}_{j,k}^{\top}-\mathbb{E}\bm{T}_{i,k}\bm{T}_{j,k}^{\top}\right)\right]\left[\sum_{i=0}^{4}\sum_{j=0}^{4-i}\left(\bm{T}_{i,k}\bm{T}_{j,k}^{\top}-\mathbb{E}\bm{T}_{i,k}\bm{T}_{j,k}^{\top}\right)\right]^{\top}\right\Vert \le C_{2}^{2}K\cdot\rho^{-2}.\label{eq:Q_Bern_var}
\end{equation}
and \end{subequations}Now invoking the truncated matrix Bernstein
inequality (see \cite{chen2021spectral}, Corollary~3.2), we have
that with probability at least $1-O(KN^{-10})$,
\[
\|\bm{Q}-\mathbb{E}\bm{Q}\|\le\frac{C_{3}\log N}{\sqrt{K}}\cdot\rho^{-1}
\]
for some large enough constant $C_{3}>0$.

We now prove \eqref{eq:norm_Q_Bern} to \eqref{eq:Q_Bern_var}. It
is easy to see that \eqref{eq:Q_Bern_var} is a direct consequence
of \eqref{eq:norm_Q_Bern}. It now suffices to show \eqref{eq:norm_Q_Bern}.
By Theorem 4.4.5 in \cite{vershynin2018high}, as long as $\log N\le C_{1}N$
for some constant $C_{1}$, with probability at least $1-N^{-11}$,
\[
\|\bm{E}_{k}\|\le C_{4}\sigma\sqrt{n\vee d_{k}}\le C_{4}\sigma\sqrt{N}=C_{4}\rho^{-1}.
\]
Since $\bm{T}_{0,k}=\bm{U}^{\star}\bm{V}^{\star\top}$ is not random,
\[
\sum_{i=0}^{4}\sum_{j=0}^{4-i}\left(\bm{T}_{i,k}\bm{T}_{j,k}^{\top}-\mathbb{E}\bm{T}_{i,k}\bm{T}_{j,k}^{\top}\right)=\sum_{(i,j):1\le i+j\le4}\left(\bm{T}_{i,k}\bm{T}_{j,k}^{\top}-\mathbb{E}\bm{T}_{i,k}\bm{T}_{j,k}^{\top}\right).
\]
From the form of $\bm{T}_{i,k}$ we discussed in Section~\ref{subsec:expression_T},
we observe that
\[
\|\bm{T}_{i,k}\bm{T}_{j,k}^{\top}\|\le\|\bm{E}_{k}\|^{i+j}\le(C_{4}\rho^{-1})^{i+j}\le C_{4}\cdot\rho^{-1}
\]
 as long as $\rho\ge1/C_{4}$. Therefore 
\[
\left\Vert \sum_{i=0}^{4}\sum_{j=0}^{4-i}\left(\bm{T}_{i,k}\bm{T}_{j,k}^{\top}-\mathbb{E}\bm{T}_{i,k}\bm{T}_{j,k}^{\top}\right)\right\Vert \le C_{2}\cdot\rho^{-1}
\]
for some constant $C_{2}>0$.

Finally \eqref{eq:Q_Bern_trunc} can be proved similar to \eqref{eq:EET-dI_trunc}.
We omit it here for conciseness.

\section{Proof of Lemma~\ref{lem:EQ} \label{sec:Analysis_Q}}

In this section we give the proof of Lemma~\ref{lem:EQ}. In particular, we compute $\mathbb{E}\bm{Q}$
by brute force, i.e., we calculate the expectation of all possible monomial and
sum them up. In Section~\ref{subsec:expression_T} we exhaust all
monomials that shows up in $\bm{T}_{i,k}$ for $i=0,1,2,3,4$ and
$k\in[K]$. Then in Section~\ref{subsec:Supporting_expectation},
we provide some useful results that assist our computation. For conciseness
in the notation, we define the projections $\mathcal{P}_{U}\coloneqq\bm{U}^{\star}\bm{U}^{\star\top}$,
$\mathcal{P}_{U}^{\perp}\coloneqq\bm{I}-\bm{U}^{\star}\bm{U}^{\star\top}$,
$\mathcal{P}_{U_{k}}=\bm{U}_{k}^{\star}\bm{U}_{k}^{\star\top}$, $\mathcal{P}_{U_{k}}^{\perp}=\bm{I}-\bm{U}_{k}^{\star}\bm{U}_{k}^{\star\top}$,
$\mathcal{P}_{W}=\bm{W}^{\star}\bm{W}^{\star\top}$, and $\mathcal{P}_{W}^{\perp}=\bm{I}-\bm{W}^{\star}\bm{W}^{\star\top}$.

Recall that 
\[
\bm{Q}\coloneqq\frac{1}{K}\sum_{k=1}^{K}\sum_{i=0}^{4}\sum_{j=0}^{4-i}\bm{T}_{i,k}\bm{T}_{j,k}^{\top}.
\]
where $\bm{T}_{0,k}$ is a sum of $i$-th degree monomials of $\bm{E}_{k}$.
For $i=j=0$, it is easy to see that $\bm{T}_{0,k}\bm{T}_{0,k}^{\top}=\bm{S}_{0}\bm{S}_{0}^{\top}=\bm{U}^{\star}\bm{U}^{\star\top}$.
When $i+j$ is odd, $\bm{T}_{i,k}\bm{T}_{j,k}^{\top}$ is the sum
of several odd degree monomial of $\bm{E}_{k}$. Therefore its expectation
is 0. Thus we only need to focus on the terms where $i+j$ is even.

In the rest of this section, we will first write out the expression
of each term. We consider the second-order terms ($i+j=2$) and fourth-order
terms ($i+j=4$) separately. For the second-order terms, the following
lemma describes their expectation. The proof is deferred to Section~\ref{subsec:expectation_2nd}.

\begin{lemma}\label{lem:second_TT} Let $\bm{T}_{i,k}$ be defined
as in \eqref{eq:def_T} for $i=0,1,2$ and $k\in[K]$. We have
\begin{align*}
\mathbb{E}\bm{T}_{1,k}\bm{T}_{1,k}^{\top} & =\sigma^{2}\left(n\mathcal{P}_{U_{k}}^{\perp}-r_{k}\mathcal{P}_{U}^{\perp}\mathcal{P}_{U_{k}}^{\perp}\right)\\
\mathbb{E}\bm{T}_{0,k}\bm{T}_{2,k}^{\top} & =0.
\end{align*}
\end{lemma}

Similarly we can have a similar result for the fourth-order terms
in the following Lemma. The proof is deferred to Section~\ref{subsec:expectation_4th}.

\begin{lemma}\label{lem:fourth_TT} Let $\bm{T}_{i,k}$ be defined
as in \eqref{eq:def_T} for $i=0,1,2,3,4$ and $k\in[K]$. We have
\begin{align*}
\mathbb{E}\bm{T}_{2,k}\bm{T}_{2,k}^{\top} & =\sigma^{4}\left[\left(2n-r-2r_{k}\right)r_{k}\bm{I}-\left(2n-r-2r_{k}\right)r_{k}\bm{U}^{\star}\bm{U}^{\star\top}\right.\\
 & \qquad\left.+\left(n^{2}-nr-4nr_{k}+2rr_{k}+3r_{k}^{2}\right)\bm{U}_{k}^{\star}\bm{U}_{k}^{\star\top}\right]\\
\mathbb{E}\bm{T}_{1,k}\bm{T}_{3,k}^{\top} & =-2\sigma^{4}(dr-r^{2})\mathcal{P}_{U}^{\perp}\mathcal{P}_{U_{k}}^{\perp}\\
\mathbb{E}\bm{T}_{0,k}\bm{T}_{4,k}^{\top} & =0.6\sigma^{4}(n-r-r_{k})(d-r_{k})\bm{U}^{\star}\bm{U}_{k}^{\star\top}.
\end{align*}

\end{lemma}Combining these terms, as long as $\sigma^{2}(\sqrt{nd}+n)\le C_{1}$
for some small enough constant $C_{1}$ and $r\le(n\wedge d)/3$,
\begin{align*}
\mathbb{E}\sum_{i=0}^{4}\sum_{j=0}^{4-i}\bm{T}_{i,k}\bm{T}_{j,k}^{\top} & =(1+\alpha_{1})\bm{U}^{\star}\bm{U}^{\star\top}+\alpha_{2}\bm{U}_{k}^{\star}\bm{U}_{k}^{\star\top}+\alpha_{3}(\bm{I}-\bm{U}^{\star}\bm{U}^{\star\top})\\
 & \quad+\alpha_{4}\left(\bm{U}^{\star}\bm{U}_{k}^{\star\top}+\bm{U}_{k}^{\star}\bm{U}^{\star\top}\right),
\end{align*}
where $\{\alpha_{i}\}_{i=1,2,3,4}$ are some scalars such that $|\alpha_{i}|\le C_{1}\sigma^{2}(\sqrt{nd}+n)$
for $i=1,2,3$ and $\alpha_{4}\in[C_{2}\sigma^{4}nd,C_{3}\sigma^{4}nd]$
for some constants $C_{1},C_{2},C_{3}>0$ and $C_{2}<C_{3}$, given
that Summing it up over $k\in[K]$ gives us \eqref{eq:EQ}. Note that
we assume $r_{k}=r$ for all $k\in[K]$.

\subsection{Decomposition of \eqref{eq:AK-UW_decomp} into monomials \label{subsec:expression_T}}

In this Section we write down all the hidden monomials \eqref{eq:AK-UW_decomp}
to facilitate further computation. For each $\bm{T}_{i,k}$, we write
it as a summation 
\[
\bm{T}_{i,k}=\sum_{j}\bm{S}_{i,j,k},
\]
where each $\bm{S}_{i,j,k}$ is an $i$-th degree monomial of $\bm{E}_{k}$.
In what follows, we list all such monomials grouped by their degree.

\subsubsection{First-order terms\label{subsec:T1}}

From \eqref{eq:def_T}, we have that 

\begin{align}
\bm{T}_{1,k} & =\bm{S}_{1,1,k}+\bm{S}_{1,2,k}.\label{eq:T1_sum}
\end{align}
We then define 
\begin{align}
\bm{S}_{1,1,k} & \coloneqq\mathcal{P}_{U_{k}}^{\perp}\bm{E}_{k};\label{eq:S_1_def}\\
\bm{S}_{1,2,k} & \coloneqq-\bm{N}_{1,k}\bm{U}_{k}^{\star}\bm{W}^{\star\top}=-\mathcal{P}_{U}^{\perp}\mathcal{P}_{U_{k}}^{\perp}\bm{E}_{k}\bm{W}^{\star}\bm{W}^{\star\top}.\nonumber 
\end{align}

\subsubsection{Second-order terms\label{subsec:T2}}

Recall from \eqref{eq:def_T} that

\[
\bm{T}_{2,k}=-\bm{N}_{1,k}\mathcal{P}_{U}^{\perp}\bm{E}_{k}-\bm{N}_{2,k}\bm{U}_{k}^{\star}\bm{W}^{\star\top}.
\]
We compute each term separately. For the first term $-\bm{N}_{1,k}\mathcal{P}_{U}^{\perp}\bm{E}_{k}$,
\[
-\bm{N}_{1,k}\mathcal{P}_{U}^{\perp}\bm{E}_{k}=\sum_{i=1}^{2}\bm{S}_{2,i,k},
\]
where
\begin{align}
\bm{S}_{2,1,k} & \coloneqq-\bm{U}_{k}^{\star}\bm{W}^{\star\top}\bm{E}_{k}^{\top}\mathcal{P}_{U}^{\perp}\mathcal{P}_{U_{k}}^{\perp}\bm{E}_{k};\label{eq:S_2_def_1}\\
\bm{S}_{2,2,k} & \coloneqq-\mathcal{P}_{U}^{\perp}\mathcal{P}_{U_{k}}^{\perp}\bm{E}_{k}\bm{W}^{\star}\bm{U}_{k}^{\star\top}\bm{E}_{k}.\nonumber 
\end{align}
For the second term $-\bm{N}_{2,k}\bm{U}_{k}^{\star}\bm{W}^{\star\top}$,
each monomial can be written as
\[
-(-1)^{|\{i:\alpha_{i}>1\}|+1}\left[\bm{P}_{k}^{-\alpha_{1}}\widetilde{\bm{E}}_{k}\bm{P}_{k}^{-\alpha_{2}}\widetilde{\bm{E}}_{k}\bm{P}_{k}^{-\alpha_{3}}\right]_{1:n,1:n}\bm{U}_{k}^{\star}\bm{W}^{\star\top}
\]
as long as $\sum_{i}\alpha_{i}=2$. When $\alpha_{3}=0$, 
\[
\left[\bm{P}_{k}^{-\alpha_{1}}\widetilde{\bm{E}}_{k}\bm{P}_{k}^{-\alpha_{2}}\widetilde{\bm{E}}_{k}\bm{P}_{k}^{-\alpha_{3}}\right]_{1:n,1:n}\bm{U}_{k}^{\star}\bm{W}^{\star\top}=\left[\bm{P}_{k}^{-\alpha_{1}}\widetilde{\bm{E}}_{k}\bm{P}_{k}^{-\alpha_{2}}\widetilde{\bm{E}}_{k}\right]_{1:n,1:n}(\bm{I}-\bm{U}_{k}^{\star}\bm{U}_{k}^{\star\top})\bm{U}_{k}^{\star}\bm{W}^{\star\top}=\bm{0}_{n\times d}.
\]
Thus we only need to consider the cases where $\alpha_{3}\neq0$.
We then have 
\[
-\bm{N}_{2,k}\bm{U}_{k}^{\star}\bm{W}^{\star\top}=\sum_{j=3}^{5}\bm{S}_{2,j,k},
\]
where
\begin{align}
\bm{S}_{2,3,k} & \coloneqq-\left[\bm{P}_{k}^{\perp}\widetilde{\bm{E}}_{k}\bm{P}_{k}^{\perp}\widetilde{\bm{E}}_{k}\bm{P}_{k}^{-2}\right]_{1:n,1:n}\bm{U}_{k}^{\star}\bm{W}^{\star\top}=-\mathcal{P}_{U}^{\perp}\mathcal{P}_{U_{k}}^{\perp}\bm{E}_{k}\mathcal{P}_{W}^{\perp}\bm{E}_{k}^{\top}\bm{U}_{k}^{\star}\bm{W}^{\star\top},\nonumber \\
\bm{S}_{2,4,k} & \coloneqq\left[\bm{P}_{k}^{\perp}\widetilde{\bm{E}}_{k}\bm{P}_{k}^{-1}\widetilde{\bm{E}}_{k}\bm{P}_{k}^{-1}\right]_{1:n,1:n}\bm{U}_{k}^{\star}\bm{W}^{\star\top}=\mathcal{P}_{U}^{\perp}\mathcal{P}_{U_{k}}^{\perp}\bm{E}_{k}\bm{W}^{\star}\bm{U}_{k}^{\star\top}\bm{E}_{k}\bm{W}^{\star}\bm{W}^{\star\top},\label{eq:S_2_def_2}\\
\bm{S}_{2,5,k} & \coloneqq\left[\bm{P}_{k}^{-1}\widetilde{\bm{E}}_{k}\bm{P}_{k}^{\perp}\widetilde{\bm{E}}_{k}\bm{P}_{k}^{-1}\right]_{1:n,1:n}\bm{U}_{k}^{\star}\bm{W}^{\star\top}=\bm{U}_{k}^{\star}\bm{W}^{\star\top}\bm{E}_{k}^{\top}\mathcal{P}_{U}^{\perp}\mathcal{P}_{U_{k}}^{\perp}\bm{E}_{k}\bm{W}^{\star}\bm{W}^{\star\top}.\nonumber 
\end{align}

\subsubsection{Third-order terms\label{subsec:T3}}

Recall from \eqref{eq:def_T} that

\[
\bm{T}_{3,k}=-\bm{N}_{2,k}\mathcal{P}_{U}^{\perp}\bm{E}_{k}-\bm{N}_{3,k}\bm{U}_{k}^{\star}\bm{W}^{\star\top}.
\]
For the first term $-\bm{N}_{2,k}\mathcal{P}_{U}^{\perp}\bm{E}_{k}$,
\begin{equation}
-\bm{N}_{2,k}\mathcal{P}_{U}^{\perp}\bm{E}_{k}=\sum_{j=1}^{6}\bm{S}_{3,j,k},\label{eq:S_third_1}
\end{equation}
where 
\begin{align*}
\bm{S}_{3,1,k} & \coloneqq-\left[\bm{P}_{k}^{-2}\widetilde{\bm{E}}_{k}\bm{P}_{k}^{\perp}\widetilde{\bm{E}}_{k}\bm{P}_{k}^{\perp}\right]_{1:n,1:n}\mathcal{P}_{U}^{\perp}\bm{E}_{k}=-\bm{U}_{k}^{\star}\bm{U}_{k}^{\star\top}\bm{E}_{k}\mathcal{P}_{W}^{\perp}\bm{E}_{k}^{\top}\mathcal{P}_{U}^{\perp}\mathcal{P}_{U_{k}}^{\perp}\bm{E}_{k}\\
\bm{S}_{3,2,k} & \coloneqq-\left[\bm{P}_{k}^{\perp}\widetilde{\bm{E}}_{k}\bm{P}_{k}^{-2}\widetilde{\bm{E}}_{k}\bm{P}_{k}^{\perp}\right]_{1:n,1:n}\mathcal{P}_{U}^{\perp}\bm{E}_{k}=-\mathcal{P}_{U}^{\perp}\mathcal{P}_{U_{k}}^{\perp}\bm{E}_{k}\bm{W}^{\star}\bm{W}^{\star\top}\bm{E}_{k}^{\top}\mathcal{P}_{U}^{\perp}\mathcal{P}_{U_{k}}^{\perp}\bm{E}_{k}\\
\bm{S}_{3,3,k} & \coloneqq-\left[\bm{P}_{k}^{\perp}\widetilde{\bm{E}}_{k}\bm{P}_{k}^{\perp}\widetilde{\bm{E}}_{k}\bm{P}_{k}^{-2}\right]_{1:n,1:n}\mathcal{P}_{U}^{\perp}\bm{E}_{k}=-\mathcal{P}_{U}^{\perp}\mathcal{P}_{U_{k}}^{\perp}\bm{E}_{k}\mathcal{P}_{W}^{\perp}\bm{E}_{k}^{\top}\bm{U}_{k}^{\star}\bm{U}_{k}^{\star\top}\bm{E}_{k}\\
\bm{S}_{3,4,k} & \coloneqq\left[\bm{P}_{k}^{\perp}\widetilde{\bm{E}}_{k}\bm{P}_{k}^{-1}\widetilde{\bm{E}}_{k}\bm{P}_{k}^{-1}\right]_{1:n,1:n}\mathcal{P}_{U}^{\perp}\bm{E}_{k}=\mathcal{P}_{U}^{\perp}\mathcal{P}_{U_{k}}^{\perp}\bm{E}\bm{W}^{\star}\bm{U}_{k}^{\star\top}\bm{E}\bm{W}^{\star}\bm{U}_{k}^{\star\top}\bm{E}_{k}\\
\bm{S}_{3,5,k} & \coloneqq\left[\bm{P}_{k}^{-1}\widetilde{\bm{E}}_{k}\bm{P}_{k}^{\perp}\widetilde{\bm{E}}_{k}\bm{P}_{k}^{-1}\right]_{1:n,1:n}\mathcal{P}_{U}^{\perp}\bm{E}_{k}=\bm{U}_{k}^{\star}\bm{W}^{\star\top}\bm{E}_{k}^{\top}\mathcal{P}_{U}^{\perp}\mathcal{P}_{U_{k}}^{\perp}\bm{E}_{k}\bm{W}^{\star}\bm{U}_{k}^{\star\top}\bm{E}_{k}\\
\bm{S}_{3,6,k} & \coloneqq\left[\bm{P}_{k}^{-1}\widetilde{\bm{E}}_{k}\bm{P}_{k}^{-1}\widetilde{\bm{E}}_{k}\bm{P}_{k}^{\perp}\right]_{1:n,1:n}\mathcal{P}_{U}^{\perp}\bm{E}_{k}=\bm{U}_{k}^{\star}\bm{W}^{\star\top}\bm{E}_{k}^{\top}\bm{U}_{k}^{\star}\bm{W}^{\star\top}\bm{E}_{k}^{\top}\mathcal{P}_{U}^{\perp}\mathcal{P}_{U_{k}}^{\perp}\bm{E}_{k}.
\end{align*}
For the second term $-\bm{N}_{3,k}\bm{U}_{k}^{\star}\bm{W}^{\star\top}$,
each monomial can be written in the form of
\[
-(-1)^{|\{i:\alpha_{i}>1\}|+1}\left[\bm{P}_{k}^{-\alpha_{1}}\widetilde{\bm{E}}_{k}\bm{P}_{k}^{-\alpha_{2}}\widetilde{\bm{E}}_{k}\bm{P}_{k}^{-\alpha_{3}}\widetilde{\bm{E}}_{k}\bm{P}_{k}^{-\alpha_{4}}\right]_{1:n,1:n}\bm{U}_{k}^{\star}\bm{W}^{\star\top}.
\]
for some $\bm{\alpha}\in\mathbb{N}^{4}$ such that $\sum_{i}\alpha_{i}=3$.
If $\alpha_{4}=0$, 
\begin{align*}
 & \left[\bm{P}_{k}^{-\alpha_{1}}\widetilde{\bm{E}}_{k}\bm{P}_{k}^{-\alpha_{2}}\widetilde{\bm{E}}_{k}\bm{P}_{k}^{-\alpha_{3}}\widetilde{\bm{E}}_{k}\bm{P}_{k}^{-\alpha_{4}}\right]_{1:n,1:n}\bm{U}_{k}^{\star}\bm{W}^{\star\top}\\
 & \quad=\left[\bm{P}_{k}^{-\alpha_{1}}\widetilde{\bm{E}}_{k}\bm{P}_{k}^{-\alpha_{2}}\widetilde{\bm{E}}_{k}\bm{P}_{k}^{-\alpha_{3}}\widetilde{\bm{E}}_{k}\right]_{1:n,1:n}(\bm{I}-\bm{U}_{k}^{\star}\bm{U}_{k}^{\star\top})\bm{U}_{k}^{\star}\bm{W}^{\star\top}=0.
\end{align*}
We consider the cases when $\alpha_{4}\neq0$. We have that 
\begin{equation}
-\bm{N}_{3,k}\bm{U}_{k}^{\star}\bm{W}^{\star\top}=\sum_{j=7}^{16}\bm{S}_{3,j,k},\label{eq:S_third_2}
\end{equation}
where $\bm{S}_{3,j,k}$ as defined below. For readability we group
the terms by the multiset (unordered set allowing repetition) of $\{\alpha_{i}\}_{i=1}^{4}$.

Term for $\{\alpha_{i}\}=\{0,0,0,3\}$:
\[
\bm{S}_{3,7,k}\coloneqq-\left[\bm{P}_{k}^{\perp}\widetilde{\bm{E}}_{k}\bm{P}_{k}^{\perp}\widetilde{\bm{E}}_{k}\bm{P}_{k}^{\perp}\widetilde{\bm{E}}_{k}\bm{P}_{k}^{-3}\right]_{1:n,1:n}\bm{U}_{k}^{\star}\bm{W}^{\star\top}=-\mathcal{P}_{U}^{\perp}\mathcal{P}_{U_{k}}^{\perp}\bm{E}_{k}\mathcal{P}_{W}^{\perp}\bm{E}_{k}^{\top}\mathcal{P}_{U}^{\perp}\mathcal{P}_{U_{k}}^{\perp}\bm{E}_{k}\bm{W}^{\star}\bm{W}^{\star\top}.
\]

Terms for $\{\alpha_{i}\}=\{0,0,1,2\}$:

\begin{align*}
\bm{S}_{3,8,k} & \coloneqq\left[\bm{P}_{k}^{-2}\widetilde{\bm{E}}_{k}\bm{P}_{k}^{\perp}\widetilde{\bm{E}}_{k}\bm{P}_{k}^{\perp}\widetilde{\bm{E}}_{k}\bm{P}_{k}^{-1}\right]_{1:n,1:n}\bm{U}_{k}^{\star}\bm{W}^{\star\top}=\bm{U}_{k}^{\star}\bm{U}_{k}^{\star\top}\bm{E}_{k}\mathcal{P}_{W}^{\perp}\bm{E}_{k}^{\top}\mathcal{P}_{U}^{\perp}\mathcal{P}_{U_{k}}^{\perp}\bm{E}_{k}\bm{W}^{\star}\bm{W}^{\star\top};\\
\bm{S}_{3,9,k} & \coloneqq\left[\bm{P}_{k}^{\perp}\widetilde{\bm{E}}_{k}\bm{P}_{k}^{-2}\widetilde{\bm{E}}_{k}\bm{P}_{k}^{\perp}\widetilde{\bm{E}}_{k}\bm{P}_{k}^{-1}\right]_{1:n,1:n}\bm{U}_{k}^{\star}\bm{W}^{\star\top}=\mathcal{P}_{U}^{\perp}\mathcal{P}_{U_{k}}^{\perp}\bm{E}_{k}\bm{W}^{\star}\bm{W}^{\star\top}\bm{E}_{k}^{\top}\mathcal{P}_{U}^{\perp}\mathcal{P}_{U_{k}}^{\perp}\bm{E}_{k}\bm{W}^{\star}\bm{W}^{\star\top};\\
\bm{S}_{3,10,k} & \coloneqq\left[\bm{P}_{k}^{\perp}\widetilde{\bm{E}}_{k}\bm{P}_{k}^{\perp}\widetilde{\bm{E}}_{k}\bm{P}_{k}^{-2}\widetilde{\bm{E}}_{k}\bm{P}_{k}^{-1}\right]_{1:n,1:n}\bm{U}_{k}^{\star}\bm{W}^{\star\top}=\mathcal{P}_{U}^{\perp}\mathcal{P}_{U_{k}}^{\perp}\bm{E}_{k}\mathcal{P}_{W}^{\perp}\bm{E}_{k}^{\top}\bm{U}_{k}^{\star}\bm{U}_{k}^{\star\top}\bm{E}_{k}\bm{W}^{\star}\bm{W}^{\star\top};\\
\bm{S}_{3,11,k} & \coloneqq\left[\bm{P}_{k}^{-1}\widetilde{\bm{E}}_{k}\bm{P}_{k}^{\perp}\widetilde{\bm{E}}_{k}\bm{P}_{k}^{\perp}\widetilde{\bm{E}}_{k}\bm{P}_{k}^{-2}\right]_{1:n,1:n}\bm{U}_{k}^{\star}\bm{W}^{\star\top}=\bm{U}_{k}^{\star}\bm{W}^{\star\top}\bm{E}_{k}^{\top}\mathcal{P}_{U}^{\perp}\mathcal{P}_{U_{k}}^{\perp}\bm{E}_{k}\mathcal{P}_{W}^{\perp}\bm{E}_{k}^{\top}\bm{U}_{k}^{\star}\bm{W}^{\star\top};\\
\bm{S}_{3,12,k} & \coloneqq\left[\bm{P}_{k}^{\perp}\widetilde{\bm{E}}_{k}\bm{P}_{k}^{-1}\widetilde{\bm{E}}_{k}\bm{P}_{k}^{\perp}\widetilde{\bm{E}}_{k}\bm{P}_{k}^{-2}\right]_{1:n,1:n}\bm{U}_{k}^{\star}\bm{W}^{\star\top}=\mathcal{P}_{U}^{\perp}\mathcal{P}_{U_{k}}^{\perp}\bm{E}_{k}\bm{W}^{\star}\bm{U}_{k}^{\star\top}\bm{E}_{k}\mathcal{P}_{W}^{\perp}\bm{E}_{k}^{\top}\bm{U}_{k}^{\star}\bm{W}^{\star\top};\\
\bm{S}_{3,13,k} & \coloneqq\left[\bm{P}_{k}^{\perp}\widetilde{\bm{E}}_{k}\bm{P}_{k}^{\perp}\widetilde{\bm{E}}_{k}\bm{P}_{k}^{-1}\widetilde{\bm{E}}_{k}\bm{P}_{k}^{-2}\right]_{1:n,1:n}\bm{U}_{k}^{\star}\bm{W}^{\star\top}=\mathcal{P}_{U}^{\perp}\mathcal{P}_{U_{k}}^{\perp}\bm{E}_{k}\mathcal{P}_{W}^{\perp}\bm{E}_{k}^{\top}\bm{U}_{k}^{\star}\bm{W}_{k}^{\star\top}\bm{E}_{k}^{\top}\bm{U}_{k}^{\star}\bm{W}^{\star\top}.
\end{align*}

Terms for $\{\alpha_{i}\}=\{0,1,1,1\}:$

\begin{align*}
\bm{S}_{3,14,k} & \coloneqq-\left[\bm{P}_{k}^{-1}\widetilde{\bm{E}}_{k}\bm{P}_{k}^{-1}\widetilde{\bm{E}}_{k}\bm{P}_{k}^{\perp}\widetilde{\bm{E}}_{k}\bm{P}_{k}^{-1}\right]_{1:n,1:n}\bm{U}_{k}^{\star}\bm{W}^{\star\top}=-\bm{U}_{k}^{\star}\bm{W}^{\star\top}\bm{E}_{k}^{\top}\bm{U}_{k}^{\star}\bm{W}^{\star\top}\bm{E}_{k}^{\top}\mathcal{P}_{U}^{\perp}\mathcal{P}_{U_{k}}^{\perp}\bm{E}_{k}\bm{W}^{\star}\bm{W}^{\star\top};\\
\bm{S}_{3,15,k} & \coloneqq-\left[\bm{P}_{k}^{-1}\widetilde{\bm{E}}_{k}\bm{P}_{k}^{\perp}\widetilde{\bm{E}}_{k}\bm{P}_{k}^{-1}\widetilde{\bm{E}}_{k}\bm{P}_{k}^{-1}\right]_{1:n,1:n}\bm{U}_{k}^{\star}\bm{W}^{\star\top}=-\bm{U}_{k}^{\star}\bm{W}^{\star\top}\bm{E}_{k}^{\top}\mathcal{P}_{U}^{\perp}\mathcal{P}_{U_{k}}^{\perp}\bm{E}_{k}\bm{W}^{\star}\bm{U}_{k}^{\star\top}\bm{E}_{k}\bm{W}^{\star}\bm{W}^{\star\top};\\
\bm{S}_{3,16,k} & \coloneqq-\left[\bm{P}_{k}^{\perp}\widetilde{\bm{E}}_{k}\bm{P}_{k}^{-1}\widetilde{\bm{E}}_{k}\bm{P}_{k}^{-1}\widetilde{\bm{E}}_{k}\bm{P}_{k}^{-1}\right]_{1:n,1:n}\bm{U}_{k}^{\star}\bm{W}^{\star\top}=-\mathcal{P}_{U}^{\perp}\mathcal{P}_{U_{k}}^{\perp}\bm{E}_{k}\bm{W}^{\star}\bm{U}_{k}^{\star\top}\bm{E}_{k}\bm{W}^{\star}\bm{U}_{k}^{\star\top}\bm{E}_{k}\bm{W}^{\star}\bm{W}^{\star\top}.
\end{align*}

\subsubsection{Fourth-order terms\label{subsec:T4}}

Recall from \eqref{eq:def_T} that

\[
\bm{T}_{4,k}=-\bm{N}_{3,k}\mathcal{P}_{U}^{\perp}\bm{E}_{k}-\bm{N}_{4}\bm{U}_{k}^{\star}\bm{W}^{\star\top}.
\]
For the first term $-\bm{N}_{3,k}\mathcal{P}_{U}^{\perp}\bm{E}_{k}$,
each monomial can be written as
\[
-(-1)^{|\{i:\alpha_{i}>1\}|+1}\left[\bm{P}_{k}^{-\alpha_{1}}\widetilde{\bm{E}}_{k}\bm{P}_{k}^{-\alpha_{2}}\widetilde{\bm{E}}_{k}\bm{P}_{k}^{-\alpha_{3}}\widetilde{\bm{E}}_{k}\bm{P}_{k}^{-\alpha_{4}}\right]_{1:n,1:n}\mathcal{P}_{U}^{\perp}\bm{E}_{k},
\]
for some $\bm{\alpha}\in\mathbb{N}^{4}$ such that $\sum_{i}\alpha_{i}=3$.
We then have
\[
-\bm{N}_{3,k}\mathcal{P}_{U}^{\perp}\bm{E}_{k}=\sum_{j=1}^{20}\bm{S}_{4,j,k},
\]
where $\bm{S}_{4,j,k}$ as defined below. For readability we group
the terms by the multiset (unordered set allowing repetition) of $\{\alpha_{i}\}_{i=1}^{4}$. 

Terms for $\{\alpha_{i}\}=\{0,0,0,3\}:$
\begin{align*}
\bm{S}_{4,1,k} & \coloneqq-\left[\bm{P}_{k}^{-3}\widetilde{\bm{E}}_{k}\bm{P}_{k}^{\perp}\widetilde{\bm{E}}_{k}\bm{P}_{k}^{\perp}\widetilde{\bm{E}}_{k}\bm{P}_{k}^{\perp}\right]_{1:n,1:n}\mathcal{P}_{U}^{\perp}\bm{E}_{k}=-\bm{U}_{k}^{\star}\bm{W}^{\star\top}\bm{E}_{k}^{\top}\mathcal{P}_{U}^{\perp}\mathcal{P}_{U_{k}}^{\perp}\bm{E}_{k}\mathcal{P}_{W}^{\perp}\bm{E}_{k}^{\top}\mathcal{P}_{U}^{\perp}\mathcal{P}_{U_{k}}^{\perp}\bm{E}_{k}\\
\bm{S}_{4,2,k} & \coloneqq-\left[\bm{P}_{k}^{\perp}\widetilde{\bm{E}}_{k}\bm{P}_{k}^{-3}\widetilde{\bm{E}}_{k}\bm{P}_{k}^{\perp}\widetilde{\bm{E}}_{k}\bm{P}_{k}^{\perp}\right]_{1:n,1:n}\mathcal{P}_{U}^{\perp}\bm{E}_{k}=-\mathcal{P}_{U}^{\perp}\mathcal{P}_{U_{k}}^{\perp}\bm{E}_{k}\bm{W}^{\star}\bm{U}_{k}^{\star\top}\bm{E}_{k}\mathcal{P}_{W}^{\perp}\bm{E}_{k}^{\top}\mathcal{P}_{U}^{\perp}\mathcal{P}_{U_{k}}^{\perp}\bm{E}_{k}\\
\bm{S}_{4,3,k} & \coloneqq-\left[\bm{P}_{k}^{\perp}\widetilde{\bm{E}}_{k}\bm{P}_{k}^{\perp}\widetilde{\bm{E}}_{k}\bm{P}_{k}^{-3}\widetilde{\bm{E}}_{k}\bm{P}_{k}^{\perp}\right]_{1:n,1:n}\mathcal{P}_{U}^{\perp}\bm{E}_{k}=-\mathcal{P}_{U}^{\perp}\mathcal{P}_{U_{k}}^{\perp}\bm{E}_{k}\mathcal{P}_{W}^{\perp}\bm{E}_{k}^{\top}\bm{U}_{k}^{\star}\bm{W}^{\star\top}\bm{E}_{k}^{\top}\mathcal{P}_{U}^{\perp}\mathcal{P}_{U_{k}}^{\perp}\bm{E}_{k}\\
\bm{S}_{4,4,k} & \coloneqq-\left[\bm{P}_{k}^{\perp}\widetilde{\bm{E}}_{k}\bm{P}_{k}^{\perp}\widetilde{\bm{E}}_{k}\bm{P}_{k}^{\perp}\widetilde{\bm{E}}_{k}\bm{P}_{k}^{-3}\right]_{1:n,1:n}\mathcal{P}_{U}^{\perp}\bm{E}_{k}=-\mathcal{P}_{U}^{\perp}\mathcal{P}_{U_{k}}^{\perp}\bm{E}_{k}\mathcal{P}_{W}^{\perp}\bm{E}_{k}^{\top}\mathcal{P}_{U}^{\perp}\mathcal{P}_{U_{k}}^{\perp}\bm{E}_{k}\bm{W}^{\star}\bm{U}_{k}^{\star\top}\bm{E}_{k}
\end{align*}

Terms for $\{\alpha_{i}\}=\{0,0,1,2\}:$
\begin{align*}
\bm{S}_{4,5,k} & \coloneqq\left[\bm{P}_{k}^{-2}\widetilde{\bm{E}}_{k}\bm{P}_{k}^{-1}\widetilde{\bm{E}}_{k}\bm{P}_{k}^{\perp}\widetilde{\bm{E}}_{k}\bm{P}_{k}^{\perp}\right]_{1:n,1:n}\mathcal{P}_{U}^{\perp}\bm{E}_{k}=\bm{U}_{k}^{\star}\bm{U}_{k}^{\star\top}\bm{E}_{k}\bm{W}^{\star}\bm{U}_{k}^{\star\top}\bm{E}_{k}\mathcal{P}_{U_{k}}^{\perp}\bm{E}_{k}^{\top}\mathcal{P}_{U}^{\perp}\mathcal{P}_{U_{k}}^{\perp}\bm{E}_{k}\\
\bm{S}_{4,6,k} & \coloneqq\left[\bm{P}_{k}^{-2}\widetilde{\bm{E}}_{k}\bm{P}_{k}^{\perp}\widetilde{\bm{E}}_{k}\bm{P}_{k}^{-1}\widetilde{\bm{E}}_{k}\bm{P}_{k}^{\perp}\right]_{1:n,1:n}\mathcal{P}_{U}^{\perp}\bm{E}_{k}=\bm{U}_{k}^{\star}\bm{U}_{k}^{\star\top}\bm{E}_{k}\mathcal{P}_{W}^{\perp}\bm{E}_{k}^{\top}\bm{U}_{k}^{\star}\bm{W}^{\star\top}\bm{E}_{k}^{\top}\mathcal{P}_{U}^{\perp}\mathcal{P}_{U_{k}}^{\perp}\bm{E}_{k}\\
\bm{S}_{4,7,k} & \coloneqq\left[\bm{P}_{k}^{-2}\widetilde{\bm{E}}_{k}\bm{P}_{k}^{\perp}\widetilde{\bm{E}}_{k}\bm{P}_{k}^{\perp}\widetilde{\bm{E}}_{k}\bm{P}_{k}^{-1}\right]_{1:n,1:n}\mathcal{P}_{U}^{\perp}\bm{E}_{k}=\bm{U}_{k}^{\star}\bm{U}_{k}^{\star\top}\bm{E}_{k}\mathcal{P}_{W}^{\perp}\bm{E}_{k}^{\top}\mathcal{P}_{U}^{\perp}\mathcal{P}_{U_{k}}^{\perp}\bm{E}_{k}\bm{W}^{\star}\bm{U}_{k}^{\star\top}\bm{E}_{k}\\
\bm{S}_{4,8,k} & \coloneqq\left[\bm{P}_{k}^{-1}\widetilde{\bm{E}}_{k}\bm{P}_{k}^{-2}\widetilde{\bm{E}}_{k}\bm{P}_{k}^{\perp}\widetilde{\bm{E}}_{k}\bm{P}_{k}^{\perp}\right]_{1:n,1:n}\mathcal{P}_{U}^{\perp}\bm{E}_{k}=\bm{U}_{k}^{\star}\bm{W}^{\star\top}\bm{E}_{k}^{\top}\bm{U}_{k}^{\star}\bm{U}_{k}^{\star\top}\bm{E}_{k}\mathcal{P}_{W}^{\perp}\bm{E}_{k}^{\top}\mathcal{P}_{U}^{\perp}\mathcal{P}_{U_{k}}^{\perp}\bm{E}_{k}\\
\bm{S}_{4,9,k} & \coloneqq\left[\bm{P}_{k}^{\perp}\widetilde{\bm{E}}_{k}\bm{P}_{k}^{-2}\widetilde{\bm{E}}_{k}\bm{P}_{k}^{-1}\widetilde{\bm{E}}_{k}\bm{P}_{k}^{\perp}\right]_{1:n,1:n}\mathcal{P}_{U}^{\perp}\bm{E}_{k}=\mathcal{P}_{U}^{\perp}\mathcal{P}_{U_{k}}^{\perp}\bm{E}_{k}\bm{W}^{\star}\bm{W}^{\star\top}\bm{E}_{k}^{\top}\bm{U}_{k}^{\star}\bm{W}^{\star\top}\bm{E}_{k}^{\top}\mathcal{P}_{U}^{\perp}\mathcal{P}_{U_{k}}^{\perp}\bm{E}_{k}\\
\bm{S}_{4,10,k} & \coloneqq\left[\bm{P}_{k}^{\perp}\widetilde{\bm{E}}_{k}\bm{P}_{k}^{-2}\widetilde{\bm{E}}_{k}\bm{P}_{k}^{\perp}\widetilde{\bm{E}}_{k}\bm{P}_{k}^{-1}\right]_{1:n,1:n}\mathcal{P}_{U}^{\perp}\bm{E}_{k}=\mathcal{P}_{U}^{\perp}\mathcal{P}_{U_{k}}^{\perp}\bm{E}_{k}\bm{W}^{\star}\bm{W}^{\star\top}\bm{E}_{k}^{\top}\mathcal{P}_{U}^{\perp}\mathcal{P}_{U_{k}}^{\perp}\bm{E}_{k}\bm{W}^{\star}\bm{U}_{k}^{\star\top}\bm{E}_{k}\\
\bm{S}_{4,11,k} & \coloneqq\left[\bm{P}_{k}^{-1}\widetilde{\bm{E}}_{k}\bm{P}_{k}^{\perp}\widetilde{\bm{E}}_{k}\bm{P}_{k}^{-2}\widetilde{\bm{E}}_{k}\bm{P}_{k}^{\perp}\right]_{1:n,1:n}\mathcal{P}_{U}^{\perp}\bm{E}_{k}=\bm{U}_{k}^{\star}\bm{W}^{\star\top}\bm{E}_{k}^{\top}\mathcal{P}_{U}^{\perp}\mathcal{P}_{U_{k}}^{\perp}\bm{E}_{k}\bm{W}^{\star}\bm{W}^{\star\top}\bm{E}_{k}^{\top}\mathcal{P}_{U}^{\perp}\mathcal{P}_{U_{k}}^{\perp}\bm{E}_{k}\\
\bm{S}_{4,12,k} & \coloneqq\left[\bm{P}_{k}^{\perp}\widetilde{\bm{E}}_{k}\bm{P}_{k}^{-1}\widetilde{\bm{E}}_{k}\bm{P}_{k}^{-2}\widetilde{\bm{E}}_{k}\bm{P}_{k}^{\perp}\right]_{1:n,1:n}\mathcal{P}_{U}^{\perp}\bm{E}_{k}=\mathcal{P}_{U}^{\perp}\mathcal{P}_{U_{k}}^{\perp}\bm{E}_{k}\bm{W}^{\star}\bm{U}_{k}^{\star\top}\bm{E}_{k}\bm{W}^{\star}\bm{W}^{\star\top}\bm{E}_{k}^{\top}\mathcal{P}_{U}^{\perp}\mathcal{P}_{U_{k}}^{\perp}\bm{E}_{k}\\
\bm{S}_{4,13,k} & \coloneqq\left[\bm{P}_{k}^{\perp}\widetilde{\bm{E}}_{k}\bm{P}_{k}^{\perp}\widetilde{\bm{E}}_{k}\bm{P}_{k}^{-2}\widetilde{\bm{E}}_{k}\bm{P}_{k}^{-1}\right]_{1:n,1:n}\mathcal{P}_{U}^{\perp}\bm{E}_{k}=\mathcal{P}_{U}^{\perp}\mathcal{P}_{U_{k}}^{\perp}\bm{E}_{k}\mathcal{P}_{W}^{\perp}\bm{E}_{k}^{\top}\bm{U}_{k}^{\star}\bm{U}_{k}^{\star\top}\bm{E}_{k}\bm{W}^{\star}\bm{U}_{k}^{\star\top}\bm{E}_{k}\\
\bm{S}_{4,14,k} & \coloneqq\left[\bm{P}_{k}^{-1}\widetilde{\bm{E}}_{k}\bm{P}_{k}^{\perp}\widetilde{\bm{E}}_{k}\bm{P}_{k}^{\perp}\widetilde{\bm{E}}_{k}\bm{P}_{k}^{-2}\right]_{1:n,1:n}\mathcal{P}_{U}^{\perp}\bm{E}_{k}=\bm{U}_{k}^{\star}\bm{W}^{\star\top}\bm{E}_{k}^{\top}\mathcal{P}_{U}^{\perp}\mathcal{P}_{U_{k}}^{\perp}\bm{E}_{k}\mathcal{P}_{W}^{\perp}\bm{E}_{k}^{\top}\bm{U}_{k}^{\star}\bm{U}_{k}^{\star\top}\bm{E}_{k}\\
\bm{S}_{4,15,k} & \coloneqq\left[\bm{P}_{k}^{\perp}\widetilde{\bm{E}}_{k}\bm{P}_{k}^{-1}\widetilde{\bm{E}}_{k}\bm{P}_{k}^{\perp}\widetilde{\bm{E}}_{k}\bm{P}_{k}^{-2}\right]_{1:n,1:n}\mathcal{P}_{U}^{\perp}\bm{E}_{k}=\mathcal{P}_{U}^{\perp}\mathcal{P}_{U_{k}}^{\perp}\bm{E}_{k}\bm{W}^{\star}\bm{U}_{k}^{\star\top}\bm{E}_{k}\mathcal{P}_{W}^{\perp}\bm{E}_{k}^{\top}\bm{U}_{k}^{\star}\bm{U}_{k}^{\star\top}\bm{E}_{k}\\
\bm{S}_{4,16,k} & \coloneqq\left[\bm{P}_{k}^{\perp}\widetilde{\bm{E}}_{k}\bm{P}_{k}^{\perp}\widetilde{\bm{E}}_{k}\bm{P}_{k}^{-1}\widetilde{\bm{E}}_{k}\bm{P}_{k}^{-2}\right]_{1:n,1:n}\mathcal{P}_{U}^{\perp}\bm{E}_{k}=\mathcal{P}_{U}^{\perp}\mathcal{P}_{U_{k}}^{\perp}\bm{E}_{k}\mathcal{P}_{W}^{\perp}\bm{E}_{k}^{\top}\bm{U}_{k}^{\star}\bm{W}_{k}^{\star\top}\bm{E}_{k}^{\top}\bm{U}_{k}^{\star}\bm{U}_{k}^{\star\top}\bm{E}_{k}
\end{align*}

Terms for $\{\alpha_{i}\}=\{0,1,1,1\}:$
\begin{align*}
\bm{S}_{4,17,k} & \coloneqq-\left[\bm{P}_{k}^{-1}\widetilde{\bm{E}}_{k}\bm{P}_{k}^{-1}\widetilde{\bm{E}}_{k}\bm{P}_{k}^{-1}\widetilde{\bm{E}}_{k}\bm{P}_{k}^{\perp}\right]_{1:n,1:n}\mathcal{P}_{U}^{\perp}\bm{E}_{k}=-\bm{U}_{k}^{\star}\bm{W}^{\star\top}\bm{E}_{k}^{\top}\mathcal{P}_{U}^{\perp}\mathcal{P}_{U_{k}}^{\perp}\bm{E}_{k}\bm{W}^{\star}\bm{U}_{k}^{\star\top}\bm{E}_{k}\bm{W}^{\star}\bm{U}_{k}^{\star\top}\bm{E}_{k}\\
\bm{S}_{4,18,k} & \coloneqq-\left[\bm{P}_{k}^{-1}\widetilde{\bm{E}}_{k}\bm{P}_{k}^{-1}\widetilde{\bm{E}}_{k}\bm{P}_{k}^{\perp}\widetilde{\bm{E}}_{k}\bm{P}_{k}^{-1}\right]_{1:n,1:n}\mathcal{P}_{U}^{\perp}\bm{E}_{k}=-\bm{U}_{k}^{\star}\bm{W}^{\star\top}\bm{E}_{k}^{\top}\bm{U}_{k}^{\star}\bm{W}^{\star\top}\bm{E}_{k}^{\top}\mathcal{P}_{U}^{\perp}\mathcal{P}_{U_{k}}^{\perp}\bm{E}_{k}\bm{W}^{\star}\bm{U}_{k}^{\star\top}\bm{E}_{k}\\
\bm{S}_{4,19,k} & \coloneqq-\left[\bm{P}_{k}^{-1}\widetilde{\bm{E}}_{k}\bm{P}_{k}^{\perp}\widetilde{\bm{E}}_{k}\bm{P}_{k}^{-1}\widetilde{\bm{E}}_{k}\bm{P}_{k}^{-1}\right]_{1:n,1:n}\mathcal{P}_{U}^{\perp}\bm{E}_{k}=-\bm{U}_{k}^{\star}\bm{W}^{\star\top}\bm{E}_{k}^{\top}\mathcal{P}_{U}^{\perp}\mathcal{P}_{U_{k}}^{\perp}\bm{E}_{k}\bm{W}^{\star}\bm{U}_{k}^{\star\top}\bm{E}_{k}\bm{W}^{\star}\bm{U}_{k}^{\star\top}\bm{E}_{k}\\
\bm{S}_{4,20,k} & \coloneqq-\left[\bm{P}_{k}^{\perp}\widetilde{\bm{E}}_{k}\bm{P}_{k}^{-1}\widetilde{\bm{E}}_{k}\bm{P}_{k}^{-1}\widetilde{\bm{E}}_{k}\bm{P}_{k}^{-1}\right]_{1:n,1:n}\mathcal{P}_{U}^{\perp}\bm{E}_{k}=-\mathcal{P}_{U}^{\perp}\mathcal{P}_{U_{k}}^{\perp}\bm{E}_{k}\bm{W}^{\star}\bm{U}_{k}^{\star\top}\bm{E}_{k}\bm{W}^{\star}\bm{U}_{k}^{\star\top}\bm{E}_{k}\bm{W}^{\star}\bm{U}_{k}^{\star\top}\bm{E}_{k}
\end{align*}

For the second term $-\bm{N}_{4}\bm{U}_{k}^{\star}\bm{W}^{\star\top}$,
each monomial can be written as
\[
-(-1)^{|\{i:\alpha_{i}>1\}|+1}\left[\bm{P}_{k}^{-\alpha_{1}}\widetilde{\bm{E}}_{k}\bm{P}_{k}^{-\alpha_{2}}\widetilde{\bm{E}}_{k}\bm{P}_{k}^{-\alpha_{3}}\widetilde{\bm{E}}_{k}\bm{P}_{k}^{-\alpha_{4}}\widetilde{\bm{E}}_{k}\bm{P}_{k}^{-\alpha_{5}}\right]_{1:n,1:n}\bm{U}_{k}^{\star}\bm{W}^{\star\top},
\]
for some $\bm{\alpha}\in\mathbb{N}^{5}$ such that $\sum_{i}\alpha_{i}=4$.
If $\alpha_{5}=0$, 
\begin{align*}
 & \left[\bm{P}_{k}^{-\alpha_{1}}\widetilde{\bm{E}}_{k}\bm{P}_{k}^{-\alpha_{2}}\widetilde{\bm{E}}_{k}\bm{P}_{k}^{-\alpha_{3}}\widetilde{\bm{E}}\bm{P}_{k}^{-\alpha_{4}}\widetilde{\bm{E}}_{k}\bm{P}_{k}^{-\alpha_{5}}\right]_{1:n,1:n}\bm{U}_{k}^{\star}\bm{W}^{\star\top}\\
 & \quad=\left[\bm{P}_{k}^{-\alpha_{1}}\widetilde{\bm{E}}_{k}\bm{P}_{k}^{-\alpha_{2}}\widetilde{\bm{E}}_{k}\bm{P}_{k}^{-\alpha_{3}}\widetilde{\bm{E}}_{k}\bm{P}_{k}^{-\alpha_{4}}\widetilde{\bm{E}}_{k}\right]_{1:n,1:n}(\bm{I}-\bm{U}_{k}^{\star}\bm{U}_{k}^{\star\top})\bm{U}_{k}^{\star}\bm{W}^{\star\top}=0.
\end{align*}
Thus we focus on the cases when $\alpha_{5}\neq0$. We then have
\[
-\bm{N}_{4,k}\bm{U}_{k}^{\star}\bm{W}^{\star\top}=\sum_{j=21}^{55}\bm{S}_{4,j,k},
\]
where $\bm{S}_{4,j,k}$ as defined below. For readability we group
the terms by the multiset (set allowing repeation) of $\{\alpha_{i}\}_{i=1}^{4}$.
We also discuss the case when $\alpha_{5}=1$ separately.

Terms for $\{\alpha_{i}\}=\{0,0,0,0,4\}$:
\begin{align*}
\bm{S}_{4,21,k}&\coloneqq-\left[\bm{P}_{k}^{\perp}\widetilde{\bm{E}}_{k}\bm{P}_{k}^{\perp}\widetilde{\bm{E}}_{k}\bm{P}_{k}^{\perp}\widetilde{\bm{E}}_{k}\bm{P}_{k}^{\perp}\widetilde{\bm{E}}_{k}\bm{P}_{k}^{-4}\right]_{1:n,1:n}\bm{U}_{k}^{\star}\bm{W}^{\star\top}
\\
&=-\mathcal{P}_{U}^{\perp}\mathcal{P}_{U_{k}}^{\perp}\bm{E}_{k}\mathcal{P}_{W}^{\perp}\bm{E}_{k}^{\top}\mathcal{P}_{U}^{\perp}\mathcal{P}_{U_{k}}^{\perp}\bm{E}_{k}\mathcal{P}_{W}^{\perp}\bm{E}_{k}^{\top}\bm{U}_{k}^{\star}\bm{W}^{\star\top}
\end{align*}

Terms for $\{\alpha_{i}\}=\{0,0,0,1,3\}$:
\begin{align*}
\bm{S}_{4,22,k} & \coloneqq\left[\bm{P}_{k}^{\perp}\widetilde{\bm{E}}_{k}\bm{P}_{k}^{\perp}\widetilde{\bm{E}}_{k}\bm{P}_{k}^{\perp}\widetilde{\bm{E}}_{k}\bm{P}_{k}^{-1}\widetilde{\bm{E}}_{k}\bm{P}_{k}^{-3}\right]_{1:n,1:n}\bm{U}_{k}^{\star}\bm{W}^{\star\top}\\
&=\mathcal{P}_{U}^{\perp}\mathcal{P}_{U_{k}}^{\perp}\bm{E}_{k}\mathcal{P}_{W}^{\perp}\bm{E}_{k}^{\top}\mathcal{P}_{U}^{\perp}\mathcal{P}_{U_{k}}^{\perp}\bm{E}_{k}\bm{W}^{\star}\bm{U}_{k}^{\star\top}\bm{E}_{k}\bm{W}^{\star}\bm{W}^{\star\top}\\
\bm{S}_{4,23,k} & \coloneqq\left[\bm{P}_{k}^{\perp}\widetilde{\bm{E}}_{k}\bm{P}_{k}^{\perp}\widetilde{\bm{E}}_{k}\bm{P}_{k}^{-1}\widetilde{\bm{E}}_{k}\bm{P}_{k}^{\perp}\widetilde{\bm{E}}_{k}\bm{P}_{k}^{-3}\right]_{1:n,1:n}\bm{U}_{k}^{\star}\bm{W}^{\star\top}\\
&=\mathcal{P}_{U}^{\perp}\mathcal{P}_{U_{k}}^{\perp}\bm{E}_{k}\mathcal{P}_{W}^{\perp}\bm{E}_{k}^{\top}\bm{U}_{k}^{\star}\bm{W}^{\star\top}\bm{E}_{k}^{\top}\mathcal{P}_{U}^{\perp}\mathcal{P}_{U_{k}}^{\perp}\bm{E}_{k}\bm{W}^{\star}\bm{W}^{\star\top}\\
\bm{S}_{4,24,k} & \coloneqq\left[\bm{P}_{k}^{\perp}\widetilde{\bm{E}}_{k}\bm{P}_{k}^{-1}\widetilde{\bm{E}}_{k}\bm{P}_{k}^{\perp}\widetilde{\bm{E}}_{k}\bm{P}_{k}^{\perp}\widetilde{\bm{E}}_{k}\bm{P}_{k}^{-3}\right]_{1:n,1:n}\bm{U}_{k}^{\star}\bm{W}^{\star\top}\\
&=\mathcal{P}_{U}^{\perp}\mathcal{P}_{U_{k}}^{\perp}\bm{E}_{k}\bm{W}^{\star}\bm{U}_{k}^{\star\top}\bm{E}_{k}\mathcal{P}_{W}^{\perp}\bm{E}_{k}^{\top}\mathcal{P}_{U}^{\perp}\mathcal{P}_{U_{k}}^{\perp}\bm{E}_{k}\bm{W}^{\star}\bm{W}^{\star\top}\\
\bm{S}_{4,25,k} & \coloneqq\left[\bm{P}_{k}^{-1}\widetilde{\bm{E}}_{k}\bm{P}_{k}^{\perp}\widetilde{\bm{E}}_{k}\bm{P}_{k}^{\perp}\widetilde{\bm{E}}_{k}\bm{P}_{k}^{\perp}\widetilde{\bm{E}}_{k}\bm{P}_{k}^{-3}\right]_{1:n,1:n}\bm{U}_{k}^{\star}\bm{W}^{\star\top}\\
&=\bm{U}_{k}^{\star}\bm{W}^{\star\top}\bm{E}_{k}^{\top}\mathcal{P}_{U}^{\perp}\mathcal{P}_{U_{k}}^{\perp}\bm{E}_{k}\mathcal{P}_{W}^{\perp}\bm{E}_{k}^{\top}\mathcal{P}_{U}^{\perp}\mathcal{P}_{U_{k}}^{\perp}\bm{E}_{k}\bm{W}^{\star}\bm{W}^{\star\top}
\end{align*}

Terms for $\{\alpha_{i}\}=\{0,0,0,2,2\}$:

\begin{align*}
\bm{S}_{4,26,k} & \coloneqq\left[\bm{P}_{k}^{\perp}\widetilde{\bm{E}}_{k}\bm{P}_{k}^{\perp}\widetilde{\bm{E}}_{k}\bm{P}_{k}^{\perp}\widetilde{\bm{E}}_{k}\bm{P}_{k}^{-2}\widetilde{\bm{E}}_{k}\bm{P}_{k}^{-2}\right]_{1:n,1:n}\bm{U}_{k}^{\star}\bm{W}^{\star\top}
\\&=\mathcal{P}_{U}^{\perp}\mathcal{P}_{U_{k}}^{\perp}\bm{E}_{k}\mathcal{P}_{W}^{\perp}\bm{E}_{k}^{\top}\mathcal{P}_{U}^{\perp}\mathcal{P}_{U_{k}}^{\perp}\bm{E}_{k}\bm{W}^{\star}\bm{W}^{\star\top}\bm{E}_{k}^{\top}\bm{U}_{k}^{\star}\bm{W}^{\star\top}\\
\bm{S}_{4,27,k} & \coloneqq\left[\bm{P}_{k}^{\perp}\widetilde{\bm{E}}_{k}\bm{P}_{k}^{\perp}\widetilde{\bm{E}}_{k}\bm{P}_{k}^{-2}\widetilde{\bm{E}}_{k}\bm{P}_{k}^{\perp}\widetilde{\bm{E}}_{k}\bm{P}_{k}^{-2}\right]_{1:n,1:n}\bm{U}_{k}^{\star}\bm{W}^{\star\top}
\\&=\mathcal{P}_{U}^{\perp}\mathcal{P}_{U_{k}}^{\perp}\bm{E}_{k}\mathcal{P}_{W}^{\perp}\bm{E}_{k}^{\top}\bm{U}_{k}^{\star}\bm{U}_{k}^{\star\top}\bm{E}_{k}\mathcal{P}_{W}^{\perp}\bm{E}_{k}^{\top}\bm{U}_{k}^{\star}\bm{W}^{\star\top}\\
\bm{S}_{4,28,k} & \coloneqq\left[\bm{P}_{k}^{\perp}\widetilde{\bm{E}}_{k}\bm{P}_{k}^{-2}\widetilde{\bm{E}}_{k}\bm{P}_{k}^{\perp}\widetilde{\bm{E}}_{k}\bm{P}_{k}^{\perp}\widetilde{\bm{E}}_{k}\bm{P}_{k}^{-2}\right]_{1:n,1:n}\bm{U}_{k}^{\star}\bm{W}^{\star\top}
\\&=\mathcal{P}_{U}^{\perp}\mathcal{P}_{U_{k}}^{\perp}\bm{E}_{k}\bm{W}^{\star}\bm{W}^{\star\top}\bm{E}_{k}^{\top}\mathcal{P}_{U}^{\perp}\mathcal{P}_{U_{k}}^{\perp}\bm{E}_{k}\mathcal{P}_{W}^{\perp}\bm{E}_{k}^{\top}\bm{U}_{k}^{\star}\bm{W}^{\star\top}\\
\bm{S}_{4,29,k} & \coloneqq\left[\bm{P}_{k}^{-2}\widetilde{\bm{E}}_{k}\bm{P}_{k}^{\perp}\widetilde{\bm{E}}_{k}\bm{P}_{k}^{\perp}\widetilde{\bm{E}}_{k}\bm{P}_{k}^{\perp}\widetilde{\bm{E}}_{k}\bm{P}_{k}^{-2}\right]_{1:n,1:n}\bm{U}_{k}^{\star}\bm{W}^{\star\top}
\\&=\bm{U}_{k}^{\star}\bm{U}_{k}^{\star\top}\bm{E}_{k}\mathcal{P}_{W}^{\perp}\bm{E}_{k}^{\top}\mathcal{P}_{U}^{\perp}\mathcal{P}_{U_{k}}^{\perp}\bm{E}_{k}\mathcal{P}_{W}^{\perp}\bm{E}_{k}^{\top}\bm{U}_{k}^{\star}\bm{W}^{\star\top}
\end{align*}

Terms for $\{\alpha_{i}\}=\{0,0,1,1,2\},\alpha_{5}\neq1$:
\begin{align*}
\bm{S}_{4,30,k} & \coloneqq-\left[\bm{P}_{k}^{-1}\widetilde{\bm{E}}_{k}\bm{P}_{k}^{-1}\widetilde{\bm{E}}_{k}\bm{P}_{k}^{\perp}\widetilde{\bm{E}}_{k}\bm{P}_{k}^{\perp}\widetilde{\bm{E}}_{k}\bm{P}_{k}^{-2}\right]_{1:n,1:n}\bm{U}_{k}^{\star}\bm{W}^{\star\top}
\\&=\bm{U}_{k}^{\star}\bm{W}^{\star\top}\bm{E}_{k}^{\top}\bm{U}_{k}^{\star}\bm{W}^{\star\top}\bm{E}_{k}^{\top}\mathcal{P}_{U}^{\perp}\mathcal{P}_{U_{k}}^{\perp}\bm{E}_{k}\mathcal{P}_{W}^{\perp}\bm{E}_{k}^{\top}\bm{U}_{k}^{\star}\bm{W}^{\star\top}\\
\bm{S}_{4,31,k} & \coloneqq-\left[\bm{P}_{k}^{-1}\widetilde{\bm{E}}_{k}\bm{P}_{k}^{\perp}\widetilde{\bm{E}}_{k}\bm{P}_{k}^{-1}\widetilde{\bm{E}}_{k}\bm{P}_{k}^{\perp}\widetilde{\bm{E}}_{k}\bm{P}_{k}^{-2}\right]_{1:n,1:n}\bm{U}_{k}^{\star}\bm{W}^{\star\top}
\\&=\bm{U}_{k}^{\star}\bm{W}^{\star\top}\bm{E}_{k}^{\top}\mathcal{P}_{U}^{\perp}\mathcal{P}_{U_{k}}^{\perp}\bm{E}_{k}\bm{W}^{\star}\bm{U}_{k}^{\star\top}\bm{E}_{k}\mathcal{P}_{W}^{\perp}\bm{E}_{k}^{\top}\bm{U}_{k}^{\star}\bm{W}^{\star\top}\\
\bm{S}_{4,32,k} & \coloneqq-\left[\bm{P}_{k}^{-1}\widetilde{\bm{E}}_{k}\bm{P}_{k}^{\perp}\widetilde{\bm{E}}_{k}\bm{P}_{k}^{\perp}\widetilde{\bm{E}}_{k}\bm{P}_{k}^{-1}\widetilde{\bm{E}}_{k}\bm{P}_{k}^{-2}\right]_{1:n,1:n}\bm{U}_{k}^{\star}\bm{W}^{\star\top}
\\&=\bm{U}_{k}^{\star}\bm{W}^{\star\top}\bm{E}_{k}^{\top}\mathcal{P}_{U}^{\perp}\mathcal{P}_{U_{k}}^{\perp}\bm{E}_{k}\mathcal{P}_{W}^{\perp}\bm{E}_{k}^{\top}\bm{U}_{k}^{\star}\bm{W}^{\star\top}\bm{E}_{k}^{\top}\bm{U}_{k}^{\star}\bm{W}^{\star\top}\\
\bm{S}_{4,33,k} & \coloneqq-\left[\bm{P}_{k}^{\perp}\widetilde{\bm{E}}_{k}\bm{P}_{k}^{-1}\widetilde{\bm{E}}_{k}\bm{P}_{k}^{-1}\widetilde{\bm{E}}_{k}\bm{P}_{k}^{\perp}\widetilde{\bm{E}}_{k}\bm{P}_{k}^{-2}\right]_{1:n,1:n}\bm{U}_{k}^{\star}\bm{W}^{\star\top}
\\&=\mathcal{P}_{U}^{\perp}\mathcal{P}_{U_{k}}^{\perp}\bm{E}_{k}\bm{W}^{\star}\bm{U}_{k}^{\star\top}\bm{E}_{k}\bm{W}^{\star}\bm{U}_{k}^{\star\top}\bm{E}_{k}\mathcal{P}_{W}^{\perp}\bm{E}_{k}^{\top}\bm{U}_{k}^{\star}\bm{W}^{\star\top}\\
\bm{S}_{4,34,k} & \coloneqq-\left[\bm{P}_{k}^{\perp}\widetilde{\bm{E}}_{k}\bm{P}_{k}^{-1}\widetilde{\bm{E}}_{k}\bm{P}_{k}^{\perp}\widetilde{\bm{E}}_{k}\bm{P}_{k}^{-1}\widetilde{\bm{E}}_{k}\bm{P}_{k}^{-2}\right]_{1:n,1:n}\bm{U}_{k}^{\star}\bm{W}^{\star\top}
\\&=\mathcal{P}_{U}^{\perp}\mathcal{P}_{U_{k}}^{\perp}\bm{E}_{k}\bm{W}^{\star}\bm{U}_{k}^{\star\top}\bm{E}_{k}\mathcal{P}_{W}^{\perp}\bm{E}_{k}^{\top}\bm{U}_{k}^{\star}\bm{W}^{\star\top}\bm{E}_{k}^{\top}\bm{U}_{k}^{\star}\bm{W}^{\star\top}\\
\bm{S}_{4,35,k} & \coloneqq-\left[\bm{P}_{k}^{\perp}\widetilde{\bm{E}}_{k}\bm{P}_{k}^{\perp}\widetilde{\bm{E}}_{k}\bm{P}_{k}^{-1}\widetilde{\bm{E}}_{k}\bm{P}_{k}^{-1}\widetilde{\bm{E}}_{k}\bm{P}_{k}^{-2}\right]_{1:n,1:n}\bm{U}_{k}^{\star}\bm{W}^{\star\top}
\\&=\mathcal{P}_{U}^{\perp}\mathcal{P}_{U_{k}}^{\perp}\bm{E}_{k}\mathcal{P}_{W}^{\perp}\bm{E}_{k}^{\top}\bm{U}_{k}^{\star}\bm{W}^{\star\top}\bm{E}_{k}^{\top}\bm{U}_{k}^{\star}\bm{W}^{\star\top}\bm{E}_{k}^{\top}\bm{U}_{k}^{\star}\bm{W}^{\star\top}
\end{align*}

Terms for $\alpha_{5}=1$: When $\alpha_{5}=1$, we have that 
\begin{align*}
 & \left[\bm{P}_{k}^{-\alpha_{1}}\widetilde{\bm{E}}_{k}\bm{P}_{k}^{-\alpha_{2}}\widetilde{\bm{E}}_{k}\bm{P}_{k}^{-\alpha_{3}}\widetilde{\bm{E}}_{k}\bm{P}_{k}^{-\alpha_{4}}\widetilde{\bm{E}}_{k}\bm{P}_{k}^{-\alpha_{5}}\right]_{1:n,1:n}\bm{U}_{k}^{\star}\bm{W}^{\star\top}\\
 & \quad=\left[\bm{P}_{k}^{-\alpha_{1}}\widetilde{\bm{E}}_{k}\bm{P}_{k}^{-\alpha_{2}}\widetilde{\bm{E}}_{k}\bm{P}_{k}^{-\alpha_{3}}\widetilde{\bm{E}}_{k}\bm{P}_{k}^{-\alpha_{4}}\right]_{1:n,1:n}\mathcal{P}_{U}^{\perp}\bm{E}_{k}\bm{W}^{\star}\bm{U}_{k}^{\star\top}\bm{U}_{k}^{\star}\bm{W}^{\star\top}\\
 & \quad=\left[\bm{P}_{k}^{-\alpha_{1}}\widetilde{\bm{E}}_{k}\bm{P}_{k}^{-\alpha_{2}}\widetilde{\bm{E}}_{k}\bm{P}_{k}^{-\alpha_{3}}\widetilde{\bm{E}}_{k}\bm{P}_{k}^{-\alpha_{4}}\right]_{1:n,1:n}\mathcal{P}_{U}^{\perp}\bm{E}_{k}\bm{W}^{\star}\bm{W}^{\star\top}.
\end{align*}
Now observe that since $\alpha_{1}+\alpha_{2}+\alpha_{3}+\alpha_{4}=3$,
\[
-(-1)^{|\{i:\alpha_{i}>1\}|+1}\left[\bm{P}_{k}^{-\alpha_{1}}\widetilde{\bm{E}}_{k}\bm{P}_{k}^{-\alpha_{2}}\widetilde{\bm{E}}_{k}\bm{P}_{k}^{-\alpha_{3}}\widetilde{\bm{E}}_{k}\bm{P}_{k}^{-\alpha_{4}}\right]_{1:n,1:n}\mathcal{P}_{U}^{\perp}\bm{E}_{k}=\bm{S}_{4,j,k}
\]
for some $j\in[20]$. We can then write $\bm{S}_{4,j,k}$ as
\begin{align}
\bm{S}_{4,j,k} & \coloneqq-(-1)^{|\{i:\alpha_{i}>1\}|+1}\left[\bm{P}_{k}^{-\alpha_{1}}\widetilde{\bm{E}}_{k}\bm{P}_{k}^{-\alpha_{2}}\widetilde{\bm{E}}_{k}\bm{P}_{k}^{-\alpha_{3}}\widetilde{\bm{E}}_{k}\bm{P}_{k}^{-\alpha_{4}}\right]_{1:n,1:n}\mathcal{P}_{U}^{\perp}\bm{E}_{k}\bm{W}^{\star}\bm{W}^{\star\top}\label{eq:S4_36}\\
 & =-\bm{S}_{4,j-35,k}\bm{W}^{\star}\bm{W}^{\star\top}\nonumber 
\end{align}
for any $j=36,\ldots,55$. This result will be convenient for our
computation of the expectation.

\subsection{Proof of Lemma~\ref{lem:second_TT} \label{subsec:expectation_2nd}}

There are two possible configurations of the second-order terms in
the expectation: $\bm{T}_{0,k}\bm{T}_{2,k}^{\top}$ (and its transpose)
and $\bm{T}_{1,k}\bm{T}_{1,k}^{\top}$. We consider each of them separately.
We have that \begin{subequations}\label{eq:TT_seconds}
\begin{align}
\mathbb{E}\bm{T}_{1,k}\bm{T}_{1,k}^{\top} & =\sigma^{2}\left(n\mathcal{P}_{U_{k}}^{\perp}-r_{k}\mathcal{P}_{U}^{\perp}\mathcal{P}_{U_{k}}^{\perp}\right);\label{eq:T1_T1}\\
\mathbb{E}\bm{T}_{0,k}\bm{T}_{2,k}^{\top} & =0.\label{eq:T0_T2}
\end{align}

\end{subequations}We defer the proof of \eqref{eq:T1_T1} to Section~\ref{subsec:Proof_T1_T1}
and the proof of \eqref{eq:T0_T2} to Section~\ref{subsec:Proof_T0_T2}.

\subsubsection{Proof of \eqref{eq:T1_T1}\label{subsec:Proof_T1_T1}}

Combining \eqref{eq:T1_sum}, \eqref{eq:S_1_def}, and Lemma~\ref{lem:expectation_2},
we have that
\begin{align*}
\mathbb{E}\bm{S}_{1,1,k}\bm{S}_{1,1,k}^{\top} & =\mathcal{P}_{U_{k}}^{\perp}\mathbb{E}\left[\bm{E}_{k}\bm{E}_{k}^{\top}\right]\mathcal{P}_{U_{k}}^{\perp}=n\sigma^{2}\mathcal{P}_{U_{k}}^{\perp}\\
\mathbb{E}\bm{S}_{1,1,k}\bm{S}_{1,2,k}^{\top} & =-\mathcal{P}_{U_{k}}^{\perp}\mathbb{E}\left[\bm{E}_{k}\bm{W}^{\star}\bm{W}^{\star\top}\bm{E}_{k}^{\top}\right]\mathcal{P}_{U}^{\perp}\mathcal{P}_{U_{k}}^{\perp}=-r_{k}\sigma^{2}\mathcal{P}_{U}^{\perp}\mathcal{P}_{U_{k}}^{\perp}\\
\mathbb{E}\bm{S}_{1,2,k}\bm{S}_{1,2,k}^{\top} & =\mathcal{P}_{U}^{\perp}\mathcal{P}_{U_{k}}^{\perp}\mathbb{E}\left[\bm{E}_{k}\bm{W}^{\star}\bm{W}^{\star\top}\bm{E}_{k}^{\top}\right]\mathcal{P}_{U}^{\perp}\mathcal{P}_{U_{k}}^{\perp}=r_{k}\sigma^{2}\mathcal{P}_{U}^{\perp}\mathcal{P}_{U_{k}}^{\perp}.
\end{align*}
Then
\begin{align*}
\mathbb{E}\bm{T}_{1,k}\bm{T}_{1,k}^{\top} & =\mathbb{E}\bm{S}_{1,1,k}\bm{S}_{1,1,k}^{\top}+\mathbb{E}\bm{S}_{1,1,k}\bm{S}_{1,2,k}^{\top}+\mathbb{E}\bm{S}_{1,2,k}\bm{S}_{1,1,k}^{\top}+\mathbb{E}\bm{S}_{1,2,k}\bm{S}_{1,2,k}^{\top}\\
 & =\sigma^{2}\left(n\mathcal{P}_{U_{k}}^{\perp}-r_{k}\mathcal{P}_{U}^{\perp}\mathcal{P}_{U_{k}}^{\perp}\right).
\end{align*}

\subsubsection{Proof of \eqref{eq:T0_T2} \label{subsec:Proof_T0_T2}}

Since $\bm{T}_{0,k}$ is a non-random quantity, it suffices to compute
$\mathbb{E}\bm{T}_{2,k}$. Recall that $\bm{T}_{0,k}=\bm{U}^{\star}\bm{V}^{\star\top}$
and $\bm{T}_{2,k}=\sum_{j=1}^{5}\bm{S}_{2,j,k}$ (see Section~\ref{subsec:T2}).
We use \eqref{eq:S_2_def_1}, \eqref{eq:S_2_def_2}, and Lemma~\ref{lem:expectation_2}
to see that
\begin{align*}
\mathbb{E}\bm{S}_{2,1,k} & =-\mathbb{E}\bm{U}_{k}^{\star}\bm{W}^{\star\top}\bm{E}_{k}^{\top}\mathcal{P}_{U}^{\perp}\mathcal{P}_{U_{k}}^{\perp}\bm{E}_{k}=-(n-r-r_{k})\sigma^{2}\bm{U}_{k}^{\star}\bm{W}^{\star\top}\\
\mathbb{E}\bm{S}_{2,2,k} & =-\mathbb{E}\mathcal{P}_{U}^{\perp}\mathcal{P}_{U_{k}}^{\perp}\bm{E}_{k}\bm{W}^{\star}\bm{U}_{k}^{\star\top}\bm{E}_{k}=0\\
\mathbb{E}\bm{S}_{2,3,k} & =-\mathbb{E}\mathcal{P}_{U}^{\perp}\mathcal{P}_{U_{k}}^{\perp}\bm{E}_{k}\mathcal{P}_{W}^{\perp}\bm{E}_{k}^{\top}\bm{U}_{k}^{\star}\bm{W}^{\star\top}=0\\
\mathbb{E}\bm{S}_{2,4,k} & =\mathbb{E}\mathcal{P}_{U}^{\perp}\mathcal{P}_{U_{k}}^{\perp}\bm{E}_{k}\bm{W}^{\star}\bm{U}_{k}^{\star\top}\bm{E}_{k}\bm{W}^{\star}\bm{W}^{\star\top}=0\\
\mathbb{E}\bm{S}_{2,5,k} & =\mathbb{E}\bm{U}_{k}^{\star}\bm{W}^{\star\top}\bm{E}_{k}^{\top}\mathcal{P}_{U}^{\perp}\mathcal{P}_{U_{k}}^{\perp}\bm{E}_{k}\bm{W}^{\star}\bm{W}^{\star\top}=(n-r-r_{k})\sigma^{2}\bm{U}_{k}^{\star}\bm{W}^{\star\top}.
\end{align*}
Summing these up, we have that 
\[
\mathbb{E}\bm{T}_{0,k}\bm{T}_{2,k}^{\top}=\bm{T}_{0,k}\mathbb{E}\bm{T}_{2,k}^{\top}=0.
\]

\subsection{Proof of Lemma~\ref{lem:fourth_TT} \label{subsec:expectation_4th}}

There are three possible configurations of the fourth order terms:
$\bm{T}_{0,k}\bm{T}_{4,k}^{\top}$, $\bm{T}_{1,k}\bm{T}_{3,k}^{\top}$,
and $\bm{T}_{2,k}\bm{T}_{2,k}^{\top}$ (and the corresponding transposes).
We will show that \begin{subequations}\label{eq:TT_fourths}
\begin{align}
\mathbb{E}\bm{T}_{2,k}\bm{T}_{2,k}^{\top} & =\sigma^{4}\left[\left(2dr-3r^{2}\right)\bm{I}-\left(2dr-3r^{2}\right)\bm{U}^{\star}\bm{U}^{\star\top}+\left(nd-4dr-nr+5r^{2}\right)\bm{U}_{k}^{\star}\bm{U}_{k}^{\star\top}\right]\label{eq:T2_T2}\\
\mathbb{E}\bm{T}_{1,k}\bm{T}_{3,k}^{\top} & =-2\sigma^{4}(dr-r^{2})\mathcal{P}_{U}^{\perp}\mathcal{P}_{U_{k}}^{\perp}\label{eq:T1_T3}\\
\mathbb{E}\bm{T}_{0,k}\bm{T}_{4,k}^{\top} & =\sigma^{4}(n-r-r_{k})(d-r_{k})\bm{U}^{\star}\bm{V}^{\star\top}\bm{W}^{\star}\bm{U}_{k}^{\star\top}.\label{eq:T0_T4}
\end{align}

\end{subequations}Then the proof of Lemma~\ref{lem:fourth_TT} is
completed by substituting $\bm{V}^{\star\top}\bm{W}^{\star}=0.6\bm{I}$.
We defer the proof of \eqref{eq:T2_T2}, \eqref{eq:T1_T3}, and \eqref{eq:T0_T4}
to Section~\ref{subsec:Proof_T2_T2}, \ref{subsec:Proof_T1_T3},
and \ref{subsec:Proof_T0_T4}, respectively. To be concise, in the
rest of this section, we compute the expectations with Lemma~\ref{lem:expectation_4}
without explicitly mentioning the lemma.

\subsubsection{Proof of \eqref{eq:T2_T2} \label{subsec:Proof_T2_T2}}

Recall that $\bm{T}_{2,k}=\sum_{j=1}^{5}\bm{S}_{2,j,k}$ (see Section~\ref{subsec:T2}).
For the symmetric terms in the form of $\bm{S}_{2,j,k}\bm{S}_{2,j,k}^{\top}$,

\begin{align*}
\mathbb{E}\bm{S}_{2,1,k}\bm{S}_{2,1,k}^{\top} & =\bm{U}_{k}^{\star}\bm{W}^{\star\top}\mathbb{E}\left[\bm{E}_{k}^{\top}\mathcal{P}_{U}^{\perp}\mathcal{P}_{U_{k}}^{\perp}\bm{E}_{k}\bm{E}_{k}^{\top}\mathcal{P}_{U}^{\perp}\mathcal{P}_{U_{k}}^{\perp}\bm{E}_{k}\right]\bm{W}^{\star}\bm{U}_{k}^{\star\top}\\
 & =\sigma^{4}(n-r-r_{k})(n+d-r-r_{k}+1)\bm{U}_{k}^{\star}\bm{U}_{k}^{\star\top}\\
\mathbb{E}\bm{S}_{2,2,k}\bm{S}_{2,2,k}^{\top} & =\mathcal{P}_{U}^{\perp}\mathcal{P}_{U_{k}}^{\perp}\mathbb{E}\left[\bm{E}_{k}\bm{W}^{\star}\bm{U}_{k}^{\star\top}\bm{E}_{k}\bm{E}_{k}^{\top}\bm{U}_{k}^{\star}\bm{W}^{\star\top}\bm{E}_{k}^{\top}\right]\mathcal{P}_{U}^{\perp}\mathcal{P}_{U_{k}}^{\perp}\\
 & =\sigma^{4}dr_{k}\mathcal{P}_{U}^{\perp}\mathcal{P}_{U_{k}}^{\perp}\\
\mathbb{E}\bm{S}_{2,3,k}\bm{S}_{2,3,k}^{\top} & =\mathbb{E}\mathcal{P}_{U}^{\perp}\mathcal{P}_{U_{k}}^{\perp}\bm{E}_{k}\mathcal{P}_{W}^{\perp}\bm{E}_{k}^{\top}\bm{U}_{k}^{\star}\bm{W}^{\star\top}\bm{W}^{\star}\bm{U}_{k}^{\star\top}\bm{E}_{k}\mathcal{P}_{W}^{\perp}\bm{E}_{k}^{\top}\mathcal{P}_{U}^{\perp}\mathcal{P}_{U_{k}}^{\perp}\\
 & =\sigma^{4}(d-r-r_{k})r_{k}\mathcal{P}_{U}^{\perp}\mathcal{P}_{U_{k}}^{\perp}\\
\mathbb{E}\bm{S}_{2,4,k}\bm{S}_{2,4,k}^{\top} & =\mathbb{E}\mathcal{P}_{U}^{\perp}\mathcal{P}_{U_{k}}^{\perp}\bm{E}_{k}\bm{W}^{\star}\bm{U}_{k}^{\star\top}\bm{E}_{k}\bm{W}^{\star}\bm{W}^{\star\top}\bm{E}_{k}^{\top}\bm{U}_{k}^{\star}\bm{W}^{\star\top}\bm{E}_{k}^{\top}\mathcal{P}_{U}^{\perp}\mathcal{P}_{U_{k}}^{\perp}\\
 & =\sigma^{4}r_{k}^{2}\mathcal{P}_{U}^{\perp}\mathcal{P}_{U_{k}}^{\perp}\\
\mathbb{E}\bm{S}_{2,5,k}\bm{S}_{2,5,k}^{\top} & =\mathbb{E}\bm{U}_{k}^{\star}\bm{W}^{\star\top}\bm{E}_{k}^{\top}\mathcal{P}_{U}^{\perp}\mathcal{P}_{U_{k}}^{\perp}\bm{E}_{k}\bm{W}^{\star}\bm{W}^{\star\top}\bm{E}_{k}^{\top}\mathcal{P}_{U}^{\perp}\mathcal{P}_{U_{k}}^{\perp}\bm{E}_{k}\bm{W}^{\star}\bm{U}_{k}^{\star\top}\\
 & =\sigma^{4}(n-r-r_{k})(n-r+1)\bm{U}_{k}^{\star}\bm{U}_{k}^{\star\top}.
\end{align*}
For the asymmetric terms,

\begin{align*}
\mathbb{E}\bm{S}_{2,1,k}\bm{S}_{2,2,k}^{\top} & =\bm{U}_{k}^{\star}\bm{W}^{\star\top}\mathbb{E}\left[\bm{E}_{k}^{\top}\mathcal{P}_{U}^{\perp}\mathcal{P}_{U_{k}}^{\perp}\bm{E}_{k}\bm{E}_{k}^{\top}\bm{U}_{k}^{\star}\bm{W}^{\star\top}\bm{E}_{k}^{\top}\right]\mathcal{P}_{U}^{\perp}\mathcal{P}_{U_{k}}^{\perp}=\bm{0}\\
\mathbb{E}\bm{S}_{2,1,k}\bm{S}_{2,3,k}^{\top} & =\mathbb{E}\bm{U}_{k}^{\star}\bm{W}^{\star\top}\bm{E}_{k}^{\top}\mathcal{P}_{U}^{\perp}\mathcal{P}_{U_{k}}^{\perp}\bm{E}_{k}\bm{W}^{\star}\bm{U}_{k}^{\star\top}\bm{E}_{k}\mathcal{P}_{W}^{\perp}\bm{E}_{k}^{\top}\mathcal{P}_{U}^{\perp}\mathcal{P}_{U_{k}}^{\perp}=\bm{0}\\
\mathbb{E}\bm{S}_{2,1,k}\bm{S}_{2,4,k}^{\top} & =-\mathbb{E}\bm{U}_{k}^{\star}\bm{W}^{\star\top}\bm{E}_{k}^{\top}\mathcal{P}_{U}^{\perp}\mathcal{P}_{U_{k}}^{\perp}\bm{E}_{k}\bm{W}^{\star}\bm{W}^{\star\top}\bm{E}_{k}^{\top}\bm{U}_{k}^{\star}\bm{W}^{\star\top}\bm{E}_{k}^{\top}\mathcal{P}_{U}^{\perp}\mathcal{P}_{U_{k}}^{\perp}=\bm{0}\\
\mathbb{E}\bm{S}_{2,1,k}\bm{S}_{2,5,k}^{\top} & =-\mathbb{E}\bm{U}_{k}^{\star}\bm{W}^{\star\top}\bm{E}_{k}^{\top}\mathcal{P}_{U}^{\perp}\mathcal{P}_{U_{k}}^{\perp}\bm{E}_{k}\bm{W}^{\star}\bm{W}^{\star\top}\bm{E}_{k}^{\top}\mathcal{P}_{U}^{\perp}\mathcal{P}_{U_{k}}^{\perp}\bm{E}_{k}\bm{W}^{\star}\bm{U}_{k}^{\star\top}\\
 & =-\sigma^{4}(n-r-r_{k})(n-r+1)\bm{U}_{k}^{\star}\bm{U}_{k}^{\star\top}\\
\mathbb{E}\bm{S}_{2,2,k}\bm{S}_{2,3,k}^{\top} & =\mathbb{E}\mathcal{P}_{U}^{\perp}\mathcal{P}_{U_{k}}^{\perp}\bm{E}_{k}\bm{W}^{\star}\bm{U}_{k}^{\star\top}\bm{E}_{k}\bm{W}^{\star}\bm{U}_{k}^{\star\top}\bm{E}_{k}\mathcal{P}_{W}^{\perp}\bm{E}_{k}^{\top}\mathcal{P}_{U}^{\perp}\mathcal{P}_{U_{k}}^{\perp}=\bm{0}\\
\mathbb{E}\bm{S}_{2,2,k}\bm{S}_{2,4,k}^{\top} & =-\mathbb{E}\mathcal{P}_{U}^{\perp}\mathcal{P}_{U_{k}}^{\perp}\bm{E}_{k}\bm{W}^{\star}\bm{U}_{k}^{\star\top}\bm{E}_{k}\bm{W}^{\star}\bm{W}^{\star\top}\bm{E}_{k}^{\top}\bm{U}_{k}^{\star}\bm{W}^{\star\top}\bm{E}_{k}^{\top}\mathcal{P}_{U}^{\perp}\mathcal{P}_{U_{k}}^{\perp}\\
 & =-\sigma^{4}r_{k}^{2}\mathcal{P}_{U}^{\perp}\mathcal{P}_{U_{k}}^{\perp}\\
\mathbb{E}\bm{S}_{2,2,k}\bm{S}_{2,5,k}^{\top} & =-\mathbb{E}\mathcal{P}_{U}^{\perp}\mathcal{P}_{U_{k}}^{\perp}\bm{E}_{k}\bm{W}^{\star}\bm{U}_{k}^{\star\top}\bm{E}_{k}\bm{W}^{\star}\bm{W}^{\star\top}\bm{E}_{k}^{\top}\mathcal{P}_{U}^{\perp}\mathcal{P}_{U_{k}}^{\perp}\bm{E}_{k}\bm{W}^{\star}\bm{U}_{k}^{\star\top}=\bm{0}\\
\mathbb{E}\bm{S}_{2,3,k}\bm{S}_{2,4,k}^{\top} & =-\mathbb{E}\mathcal{P}_{U}^{\perp}\mathcal{P}_{U_{k}}^{\perp}\bm{E}_{k}\mathcal{P}_{W}^{\perp}\bm{E}_{k}^{\top}\bm{U}_{k}^{\star}\bm{W}^{\star\top}\bm{E}_{k}^{\top}\bm{U}_{k}^{\star}\bm{W}^{\star\top}\bm{E}_{k}^{\top}\mathcal{P}_{U}^{\perp}\mathcal{P}_{U_{k}}^{\perp}=\bm{0}\\
\mathbb{E}\bm{S}_{2,3,k}\bm{S}_{2,5,k}^{\top} & =-\mathbb{E}\mathcal{P}_{U}^{\perp}\mathcal{P}_{U_{k}}^{\perp}\bm{E}_{k}\mathcal{P}_{W}^{\perp}\bm{E}_{k}^{\top}\bm{U}_{k}^{\star}\bm{W}^{\star\top}\bm{E}_{k}^{\top}\mathcal{P}_{U}^{\perp}\mathcal{P}_{U_{k}}^{\perp}\bm{E}_{k}\bm{W}^{\star}\bm{U}_{k}^{\star\top}=\bm{0}\\
\mathbb{E}\bm{S}_{2,4,k}\bm{S}_{2,5,k}^{\top} & =\mathbb{E}\mathcal{P}_{U}^{\perp}\mathcal{P}_{U_{k}}^{\perp}\bm{E}_{k}\bm{W}^{\star}\bm{U}_{k}^{\star\top}\bm{E}_{k}\bm{W}^{\star}\bm{W}^{\star\top}\bm{E}_{k}^{\top}\mathcal{P}_{U}^{\perp}\mathcal{P}_{U_{k}}^{\perp}\bm{E}_{k}\bm{W}^{\star}\bm{U}_{k}^{\star\top}=\bm{0}.
\end{align*}
Combining all these terms and the fact that $r=r_{k}$ (note that
cross terms need to include their transpose), we have that
\begin{align*}
\mathbb{E}\bm{T}_{2,k}\bm{T}_{2,k}^{\top} & =\mathbb{E}\left(\sum_{i=1}^{5}\sum_{j=1}^{5}\bm{S}_{2,i,k}\bm{S}_{2,j,k}^{\top}\right)\\
 & =\sigma^{4}\left[\left(2dr-3r^{2}\right)\bm{I}-\left(2dr-3r^{2}\right)\bm{U}^{\star}\bm{U}^{\star\top}+\left(nd-4dr-nr+5r^{2}\right)\bm{U}_{k}^{\star}\bm{U}_{k}^{\star\top}\right].
\end{align*}

\subsubsection{Proof of \eqref{eq:T1_T3} \label{subsec:Proof_T1_T3}}

Recall that $\bm{T}_{1,k}=\sum_{j=1}^{2}\bm{S}_{1,j,k}$, $\bm{T}_{3,k}=\sum_{j=1}^{16}\bm{S}_{3,j,k}$
(see Section~\ref{subsec:T1} and \ref{subsec:T3}). For each term
$\bm{S}_{1,i,k}\bm{S}_{3,j,k}^{\top}$, we first will discuss the
case of $j=1,\ldots,6$ and $j=7,\ldots,16$ separately. 

We start with the case $j=1,\ldots,6$. 
\begin{align*}
\mathbb{E}\bm{S}_{3,1,k}\bm{S}_{1,1,k}^{\top} & =-\mathbb{E}\bm{U}_{k}^{\star}\bm{U}_{k}^{\star\top}\bm{E}_{k}\mathcal{P}_{W}^{\perp}\bm{E}_{k}^{\top}\mathcal{P}_{U}^{\perp}\mathcal{P}_{U_{k}}^{\perp}\bm{E}_{k}\bm{E}_{k}^{\top}\mathcal{P}_{U_{k}}^{\perp}=\bm{0}\\
\mathbb{E}\bm{S}_{3,1,k}\bm{S}_{1,2,k}^{\top} & =\mathbb{E}\bm{U}_{k}^{\star}\bm{U}_{k}^{\star\top}\bm{E}_{k}\mathcal{P}_{W}^{\perp}\bm{E}_{k}^{\top}\mathcal{P}_{U}^{\perp}\mathcal{P}_{U_{k}}^{\perp}\bm{E}_{k}\bm{W}^{\star}\bm{W}^{\star\top}\bm{E}_{k}^{\top}\mathcal{P}_{U}^{\perp}\mathcal{P}_{U_{k}}^{\perp}=\bm{0}\\
\mathbb{E}\bm{S}_{3,2,k}\bm{S}_{1,1,k}^{\top} & =-\mathbb{E}\mathcal{P}_{U}^{\perp}\mathcal{P}_{U_{k}}^{\perp}\bm{E}_{k}\bm{W}^{\star}\bm{W}^{\star\top}\bm{E}_{k}^{\top}\mathcal{P}_{U}^{\perp}\mathcal{P}_{U_{k}}^{\perp}\bm{E}_{k}\bm{E}_{k}^{\top}\mathcal{P}_{U_{k}}^{\perp}\\
 & =-\sigma^{4}(n+d-r-r_{k}+1)r_{k}\mathcal{P}_{U}^{\perp}\mathcal{P}_{U_{k}}^{\perp}\\
\mathbb{E}\bm{S}_{3,2,k}\bm{S}_{1,2,k}^{\top} & =\mathbb{E}\mathcal{P}_{U}^{\perp}\mathcal{P}_{U_{k}}^{\perp}\bm{E}_{k}\bm{W}^{\star}\bm{W}^{\star\top}\bm{E}_{k}^{\top}\mathcal{P}_{U}^{\perp}\mathcal{P}_{U_{k}}^{\perp}\bm{E}_{k}\bm{W}^{\star}\bm{W}^{\star\top}\bm{E}_{k}^{\top}\mathcal{P}_{U}^{\perp}\mathcal{P}_{U_{k}}^{\perp}\\
 & =\sigma^{4}(n-r+1)r_{k}\mathcal{P}_{U}^{\perp}\mathcal{P}_{U_{k}}^{\perp}\\
\mathbb{E}\bm{S}_{3,3,k}\bm{S}_{1,1,k}^{\top} & =-\mathbb{E}\mathcal{P}_{U}^{\perp}\mathcal{P}_{U_{k}}^{\perp}\bm{E}_{k}\mathcal{P}_{W}^{\perp}\bm{E}_{k}^{\top}\bm{U}_{k}^{\star}\bm{U}_{k}^{\star\top}\bm{E}_{k}\bm{E}_{k}^{\top}\mathcal{P}_{U_{k}}^{\perp}\\
 & =-\sigma^{4}(d-r_{k})r_{k}\mathcal{P}_{U}^{\perp}\mathcal{P}_{U_{k}}^{\perp}\\
\mathbb{E}\bm{S}_{3,3,k}\bm{S}_{1,2,k}^{\top} & =\mathbb{E}\mathcal{P}_{U}^{\perp}\mathcal{P}_{U_{k}}^{\perp}\bm{E}_{k}\mathcal{P}_{W}^{\perp}\bm{E}_{k}^{\top}\bm{U}_{k}^{\star}\bm{U}_{k}^{\star\top}\bm{E}_{k}\bm{W}^{\star}\bm{W}^{\star\top}\bm{E}_{k}^{\top}\mathcal{P}_{U}^{\perp}\mathcal{P}_{U_{k}}^{\perp}=\bm{0}\\
\mathbb{E}\bm{S}_{3,4,k}\bm{S}_{1,1,k}^{\top} & =\mathbb{E}\mathcal{P}_{U}^{\perp}\mathcal{P}_{U_{k}}^{\perp}\bm{E}_{k}\bm{W}^{\star}\bm{U}_{k}^{\star\top}\bm{E}_{k}\bm{W}^{\star}\bm{U}_{k}^{\star\top}\bm{E}_{k}\bm{E}_{k}^{\top}\mathcal{P}_{U_{k}}^{\perp}\\
 & =\sigma^{4}r_{k}\mathcal{P}_{U}^{\perp}\mathcal{P}_{U_{k}}^{\perp}\\
\mathbb{E}\bm{S}_{3,4,k}\bm{S}_{1,2,k}^{\top} & =-\mathbb{E}\mathcal{P}_{U}^{\perp}\mathcal{P}_{U_{k}}^{\perp}\bm{E}_{k}\bm{W}^{\star}\bm{U}_{k}^{\star\top}\bm{E}_{k}\bm{W}^{\star}\bm{U}_{k}^{\star\top}\bm{E}_{k}\bm{W}^{\star}\bm{W}^{\star\top}\bm{E}_{k}^{\top}\mathcal{P}_{U}^{\perp}\mathcal{P}_{U_{k}}^{\perp}\\
 & =-\sigma^{4}r_{k}\mathcal{P}_{U}^{\perp}\mathcal{P}_{U_{k}}^{\perp}\\
\mathbb{E}\bm{S}_{3,5,k}\bm{S}_{1,1,k}^{\top} & =\mathbb{E}\bm{U}_{k}^{\star}\bm{W}^{\star\top}\bm{E}_{k}^{\top}\mathcal{P}_{U}^{\perp}\mathcal{P}_{U_{k}}^{\perp}\bm{E}_{k}\bm{W}^{\star}\bm{U}_{k}^{\star\top}\bm{E}_{k}\bm{E}_{k}^{\top}\mathcal{P}_{U_{k}}^{\perp}=\bm{0}\\
\mathbb{E}\bm{S}_{3,5,k}\bm{S}_{1,2,k}^{\top} & =-\mathbb{E}\bm{U}_{k}^{\star}\bm{W}^{\star\top}\bm{E}_{k}^{\top}\mathcal{P}_{U}^{\perp}\mathcal{P}_{U_{k}}^{\perp}\bm{E}_{k}\bm{W}^{\star}\bm{U}_{k}^{\star\top}\bm{E}_{k}\bm{W}^{\star}\bm{W}^{\star\top}\bm{E}_{k}^{\top}\mathcal{P}_{U}^{\perp}\mathcal{P}_{U_{k}}^{\perp}=\bm{0}\\
\mathbb{E}\bm{S}_{3,6,k}\bm{S}_{1,1,k}^{\top} & =\mathbb{E}\bm{U}_{k}^{\star}\bm{W}^{\star\top}\bm{E}_{k}^{\top}\bm{U}_{k}^{\star}\bm{W}^{\star\top}\bm{E}_{k}^{\top}\mathcal{P}_{U}^{\perp}\mathcal{P}_{U_{k}}^{\perp}\bm{E}_{k}\bm{E}_{k}^{\top}\mathcal{P}_{U_{k}}^{\perp}=\bm{0}\\
\mathbb{E}\bm{S}_{3,6,k}\bm{S}_{1,2,k}^{\top} & =-\mathbb{E}\bm{U}_{k}^{\star}\bm{W}^{\star\top}\bm{E}_{k}^{\top}\bm{U}_{k}^{\star}\bm{W}^{\star\top}\bm{E}_{k}^{\top}\mathcal{P}_{U}^{\perp}\mathcal{P}_{U_{k}}^{\perp}\bm{E}_{k}\bm{W}^{\star}\bm{W}^{\star\top}\bm{E}_{k}^{\top}\mathcal{P}_{U}^{\perp}\mathcal{P}_{U_{k}}^{\perp}=\bm{0}
\end{align*}
Now we discuss the case of $j=7,\ldots,16$. All the terms in this
category ends with $\bm{W}^{\star\top}$. This is convenient for us
since then 
\begin{align*}
\bm{S}_{3,j,k}(\bm{S}_{1,1,k}+\bm{S}_{1,2,k})^{\top} & =\bm{S}_{3,j,k}\bm{E}_{k}^{\top}\mathcal{P}_{U_{k}}^{\perp}-\bm{S}_{3,j,k}\bm{E}_{k}^{\top}\mathcal{P}_{U}^{\perp}\mathcal{P}_{U_{k}}^{\perp}\\
 & =\bm{S}_{3,j,k}\bm{E}_{k}^{\top}\bm{U}^{\star}\bm{U}^{\star\top}.
\end{align*}
We can use it to compute that
\begin{align*}
\mathbb{E}\bm{S}_{3,7,k}(\bm{S}_{1,1,k}+\bm{S}_{1,2,k})^{\top} & =-\mathbb{E}\mathcal{P}_{U}^{\perp}\mathcal{P}_{U_{k}}^{\perp}\bm{E}_{k}\mathcal{P}_{W}^{\perp}\bm{E}_{k}^{\top}\mathcal{P}_{U}^{\perp}\mathcal{P}_{U_{k}}^{\perp}\bm{E}_{k}\bm{W}^{\star}\bm{W}^{\star\top}\bm{E}_{k}^{\top}\bm{U}^{\star}\bm{U}^{\star\top}=\bm{0}\\
\mathbb{E}\bm{S}_{3,8,k}(\bm{S}_{1,1,k}+\bm{S}_{1,2,k})^{\top} & =\mathbb{E}\bm{U}_{k}^{\star}\bm{U}_{k}^{\star\top}\bm{E}_{k}\mathcal{P}_{W}^{\perp}\bm{E}_{k}^{\top}\mathcal{P}_{U}^{\perp}\mathcal{P}_{U_{k}}^{\perp}\bm{E}_{k}\bm{W}^{\star}\bm{W}^{\star\top}\bm{E}_{k}^{\top}\bm{U}^{\star}\bm{U}^{\star\top}=\bm{0}\\
\mathbb{E}\bm{S}_{3,9,k}(\bm{S}_{1,1,k}+\bm{S}_{1,2,k})^{\top} & =\mathbb{E}\mathcal{P}_{U}^{\perp}\mathcal{P}_{U_{k}}^{\perp}\bm{E}_{k}\bm{W}^{\star}\bm{W}^{\star\top}\bm{E}_{k}^{\top}\mathcal{P}_{U}^{\perp}\mathcal{P}_{U_{k}}^{\perp}\bm{E}_{k}\bm{W}^{\star}\bm{W}^{\star\top}\bm{E}_{k}^{\top}\bm{U}^{\star}\bm{U}^{\star\top}=\bm{0}\\
\mathbb{E}\bm{S}_{3,10,k}(\bm{S}_{1,1,k}+\bm{S}_{1,2,k})^{\top} & =\mathbb{E}\mathcal{P}_{U}^{\perp}\mathcal{P}_{U_{k}}^{\perp}\bm{E}_{k}\mathcal{P}_{W}^{\perp}\bm{E}_{k}^{\top}\bm{U}_{k}^{\star}\bm{U}_{k}^{\star\top}\bm{E}_{k}\bm{W}^{\star}\bm{W}^{\star\top}\bm{E}_{k}^{\top}\bm{U}^{\star}\bm{U}^{\star\top}=\bm{0}\\
\mathbb{E}\bm{S}_{3,11,k}(\bm{S}_{1,1,k}+\bm{S}_{1,2,k})^{\top} & =\mathbb{E}\bm{U}_{k}^{\star}\bm{W}^{\star\top}\bm{E}_{k}^{\top}\mathcal{P}_{U}^{\perp}\mathcal{P}_{U_{k}}^{\perp}\bm{E}_{k}\mathcal{P}_{W}^{\perp}\bm{E}_{k}^{\top}\bm{U}_{k}^{\star}\bm{W}^{\star\top}\bm{E}_{k}^{\top}\bm{U}^{\star}\bm{U}^{\star\top}=\bm{0}\\
\mathbb{E}\bm{S}_{3,12,k}(\bm{S}_{1,1,k}+\bm{S}_{1,2,k})^{\top} & =\mathbb{E}\mathcal{P}_{U}^{\perp}\mathcal{P}_{U_{k}}^{\perp}\bm{E}_{k}\bm{W}^{\star}\bm{U}_{k}^{\star\top}\bm{E}_{k}\mathcal{P}_{W}^{\perp}\bm{E}_{k}^{\top}\bm{U}_{k}^{\star}\bm{W}^{\star\top}\bm{E}_{k}^{\top}\bm{U}^{\star}\bm{U}^{\star\top}=\bm{0}\\
\mathbb{E}\bm{S}_{3,13,k}(\bm{S}_{1,1,k}+\bm{S}_{1,2,k})^{\top} & =\mathbb{E}\mathcal{P}_{U}^{\perp}\mathcal{P}_{U_{k}}^{\perp}\bm{E}_{k}\mathcal{P}_{W}^{\perp}\bm{E}_{k}^{\top}\bm{U}_{k}^{\star}\bm{W}_{k}^{\star\top}\bm{E}_{k}^{\top}\bm{U}_{k}^{\star}\bm{W}^{\star\top}\bm{E}_{k}^{\top}\bm{U}^{\star}\bm{U}^{\star\top}=\bm{0}\\
\mathbb{E}\bm{S}_{3,14,k}(\bm{S}_{1,1,k}+\bm{S}_{1,2,k})^{\top} & =-\mathbb{E}\bm{U}_{k}^{\star}\bm{W}^{\star\top}\bm{E}_{k}^{\top}\bm{U}_{k}^{\star}\bm{W}^{\star\top}\bm{E}_{k}^{\top}\mathcal{P}_{U}^{\perp}\mathcal{P}_{U_{k}}^{\perp}\bm{E}_{k}\bm{W}^{\star}\bm{W}^{\star\top}\bm{E}_{k}^{\top}\bm{U}^{\star}\bm{U}^{\star\top}=\bm{0}\\
\mathbb{E}\bm{S}_{3,15,k}(\bm{S}_{1,1,k}+\bm{S}_{1,2,k})^{\top} & =-\mathbb{E}\bm{U}_{k}^{\star}\bm{W}^{\star\top}\bm{E}_{k}^{\top}\mathcal{P}_{U}^{\perp}\mathcal{P}_{U_{k}}^{\perp}\bm{E}_{k}\bm{W}^{\star}\bm{U}_{k}^{\star\top}\bm{E}_{k}\bm{W}^{\star}\bm{W}^{\star\top}\bm{E}_{k}^{\top}\bm{U}^{\star}\bm{U}^{\star\top}=\bm{0}\\
\mathbb{E}\bm{S}_{3,16,k}(\bm{S}_{1,1,k}+\bm{S}_{1,2,k})^{\top} & =-\mathbb{E}\mathcal{P}_{U}^{\perp}\mathcal{P}_{U_{k}}^{\perp}\bm{E}_{k}\bm{W}^{\star}\bm{U}_{k}^{\star\top}\bm{E}_{k}\bm{W}^{\star}\bm{U}_{k}^{\star\top}\bm{E}_{k}\bm{W}^{\star}\bm{W}^{\star\top}\bm{E}_{k}^{\top}\bm{U}^{\star}\bm{U}^{\star\top}=\bm{0}.
\end{align*}
Combining all these terms and the fact that $r_{k}=r$, we have that
\begin{align*}
\mathbb{E}\bm{T}_{1,k}\bm{T}_{3,k}^{\top} & =\mathbb{E}\left(\sum_{i=1}^{5}\sum_{j=1}^{5}\bm{S}_{1,i,k}\bm{S}_{3,j,k}^{\top}\right)\\
 & =-2\sigma^{4}(dr-r^{2})\mathcal{P}_{U}^{\perp}\mathcal{P}_{U_{k}}^{\perp}.
\end{align*}

\subsubsection{Proof of \eqref{eq:T0_T4} \label{subsec:Proof_T0_T4}}

As $\bm{T}_{0,k}$ is fixed, it suffices to show $\mathbb{E}[\bm{T}_{4,k}]$.
Recall from that $\bm{T}_{4,k}=\sum_{j=1}^{55}\bm{S}_{4,j,k}$ (see
Section~\ref{subsec:T4}). We start with the group $j=1,\ldots,20$.

Terms for $j=1,\ldots,4$:
\begin{align*}
\mathbb{E}\bm{S}_{4,1,k} & =-\mathbb{E}\bm{U}_{k}^{\star}\bm{W}^{\star\top}\bm{E}_{k}^{\top}\mathcal{P}_{U}^{\perp}\mathcal{P}_{U_{k}}^{\perp}\bm{E}_{k}\mathcal{P}_{W}^{\perp}\bm{E}_{k}^{\top}\mathcal{P}_{U}^{\perp}\mathcal{P}_{U_{k}}^{\perp}\bm{E}_{k}\\
 & =-\sigma^{4}(n-r-r_{k})(d-r_{k})\bm{U}_{k}^{\star}\bm{W}^{\star\top}\\
\mathbb{E}\bm{S}_{4,2,k} & =-\mathbb{E}\mathcal{P}_{U}^{\perp}\mathcal{P}_{U_{k}}^{\perp}\bm{E}_{k}\bm{W}^{\star}\bm{U}_{k}^{\star\top}\bm{E}_{k}\mathcal{P}_{W}^{\perp}\bm{E}_{k}^{\top}\mathcal{P}_{U}^{\perp}\mathcal{P}_{U_{k}}^{\perp}\bm{E}_{k}=\bm{0}\\
\mathbb{E}\bm{S}_{4,3,k} & =-\mathbb{E}\mathcal{P}_{U}^{\perp}\mathcal{P}_{U_{k}}^{\perp}\bm{E}_{k}\mathcal{P}_{W}^{\perp}\bm{E}_{k}^{\top}\bm{U}_{k}^{\star}\bm{W}^{\star\top}\bm{E}_{k}^{\top}\mathcal{P}_{U}^{\perp}\mathcal{P}_{U_{k}}^{\perp}\bm{E}_{k}=\bm{0}\\
\mathbb{E}\bm{S}_{4,4,k} & =-\mathbb{E}\mathcal{P}_{U}^{\perp}\mathcal{P}_{U_{k}}^{\perp}\bm{E}_{k}\mathcal{P}_{W}^{\perp}\bm{E}_{k}^{\top}\mathcal{P}_{U}^{\perp}\mathcal{P}_{U_{k}}^{\perp}\bm{E}_{k}\bm{W}^{\star}\bm{U}_{k}^{\star\top}\bm{E}_{k}=\bm{0}.
\end{align*}

Terms for $j=5,\ldots,16$:
\begin{align*}
\mathbb{E}\bm{S}_{4,5,k} & =\mathbb{E}\bm{U}_{k}^{\star}\bm{U}_{k}^{\star\top}\bm{E}_{k}\bm{W}^{\star}\bm{U}_{k}^{\star\top}\bm{E}_{k}\mathcal{P}_{W}^{\perp}\bm{E}_{k}^{\top}\mathcal{P}_{U}^{\perp}\mathcal{P}_{U_{k}}^{\perp}\bm{E}_{k}=\bm{0}\\
\mathbb{E}\bm{S}_{4,6,k} & =\mathbb{E}\bm{U}_{k}^{\star}\bm{U}_{k}^{\star\top}\bm{E}_{k}\mathcal{P}_{W}^{\perp}\bm{E}_{k}^{\top}\bm{U}_{k}^{\star}\bm{W}^{\star\top}\bm{E}_{k}^{\top}\mathcal{P}_{U}^{\perp}\mathcal{P}_{U_{k}}^{\perp}\bm{E}_{k}\\
 & =\sigma^{4}(n-r-r_{k})(d-r_{k})\bm{U}_{k}^{\star}\bm{W}^{\star\top}\\
\mathbb{E}\bm{S}_{4,7,k} & =\mathbb{E}\bm{U}_{k}^{\star}\bm{U}_{k}^{\star\top}\bm{E}_{k}\mathcal{P}_{W}^{\perp}\bm{E}_{k}^{\top}\mathcal{P}_{U}^{\perp}\mathcal{P}_{U_{k}}^{\perp}\bm{E}_{k}\bm{W}^{\star}\bm{U}_{k}^{\star\top}\bm{E}_{k}=\bm{0}\\
\mathbb{E}\bm{S}_{4,8,k} & =\mathbb{E}\bm{U}_{k}^{\star}\bm{W}^{\star\top}\bm{E}_{k}^{\top}\bm{U}_{k}^{\star}\bm{U}_{k}^{\star\top}\bm{E}_{k}\mathcal{P}_{W}^{\perp}\bm{E}_{k}^{\top}\mathcal{P}_{U}^{\perp}\mathcal{P}_{U_{k}}^{\perp}\bm{E}_{k}=\bm{0}\\
\mathbb{E}\bm{S}_{4,9,k} & =\mathbb{E}\mathcal{P}_{U}^{\perp}\mathcal{P}_{U_{k}}^{\perp}\bm{E}_{k}\bm{W}^{\star}\bm{W}^{\star\top}\bm{E}_{k}^{\top}\bm{U}_{k}^{\star}\bm{W}^{\star\top}\bm{E}_{k}^{\top}\mathcal{P}_{U}^{\perp}\mathcal{P}_{U_{k}}^{\perp}\bm{E}_{k}=\bm{0}\\
\mathbb{E}\bm{S}_{4,10,k} & =\mathbb{E}\mathcal{P}_{U}^{\perp}\mathcal{P}_{U_{k}}^{\perp}\bm{E}_{k}\bm{W}^{\star}\bm{W}^{\star\top}\bm{E}_{k}^{\top}\mathcal{P}_{U}^{\perp}\mathcal{P}_{U_{k}}^{\perp}\bm{E}_{k}\bm{W}^{\star}\bm{U}_{k}^{\star\top}\bm{E}_{k}=\bm{0}\\
\mathbb{E}\bm{S}_{4,11,k} & =\mathbb{E}\bm{U}_{k}^{\star}\bm{W}^{\star\top}\bm{E}_{k}^{\top}\mathcal{P}_{U}^{\perp}\mathcal{P}_{U_{k}}^{\perp}\bm{E}_{k}\bm{W}^{\star}\bm{W}^{\star\top}\bm{E}_{k}^{\top}\mathcal{P}_{U}^{\perp}\mathcal{P}_{U_{k}}^{\perp}\bm{E}_{k}\\
 & =\sigma^{4}(n-r-r_{k})(n-r+1)\bm{U}_{k}^{\star}\bm{W}^{\star\top}\\
\mathbb{E}\bm{S}_{4,12,k} & =\mathbb{E}\mathcal{P}_{U}^{\perp}\mathcal{P}_{U_{k}}^{\perp}\bm{E}_{k}\bm{W}^{\star}\bm{U}_{k}^{\star\top}\bm{E}_{k}\bm{W}^{\star}\bm{W}^{\star\top}\bm{E}_{k}^{\top}\mathcal{P}_{U}^{\perp}\mathcal{P}_{U_{k}}^{\perp}\bm{E}_{k}=\bm{0}\\
\mathbb{E}\bm{S}_{4,13,k} & =\mathbb{E}\mathcal{P}_{U}^{\perp}\mathcal{P}_{U_{k}}^{\perp}\bm{E}_{k}\mathcal{P}_{W}^{\perp}\bm{E}_{k}^{\top}\bm{U}_{k}^{\star}\bm{U}_{k}^{\star\top}\bm{E}_{k}\bm{W}^{\star}\bm{U}_{k}^{\star\top}\bm{E}_{k}=\bm{0}\\
\mathbb{E}\bm{S}_{4,14,k} & =\mathbb{E}\bm{U}_{k}^{\star}\bm{W}^{\star\top}\bm{E}_{k}^{\top}\mathcal{P}_{U}^{\perp}\mathcal{P}_{U_{k}}^{\perp}\bm{E}_{k}\mathcal{P}_{W}^{\perp}\bm{E}_{k}^{\top}\bm{U}_{k}^{\star}\bm{U}_{k}^{\star\top}\bm{E}_{k}=\bm{0}\\
\mathbb{E}\bm{S}_{4,15,k} & =\mathbb{E}\mathcal{P}_{U}^{\perp}\mathcal{P}_{U_{k}}^{\perp}\bm{E}_{k}\bm{W}^{\star}\bm{U}_{k}^{\star\top}\bm{E}_{k}\mathcal{P}_{W}^{\perp}\bm{E}_{k}^{\top}\bm{U}_{k}^{\star}\bm{U}_{k}^{\star\top}\bm{E}_{k}=\bm{0}\\
\mathbb{E}\bm{S}_{4,16,k} & =\mathbb{E}\mathcal{P}_{U}^{\perp}\mathcal{P}_{U_{k}}^{\perp}\bm{E}_{k}\mathcal{P}_{W}^{\perp}\bm{E}_{k}^{\top}\bm{U}_{k}^{\star}\bm{W}_{k}^{\star\top}\bm{E}_{k}^{\top}\bm{U}_{k}^{\star}\bm{U}_{k}^{\star\top}\bm{E}_{k}=\bm{0}.
\end{align*}

Terms for $j=17,\ldots,20$:

\begin{align*}
\mathbb{E}\bm{S}_{4,17,k} & =-\mathbb{E}\bm{U}_{k}^{\star}\bm{W}^{\star\top}\bm{E}_{k}^{\top}\bm{U}_{k}^{\star}\bm{W}^{\star\top}\bm{E}_{k}^{\top}\bm{U}_{k}^{\star}\bm{W}^{\star\top}\bm{E}_{k}^{\top}\mathcal{P}_{U}^{\perp}\mathcal{P}_{U_{k}}^{\perp}\bm{E}_{k}\\
 & =-\sigma^{4}(n-r-r_{k})\bm{U}_{k}^{\star}\bm{W}^{\star\top}\\
\mathbb{E}\bm{S}_{4,18,k} & =-\mathbb{E}\bm{U}_{k}^{\star}\bm{W}^{\star\top}\bm{E}_{k}^{\top}\bm{U}_{k}^{\star}\bm{W}^{\star\top}\bm{E}_{k}^{\top}\mathcal{P}_{U}^{\perp}\mathcal{P}_{U_{k}}^{\perp}\bm{E}_{k}\bm{W}^{\star}\bm{U}_{k}^{\star\top}\bm{E}_{k}\\
 & =-\sigma^{4}(n-r-r_{k})r_{k}\bm{U}_{k}^{\star}\bm{W}^{\star\top}\\
\mathbb{E}\bm{S}_{4,19,k} & =-\mathbb{E}\bm{U}_{k}^{\star}\bm{W}^{\star\top}\bm{E}_{k}^{\top}\mathcal{P}_{U}^{\perp}\mathcal{P}_{U_{k}}^{\perp}\bm{E}_{k}\bm{W}^{\star}\bm{U}_{k}^{\star\top}\bm{E}_{k}\bm{W}^{\star}\bm{U}_{k}^{\star\top}\bm{E}_{k}\\
 & =-\sigma^{4}(n-r-r_{k})\bm{U}_{k}^{\star}\bm{W}^{\star\top}\\
\mathbb{E}\bm{S}_{4,20,k} & =-\mathbb{E}\mathcal{P}_{U}^{\perp}\mathcal{P}_{U_{k}}^{\perp}\bm{E}_{k}\bm{W}^{\star}\bm{U}_{k}^{\star\top}\bm{E}_{k}\bm{W}^{\star}\bm{U}_{k}^{\star\top}\bm{E}_{k}\bm{W}^{\star}\bm{U}_{k}^{\star\top}\bm{E}_{k}=\bm{0}.
\end{align*}

It turns out that for any $j=1,\ldots,20$. $\mathbb{E}\bm{S}_{4,j,k}=c\bm{U}_{k}^{\star}\bm{W}^{\star\top}$
for some scalar $c$. Then applying \eqref{eq:S4_36}, we have that
\begin{align*}
\mathbb{E}\left[\bm{S}_{4,j,k}+\bm{S}_{4,j+35,k}\right] & =\mathbb{E}\bm{S}_{4,j,k}+\mathbb{E}\bm{S}_{4,j+35,k}\\
 & =\mathbb{E}\bm{S}_{4,j,k}+\mathbb{E}-\bm{S}_{4,j,k}\bm{W}^{\star}\bm{W}^{\star\top}\\
 & =\bm{0},
\end{align*}
where the last equality holds since $\mathbb{E}\bm{S}_{4,j,k}=c\bm{U}_{k}^{\star}\bm{W}^{\star\top}$.
This implies that 
\[
\sum_{j=1}^{20}\mathbb{E}\bm{S}_{4,j,k}+\sum_{j=36}^{55}\mathbb{E}\bm{S}_{4,j,k}=\bm{0}.
\]
We continue with $j=21,\ldots,35$. 

Terms for $j=21,\ldots,25$:
\begin{align*}
\mathbb{E}\bm{S}_{4,21,k} & =-\mathbb{E}\mathcal{P}_{U}^{\perp}\mathcal{P}_{U_{k}}^{\perp}\bm{E}_{k}\mathcal{P}_{W}^{\perp}\bm{E}_{k}^{\top}\mathcal{P}_{U}^{\perp}\mathcal{P}_{U_{k}}^{\perp}\bm{E}_{k}\mathcal{P}_{W}^{\perp}\bm{E}_{k}^{\top}\bm{U}_{k}^{\star}\bm{W}^{\star\top}=\bm{0}\\
\mathbb{E}\bm{S}_{4,22,k} & =\mathbb{E}\mathcal{P}_{U}^{\perp}\mathcal{P}_{U_{k}}^{\perp}\bm{E}_{k}\mathcal{P}_{W}^{\perp}\bm{E}_{k}^{\top}\mathcal{P}_{U}^{\perp}\mathcal{P}_{U_{k}}^{\perp}\bm{E}_{k}\bm{W}^{\star}\bm{U}_{k}^{\star\top}\bm{E}_{k}\bm{W}^{\star}\bm{W}^{\star\top}=\bm{0}\\
\mathbb{E}\bm{S}_{4,23,k} & =\mathbb{E}\mathcal{P}_{U}^{\perp}\mathcal{P}_{U_{k}}^{\perp}\bm{E}_{k}\mathcal{P}_{W}^{\perp}\bm{E}_{k}^{\top}\bm{U}_{k}^{\star}\bm{W}^{\star\top}\bm{E}_{k}^{\top}\mathcal{P}_{U}^{\perp}\mathcal{P}_{U_{k}}^{\perp}\bm{E}_{k}\bm{W}^{\star}\bm{W}^{\star\top}=\bm{0}\\
\mathbb{E}\bm{S}_{4,24,k} & =\mathbb{E}\mathcal{P}_{U}^{\perp}\mathcal{P}_{U_{k}}^{\perp}\bm{E}_{k}\bm{W}^{\star}\bm{U}_{k}^{\star\top}\bm{E}_{k}\mathcal{P}_{W}^{\perp}\bm{E}_{k}^{\top}\mathcal{P}_{U}^{\perp}\mathcal{P}_{U_{k}}^{\perp}\bm{E}_{k}\bm{W}^{\star}\bm{W}^{\star\top}=\bm{0}\\
\mathbb{E}\bm{S}_{4,25,k} & =\mathbb{E}\bm{U}_{k}^{\star}\bm{W}^{\star\top}\bm{E}_{k}^{\top}\mathcal{P}_{U}^{\perp}\mathcal{P}_{U_{k}}^{\perp}\bm{E}_{k}\mathcal{P}_{W}^{\perp}\bm{E}_{k}^{\top}\mathcal{P}_{U}^{\perp}\mathcal{P}_{U_{k}}^{\perp}\bm{E}_{k}\bm{W}^{\star}\bm{W}^{\star\top}\\
 & =\sigma^{4}(n-r-r_{k})(d-r_{k})\bm{U}_{k}^{\star}\bm{W}^{\star\top}.
\end{align*}

Terms for $j=26,\ldots,29$:

\begin{align*}
\mathbb{E}\bm{S}_{4,26,k} & =\mathbb{E}\mathcal{P}_{U}^{\perp}\mathcal{P}_{U_{k}}^{\perp}\bm{E}_{k}\mathcal{P}_{W}^{\perp}\bm{E}_{k}^{\top}\mathcal{P}_{U}^{\perp}\mathcal{P}_{U_{k}}^{\perp}\bm{E}_{k}\bm{W}^{\star}\bm{W}^{\star\top}\bm{E}_{k}^{\top}\bm{U}_{k}^{\star}\bm{W}^{\star\top}=0\\
\mathbb{E}\bm{S}_{4,27,k} & =\mathbb{E}\mathcal{P}_{U}^{\perp}\mathcal{P}_{U_{k}}^{\perp}\bm{E}_{k}\mathcal{P}_{W}^{\perp}\bm{E}_{k}^{\top}\bm{U}_{k}^{\star}\bm{U}_{k}^{\star\top}\bm{E}_{k}\mathcal{P}_{W}^{\perp}\bm{E}_{k}^{\top}\bm{U}_{k}^{\star}\bm{W}^{\star\top}=0\\
\mathbb{E}\bm{S}_{4,28,k} & =\mathbb{E}\mathcal{P}_{U}^{\perp}\mathcal{P}_{U_{k}}^{\perp}\bm{E}_{k}\bm{W}^{\star}\bm{W}^{\star\top}\bm{E}_{k}^{\top}\mathcal{P}_{U}^{\perp}\mathcal{P}_{U_{k}}^{\perp}\bm{E}_{k}\mathcal{P}_{W}^{\perp}\bm{E}_{k}^{\top}\bm{U}_{k}^{\star}\bm{W}^{\star\top}=0\\
\mathbb{E}\bm{S}_{4,29,k} & =\mathbb{E}\bm{U}_{k}^{\star}\bm{U}_{k}^{\star\top}\bm{E}_{k}\mathcal{P}_{W}^{\perp}\bm{E}_{k}^{\top}\mathcal{P}_{U}^{\perp}\mathcal{P}_{U_{k}}^{\perp}\bm{E}_{k}\mathcal{P}_{W}^{\perp}\bm{E}_{k}^{\top}\bm{U}_{k}^{\star}\bm{W}^{\star\top}\\
 & =\sigma^{4}(n-r-r_{k})(d-r_{k})\bm{U}_{k}^{\star}\bm{W}^{\star\top}.
\end{align*}

Terms for $j=30,\ldots,35$:
\begin{align*}
\mathbb{E}\bm{S}_{4,30,k} & =-\mathbb{E}\bm{U}_{k}^{\star}\bm{W}^{\star\top}\bm{E}_{k}^{\top}\bm{U}_{k}^{\star}\bm{W}^{\star\top}\bm{E}_{k}^{\top}\mathcal{P}_{U}^{\perp}\mathcal{P}_{U_{k}}^{\perp}\bm{E}_{k}\mathcal{P}_{W}^{\perp}\bm{E}_{k}^{\top}\bm{U}_{k}^{\star}\bm{W}^{\star\top}=\bm{0}\\
\mathbb{E}\bm{S}_{4,31,k} & =-\mathbb{E}\bm{U}_{k}^{\star}\bm{W}^{\star\top}\bm{E}_{k}^{\top}\mathcal{P}_{U}^{\perp}\mathcal{P}_{U_{k}}^{\perp}\bm{E}_{k}\bm{W}^{\star}\bm{U}_{k}^{\star\top}\bm{E}_{k}\mathcal{P}_{W}^{\perp}\bm{E}_{k}^{\top}\bm{U}_{k}^{\star}\bm{W}^{\star\top}\\
 & =-\sigma^{4}(n-r-r_{k})(d-r_{k})\bm{U}_{k}^{\star}\bm{W}^{\star\top}\\
\mathbb{E}\bm{S}_{4,32,k} & =-\mathbb{E}\bm{U}_{k}^{\star}\bm{W}^{\star\top}\bm{E}_{k}^{\top}\mathcal{P}_{U}^{\perp}\mathcal{P}_{U_{k}}^{\perp}\bm{E}_{k}\mathcal{P}_{W}^{\perp}\bm{E}_{k}^{\top}\bm{U}_{k}^{\star}\bm{W}^{\star\top}\bm{E}_{k}^{\top}\bm{U}_{k}^{\star}\bm{W}^{\star\top}=\bm{0}\\
\mathbb{E}\bm{S}_{4,33,k} & =-\mathbb{E}\mathcal{P}_{U}^{\perp}\mathcal{P}_{U_{k}}^{\perp}\bm{E}_{k}\bm{W}^{\star}\bm{U}_{k}^{\star\top}\bm{E}_{k}\bm{W}^{\star}\bm{U}_{k}^{\star\top}\bm{E}_{k}\mathcal{P}_{W}^{\perp}\bm{E}_{k}^{\top}\bm{U}_{k}^{\star}\bm{W}^{\star\top}=\bm{0}\\
\mathbb{E}\bm{S}_{4,34,k} & =-\mathbb{E}\mathcal{P}_{U}^{\perp}\mathcal{P}_{U_{k}}^{\perp}\bm{E}_{k}\bm{W}^{\star}\bm{U}_{k}^{\star\top}\bm{E}_{k}\mathcal{P}_{W}^{\perp}\bm{E}_{k}^{\top}\bm{U}_{k}^{\star}\bm{W}^{\star\top}\bm{E}_{k}^{\top}\bm{U}_{k}^{\star}\bm{W}^{\star\top}=\bm{0}\\
\mathbb{E}\bm{S}_{4,35,k} & =-\mathbb{E}\mathcal{P}_{U}^{\perp}\mathcal{P}_{U_{k}}^{\perp}\bm{E}_{k}\mathcal{P}_{W}^{\perp}\bm{E}_{k}^{\top}\bm{U}_{k}^{\star}\bm{W}^{\star\top}\bm{E}_{k}^{\top}\bm{U}_{k}^{\star}\bm{W}^{\star\top}\bm{E}_{k}^{\top}\bm{U}_{k}^{\star}\bm{W}^{\star\top}=\bm{0}.
\end{align*}
In summary, we have that 
\[
\mathbb{E}\bm{T}_{4,k}=\sum_{j=1}^{55}\mathbb{E}\bm{S}_{4,j,k}=\sigma^{4}(n-r-r_{k})(d-r_{k})\bm{U}_{k}^{\star}\bm{W}^{\star\top}.
\]
Then 

\[
\mathbb{E}\bm{T}_{0,k}\bm{T}_{4,k}^{\top}=\sigma^{4}(n-r-r_{k})(d-r_{k})\bm{U}^{\star}\bm{V}^{\star\top}\bm{W}^{\star}\bm{U}_{k}^{\star\top}.
\]

\end{document}